\documentclass[ds, nosecondcompanypicture, nofrontpicture, nolinenumbering, nofirstcompanypicture]{mscthesis}

\title{Predicting Delayed Train Trajectories on the Dutch Railway Network: An Explainable AI Evaluation of Topological, Operational, and Weather Features Using Tree-Based Ensemble Methods}

\author{Jia Long Bao}
\email{jialongb@gmail.com}
\affiliation{
  \institution{Informatics Institute, University of Amsterdam}
  \city{Amsterdam}
  \country{The Netherlands}
}

\author{Ali Mohammed Mansoor Alsahag}
\authornote{Corresponding author.}
\email{a.m.m.alsahag@uva.nl}
\affiliation{
  \institution{Informatics Institute, University of Amsterdam}
  \city{Amsterdam}
  \country{The Netherlands}
}

\author{Seyed Sahand Mohammadi Ziabari}
\email{s.s.mohammadiziabari@uva.nl}
\affiliation{
  \institution{Informatics Institute, University of Amsterdam}
  \city{Amsterdam}
  \country{The Netherlands}
}

\keywords{Network-Wide Railway Delay Prediction, Spatiotemporal Data, Explainable Artificial Intelligence, Tree-Based Ensemble Models, Supervised Classification}
\begin{document}
\pagestyle{plain}

\fixemptypage
\setcounter{page}{0}

\begin{abstract} 

The reliable prediction of passenger train delays is a critical component of railway management. While contemporary research frequently attempts to maximize absolute accuracy by deploying opaque deep learning architectures, the underlying data mechanics driving longitudinal predictive decay remain underexplored. Consequently, this study provides an explainable temporal robustness analysis of network-wide railway delay prediction. Focusing on the Dutch railway network, this research utilizes interpretable tree-based ensembles to integrate granular topological, environmental, and operational features. The overarching finding establishes that while feature-rich tree-based models improve simultaneous (within-month) prediction, predictive performance systematically degrades when evaluated across non-simultaneous (future) months. Furthermore, multi-horizon SHAP and dispersion analyses associate this degradation with environmental feature volatility and instability within the statistical target definition. Ultimately, this study demonstrates that richer feature sets alone are insufficient to resolve long-term forecasting constraints, underscoring the necessity to transition toward dynamic, season-aware architectures anchored by absolute operational boundaries.
\end{abstract}

\maketitle

\section{Introduction}
\label{sec:introduction}

This section examines the economic consequences of railway delays, identifies research gaps in network-wide delay prediction, and defines the guiding research questions.

\subsection{Economic and Operational Context}

\subsubsection{Financial Costs of Delays}
In 2025, the Dutch national railway operator, Nederlandse Spoorwegen (NS), distributed €9.7 million in annual compensation and refunds \cite{ns_ns_2026}. Beyond direct fiscal burdens, schedule deviations result in cumulative time losses costing the Dutch economy over €400 million annually \cite{savelberg_estimation_2017}.

\subsubsection{Systemic Vulnerabilities}
The Netherlands maintains a high-density rail network where high-frequency scheduling amplifies sensitivities to operational failures and environmental factors, such as winter weather or seasonal track debris \cite{noauthor_about_2026, wilde_time_2025}. Managing this complexity requires a continuous trade-off between mitigating cascading delays and maintaining individual service punctuality \cite{konig_review_2020}.

\subsubsection{Regulatory AI Transparency in Public Infrastructure}
The European Union Artificial Intelligence Act (EU AI Act) mandates transparency and human oversight for automated decision-making systems deployed within high-risk infrastructure domains, such as the Dutch national railway \cite{eu_ai_act_2024}. With transparency rules scheduled for enforcement by August 2026, developing interpretable machine learning pipelines is a practical regulatory necessity \cite{european_commission_ai_2026}. This research utilizes SHapley Additive exPlanations (SHAP) to align network-wide delay modeling with these emerging regulatory requirements \cite{lundberg_unified_2017, lundberg_local_2020}.

\subsection{Research Gaps in Network-Wide Train Delay Prediction}

\subsubsection{The Interpretability and Temporal Robustness Gap}
Recent literature indicates a paradigm shift toward short-term, real-time prediction methodologies utilizing complex hybrid architectures to maximize immediate operational accuracy \cite{tiong_review_2023, wu_hybrid_2025, chang_sta-gcn_2023, zong_tsta-gcn_2025, wandelt_flight_2025, li_train_2026}. 
However, this trajectory frequently neglects spatiotemporal data representation within aggregated, long-term network-wide predictive frameworks \cite{tiong_review_2023}. While topological features extracted from spatial graphs have facilitated effective link prediction in alternative domains \cite{lei_forecasting_2022}, adapting these techniques to railway networks reveals significant empirical generalization challenges across unseen temporal horizons \cite{kampere_predicting_2025}.

\subsubsection{Positioning Within Contemporary Literature}
This research diverges from recent foundational frameworks across several key dimensions. Compared to the baseline established by Kämper \cite{kampere_predicting_2025}, this study adopts a highly granular stop-to-stop network representation, incorporates engineered operational features, and integrates canceled services into the delay target. Unlike prior research utilizing regional weather data \cite{brakenhoff_dynamic_2026}, this study employs exact station coordinates to minimize spatial noise. Finally, in contrast to black-box deep learning architectures \cite{van_der_maas_predicting_2025}, this research prioritizes diagnostic interpretability, utilizing tree-based ensembles to enable explicit SHAP value analysis.

Ultimately, the central contribution of this research is not exclusively the absolute prediction of delays, but an explainable temporal robustness analysis of network-wide railway delay prediction. By trading the raw predictive density of deep learning for the diagnostic capability of tree-based models and SHAP analysis, this approach systematically links predictive degradation to environmental feature volatility and target instability.

\subsection{Methodological Approach and Research Objectives}
Tree-based ensembles are selected for this network-wide analysis because they bypass strict statistical distribution assumptions, capture non-linear feature interactions, and generate explicit feature importance rankings \cite{tiong_review_2023}. A network-wide spatial scope is necessary because delays within dense infrastructures rarely remain isolated; this approach ensures cascading delays are accurately modeled across the entire topology.

\subsubsection{Temporal Evaluation Framework}
To evaluate long-term temporal generalization, operational data is partitioned into discrete monthly snapshots. While this isolates macroscopic network states from real-time operational noise \cite{tiong_review_2023, lei_forecasting_2022, kampere_predicting_2025}, it introduces inherent methodological trade-offs. Specifically, defining a binary target variable across aggregated monthly periods introduces vulnerability to temporal label shift as seasonal operational frequencies fluctuate \cite{kampere_predicting_2025}. To mitigate this structural limitation, this research prioritizes temporal robustness by incorporating a sensitivity analysis to validate the target definition across alternative operational thresholds. Furthermore, rather than solely optimizing predictive accuracy, this study prioritizes temporal robustness through a comparative evaluation of topological, operational, and environmental feature sets. Finally, SHAP values are utilized to diagnose feature volatility, isolate the primary factors underlying predictive stability, and critically assess the underlying assumptions of the monthly aggregation framework \cite{lundberg_unified_2017}.

This study evaluates whether integrating topological, environmental, and operational features into tree-based ensemble models supports temporal generalization for network-wide passenger train delay prediction across monthly snapshots. Specifically, it compares balanced accuracy between simultaneous (within-month) and non-simultaneous (future-month) evaluations using paired Wilcoxon tests. It also examines how the relative importance of these feature categories changes across future-month horizons, using SHAP rankings from the primary XGBoost model with a trajectory-based split, and identifies the variables whose importance is most stable over time.

The remainder of this paper reviews the relevant literature, details the methodology and experimental setup, and concludes with a discussion of the results.

\section{Related Work}
\label{sec:related_work}

The prediction of long-term transportation dynamics increasingly relies on topological feature extraction and graph-based learning. While effective in aviation, its application to railway delay prediction reveals a fundamental gap between representational richness, temporal generalization, and model interpretability. This section evaluates prior methodologies to ground the diagnostic framework required for \textbf{SRQ1} and \textbf{SRQ2}.

\subsection{Topological Representation and Network Dynamics}

Modeling spatial dependencies remains a central challenge in machine learning \cite{nikparvar_machine_2021}. To circumvent the high-dimensional feature spaces generated by traditional encoding, recent methodologies extract topological features directly from spatial graphs for integration into tabular models \cite{lei_forecasting_2022, kampere_predicting_2025}.  

\subsubsection{Foundational Representations and Railway Limitations}

Dynamic graph topology serves as an effective proxy for spatial relationships. Lei demonstrated that localized topological features, particularly operational edge weights, facilitate effective tabular link prediction in aviation \cite{lei_forecasting_2022}. However, Lei noted severe generalization degradation under non-simultaneous testing in alternative domains such as the Brazil bus network, suggesting topological features struggle under temporal variation.

Adapting this framework to the Dutch railway network, Kämper observed similarly restricted predictive generalization across all temporal testing scenarios \cite{kampere_predicting_2025}. This failure was linked to this limited generalization to two primary limitations. First, trajectories were aggregated to origin-destination pairs, overlooking localized network behavior and inducing severe data sparsity. Second, operational edge weights were explicitly excluded to prevent target leakage. Despite employing node centrality metrics and hyperparameter tuning, performance remained near random baseline levels. Kämper suggested this failure likely associated with data scarcity and a lack of external feature richness rather than inherent methodological flaws.

\subsection{Model Complexity and Data Integrity}
\subsubsection{The Interpretability Trade-off}
Within the context of prior work on Dutch railway delay prediction, three complementary research directions emerge. Kämper employed a topological, tabular framework on monthly aggregated data, demonstrating limited generalization under simultaneous and non-simultaneous evaluation constrained by data sparsity and restricted feature richness \cite{kampere_predicting_2025}. Brakenhoff adapted this approach by incorporating operational and environmental features, improving predictive performance for disruption and cancellation tasks \cite{brakenhoff_dynamic_2026}. In contrast, van der Maas introduced hybrid spatio-temporal deep learning architectures (GAT-GRU and GAT-LSTM), achieving substantially stronger non-simultaneous predictive performance through richer modeling of temporal dependencies \cite{van_der_maas_predicting_2025}.

Building upon these contributions, this study prioritizes interpretability and feature stability analysis over absolute model complexity. While hybrid deep learning architectures (GATs) capture complex spatiotemporal dependencies, their attention weights function primarily as structural proxies and rarely provide explicit feature explainability regarding why a specific prediction is made \cite{jain_attention_2019, wiegreffe_attention_2019, serrano_is_2019, bibal_is_2022, hassija_interpreting_2024}. Because deployed models within critical public infrastructure must satisfy strict transparency requirements under the EU AI Act \cite{eu_ai_act_2024}, this research utilizes tree-based ensembles integrated with SHAP to calculate specific feature attributions \cite{lundberg_unified_2017, lundberg_local_2020, bastings_elephant_2020}. While SHAP values quantify correlational predictive reliance rather than causality, they provide the necessary diagnostic transparency for regulatory compliance \cite{bastings_elephant_2020, grinsztajn_why_2022}. By trading the raw predictive density of deep learning for the diagnostic capability of SHAP, this approach rigorously evaluates feature robustness over time. 

Furthermore, while prior work demonstrated the efficacy of multi-year seasonal models to capture recurring patterns \cite{brakenhoff_dynamic_2026}, this study explicitly restricts its scope to consecutive cross-month generalization to strictly isolate the immediate temporal degradation of the feature space. 

\subsubsection{Environmental Data Integrity}
Prior methodologies also exhibited environmental data limitations. Previous studies utilizing nearest-neighbor approximations of relocated KNMI weather station data likely introduced structural measurement noise and spatial inconsistencies \cite{brakenhoff_dynamic_2026, knmi_knmi_2026}. Consequently, the limited predictive contribution of environmental features in prior work may reflect data aggregation flaws rather than an absence of meteorological signal. To mitigate this noise, this methodology extracts historical weather data utilizing exact station coordinates \cite{van_der_maas_predicting_2025, wilde_time_2025}.

\subsection{Temporal Instability and Explainability}
Beyond representational constraints, temporal instability remains a critical challenge. Lei attributed near-random non-simultaneous performance to unstable feature importance for the Brazil bus dataset, where models overfit to the structural patterns of individual training snapshots \cite{lei_forecasting_2022}. To systematically diagnose this degradation, SHAP values are employed as a post-hoc interpretability tool to track feature stability across temporal contexts. Crucially, SHAP is restricted to interpretation rather than feature selection, as its use in selection can artificially degrade predictive performance \cite{fryer_shapley_2021, brakenhoff_dynamic_2026}. Recent state-of-the-art studies continue to emphasize SHAP as a critical tool for interpreting model behavior across diverse domains \cite{leneman_explainable_2026, chen_analytics_2025}.

\subsection{Research Gap Summary}
The existing literature demonstrates that long-term network-wide delay prediction is constrained by data sparsity and feature exclusion \cite{kampere_predicting_2025}. Conversely, deep learning methodologies that successfully capture these complex spatiotemporal dependencies sacrifice the diagnostic interpretability required for deployment in critical public infrastructure \cite{van_der_maas_predicting_2025}. 

To address this fundamental gap, this research diverges from prior static frameworks by proposing a highly granular, multi-layered tabular approach. By integrating dynamic topological graphs, operational edge weights, and exact-coordinate environmental data within interpretable tree-based ensembles, this study navigates the trade-off between representational richness and diagnostic transparency \cite{li_train_2026, brakenhoff_dynamic_2026, zhong_prediction_2025, van_der_maas_predicting_2025, biswas_spatio-temporal_2024}.

To statistically evaluate the predictive capacities of this proposed architecture, the following hypotheses are formulated:
\begin{itemize}
    \item \textbf{H1 (Simultaneous Validity):} Under simultaneous testing (trajectory-based split), tabular ensemble performance will significantly exceed the 50\% null baseline, establishing the discriminative validity of the engineered features.
    \item \textbf{H2 (Temporal Generalization):} Under non-simultaneous testing (trajectory-based split), performance will exhibit a statistically significant degradation compared to simultaneous evaluations.
    \item \textbf{H3 (Structural Degradation):} Under a strict time-based split, performance will significantly degrade across all evaluated models during non-simultaneous horizons, suggesting that temporal instability is an inherent operational phenomenon rather than an artifact of sample size fluctuations.
\end{itemize}

While these hypotheses address the statistical significance of performance shifts (SRQ1), the internal feature dynamics underlying these shifts (SRQ2) are subsequently evaluated via post-hoc SHAP analysis.
\section{Methodology}
\label{sec:methodology}

\begin{figure*}[t]
    \centering
    \includegraphics[width=\textwidth]{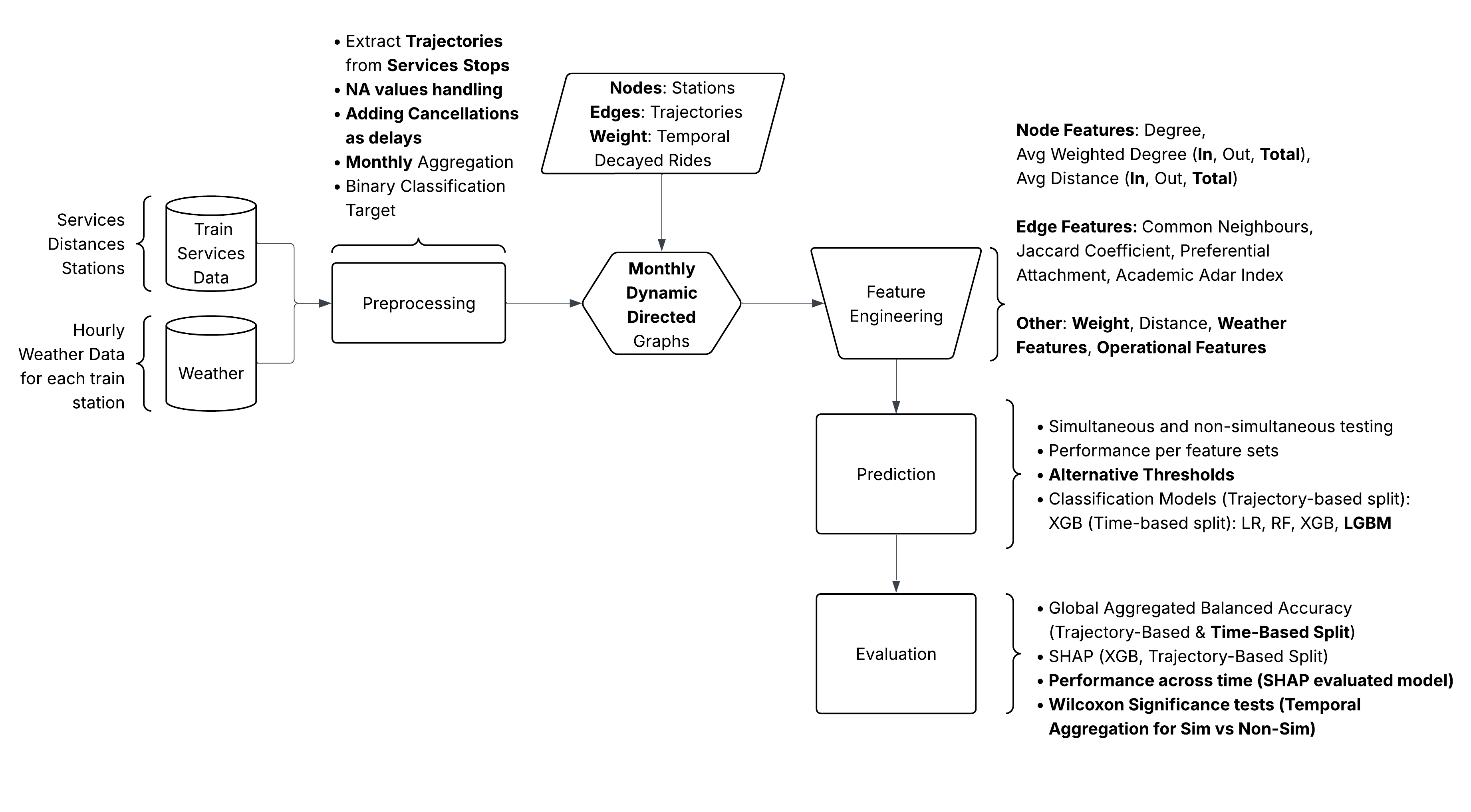}
    \caption{Research Scope and Methodologies}
    \label{fig:methodology_proposed}
\end{figure*}

This section details the analytical framework and feature engineering procedures used to evaluate the temporal predictability of network-wide train delays. Rather than optimizing strictly for absolute algorithmic accuracy, the methodology utilizes comparative evaluation designs to isolate feature performance across consecutive monthly horizons. A comprehensive overview is visualized in  Figure \ref{fig:methodology_proposed}.

\subsection{Data Acquisition and Pre-processing}
\label{subsec:data_acquisition}

\subsubsection{Railway Operations Data and Spatial Filtering}
The operational dataset, sourced from RijdenDeTreinen \cite{noauthor_open_nodate}, comprises historical services from 2019 to 2024, a station distance matrix, and reference spatial data. Freight and international services were excluded to isolate standard domestic passenger operations, yielding an initial dataset of 83,810,933 raw station stop records. 

These records were chronologically sequenced into directed trajectories (scheduled connections between consecutive stations), adopting the high-density spatial formulation utilized by Brakenhoff \cite{brakenhoff_dynamic_2026} to systematically increase data density. To ensure topological validity, bidirectional spatial filtering was applied: trajectories containing unreferenced stations, and reference stations lacking historical trajectories, were dropped. This guarantees all topological features rely exclusively on verified spatial distances within the Netherlands (Appendix \ref{sec:apx:eda_supplementary} Figure \ref{fig:eda_stations_noservices}).

\subsubsection{Imputation and Service Cancellations}
\label{subsec:cancellations}
Missing terminal timestamps were imputed deterministically based on physical network constraints: missing arrivals for starting services and missing departures for terminating services were set to zero delay. Both partially and fully canceled services were retained and assigned an arrival delay penalty ($>0$). As established in Section \ref{sec:introduction}, this framework explicitly retains both partially and fully canceled services within the predictive target, diverging from prior baselines \cite{kampere_predicting_2025}. This inclusive definition recognizes a cancellation as a literal topological edge removal, thereby capturing actual network-wide disruption \cite{koninkrijksrelaties_wet_2024, eu_rail_passengers_2021}. Furthermore, current operational data lacks the 'slowly changing dimensions' to track the change of the cancellation status over time, which is necessary to differentiate ad-hoc operational failures from planned maintenance cancellations (e.g., exact announcement timestamps). Given this limitation, all cancellation instances are conservatively treated as delays.


\subsubsection{Environmental Data Acquisition and Cleaning}
Historical hourly weather data for exact station coordinates was acquired via the Open-Meteo API \cite{Zippenfenig_Open-Meteo}. Duplicate timestamp entries, occurring annually during daylight saving time (DST) transitions, were retained. Because the underlying meteorological values remain distinct despite identical timestamps, retaining these sequential measurements preserves the statistical integrity of subsequent monthly aggregations. However, this assumption could be revisited in future work.

\subsubsection{Temporal Aggregation and Target Definition}
Trajectory records were aggregated into discrete monthly snapshots to capture macroscopic network states. An activity threshold of at least four scheduled rides per month was applied to filter anomalous routing noise \cite{kampere_predicting_2025}. 

To establish baseline operational reliability, the delay ratio for a given trajectory $T$ was defined as:
$$
\text{ratio\_arrival\_delay}_T = \frac{\text{scheduled rides with arrival delay } > 0}{\text{total scheduled rides}}
$$
To mitigate majority-class bias during model training, the binary target label $Y_T$ ('Is Significantly Delayed') was established via a global median threshold. The target definition was adapted for comparability with prior work and controlled class balance \cite{kampere_predicting_2025}:
$$
Y_T = 
\begin{cases} 
1 & \text{if } \text{ratio\_arrival\_delay}_T > 0.2629 \\
0 & \text{if } \text{ratio\_arrival\_delay}_T \leq 0.2629
\end{cases}
$$

While the initial median-based threshold ensures algorithmic class balance, it introduces vulnerability to temporal label shift as actual delay ratios fluctuate (Appendix \ref{sec:apx:eda_supplementary} Figure \ref{fig:eda_class_balance_OT}). To verify that temporal degradation is not solely an artifact of this static definition, a sensitivity analysis utilizing XGBoost (under the trajectory-based splitting schema) evaluates two alternative thresholds: an Absolute Operational Threshold (fixed 80\% punctuality, in other words a 0.20 threshold), chosen in alignment with railway practice as a conservative variant of the NS 3-minute minimum floor (84.4\%) \cite{ns_ns_2026}, and a Month Relative Historical Seasonal Threshold. Evaluating predictive degradation across these varied definitions enables a diagnosis of whether temporal instability is driven by the global median target formulation itself.

Furthermore, to enhance operational utility, the experimental framework evaluates a regression model across two operational resolution levels as an additional sensitivity analysis: Schedule Adherence (0-minute threshold) and Regulatory Punctuality (3-minute threshold). These regression targets provide a granular assessment of service reliability, complementing the classification tasks. The underlying regression models are optimized using a logistic loss objective function (reg:logistic). This approach applies a logistic transformation to the model outputs, thereby ensuring that all continuous punctuality and adherence ratio predictions are naturally bounded within the interval $[0, 1]$. Model parameters are subsequently learned by minimizing the negative log-likelihood, providing a mathematically robust framework for fractional response estimation. By comparing predictive performance across these diverse classification and regression formulations, this study demonstrates whether the model’s efficacy is robust regardless of the chosen operational strictness or target architecture.

To evaluate the operational utility of these continuous formulations, the models were assessed using Mean Absolute Error (MAE) and Root Mean Squared Error (RMSE). These metrics were contrasted against a Null Baseline (predicting the mean target value of the training set) to confirm that the models retain continuous predictive signal extraction capabilities alongside their classification boundaries.

While exploratory time-series analysis indicates that pure delays and outright cancellations exhibit distinct volumetric trends (Appendix \ref{sec:apx:operational_metrics} Figure \ref{fig:pure_eda}), their combined delays to structural routing integrity necessitates a unified predictive target for macroscopic evaluation.


\subsection{Feature Engineering}
\label{subsec:feature_engineering}
Predictive features across operational, spatial, and environmental domains were engineered at the aggregated monthly trajectory level to align with long-term predictive horizons.

\subsubsection{Operational Feature Engineering}
\label{subsubsec:operational_features}
Operational context was established using the total service volume of the last active month and the elapsed months since the previous service. Inspired by the network operational features introduced by Brakenhoff \cite{brakenhoff_dynamic_2026}, a decayed operational edge weight ($W_{\text{decayed}}$) was calculated to approximate sequential pattern recognition without inducing target leakage, as it relies exclusively on past information:
$$
W_{\text{decayed}} = V_{t-1}\frac{1}{\Delta m} \quad \text{for } 1 \leq \Delta m \leq 12
$$
where $V_{t-1}$ is the most recent active service volume and $\Delta m$ is the months elapsed. A maximum 12-month lookback aligns with the annual NS scheduling cycle. The feature behaves as a proxy for the recent service rate because it anchors the model's focus on the most recent active month while proportionately scaling down its relevance based on elapsed inactivity. If no prior active month was identified, $\Delta m$ was imputed to 12.

\subsubsection{Spatial and Topological Feature Engineering}
\label{subsubsec:spatial_features}
Dynamic, directed network graphs $G_t = (V_t, E_t)$ were constructed for each month (Appendix \ref{sec:apx:eda_supplementary} Figure \ref{fig:eda_map}). Static graphs were avoided to capture seasonal topological shifts. Using physical distances and the decayed operational edge weights ($W_{\text{decayed}}$), node-level features (e.g., weighted outbound/inbound degrees, average distances) were extracted to quantify a station's outward distribution capabilities and localized congestion potential, expanding upon the baseline features established by Brakenhoff \cite{brakenhoff_dynamic_2026}. Concurrently, trajectory-level features (e.g., Jaccard Coefficient, Adamic-Adar Index) were calculated to evaluate routing redundancy (Appendix \ref{sec:apx:feature_hyperparams} Table \ref{tab:topological_features}).

\subsubsection{Environmental Weather Features}
\label{subsubsec:weather_features}
Hourly station-level historical weather data was aggregated into monthly station-level features (e.g., air temperature, rain, wind speed, wind gusts, snowfall, snow depth, and soil temperature) to scale the daily predictive validity of these specific variables established by van der Maas \cite{van_der_maas_predicting_2025} into macroscopic seasonal network vulnerabilities (Appendix \ref{sec:apx:feature_hyperparams} Table \ref{tab:feature_sets}).

\subsubsection{Feature Set Construction and Ablation Logic}
\label{subsubsec:ablation_logic}
To isolate specific predictive drivers, a comprehensive ablation logic was established. Feature sets (Topological, Weight, Weather, Operational) are first evaluated independently to establish baselines, followed by incremental integration (e.g., Topology + Weight, Topology + Weight + Weather, Topology + Weight + Operational, All Features) to assess complex interactions. 

Furthermore, explicit feature selection prior to training was omitted. While this risks retaining highly collinear variables, particularly spatially proximate environmental metrics and could increase overfitting under temporal shift, it is a deliberate diagnostic choice. Retaining the unfiltered feature space allows tree-based algorithms to inherently manage multicollinearity during node splitting, while enabling post-hoc SHAP analysis to identify exactly which variables introduce noise or degrade over time. Consequently, aggressive feature pruning optimized strictly for raw accuracy is designated as future work.

\subsubsection{Exploratory Data Analysis (EDA)}
The aggregated dataset comprises 49,793 stable trajectories across 43 engineered features (Appendix \ref{sec:apx:feature_hyperparams} Table \ref{tab:feature_sets}). Structurally, the network maintains an average of 247 station nodes and 692 active trajectory edges per monthly snapshot. While scheduled passenger service volumes exhibited minimal variance across the temporal horizon (Appendix \ref{sec:apx:eda_supplementary} Figure \ref{fig:eda_total_services}), an upward trend in unique trajectory edges was observed from 2022 onward (Appendix \ref{sec:apx:eda_supplementary} Figure \ref{fig:eda_target_unique_edges}). Finally, although the overarching target class distribution is mathematically balanced at a 50/50 ratio (Appendix \ref{sec:apx:eda_supplementary} Figure \ref{fig:class_balance_pie}), the underlying proportion of significantly delayed trajectories exhibits natural operational volatility, fluctuating between 10\% and 90\% across individual months (Appendix \ref{sec:apx:eda_supplementary} Figure \ref{fig:eda_target_unique_edges}).

\subsection{Evaluation Framework and Experimental Setup}
\label{subsec:evaluation_framework}
This subsection details the validation methodologies, dataset splitting strategies, and model configurations utilized to systematically benchmark predictive temporal generalizability. A universal random seed was uniformly established across all computational environments for reproducibility.

\subsubsection{Temporal Validation Phases and Data Splitting}
To assess out-of-sample generalizability, models were first evaluated under Simultaneous Testing (trained and tested within the same monthly snapshot) and Non-Simultaneous Testing (trained on historical monthly snapshot, tested on unseen future months) \cite{lei_forecasting_2022, kampere_predicting_2025}. 

These phases utilized two data splitting schemas  Table \ref{tab:experimental_baselines}. A strict time-based split (70/30 chronological division) isolated temporal degradation while controlling for sample size. Additionally, a trajectory-based split with random under-sampling was applied exclusively to the XGBoost model to maintain alignment with prior literature and standardize SHAP evaluations \cite{lei_forecasting_2022}.

\subsubsection{Model Configuration}
While Deep Learning architectures (e.g., GNNs) might offer high predictive accuracy in spatiotemporal tasks, tree-based ensembles were selected as the primary methodology to prioritize feature interpretability and diagnostic transparency, which are often obscured in black-box deep learning models. The models evaluated include Logistic Regression (serving as a linear baseline) alongside three tree-based ensembles: Random Forest, XGBoost, and LightGBM. These specific non-linear models were selected for their native capacity to manage the complex feature interactions and spatial dependencies inherent in transportation networks \cite{nikparvar_machine_2021, tiong_review_2023}. Prior to training, standard scaling normalized the continuous feature space to prevent variables with expansive numerical ranges from disproportionately dominating the linear baseline.

Default hyperparameters were predominantly preserved to prioritize feature evaluation over algorithmic optimization, with minor operational deviations (e.g., max iterations, binary objectives) detailed in Appendix \ref{sec:apx:feature_hyperparams} Table \ref{tab:classifier_hyperparameters}. To show baseline conclusions did not hinge on extensive tuning, a randomized search cross-validation was executed on the SHAP-evaluated model (XGBoost, trajectory-based split). To prevent look-ahead bias, this sanity check utilized a Time Series Split (TSCV) constrained to 20 iterations across 5 folds. The search space was defined as \texttt{subsample}: [0.6, 0.8, 1.0], \texttt{n\_estimators}: [100, 200, 300], \texttt{max\_depth}: [3, 5, 7], \texttt{learning\_rate}: [0.01, 0.05, 0.1], and \texttt{colsample\_bytree}: [0.6, 0.8, 1.0].

\subsubsection{Metrics and Statistical Inference}
\label{subsubsec:evaluation_metrics}
Balanced Accuracy serves as the primary evaluation metric. By computing the unweighted average recall across classes, it prevents artificial performance inflation and directly addresses the inherent stochasticity of railway delays:
$$
    \text{Balanced Accuracy} = \frac{1}{2} \left( \frac{TP}{TP + FN} + \frac{TN}{TN + FP} \right)
$$
In the highly stochastic domain of railway operations, where predictability is inherently constrained by unobserved external variables, even marginal improvements of a few hundredths in balanced accuracy hold significant operational meaning, provided they demonstrate statistical stability over time.
F1 Score and Receiver Operating Characteristic Area Under the Curve (ROC AUC) were calculated as supplementary metrics specifically within the time-based splitting configuration to facilitate comprehensive model benchmarking. Conversely, the trajectory-based evaluation focused primarily on Balanced Accuracy to provide a consistent, unweighted diagnostic signal during the SHAP-based feature stability analysis and target threshold sensitivity tests. Furthermore, to isolate genuine model learning from overfitting, a Null Baseline Model was established by randomly shuffling the target labels prior to training to generate a strict, non-informative performance floor equating to random chance.

Predictive performance was evaluated via Global Aggregation (micro-averaging) to establish holistic benchmarks comparable to prior foundational frameworks \cite{lei_forecasting_2022, kampere_predicting_2025}, and Temporal Aggregation (macro-averaging per month) for statistical testing. 

To ensure valid paired statistical testing between simultaneous and non-simultaneous performance, results required temporal alignment. For each test month $t$, the simultaneous configuration yielded a single score ($S_t$), whereas the non-simultaneous configuration yielded multiple scores from preceding training months. To resolve this dimensionality mismatch, non-simultaneous scores were aggregated via the arithmetic mean ($\bar{N}_t = \frac{1}{n} \sum_{i=1}^{n} N_{t,i}$), establishing the formal paired observation ($S_t$, $\bar{N}_t$) for each test month $t$. Because subsequent Shapiro-Wilk tests yielded inconsistent normality results across various model and feature configurations, the assumption of a universal normal distribution was rejected. Consequently, the non-parametric Wilcoxon signed-rank test ($\alpha = 0.05$) was uniformly adopted \cite{demsar_statistical_2006}, ensuring observed temporal degradation is statistically verifiable. 

Finally, to facilitate temporal analysis, temporal aggregation was utilized to track distribution shifts in both predictive performance and the top three SHAP-identified features. These shifts were quantified using the Interquartile Range (IQR, bounded by the 25th and 75th percentiles). This explicitly isolates the true within-month dispersion of the feature space, while concurrently quantifying the variance in historical predictive stability for each target month across the sequential testing horizons.

While the IQR quantifies within-month dispersion, explicitly evaluating the temporal robustness of the model explanations required further non-parametric analysis. To this end, the Kendall rank correlation coefficient ($\tau$) was employed to quantify the ordinal consistency of SHAP feature importance rankings across sequential temporal transitions (period $t$ to $t+1$). This established whether the fundamental hierarchy of feature prioritization remained stable over time, with statistical significance evaluated against an alpha threshold of $\alpha = 0.05$. 

Furthermore, to isolate the underlying mechanisms driving any observed instability in the SHAP values, a two-sample Kolmogorov-Smirnov (K-S) test was concurrently incorporated into the analytical framework. By applying the K-S test directly to the raw input features across the corresponding temporal periods, the methodology quantifies actual distributional shifts within the environment. This dual-testing strategy systematically differentiates between stochastic algorithmic variance and true concept drift within the underlying railway data structures.

\subsubsection{Temporal Baselines}
To rigorously benchmark the generalization capabilities of the models, two temporal baselines were introduced: a Naive baseline (persistence based on the prior month, $t-1$) and a Seasonal Naive baseline (SNaive, utilizing the identical calendar month from the prior year, $t-12$). Because the railway network topology is dynamic and edges may not operate every month, these baselines encounter a cold-start problem. Consistent with standard practices in temporal network analysis, missing observations were handled via zero imputation, conservatively predicting ``no delay'' for trajectories without historical precedent.

Approximately $3.5 \%$ of Naive and $5.2\%$ of SNaive predictions rely on this default imputation (Appendix \ref{sec:apx:temporal_baselines} Table \ref{tab:temporal_baseline_survivorship}) and Figure \ref{fig:baseline_survivorship}. Because defaulting to zero artificially inflates True Negatives, Balanced Accuracy is specifically utilized to mitigate survivorship bias by equally weighting both classes. To ensure a fair, equal-$N$ comparison ($N=60$), the first 12 months of the dataset were truncated for these temporal baseline evaluations (Appendix \ref{sec:apx:temporal_baselines} Table \ref{tab:temporal_baseline_performance}).

\subsection{Hardware and Software Requirements}
Experiments were executed on the Snellius supercomputing cluster utilizing NVIDIA A100 GPUs (40GB memory) via the University of Amsterdam \cite{surf_snellius_2025}. Data preprocessing utilized the DuckDB Python API \cite{raasveldt_duckdb_2019}. All model dependencies were version-controlled and specified in the environment configuration file.
\section{Results}
\label{sec:results}


This section details the empirical evaluation of the network-wide delay prediction models. The findings sequentially address temporal generalization capabilities across distinct testing paradigms (SRQ1) and evaluate temporal feature stability via SHAP diagnostics (SRQ2).

\paragraph{Sample Size Variations Across Configurations}
The final number of temporally aligned pairs ($N$) varies marginally across experimental configurations due to pipeline constraints during non-simultaneous forecasting aggregation. Baseline evaluations within the simultaneous context retain the full dataset of 72 monthly horizons. For paired simultaneous versus non-simultaneous evaluations, the trajectory-based data split yields 71 aligned pairs, whereas the time-based algorithmic comparisons yield 66 aligned pairs. Because the Wilcoxon signed-rank tests are applied independently within each configuration, these minor variations do not compromise statistical validity.

\subsection{SRQ1: Evaluation of Temporal Generalization}
\label{subsec:results_srq1}
This subsection evaluates the predictive performance of the engineered feature sets across simultaneous and non-simultaneous testing horizons to address Hypotheses 1, 2, and 3.

\subsubsection{Hypothesis 1: Simultaneous Predictive Performance}
\label{subsubsec:results_h1}
Under trajectory-based simultaneous testing, all XGBoost feature configurations demonstrated discriminative predictive capacity significantly exceeding the null baseline (Table \ref{tab:xgb_sim_base} and Appendix \ref{sec:apx:performance_distributions} Figure \ref{fig:xgb_bp_sim}). One-sided Wilcoxon signed-rank tests confirmed the predictive validity across all evaluated feature subsets ($p < .001$, Appendix \ref{sec:apx:statistical_tests} Table \ref{tab:wilcoxon_results_sim_vs_null}), supporting Hypothesis 1. Notably, the isolated topological features generated a significant median accuracy gain of 0.075 over the null expectation ($W = 2599.0, p < .001, r = 0.601$). The fully integrated feature sets (All Features) expanded this median gain to 0.127 ($W = 2595.0, p < .001, r = 0.599$), demonstrating fundamental predictive validity before temporal evaluation.

\subsubsection{Hypothesis 2: Temporal Generalization Degradation}
\label{subsubsec:results_h2}
Transitioning to non-simultaneous generalization testing, predictive performance exhibited a structural degradation pattern across most evaluated feature configurations (Table \ref{tab:xgb_nonsim_base} and Appendix \ref{sec:apx:performance_distributions} Figure \ref{fig:xgb_bp_nonsim}). One-sided paired Wilcoxon signed-rank tests confirmed that these temporal performance decreases were statistically significant across all configurations ($p < .001$, Appendix \ref{sec:apx:statistical_tests} Table \ref{tab:wilcoxon_results_sim_vs_nonsim}). Notably, the fully integrated architecture (All Features) experienced a significant median accuracy loss of ~0.069 when shifted from simultaneous to non-simultaneous evaluation ($W = 2529.0, p < .001, r = 0.602$). 

However, this degradation is not universal. Under strict time-based splits, isolated topological feature sets successfully resisted temporal decay across the tree-based models, including XGBoost ($W=384, p=1.000, r = 0.401$), RandomForest ($W=372.0, p<1.00, r = 0.408$), and LGBM ($W=557.0, p=1.00, r = 0.305$) (Appendix \ref{sec:apx:statistical_tests} Table \ref{tab:wilcoxon_results_all_models_timebased}). All remaining configurations, including integrated feature sets, logistic regression baselines, trajectory-based XGBoost models, hyperparameter-tuned variants, and alternative threshold formulations, consistently exhibited statistically significant degradation ($p < .001$), while several intermediate models (e.g., Topological + Weight of XGBoost and RandomForest) demonstrated no significant change (Appendix \ref{sec:apx:statistical_tests} Tables \ref{tab:wilcoxon_results_all_models_timebased} \ref{tab:wilcoxon_results_absolute} \ref{tab:wilcoxon_results_relative} \ref{tab:wilcoxon_results_tuned}).

Incremental ablation analysis within the trajectory-based XGBoost configuration further clarifies these architectural dynamics (Appendix \ref{sec:apx:statistical_tests} Table \ref{tab:wilcoxon_incremental_features}). The integration of weight-based features into the baseline topological framework yielded a statistically significant performance improvement, generating a median accuracy gain of 0.018 ($W = 368.0, p < .001, r = 0.611$). Conversely, the inclusion of environmental variables (weather features) actively degraded predictive capacity; the enhancement hypothesis was completely rejected ($p = 1.000$) as the median accuracy decreased. In contrast, the addition of operational features consistently enhanced predictive stability when combined with the topological and weight representations ($W = 699.0, p < .001, r = 0.278$). Ultimately, the comparison between the base topological model and the optimal Topological + Weight + Operational architecture confirmed a statistically significant aggregate improvement ($W = 21.0, p < .001, r = 0.604$). These findings partially support Hypothesis 2, indicating that temporal robustness is likely configuration-dependent and compromised by meteorological noise.

\begin{figure}[htbp]
\centering
\caption{Trajectory-Based Split Results: XGBoost Performance Across Testing Paradigms (Global Aggregation)}
\label{fig:xgb_trajectory_combined}

\begin{subfigure}{\columnwidth}
\centering
\resizebox{\columnwidth}{!}{%
\begin{tabular}{lcc}
\toprule
\textbf{Feature Set} & \textbf{Balanced Acc} & \textbf{Balanced Acc Null} \\ 
\midrule
Topological Features & 0.584 & 0.500 \\
Weight Features & 0.606 & 0.495 \\
Weather Features & 0.573 & 0.495 \\
Operational Features & 0.592 & 0.501 \\
Topological + Weight & 0.611 & 0.511 \\
Topological + Weight + Weather & 0.630 & 0.506 \\
Topological + Weight + Operational & 0.624 & 0.494 \\
All Features & \textbf{0.634} & 0.504 \\
\bottomrule
\end{tabular}%
}
\caption{Simultaneous Testing vs. Null Accuracy}
\label{tab:xgb_sim_base}
\end{subfigure}

\vspace{1.5em} 

\begin{subfigure}{\columnwidth}
\centering
\resizebox{\columnwidth}{!}{%
\begin{tabular}{lcc}
\toprule
\textbf{Feature Set} & \textbf{Balanced Acc} & \textbf{Balanced Acc Null} \\ 
\midrule
Topological Features & 0.556 & 0.490 \\
Weight Features & 0.570 & 0.498 \\
Weather Features & 0.501 & 0.500 \\
Operational Features & 0.541 & 0.495 \\
Topological + Weight & 0.574 & 0.492 \\
Topological + Weight + Weather & 0.556 & 0.502 \\
Topological + Weight + Operational & \textbf{0.582} & 0.506 \\
All Features & 0.565 & 0.502 \\
\bottomrule
\end{tabular}%
}
\caption{Non-Simultaneous Testing vs. Null Accuracy}
\label{tab:xgb_nonsim_base}
\end{subfigure}

\vspace{1.5em} 

\begin{subfigure}{\columnwidth}
\centering
\resizebox{\columnwidth}{!}{%
\begin{tabular}{lcc}
\toprule
\textbf{Feature Set} & \textbf{Balanced Acc} & \textbf{Balanced Acc Null} \\ 
\midrule
All Features (Simultaneous) & 0.653 & 0.499 \\
All Features (Non-Simultaneous) & 0.577 & 0.506 \\
\bottomrule
\end{tabular}%
}
\caption{Hyperparameter Tuned (All Features Set Configuration)}
\label{fig:xgb_tuned_all}
\end{subfigure}

\end{figure}

\subsubsection{Methodological Sanity Check: Hyperparameter Optimization}
\label{subsubsec:results_tuning_check}
targeted hyperparameter optimization was applied to the fully integrated feature set. The tuned configuration yielded a comparable significant degradation pattern between simultaneous and non-simultaneous testing paradigms (Figure \ref{fig:xgb_tuned_all} and Appendix \ref{sec:apx:statistical_tests} Table \ref{tab:wilcoxon_results_tuned}), confirming that performance decay is not attributable to model mis-specification.

\subsubsection{Hypothesis 3: Temporal Degradation Across Time-Based Splits}
\label{subsubsec:results_h3}
To isolate temporal drift from sampling effects, a strict time-based split was implemented across multiple models. While global aggregations suggest an overall reduction in predictive performance under non-simultaneous testing (Tables \ref{tab:sim_bal_acc} and \ref{tab:nonsim_bal_acc}), Wilcoxon tests reveal a more nuanced pattern. 

Statistically significant degradation is observed across the majority of feature configurations and algorithms, particularly within highly integrated sets heavily reliant on environmental data (e.g., All Features, $p < .001$). However, isolated topological representations completely resisted this temporal decay across all evaluated tree-based algorithms (LGBM, Random Forest, and XGBoost); non-simultaneous performance marginally exceeded simultaneous baselines, yielding a complete absence of degradation ($p = 1.000$). Furthermore, specific hybrid configurations, such as the Topological + Weight + Operational architecture, demonstrated robust temporal stability within the tree-based frameworks, yielding no statistically significant degradation (e.g., XGBoost $p = .696$, LGBM $p = .600$) (Appendix \ref{sec:apx:statistical_tests} Table \ref{tab:wilcoxon_results_all_models_timebased}).

These findings partially support Hypothesis 3, indicating that temporal generalization decay is not uniform but instead depends critically on feature composition and model class.

\begin{table}[htbp]
\centering
\caption{Time-Based Split Results: Balanced Accuracy across Simultaneous and Non-Simultaneous Testing (Global Aggregation). 
}
\label{fig:combined_bal_acc}

\begin{subtable}{\columnwidth}
\centering
\caption{Simultaneous Testing Results}
\label{tab:sim_bal_acc}
\resizebox{\columnwidth}{!}{%
\begin{tabular}{lcccccccc}
\toprule
\textbf{Classifier} & \textbf{TOP} & \textbf{WGT} & \textbf{WTH} & \textbf{OPS} & \textbf{TW} & \textbf{TWCW} & \textbf{TWCO} & \textbf{ALLF} \\ 
\midrule
LGBMClassifier & 0.639 & 0.673 & 0.637 & 0.644 & 0.683 & 0.697 & 0.693 & 0.705 \\
LogisticRegression & 0.579 & 0.606 & 0.575 & 0.586 & 0.603 & 0.627 & 0.628 & 0.642 \\
RandomForest & \textbf{0.675} & \textbf{0.697} & \textbf{0.665} & \textbf{0.673} & \textbf{0.707} & \textbf{0.716} & \textbf{0.717} & \textbf{0.723} \\
XGBClassifier & 0.646 & 0.676 & 0.638 & 0.654 & 0.680 & 0.695 & 0.692 & 0.702 \\
\bottomrule
\end{tabular}%
}
\end{subtable}

\vspace{1.5em} 

\begin{subtable}{\columnwidth}
\centering
\caption{Non-Simultaneous Testing Results}
\label{tab:nonsim_bal_acc}
\resizebox{\columnwidth}{!}{%
\begin{tabular}{lcccccccc}
\toprule
\textbf{Classifier} & \textbf{TOP} & \textbf{WGT} & \textbf{WTH} & \textbf{OPS} & \textbf{TW} & \textbf{TWCW} & \textbf{TWCO} & \textbf{ALLF} \\ 
\midrule
LGBMClassifier & 0.593 & \textbf{0.583} & 0.495 & 0.558 & \textbf{0.594} & \textbf{0.555} & \textbf{0.600} & 0.563 \\
LogisticRegression & 0.541 & 0.548 & 0.483 & \textbf{0.559} & 0.550 & 0.492 & 0.573 & 0.497 \\
RandomForest & 0.592 & 0.577 & \textbf{0.499} & 0.554 & 0.588 & 0.535 & 0.592 & 0.542 \\
XGBClassifier & \textbf{0.595} & 0.580 & 0.499 & 0.557 & 0.593 & 0.554 & 0.599 & \textbf{0.563} \\
\bottomrule
\end{tabular}%
}
\end{subtable}

\end{table}


\subsubsection{Temporal Baseline Comparisons.} The evaluation of the temporal baselines explicitly highlights the difficulty of non-simultaneous forecasting. Over the $N=60$ evaluated months, the simple Naive persistence model achieved a $Balanced Accuracy = 0.8080$ ($Median = 0.8246$) (Appendix \ref{sec:apx:temporal_baselines} Table \ref{tab:temporal_baseline_performance} and Figures \ref{fig:baseline_acc_calendar}, \ref{fig:baseline_acc_time}). The Seasonal Naive model achieved a mean $Balanced Accuracy = 0.7190$ ($Median=0.7257$). The strength of the immediate $t-1$ Naive baseline suggests that underlying network states are highly auto-correlated on a month-to-month basis. While the tabular XGBoost architectures excel at isolating complex interactions within localized simultaneous snapshots, the raw predictive floor established by simple temporal persistence underscores the likely necessity of anchoring future predictive frameworks in dynamic, season-aware memory rather than static tabular horizons.

\subsection{SRQ2: Model Behavior and Temporal Feature Stability}
\label{subsec:results_srq2}
Addressing SRQ2, SHAP values are utilized strictly as a correlational diagnostic tool to quantify temporal shifts in predictive reliance, rather than to infer physical network causality.

\subsubsection{Static Feature Importance: A Single-Month Snapshot}
\label{subsubsec:results_shap_snapshot}
Single-month analysis (Appendix \ref{sec:apx:shap_analysis} Figure \ref{fig:xgb_shap_allf}, January 2024) reveals diverse feature correlations with predictions. The operational metric \textit{Target Weighted Outbound Degree} emerges as a primary predictor, with higher values associated with increased predicted delay likelihood. Environmental variables, such as \textit{Target Mean Temperature (2m)}, also show strong positive associations within this specific winter snapshot, while topological features exhibit mixed directional effects.

\subsubsection{Feature Volatility and Long-Term Stability}
\label{subsubsec:results_shap_temporal}
Across multiple temporal horizons, feature importance rankings exhibit substantial instability (Appendix \ref{sec:apx:yoy_feature_importance} Figure \ref{fig:xgb_yoy_core_grid}). All isolated feature sets demonstrate fluctuating importance patterns, with environmental features showing the highest volatility and operational features displaying comparatively greater consistency.

Within the fully integrated model (Appendix \ref{sec:apx:yoy_feature_importance} Figure \ref{fig:xgb_yoy_allf}), the feature \textit{Decayed edge weight} exhibits the highest relative stability in SHAP rankings, followed by \textit{Target weighted outbound degree} and \textit{Source mean soil temperature}. Nevertheless, even these top-ranked features display notable temporal variability, indicating the absence of strictly stable predictors.

Temporal analysis of the top three features further highlights this distinction (Appendix \ref{sec:apx:eda_supplementary} Figure \ref{fig:eda_features_ot}). \textit{Decayed edge weight} (1493.23 [1026.00–1999.00]) and \textit{Target weighted outbound degree} (6122.92 [2437.00–8304.00]) maintain relatively stable temporal means while exhibiting substantial within-month dispersion. In contrast, \textit{Source mean soil temperature} (11.60 [6.48–16.85]) demonstrates pronounced temporal shifts in its mean with minimal within-month variance.

These findings indicate that structural and operational features provide comparatively more stable predictive signals, whereas environmental features are disproportionately associated with temporal drift, aligning with the observed degradation patterns in SRQ1.

\subsubsection{Quantitative Assessment of Explanation Stability and Concept Drift}
\label{subsubsec:results_quantitative_drift}

To formalize the descriptive observations of feature volatility, Kendall rank correlation coefficients ($\tau$) were calculated across the 71 sequential temporal transitions for each feature set utilizing the trajectory based XGBoost base model configuration. This analysis reveals substantial variance in explanation stability across the evaluated feature sets, confirming the visual diagnostics.

As detailed in Appendix \ref{sec:apx:kendall} Table \ref{tab:kendall_shap_stability}, the \texttt{OPS} configuration demonstrated the highest temporal stability (mean $\tau = 0.831$), with all sequential transitions exhibiting statistical significance ($p < 0.05$). This suggests a consistent prioritization of the operational feature hierarchy over the observation period. Conversely, configurations such as \texttt{WGT} and \texttt{WTH} exhibited severe rank instability. These sets were characterized by lower mean correlations and isolated periods of negative ordinal association, indicating complete inversions of feature importance. Furthermore, the \texttt{WGT} configuration achieved statistical significance in only 16.9\% (12 of 71) of transitions, suggesting that the observed ranking variations within this subset are largely attributable to algorithmic noise rather than systematic shifting. Configurations encompassing larger feature spaces, specifically \texttt{TWCO} and \texttt{ALLF}, maintained moderate average stability while demonstrating nearly universal statistical significance across transitions. 

To contextualize this explanatory instability, the two-sample Kolmogorov-Smirnov (K-S) test results were evaluated to quantify actual distributional shifts within the raw feature space (Appendix \ref{sec:apx:k-s} Tables \ref{tab:ks_macro_summary}, \ref{tab:ks_diagnostic_part1} and \ref{tab:ks_diagnostic_part2}). This macroscopic analysis revealed a stark categorical divide in environmental stability. Weather features (WTH) exhibited extreme volatility, with statistically significant distributional drift occurring in 84.8\% of all sequential temporal transitions. In contrast, Operational (OPS), Topological (TOP), and Weight (WGT) features demonstrated high environmental stability, exhibiting significant drift in only 3.29\%, 6.94\%, and 1.41\% of transitions, respectively. By pairing these explanatory metrics with the distributional analysis, a distinct pattern emerges: periods of significant SHAP rank inversion frequently align with statistically significant distributional shifts in the underlying feature space. Therefore, the explanatory instability recorded in specific configurations, particularly those reliant on environmental variables, does not inherently indicate algorithmic fragility. Instead, these findings show temporal changes in feature distributions and in the model's predictive reliance on those features. Together, the distributional tests and SHAP analyses characterize temporal instability, but do not by themselves establish a change in the conditional relationship between predictors and the target.

\subsection{Robustness Check: Alternative Target Thresholds}
\label{subsec:robustness_check}
Across all threshold definitions, models retained strong simultaneous discriminative performance (e.g., 0.643 to 0.646 for the fully integrated configuration), confirming that predictive signal extraction remains valid within localized temporal snapshots.

However, non-simultaneous performance patterns remained highly dependent on the chosen threshold definition. Under the \textit{Month Relative Threshold}, which preserves class balance within each calendar month, the fully integrated model continued to exhibit substantial and statistically significant degradation (0.646 to 0.556, $\Delta = -0.090$, Appendix \ref{sec:apx:statistical_tests} Table \ref{tab:wilcoxon_results_relative}), indicating that this threshold formulation did not eliminate the observed performance degradation.

Under the \textit{Absolute Threshold} ($>20\%$ delayed rides), the Topological configuration increased from 0.570 to 0.580 in aggregate balanced accuracy ($\Delta = +0.010$), while the Weight configuration decreased slightly from 0.624 to 0.618 ($\Delta = -0.006$). However, the paired monthly Wilcoxon tests detected no statistically significant degradation for Topological features ($p = 0.647$), whereas Weight features exhibited statistically significant degradation ($p < 0.001$) (Appendix \ref{sec:apx:statistical_tests}, Table \ref{tab:wilcoxon_results_absolute}). These results indicate that temporal sensitivity under the absolute threshold differs between structural feature sets.

Nevertheless, this effect did not generalize across all feature categories. Isolated Weather features continued to exhibit significant degradation (0.575 to 0.501, $\Delta = -0.074$), and the fully integrated model also demonstrated a statistically significant performance decline under the absolute definition (0.643 to 0.604, $\Delta = -0.039$). 

Overall, these findings indicate that while alternative target definitions can reduce apparent temporal sensitivity for specific structural feature sets, they do not eliminate generalization decay in more complex or environmentally enriched configurations.

\subsection{Robustness Check: Regression Formulations}
To verify that the predictive validity of the engineered features is not merely an artifact of binary thresholding, model performance was evaluated within a continuous regression space targeting Schedule Adherence (0-minute delay) and Regulatory Punctuality (3-minute delay).

Under simultaneous testing, the models successfully extracted continuous predictive signals, consistently outperforming the null baseline (Appendix \ref{sec:apx:regression} Tables \ref{tab:reg_regulatory} and \ref{tab:reg_schedule}). For Regulatory Punctuality, the Topological + Weight + Operational configuration achieved an $MAE = 0.0532$ and an $RMSE = 0.1073$, notably improving upon the Null baseline ($MAE = 0.0939$, RMSE $RMSE = 0.1532$). However, mirroring the classification results, predictive decay occurred under non-simultaneous horizons. For Regulatory Punctuality, the same configuration degraded to an $MAE= 0.0944$ and an $RMSE=0.1571$, closing the gap with the Null baseline ($MAE = 0.1062$, $RMSE = 0.1681$). A similar trend was observed under the stricter Schedule Adherence threshold. Ultimately, this indicates that while the feature sets possess continuous predictive capacity within-month, they remain structurally vulnerable to temporal drift across future horizons.
\section{Discussion}
\label{sec:discussion}


This section contextualizes the empirical findings within the current state of railway delay prediction. To clarify predictive and interpretability trade-offs, the methodology is positioned against foundational literature. Subsequently, a structural diagnosis of temporal degradation is provided, empirically demonstrating that while the models learn meaningful signals within-month, this capacity systematically degrades across future months. Crucially, this degradation is diagnosed not as a parametric tuning artifact, but as the consequence of label shift, covariate shift, and concept drift. Finally, methodological limitations regarding validity and generalizability are assessed to guide future research trajectories.

\subsection{Positioning within the State-of-the-Art}
\label{subsec:disc_sota}

This research implemented theoretical recommendations to evaluate whether granular feature engineering could mitigate the temporal generalization challenges inherent in railway networks. Specifically, prior work suggested that integrating external operational and environmental variables within a stop-to-stop topology would yield a more robust predictive model \cite{kampere_predicting_2025, brakenhoff_dynamic_2026}. By adopting this multi-layered framework and expanding the target definition to encompass service cancellations, the proposed methodology successfully elevated simultaneous (within-month) predictive capabilities relative to previous baselines, establishing an average balanced accuracy of 0.650 \cite{kampere_predicting_2025}.

However, non-simultaneous generalization remains inferior to contemporary deep learning approaches. While the tabular XGBoost models achieved modest non-simultaneous balanced accuracies, they remain inferior to the robust thresholds ($\approx 0.75$) demonstrated by hybrid Graph Attention Networks (GAT-LSTM/GRU) \cite{van_der_maas_predicting_2025}. Because these hybrid architectures natively model sequential spatiotemporal flows, the performance gap suggests that flattening dynamic railway operations into static monthly snapshots omits the critical temporal sequencing required for high-fidelity prediction.

\paragraph{The Diagnostic Value of Interpretability}
While contemporary hybrid deep learning architectures exhibit robust absolute predictive accuracy thresholds for spatiotemporal forecasting, they frequently obscure the underlying mechanics of temporal degradation. Consequently, the primary academic contribution of this manuscript is not the maximization of raw predictive density, but the provision of an explainable temporal robustness analysis of network-wide railway delay prediction. As the Dutch railway network functions as critical public infrastructure, automated decision-making systems within this domain are subject to imminent regulatory scrutiny, necessitating high diagnostic transparency.

By prioritizing interpretable tree-based ensembles and SHAP analysis over opaque hybrid architectures, this methodology establishes a rigorous diagnostic lens for characterizing model degradation. Rather than identifying a singular algorithmic limitation, this transparent framework provides empirical evidence that systemic temporal drift is a multifaceted challenge. Ultimately, this approach demonstrates that feature-rich tree-based models improve within-month prediction, but their performance degrades across future horizons. Crucially, this framework explicitly links this degradation to environmental feature volatility and instability in the target definition.

\subsection{Diagnosing Temporal Degradation}
\label{subsec:disc_reflection}
The empirical findings suggest that the temporal generalization failure observed during non-simultaneous testing is not an artifact of hyperparameter tuning limitation, but rather a dual-faceted data vulnerability. Specifically, the predictive degradation appears to be systematically linked to environmental feature volatility and instability in the target definition.

Notably, time-based evaluations revealed that isolated topological configurations completely resisted temporal decay across the primary tree-based ensemble models. Non-simultaneous evaluations marginally outperformed simultaneous baselines, confirming a complete absence of degradation for XGBoost ($W=384.0, p=1.000, r=0.401$), Random Forest ($W=372.0, p=1.000, r=0.408$), and LightGBM ($W=557.0, p=1.000, r=0.305$) (Appendix \ref{sec:apx:statistical_tests} Table \ref{tab:wilcoxon_results_all_models_timebased}). This indicates that the foundational spatial network structure possesses inherent temporal robustness. However, highly integrated feature sets and alternative threshold configurations overwhelmingly exhibited statistically significant temporal degradation ($p < .001$) or yielded mixed robustness (Appendix \ref{sec:apx:statistical_tests} Tables \ref{tab:wilcoxon_results_sim_vs_nonsim} \ref{tab:wilcoxon_results_all_models_timebased} \ref{tab:wilcoxon_results_absolute} \ref{tab:wilcoxon_results_relative} \ref{tab:wilcoxon_results_tuned}). 
Incremental ablation testing explicitly isolates the source of this volatility. Supplementing the topological baseline with edge weights and operational metrics does not merely resist degradation; it yields statistically significant performance improvements ($p < .001$). In contrast, the integration of environmental (weather) features actively compromises the architecture. The addition of meteorological data did not enhance predictive capacity ($p = 1.000$), instead precipitating a measurable drop in median accuracy, which is associated with a statistically significant reduction in non-simultaneous balanced accuracy observed in the fully integrated models (Appendix \ref{sec:apx:statistical_tests} Table \ref{tab:wilcoxon_incremental_features}).

This performance decrease during non-simultaneous testing is strongly associated with temporal label shift. Prior studies utilized a static, global median threshold across multi-year datasets to classify significantly delayed trajectories \cite{kampere_predicting_2025}. However, exploratory analysis reveals that actual class prevalence fluctuates between 10\% and 90\% monthly (Appendix \ref{sec:apx:eda_supplementary} Figure \ref{fig:eda_class_balance_OT}). Applying a static percentile threshold to this volatile environment induces severe label shift, as the operational meaning of the positive class changes over time.

This target instability is explicitly supported by the threshold sensitivity analysis (Appendix \ref{sec:apx:statistical_tests} Tables \ref{tab:wilcoxon_results_sim_vs_null}, \ref{tab:wilcoxon_results_absolute}, and \ref{tab:wilcoxon_results_relative}). While utilizing the absolute threshold (a static 0.20 delay ratio) did not entirely eliminate temporal decay across the broader feature sets, it uniquely stabilized the core network representations, resulting in no statistically significant predictive change. In contrast, configurations reliant on volatile environmental variables continued to suffer statistically significant collapse ($p < 0.001$). This divergence suggests that statistical target definitions artificially compound predictive decay by disproportionately penalizing otherwise stable structural topologies.

\subsubsection{Covariate Shift and Concept Drift}
Beyond target formulation, the observed predictive variance is further explained by the interplay between covariate shift and concept drift within the external variables. Temporal analysis revealed that the predictive reliance on all features is highly unstable (Appendix \ref{sec:apx:yoy_feature_importance} Figure \ref{fig:xgb_shap_allf} and Appendix \ref{sec:apx:eda_supplementary} Figure \ref{fig:eda_features_ot}).

Conversely, the primary environmental feature, \textit{Source mean soil temperature}, exhibited notable covariate shift; while the mean fluctuated seasonally, the within-month variance remained entirely collapsed. This spatial uniformity implies that while the input distribution changes over time, the localized variance required for distinct trajectory prediction is absent, causing the models to overfit to the specific monthly snapshot.

Temporal variation in SHAP importance rankings across consecutive horizons (Section \ref{subsec:results_srq2}) indicates changes in the model's reliance on individual features. Environmental features show varying predictive importance across months, suggesting that their contribution is sensitive to the temporal context.

To contextualize this realignment, the evaluation of explanation stability via the Kendall rank correlation ($\tau$) revealed clear statistical divergence across model configurations. While configurations exhibiting high temporal stability, notably \texttt{OPS}, indicate robust predictive mechanisms, the rank instability observed in configurations heavily reliant on external variables (such as \texttt{WGT}, \texttt{WTH}, and \texttt{TW}) required further diagnosis.

The application of the two-sample K-S test to the raw input features clarifies the source of this instability. As detailed in Appendix \ref{sec:apx:k-s}, meteorological features underwent statistically significant distributional shifts in nearly 85\% of evaluated temporal transitions, whereas core operational and structural network features remained statistically stable in over 93\% of transitions. By pairing the explanatory metrics with this distributional analysis, a distinct pattern emerges. Periods of significant SHAP rank inversion frequently align with these measured distributional shifts in the underlying feature space. Together, the distributional tests and SHAP analyses show that environmental feature distributions and their predictive importance vary over time. These findings are consistent with the poorer cross-month performance observed in weather-enriched configurations, although they do not establish the cause of that degradation.

Overall, environmental features exhibit greater temporal distributional variation than structural and operational features. Combined with the feature-set comparisons, these findings suggest that adding weather information does not consistently improve cross-month generalization.

\subsection{ Limitations and Future Work}
\label{subsec:disc_limitations}

The findings concern monthly predictions for domestic NS passenger services. The target combines delays and cancellations, as the available records do not distinguish planned from unexpected cancellations. Weather inputs are historical observations, so prospective use requires evaluation with information available at prediction time. Future work could examine separate disruption targets and lagged delay features, with comparisons against persistence baselines under matched evaluation conditions.

\section{Conclusion}
\label{sec:conclusion}


This section synthesizes the empirical findings to establish the primary contribution of this research: an explainable temporal robustness analysis of network-wide railway delay prediction. While contemporary state-of-the-art architectures often obscure temporal degradation mechanics within opaque deep learning structures, this study presents a transparent diagnostic evaluation of predictive instability. The overarching finding establishes that feature-rich tree-based models improve within-month prediction, but their performance experiences a statistically significant reduction when evaluated on future months. Crucially, this degradation is explicitly linked to environmental feature volatility and instability in the statistical target definition.

By utilizing a comprehensive tabular dataset of domestic passenger trajectories, this diagnostic evaluation establishes four supporting operational realities. First, integrating granular external features successfully captures a meaningful predictive signal within localized monthly snapshots, achieving robust baseline balanced accuracies. Second, paired Wilcoxon statistical tests reveal a divided generalization capacity; isolated topological features demonstrate temporal resilience in tree-based models, whereas the broader integration of external features systematically degrades out-of-sample predictive performance. Third, temporal performance degradation persisted in the evaluated hyperparameter-tuned configuration, indicating that the tuning procedure used here did not eliminate it. Finally, the threshold sensitivity analyses show that measured temporal performance depends on the target definition, while the SHAP and distributional analyses reveal temporal variation in environmental features and their predictive importance.

Ultimately, this thesis demonstrates that richer feature sets alone are insufficient to resolve the constraints of long-term network-wide railway delay prediction. The primary barrier to predictive robustness is the highly unstable behavior of both the statistical target definition and the external environment across time. Therefore, achieving robust temporal generalization necessitates transitioning beyond the mere accumulation of tabular variables toward dynamic, season-aware modeling architectures anchored by absolute, operationally grounded target designs.

\bibliographystyle{ACM-Reference-Format}
\bibliography{bibliographies/references}

\newpage
\onecolumn

\begin{appendices}

\begin{table}[htbp]
\centering
\caption{Summary of experimental configurations detailing split type, imbalance handling, and purpose.}
\label{tab:experimental_baselines}
\resizebox{\textwidth}{!}{
\begin{tabular}{llp{10cm}}
\toprule
\textbf{Split Type} & \textbf{Imbalance Handling} & \textbf{Purpose} \\
\midrule
Time-Based (All Models) & Time-Based Split & Primary temporal generalizability evaluation. Ensures 70\%/30\% train data split volume for each month.\\
Trajectory-Based (XGBoost) & Random Under-Sampling & Secondary baseline for strict literature comparability and standardized SHAP evaluation. Ensures 70\%/30\% train data split of trajectories across the whole dataset \\
Trajectory-Based Null Model & Random Under-Sampling & Establishes a theoretical performance floor (shuffled labels). If the Null Model performs similar to non-shuffled data, then the non-shuffled data is likely a result of overfitting.\\
\bottomrule
\end{tabular}
}
\end{table}

\section{Supplementary Model Performance Metrics (Global Aggregation) 
Feature sets abbreviations are TOP = topology, WGT = Weight, WTH = Weather, OPS = Operational, TW = TOP + WGT, TWCW = TOP + WGT + WTH, TWCO = TOP + WGT + OPS, ALLF = TOP + WGT + WTH + OPS}
\label{sec:apx:supplementary_performance}

\begin{table}[htbp]
\centering
\caption{Time-Based Split Results: F1 Score across Simultaneous and Non-Simultaneous Testing}
\label{tab:combined_f1}

\begin{subtable}{\textwidth}
\centering
\caption{Simultaneous Testing Results}
\label{tab:sim_f1}
\begin{tabular}{lcccccccc}
\toprule
\textbf{Classifier} & \textbf{TOP} & \textbf{WGT} & \textbf{WTH} & \textbf{OPS} & \textbf{TW} & \textbf{TWCW} & \textbf{TWCO} & \textbf{ALLF} \\ 
\midrule
LGBMClassifier & 0.640 & 0.680 & 0.637 & 0.641 & 0.687 & 0.699 & 0.694 & 0.707 \\
LogisticRegression & 0.548 & 0.580 & 0.577 & 0.537 & 0.585 & 0.622 & 0.608 & 0.634 \\
RandomForest & \textbf{0.675} & \textbf{0.696} & \textbf{0.664} & \textbf{0.675} & \textbf{0.708} & \textbf{0.714} & \textbf{0.717} & \textbf{0.722} \\
XGBClassifier & 0.645 & 0.673 & 0.637 & 0.649 & 0.682 & 0.695 & 0.693 & 0.702 \\
\bottomrule
\end{tabular}
\end{subtable}

\vspace{1.5em} 

\begin{subtable}{\textwidth}
\centering
\caption{Non-Simultaneous Testing Results}
\label{tab:nonsim_f1}
\begin{tabular}{lcccccccc}
\toprule
\textbf{Classifier} & \textbf{TOP} & \textbf{WGT} & \textbf{WTH} & \textbf{OPS} & \textbf{TW} & \textbf{TWCW} & \textbf{TWCO} & \textbf{ALLF} \\ 
\midrule
LGBMClassifier & \textbf{0.596} & \textbf{0.579} & 0.452 & 0.550 & \textbf{0.592} & \textbf{0.570} & \textbf{0.596} & \textbf{0.565} \\
LogisticRegression & 0.504 & 0.517 & \textbf{0.499} & 0.507 & 0.525 & 0.518 & 0.549 & 0.515 \\
RandomForest & 0.593 & 0.573 & 0.484 & \textbf{0.553} & 0.586 & 0.545 & 0.590 & 0.542 \\
XGBClassifier & 0.595 & 0.573 & 0.443 & 0.544 & 0.588 & 0.563 & 0.595 & 0.561 \\
\bottomrule
\end{tabular}
\end{subtable}

\end{table}

\begin{table}[htbp]
\centering
\caption{Time-Based Split Results: ROC AUC across Simultaneous and Non-Simultaneous Testing}
\label{tab:combined_roc_auc}

\begin{subtable}{\textwidth}
\centering
\caption{Simultaneous Testing Results}
\label{tab:sim_roc_auc}
\begin{tabular}{lcccccccc}
\toprule
\textbf{Classifier} & \textbf{TOP} & \textbf{WGT} & \textbf{WTH} & \textbf{OPS} & \textbf{TW} & \textbf{TWCW} & \textbf{TWCO} & \textbf{ALLF} \\ 
\midrule
LGBMClassifier & 0.700 & 0.742 & 0.697 & 0.702 & 0.750 & 0.767 & 0.762 & 0.777 \\
LogisticRegression & 0.608 & 0.653 & 0.599 & 0.638 & 0.647 & 0.673 & 0.679 & 0.695 \\
RandomForest & \textbf{0.730} & \textbf{0.765} & \textbf{0.726} & \textbf{0.738} & \textbf{0.770} & \textbf{0.781} & \textbf{0.783} & \textbf{0.793} \\
XGBClassifier & 0.700 & 0.741 & 0.695 & 0.710 & 0.742 & 0.759 & 0.755 & 0.770 \\
\bottomrule
\end{tabular}
\end{subtable}

\vspace{1.5em} 

\begin{subtable}{\textwidth}
\centering
\caption{Non-Simultaneous Testing Results}
\label{tab:nonsim_roc_auc}
\begin{tabular}{lcccccccc}
\toprule
\textbf{Classifier} & \textbf{TOP} & \textbf{WGT} & \textbf{WTH} & \textbf{OPS} & \textbf{TW} & \textbf{TWCW} & \textbf{TWCO} & \textbf{ALLF} \\ 
\midrule
LGBMClassifier & 0.619 & \textbf{0.608} & 0.494 & \textbf{0.576} & \textbf{0.620} & \textbf{0.575} & \textbf{0.627} & \textbf{0.585} \\
LogisticRegression & 0.561 & 0.579 & 0.474 & 0.565 & 0.577 & 0.483 & 0.601 & 0.489 \\
RandomForest & 0.613 & 0.603 & \textbf{0.500} & 0.573 & 0.615 & 0.551 & 0.620 & 0.561 \\
XGBClassifier & \textbf{0.620} & 0.607 & 0.498 & 0.573 & 0.620 & 0.573 & \textbf{0.627} & 0.585 \\
\bottomrule
\end{tabular}
\end{subtable}

\end{table}

\begin{table}[htbp]
    \centering
    \caption{Sensitivity Analysis: XGBoost Balanced Accuracy across Alternative Target Thresholds. \textit{Note: The average null baseline across all tested configurations remained mathematically stable at approximately 0.500 and is omitted for clarity.}}
    \label{tab:appendix_sensitivity_results}
    \renewcommand{\arraystretch}{1.2}
    \resizebox{\textwidth}{!}{%
    \begin{tabular}{l c c c c}
        \toprule
        & \multicolumn{2}{c}{\textbf{Absolute Threshold (0.20)}} & \multicolumn{2}{c}{\textbf{Historical Seasonal Threshold}} \\
        \cmidrule(lr){2-3} \cmidrule(lr){4-5}
        \textbf{Feature Set} & \textbf{Simultaneous} & \textbf{Non-Simultaneous} & \textbf{Simultaneous} & \textbf{Non-Simultaneous} \\
        \midrule
        Topological Features                 & 0.570 & 0.580 & 0.577 & 0.546 \\
        Weight Features                      & 0.624 & 0.618 & 0.613 & 0.562 \\
        Weather Features                     & 0.575 & 0.501 & 0.577 & 0.500 \\
        Operational Features                 & 0.581 & 0.538 & 0.592 & 0.533 \\
        Topological + Weight                 & 0.612 & 0.608 & 0.621 & 0.564 \\
        Topological + Weight + Weather       & 0.626 & 0.596 & 0.632 & 0.559 \\
        Topological + Weight + Operational   & 0.625 & 0.621 & 0.629 & 0.569 \\
        All Features                         & 0.643 & 0.604 & 0.646 & 0.556 \\
        \bottomrule
    \end{tabular}%
    }
\end{table}

\begin{figure}[htbp]
    \centering
    \caption{Sensitivity Analysis Boxplots: Distribution of XGBoost Balanced Accuracy across alternative target thresholds and temporal configurations. These distributions correspond to the mean values reported in Table~\ref{tab:appendix_sensitivity_results}.}
    \label{fig:appendix_sensitivity_boxplots}
    \begin{subfigure}[b]{0.48\textwidth}
        \centering
        \includegraphics[width=\textwidth]{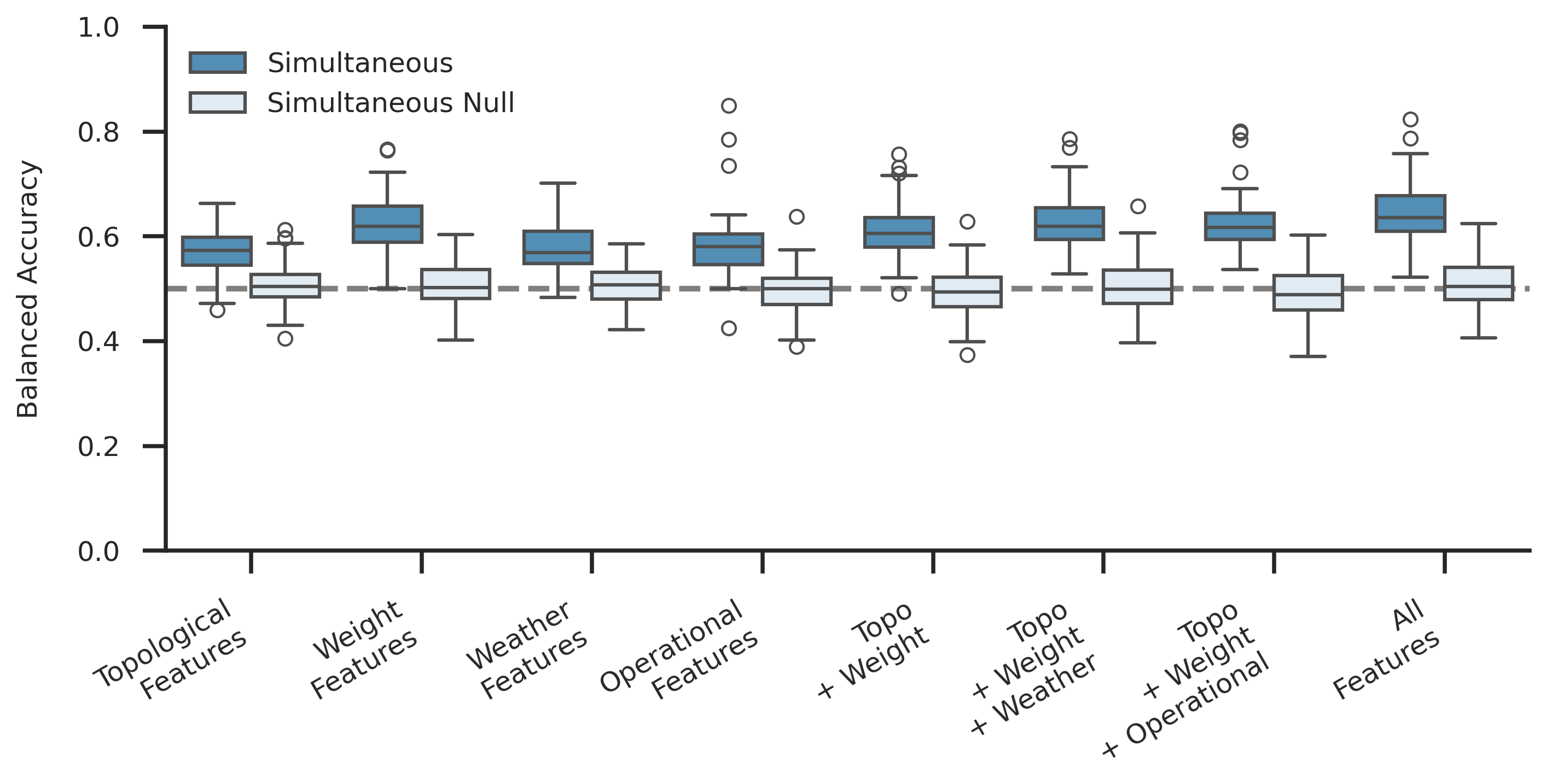}
        \caption{Absolute Threshold: Simultaneous}
        \label{fig:boxplot_sim_abs}
    \end{subfigure}
    \hfill
    \begin{subfigure}[b]{0.48\textwidth}
        \centering
        \includegraphics[width=\textwidth]{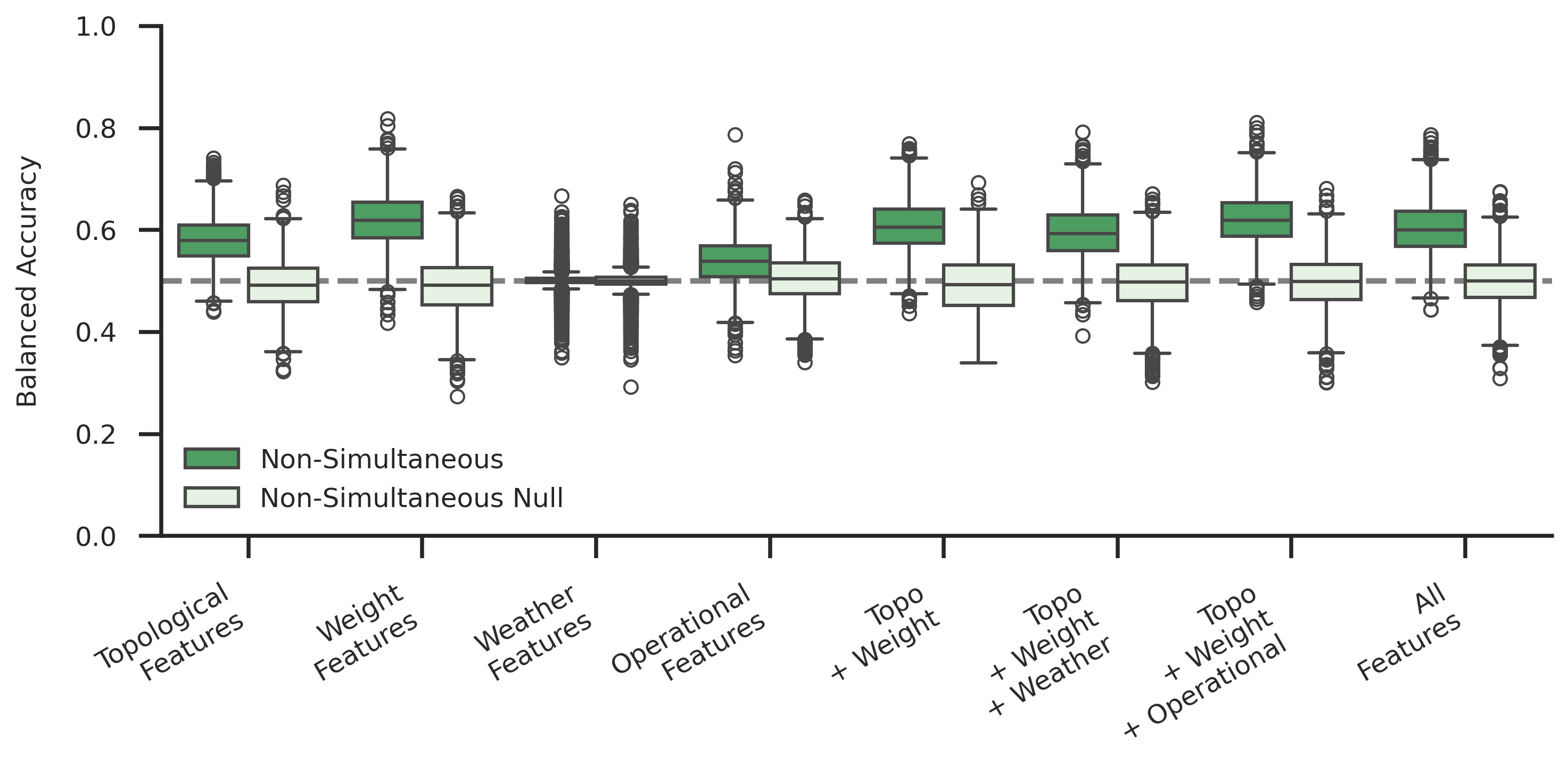}
        \caption{Absolute Threshold: Non-Simultaneous}
        \label{fig:boxplot_nonsim_abs}
    \end{subfigure}
    
    \vspace{1em} 
    
    \begin{subfigure}[b]{0.48\textwidth}
        \centering
        \includegraphics[width=\textwidth]{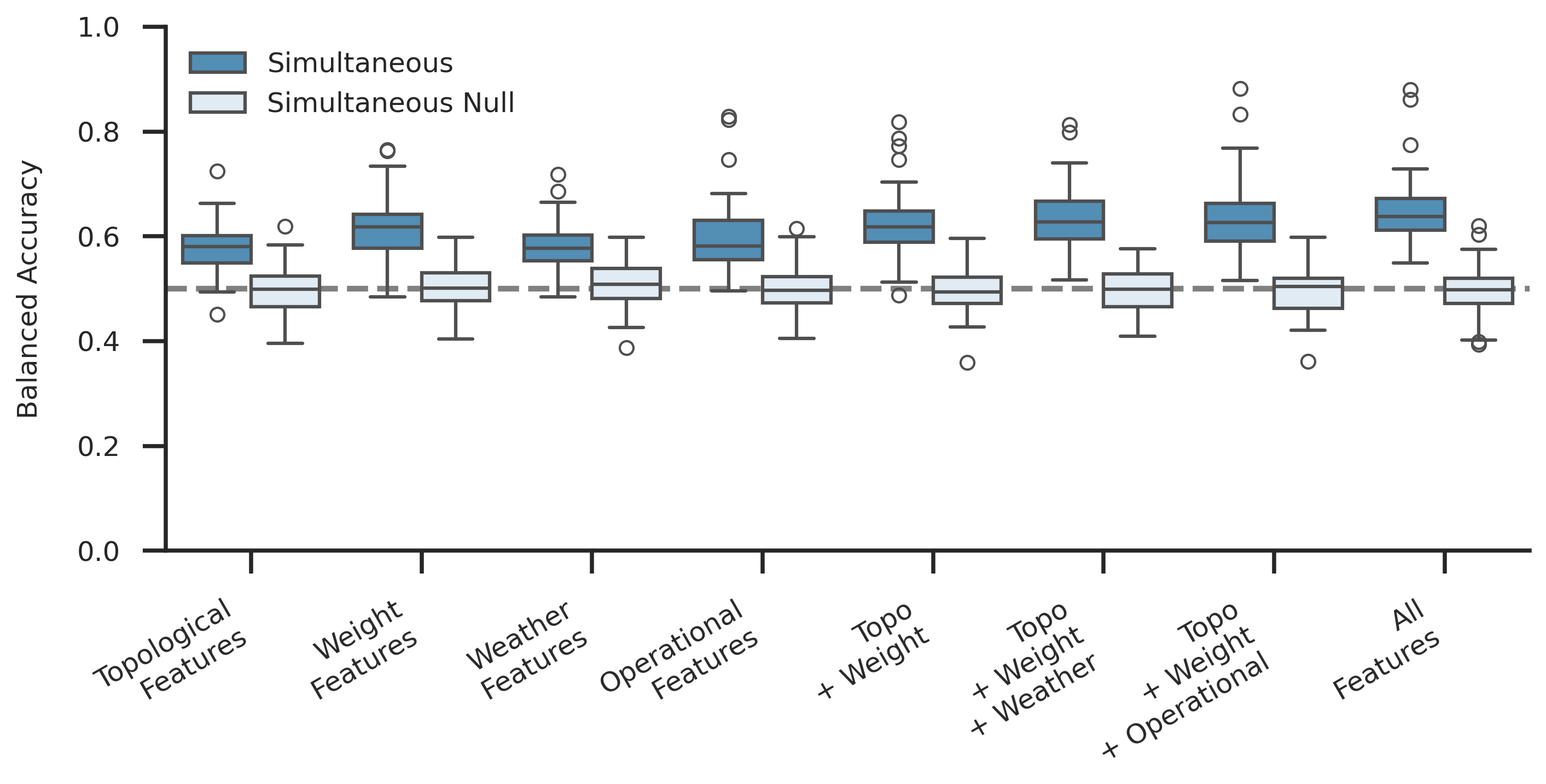}
        \caption{Historical Seasonal: Simultaneous}
        \label{fig:boxplot_sim_rel}
    \end{subfigure}
    \hfill
    \begin{subfigure}[b]{0.48\textwidth}
        \centering
        \includegraphics[width=\textwidth]{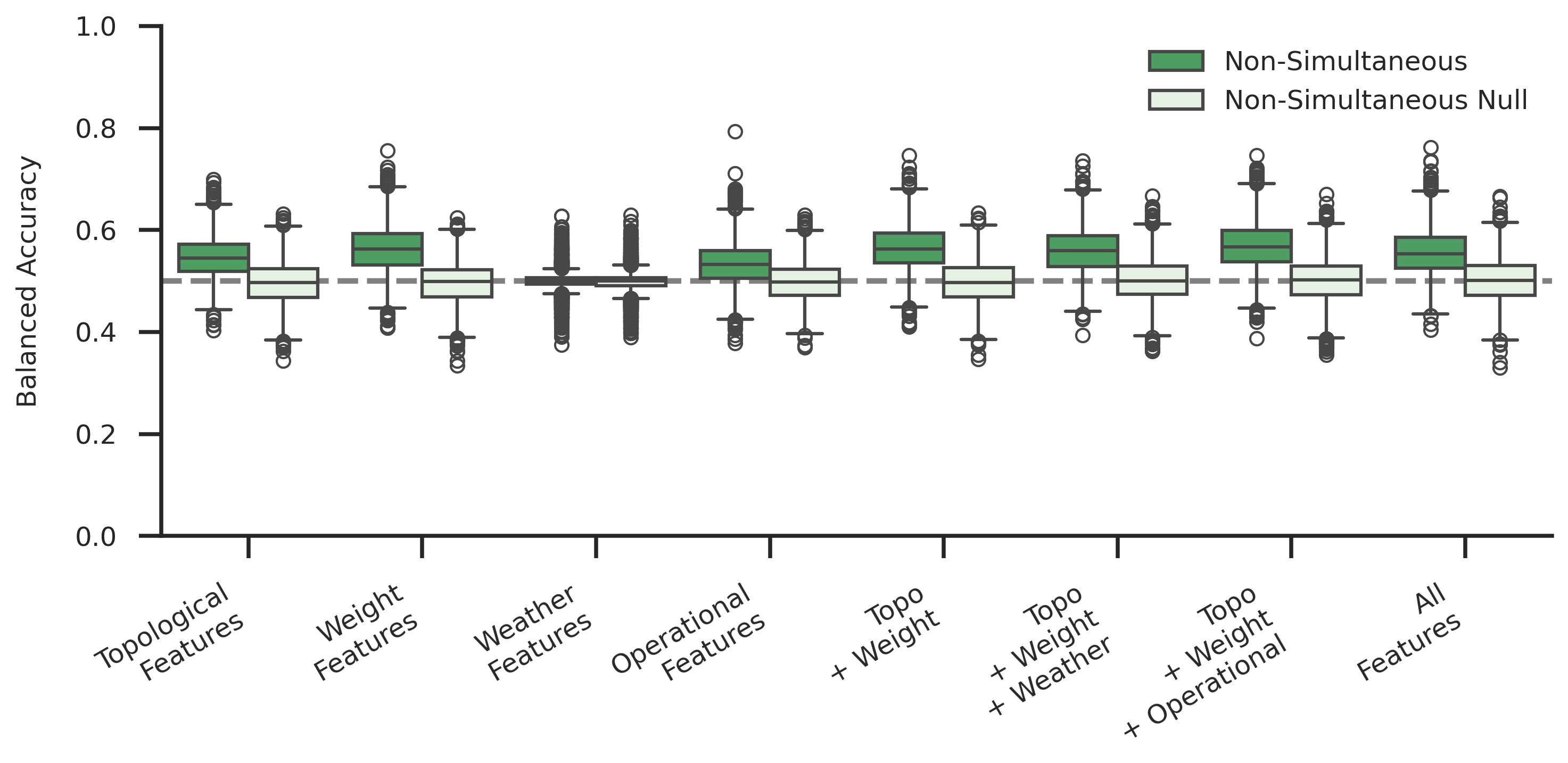}
        \caption{Historical Seasonal: Non-Simultaneous}
        \label{fig:boxplot_nonsim_rel}
    \end{subfigure}

\end{figure}

\clearpage
\section{Statistical Significance Tests (Temporal Aggregation)}
\label{sec:apx:statistical_tests}


\begin{table}[htbp]
    \centering
    \caption{Wilcoxon Signed-Rank Test results comparing simultaneous XGBoost performance against the null baseline (Trajectory-based split, $\alpha=0.05$, $N=72$ per feature set).}
    \label{tab:wilcoxon_results_sim_vs_null}
    \begin{tabular}{l c c c c c c c}
        \toprule
        \textbf{Feature Set} & \textbf{Direction} & \textbf{Mdn (Sim)} & \textbf{Mdn (Null)} & \textbf{Mdn Diff} & \textbf{W} & \textbf{p-value} & \textbf{Effect ($r$)} \\
        \midrule
        Topological Features & Sim $>$ Null & 0.582 & 0.501 & +0.075 & 2599.0 & $< 0.001$ & 0.601 \\
        Weight Features & Sim $>$ Null & 0.601 & 0.496 & +0.107 & 2538.0 & $< 0.001$ & 0.602 \\
        Weather Features & Sim $>$ Null & 0.571 & 0.490 & +0.082 & 2582.0 & $< 0.001$ & 0.593 \\
        Operational Features & Sim $>$ Null & 0.584 & 0.496 & +0.090 & 2520.0 & $< 0.001$ & 0.564 \\
        Topological + Weight & Sim $>$ Null & 0.605 & 0.513 & +0.104 & 2593.5 & $< 0.001$ & 0.598 \\
        Topological + Weight + Weather & Sim $>$ Null & 0.630 & 0.512 & +0.106 & 2627.0 & $< 0.001$ & 0.614 \\
        Topological + Weight + Operational & Sim $>$ Null & 0.629 & 0.499 & +0.132 & 2619.0 & $< 0.001$ & 0.610 \\
        All Features & Sim $>$ Null & 0.629 & 0.503 & +0.127 & 2595.0 & $< 0.001$ & 0.599 \\
        \bottomrule
    \end{tabular}
\end{table}


\begin{table}[htbp]
    \centering
    \caption{Wilcoxon Signed-Rank Test results comparing simultaneous and non-simultaneous XGBoost performance (Trajectory-based split, $\alpha=0.05$, $N=71$ per feature set).}
    \label{tab:wilcoxon_results_sim_vs_nonsim}
    \begin{tabular}{l c c c c c c c}
        \toprule
        \textbf{Model} & \textbf{Direction} & \textbf{Mdn (Sim)} & \textbf{Mdn (Non-Sim)} & \textbf{Mdn Diff} & \textbf{W} & \textbf{p-value} & \textbf{Effect ($r$)} \\
        \midrule
        Topological Features & Sim $>$ Non-Sim & 0.583 & 0.556 & +0.033 & 2197.0 & $< 0.001$ & 0.442 \\
        Weight Features & Sim $>$ Non-Sim & 0.601 & 0.569 & +0.041 & 2305.0 & $< 0.001$ & 0.494 \\
        Weather Features & Sim $>$ Non-Sim & 0.571 & 0.500 & +0.069 & 2556.0 & $< 0.001$ & 0.615 \\
        Operational Features & Sim $>$ Non-Sim & 0.584 & 0.540 & +0.045 & 2394.0 & $< 0.001$ & 0.537 \\
        Topological + Weight & Sim $>$ Non-Sim & 0.606 & 0.571 & +0.037 & 2308.0 & $< 0.001$ & 0.495 \\
        Topological + Weight + Weather & Sim $>$ Non-Sim & 0.630 & 0.553 & +0.074 & 2539.0 & $< 0.001$ & 0.606 \\
        Topological + Weight + Operational & Sim $>$ Non-Sim & 0.630 & 0.578 & +0.043 & 2409.0 & $< 0.001$ & 0.544 \\
        All Features & Sim $>$ Non-Sim & 0.629 & 0.561 & +0.069 & 2529.0 & $< 0.001$ & 0.602 \\
        \bottomrule
    \end{tabular}
\end{table}

\begin{table}[htbp]
    \centering
    \caption{Wilcoxon Signed-Rank Test results comparing simultaneous and non-simultaneous performance across all models (Time-based split, $\alpha=0.05$, N=66 per feature set.}
    \label{tab:wilcoxon_results_all_models_timebased}
    \resizebox{\textwidth}{!}{%
    \begin{tabular}{l l c c c c c c c}
        \toprule
        \textbf{Classifier} & \textbf{Feature Set} & \textbf{Direction} & \textbf{Mdn (Sim)} & \textbf{Mdn (Non-Sim)} & \textbf{Mdn Diff} & \textbf{W} & \textbf{p-value} & \textbf{Effect ($r$)} \\
        \midrule
        
        \textbf{LGBM} 
        & Topological Features & Sim $<$ Non-Sim & 0.576 & 0.609 & -0.029 & 557.0 & $1.000$ & 0.305 \\
        & Weight Features & Sim $>$ Non-Sim & 0.619 & 0.601 & +0.019 & 1363.0 & $0.050$ & 0.143 \\
        & Weather Features & Sim $>$ Non-Sim & 0.584 & 0.500 & +0.079 & 2179.0 & $< 0.001$ & 0.597 \\
        & Operational Features & Sim $>$ Non-Sim & 0.592 & 0.574 & +0.014 & 1435.0 & $0.018$ & 0.183 \\
        & Topological + Weight & Sim $>$ Non-Sim & 0.616 & 0.616 & +0.000 & 1058.0 & $0.619$ & 0.026 \\
        & Top + Weight + Weather & Sim $>$ Non-Sim & 0.632 & 0.568 & +0.062 & 1998.0 & $< 0.001$ & 0.496 \\
        & Top + Weight + Operational & Sim $>$ Non-Sim & 0.627 & 0.623 & +0.008 & 1066.0 & $0.600$ & 0.022 \\
        & All Features & Sim $>$ Non-Sim & 0.642 & 0.579 & +0.062 & 1891.0 & $< 0.001$ & 0.437 \\
        \midrule
        
        \textbf{Logistic Regression} 
        & Topological Features & Sim $>$ Non-Sim & 0.551 & 0.544 & +0.004 & 1311.0 & $0.060$ & 0.136 \\
        & Weight Features & Sim $>$ Non-Sim & 0.582 & 0.558 & +0.025 & 1919.0 & $< 0.001$ & 0.452 \\
        & Weather Features & Sim $>$ Non-Sim & 0.560 & 0.500 & +0.059 & 2175.0 & $< 0.001$ & 0.595 \\
        & Operational Features & Sim $>$ Non-Sim & 0.588 & 0.574 & +0.015 & 1738.0 & $< 0.001$ & 0.352 \\
        & Topological + Weight & Sim $>$ Non-Sim & 0.577 & 0.559 & +0.019 & 1690.0 & $< 0.001$ & 0.325 \\
        & Top + Weight + Weather & Sim $>$ Non-Sim & 0.611 & 0.505 & +0.106 & 2207.0 & $< 0.001$ & 0.612 \\
        & Top + Weight + Operational & Sim $>$ Non-Sim & 0.612 & 0.587 & +0.017 & 1754.0 & $< 0.001$ & 0.361 \\
        & All Features & Sim $>$ Non-Sim & 0.618 & 0.511 & +0.108 & 2211.0 & $< 0.001$ & 0.615 \\
        \midrule
        
        \textbf{Random Forest} 
        & Topological Features & Sim $<$ Non-Sim & 0.575 & 0.605 & -0.033 & 372.0 & $1.000$ & 0.408 \\
        & Weight Features & Sim $>$ Non-Sim & 0.598 & 0.595 & +0.005 & 1040.0 & $0.662$ & 0.036 \\
        & Weather Features & Sim $>$ Non-Sim & 0.567 & 0.500 & +0.065 & 2193.0 & $< 0.001$ & 0.605 \\
        & Operational Features & Sim $<$ Non-Sim & 0.566 & 0.572 & -0.000 & 1080.0 & $0.565$ & 0.014 \\
        & Topological + Weight & Sim $<$ Non-Sim & 0.598 & 0.611 & -0.011 & 804.0 & $0.973$ & 0.168 \\
        & Top + Weight + Weather & Sim $>$ Non-Sim & 0.619 & 0.540 & +0.078 & 2091.0 & $< 0.001$ & 0.548 \\
        & Top + Weight + Operational & Sim $>$ Non-Sim & 0.613 & 0.613 & +0.001 & 977.0 & $0.794$ & 0.071 \\
        & All Features & Sim $>$ Non-Sim & 0.623 & 0.548 & +0.075 & 1982.0 & $< 0.001$ & 0.487 \\
        \midrule
        
        \textbf{XGBoost} 
        & Topological Features & Sim $<$ Non-Sim & 0.580 & 0.611 & -0.033 & 384.0 & $1.000$ & 0.401 \\
        & Weight Features & Sim $>$ Non-Sim & 0.616 & 0.596 & +0.030 & 1286.0 & $0.124$ & 0.100 \\
        & Weather Features & Sim $>$ Non-Sim & 0.571 & 0.499 & +0.075 & 2192.0 & $< 0.001$ & 0.604 \\
        & Operational Features & Sim $>$ Non-Sim & 0.587 & 0.570 & +0.003 & 1340.0 & $0.067$ & 0.130 \\
        & Topological + Weight & Sim $<$ Non-Sim & 0.605 & 0.611 & -0.002 & 942.0 & $0.852$ & 0.091 \\
        & Top + Weight + Weather & Sim $>$ Non-Sim & 0.628 & 0.567 & +0.064 & 2028.0 & $< 0.001$ & 0.513 \\
        & Top + Weight + Operational & Sim $>$ Non-Sim & 0.622 & 0.620 & +0.004 & 1025.0 & $0.696$ & 0.045 \\
        & All Features & Sim $>$ Non-Sim & 0.634 & 0.580 & +0.057 & 1832.0 & $< 0.001$ & 0.404 \\
        
        \bottomrule
    \end{tabular}%
    }
\end{table}

\begin{table}[htbp]
    \centering
    \caption{Wilcoxon Signed-Rank Test results comparing simultaneous and non-simultaneous XGBoost performance under the Absolute Threshold (Trajectory-based split, $\alpha=0.05$, $N=71$ per feature set).}
    \label{tab:wilcoxon_results_absolute}
    \resizebox{\textwidth}{!}{%
    \begin{tabular}{l c c c c c c c}
        \toprule
        \textbf{Model} & \textbf{Direction} & \textbf{Mdn (Sim)} & \textbf{Mdn (Non-Sim)} & \textbf{Mdn Diff} & \textbf{W} & \textbf{p-value} & \textbf{Effect ($r$)} \\
        \midrule
        Topological Features & Sim $<$ Non-Sim & 0.574 & 0.570 & -0.000 & 1212.0 & $0.647$ & 0.032 \\
        Weight Features & Sim $>$ Non-Sim & 0.620 & 0.607 & +0.024 & 1876.0 & $< 0.001$ & 0.288 \\
        Weather Features & Sim $>$ Non-Sim & 0.569 & 0.502 & +0.068 & 2538.0 & $< 0.001$ & 0.606 \\
        Operational Features & Sim $>$ Non-Sim & 0.581 & 0.543 & +0.039 & 2318.0 & $< 0.001$ & 0.500 \\
        Topological + Weight & Sim $>$ Non-Sim & 0.606 & 0.595 & +0.010 & 1695.0 & $0.008$ & 0.201 \\
        Topological + Weight + Weather & Sim $>$ Non-Sim & 0.619 & 0.588 & +0.032 & 2314.0 & $< 0.001$ & 0.498 \\
        Topological + Weight + Operational & Sim $>$ Non-Sim & 0.618 & 0.614 & +0.007 & 1614.0 & $0.027$ & 0.162 \\
        All Features & Sim $>$ Non-Sim & 0.636 & 0.595 & +0.049 & 2408.0 & $< 0.001$ & 0.543 \\
        \bottomrule
    \end{tabular}%
    }
\end{table}

\begin{table}[htbp]
    \centering
    \caption{Wilcoxon Signed-Rank Test results comparing simultaneous and non-simultaneous XGBoost performance under the Month-Relative Threshold (Trajectory-based split, $\alpha=0.05$, $N=71$ per feature set).}
    \label{tab:wilcoxon_results_relative}
    \resizebox{\textwidth}{!}{%
    \begin{tabular}{l c c c c c c c}
        \toprule
        \textbf{Model} & \textbf{Direction} & \textbf{Mdn (Sim)} & \textbf{Mdn (Non-Sim)} & \textbf{Mdn Diff} & \textbf{W} & \textbf{p-value} & \textbf{Effect ($r$)} \\
        \midrule
        Topological Features & Sim $>$ Non-Sim & 0.580 & 0.547 & +0.034 & 2178.0 & $< 0.001$ & 0.433 \\
        Weight Features & Sim $>$ Non-Sim & 0.618 & 0.562 & +0.053 & 2400.0 & $< 0.001$ & 0.539 \\
        Weather Features & Sim $>$ Non-Sim & 0.580 & 0.501 & +0.080 & 2495.0 & $< 0.001$ & 0.585 \\
        Operational Features & Sim $>$ Non-Sim & 0.582 & 0.532 & +0.056 & 2505.0 & $< 0.001$ & 0.590 \\
        Topological + Weight & Sim $>$ Non-Sim & 0.619 & 0.565 & +0.055 & 2406.0 & $< 0.001$ & 0.542 \\
        Topological + Weight + Weather & Sim $>$ Non-Sim & 0.628 & 0.556 & +0.075 & 2489.0 & $< 0.001$ & 0.582 \\
        Topological + Weight + Operational & Sim $>$ Non-Sim & 0.629 & 0.565 & +0.058 & 2454.0 & $< 0.001$ & 0.565 \\
        All Features & Sim $>$ Non-Sim & 0.639 & 0.553 & +0.090 & 2555.0 & $< 0.001$ & 0.614 \\
        \bottomrule
    \end{tabular}%
    }
\end{table}

\begin{table}[htbp]
    \centering
    \caption{Wilcoxon Signed-Rank Test results evaluating incremental feature engineering configurations (XGBoost Trajectory based split, Non-simultaneous testing, $\alpha=0.05$).}
    \label{tab:wilcoxon_incremental_features}
    \resizebox{\textwidth}{!}{%
\begin{tabular}{l l l c c c c c c c}
        \toprule
        \textbf{Step} & \textbf{Base Model} & \textbf{Enhanced Model} & \textbf{Direction} & \textbf{Mdn (Base)} & \textbf{Mdn (Enh)} & \textbf{Mdn Diff} & \textbf{W} & \textbf{p-value} & \textbf{Effect ($r$)} \\
        \midrule
        Adding Weights & Topological Features & Topological + Weight & Enh $>$ Base & 0.547 & 0.565 & +0.018 & 368.0 & $< 0.001$ & 0.611 \\
        Adding Weather & Topological + Weight & Top + Weight + Weather & Enh $<$ Base & 0.565 & 0.556 & -0.006 & 1931.0 & $1.000$ & 0.314 \\
        Adding Operational & Topological + Weight & Top + Weight + Op & Enh $>$ Base & 0.565 & 0.565 & +0.003 & 699.0 & $< 0.001$ & 0.278 \\
        Adding Op. to Weather & Top + Weight + Weather & All Features & Enh $<$ Base & 0.556 & 0.553 & -0.006 & 1831.0 & $0.999$ & 0.266 \\
        Adding Wth. to Op. & Top + Weight + Op & All Features & Enh $<$ Base & 0.565 & 0.553 & -0.014 & 2528.0 & $1.000$ & 0.601 \\
        Base vs. Best & Topological Features & Top + Weight + Op & Enh $>$ Base & 0.547 & 0.565 & +0.023 & 21.0 & $< 0.001$ & 0.604 \\
        \bottomrule
    \end{tabular}%
    }
\end{table}

\begin{table}[htbp]
    \centering
    \caption{Wilcoxon Signed-Rank Test results comparing simultaneous and non-simultaneous tuned XGBoost performance (Trajectory-based split, $\alpha=0.05$, $N=71$).}
    \label{tab:wilcoxon_results_tuned}
    \resizebox{\textwidth}{!}{%
    \begin{tabular}{l c c c c c c c}
        \toprule
        \textbf{Model} & \textbf{Direction} & \textbf{Mdn (Sim)} & \textbf{Mdn (Non-Sim)} & \textbf{Mdn Diff} & \textbf{W} & \textbf{p-value} & \textbf{Effect ($r$)} \\
        \midrule
        All Features & Sim $>$ Non-Sim & 0.650 & 0.572 & +0.071 & 2555.0 & $< 0.001$ & 0.614 \\
        \bottomrule
    \end{tabular}%
    }
\end{table}


\clearpage
\section{Regression Metrics (Regulatory Punctuality \& Schedule Adherence)}
\label{sec:apx:regression}

\begin{table}[htbp]
\centering
\caption{Regression Performance: Regulatory Punctuality Threshold (3-minute delay margin). Performance metrics are evaluated via Mean Absolute Error (MAE) and Root Mean Squared Error (RMSE) across both testing paradigms.}
\label{tab:reg_regulatory}
\begin{tabular}{lcccc}
\toprule
\textbf{Feature Set} & \textbf{MAE} & \textbf{RMSE} & \textbf{MAE (Null)} & \textbf{RMSE (Null)} \\
\midrule
\multicolumn{5}{c}{\textit{(a) Simultaneous Testing}} \\
\midrule
Topological Features                 & 0.0758 & 0.1326 & 0.0961 & 0.1574 \\
Weight Features                      & 0.0600 & 0.1196 & 0.0942 & 0.1542 \\
Weather Features                     & 0.0771 & 0.1428 & 0.0978 & 0.1622 \\
Operational Features                 & 0.0553 & 0.1145 & 0.0936 & 0.1548 \\
Topological + Weight                 & 0.0585 & 0.1141 & 0.0934 & 0.1529 \\
Topological + Weight + Weather       & 0.0542 & 0.1090 & 0.0935 & 0.1550 \\
Topological + Weight + Operational   & 0.0532 & 0.1073 & 0.0939 & 0.1532 \\
All Features                         & 0.0498 & 0.1031 & 0.0930 & 0.1548 \\
\midrule
\multicolumn{5}{c}{\textit{(b) Non-Simultaneous Testing}} \\
\midrule
Topological Features                 & 0.1016 & 0.1655 & 0.1093 & 0.1721 \\
Weight Features                      & 0.0970 & 0.1602 & 0.1077 & 0.1706 \\
Weather Features                     & 0.1179 & 0.1739 & 0.1139 & 0.1670 \\
Operational Features                 & 0.0982 & 0.1633 & 0.1081 & 0.1707 \\
Topological + Weight                 & 0.0961 & 0.1584 & 0.1080 & 0.1703 \\
Topological + Weight + Weather       & 0.1032 & 0.1621 & 0.1143 & 0.1716 \\
Topological + Weight + Operational   & 0.0944 & 0.1571 & 0.1062 & 0.1681 \\
All Features                         & 0.0974 & 0.1582 & 0.1097 & 0.1677 \\
\bottomrule
\end{tabular}
\end{table}

\begin{table}[htbp]
\centering
\caption{Regression Performance: Schedule Adherence Threshold (0-minute delay margin). Performance metrics are evaluated via Mean Absolute Error (MAE) and Root Mean Squared Error (RMSE) across both testing paradigms.}
\label{tab:reg_schedule}
\begin{tabular}{lcccc}
\toprule
\textbf{Feature Set} & \textbf{MAE} & \textbf{RMSE} & \textbf{MAE (Null)} & \textbf{RMSE (Null)} \\
\midrule
\multicolumn{5}{c}{\textit{(a) Simultaneous Testing}} \\
\midrule
Topological Features                 & 0.1366 & 0.1864 & 0.1645 & 0.2200 \\
Weight Features                      & 0.1228 & 0.1758 & 0.1617 & 0.2169 \\
Weather Features                     & 0.1486 & 0.2028 & 0.1673 & 0.2241 \\
Operational Features                 & 0.1290 & 0.1827 & 0.1588 & 0.2123 \\
Topological + Weight                 & 0.1209 & 0.1721 & 0.1620 & 0.2185 \\
Topological + Weight + Weather       & 0.1196 & 0.1711 & 0.1622 & 0.2182 \\
Topological + Weight + Operational   & 0.1191 & 0.1701 & 0.1618 & 0.2169 \\
All Features                         & 0.1172 & 0.1677 & 0.1620 & 0.2173 \\
\midrule
\multicolumn{5}{c}{\textit{(b) Non-Simultaneous Testing}} \\
\midrule
Topological Features                 & 0.1683 & 0.2243 & 0.1800 & 0.2366 \\
Weight Features                      & 0.1631 & 0.2226 & 0.1790 & 0.2353 \\
Weather Features                     & 0.1831 & 0.2376 & 0.1798 & 0.2311 \\
Operational Features                 & 0.1723 & 0.2297 & 0.1785 & 0.2350 \\
Topological + Weight                 & 0.1639 & 0.2222 & 0.1795 & 0.2361 \\
Topological + Weight + Weather       & 0.1687 & 0.2243 & 0.1794 & 0.2337 \\
Topological + Weight + Operational   & 0.1631 & 0.2216 & 0.1785 & 0.2347 \\
All Features                         & 0.1658 & 0.2225 & 0.1804 & 0.2347 \\
\bottomrule
\end{tabular}
\end{table}


\clearpage
\section{Temporal Baselines (Performance \& Surviorship Bias)}
\label{sec:apx:temporal_baselines}

\begin{table}[htbp]
\centering
\caption{Temporal Baselines: Predictive Performance Summary. Evaluated over $N=60$ truncated months to ensure an equal comparison window between Naive ($t-1$) and Seasonal Naive ($t-12$) models.}
\label{tab:temporal_baseline_performance}
\begin{tabular}{llccc}
\toprule
\textbf{Baseline Model} & \textbf{Statistic} & \textbf{Balanced Accuracy} & \textbf{F1 Score} & \textbf{ROC AUC} \\
\midrule
Naive ($t-1$)
 & Mean   & 0.8080 & 0.7798 & 0.8080 \\
 & Median & 0.8246 & 0.8147 & 0.8246 \\
 & Q1     & 0.7807 & 0.7172 & 0.7807 \\
 & Q3     & 0.8459 & 0.8703 & 0.8459 \\
\midrule
Seasonal Naive ($t-12$)
 & Mean   & 0.7190 & 0.6531 & 0.7190 \\
 & Median & 0.7257 & 0.6612 & 0.7257 \\
 & Q1     & 0.6739 & 0.5352 & 0.6739 \\
 & Q3     & 0.7778 & 0.7868 & 0.7778 \\
\bottomrule
\end{tabular}
\end{table}

\begin{table}[htbp]
\centering
\caption{Temporal Baselines: Survivorship-Bias and Imputation Summary ($N=60$). This table details the percentage of predictions relying on zero imputation (Default Rate) due to missing historical trajectories, alongside the resulting Delay Rate Gap and Non-Defaulted Accuracy.}
\label{tab:temporal_baseline_survivorship}
\resizebox{\textwidth}{!}{%
\begin{tabular}{llccccc}
\toprule
\textbf{Baseline Model} & \textbf{Metric} & \textbf{Default Rate} & \textbf{Default Delay Rate} & \textbf{Non-Def. Delay Rate} & \textbf{Delay Rate Gap} & \textbf{Non-Def. Bal. Acc.} \\
\midrule
Naive          & Mean   & 0.0356 & 0.6983 & 0.4953 & 0.2030 & 0.8274 \\
               & Median & 0.0329 & 0.7208 & 0.5355 & 0.2155 & 0.8453 \\
\midrule
Seasonal Naive & Mean   & 0.0525 & 0.6215 & 0.4962 & 0.1254 & 0.7377 \\
               & Median & 0.0466 & 0.6250 & 0.5396 & 0.1760 & 0.7396 \\
\bottomrule
\end{tabular}%
}
\end{table}

\begin{figure}[htbp]
    \centering
    \includegraphics[width=\textwidth]{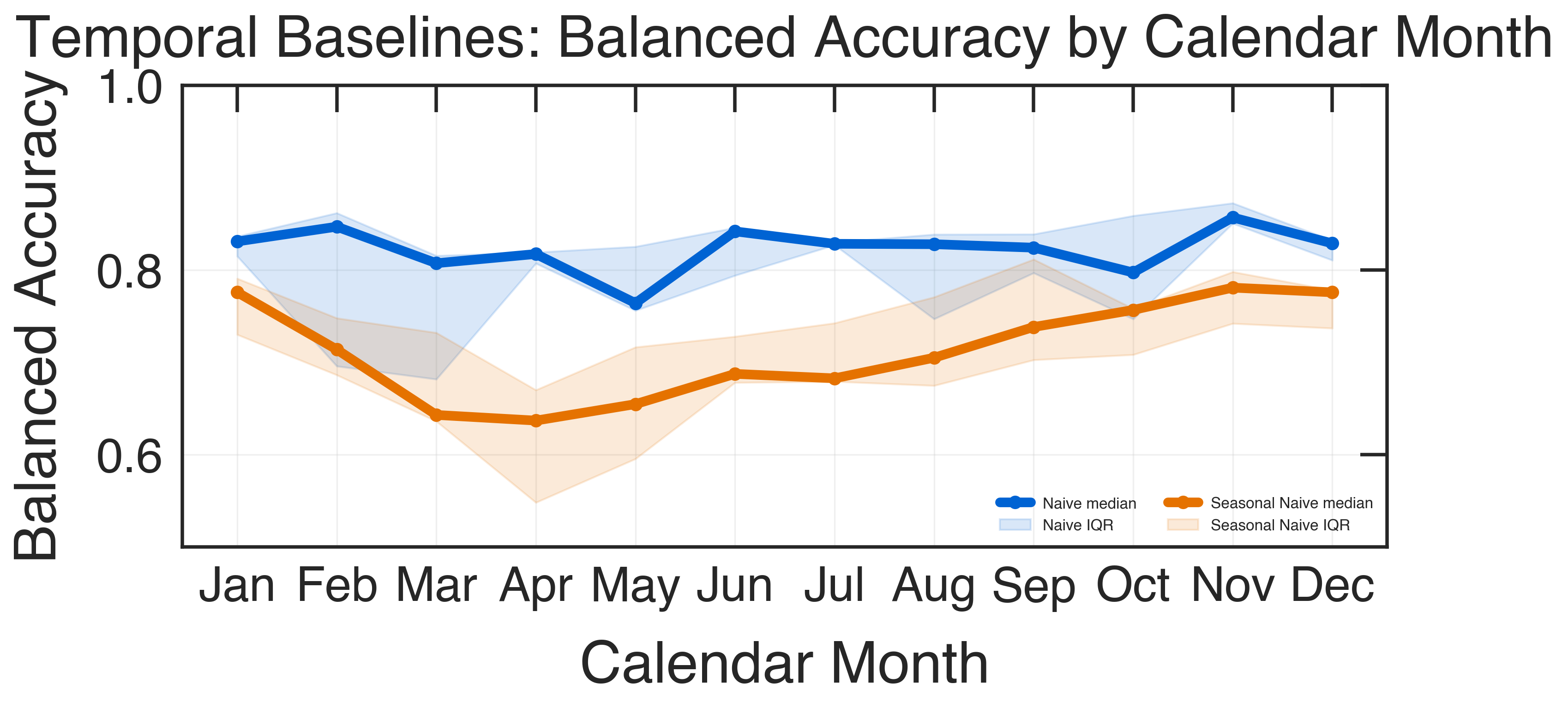}
    \caption{Temporal Baselines: Balanced Accuracy by Calendar Month. Illustrates the aggregated seasonal variance and interquartile range for the Naive and Seasonal Naive baselines.}
    \label{fig:baseline_acc_calendar}
\end{figure}

\begin{figure}[htbp]
    \centering
    \includegraphics[width=\textwidth]{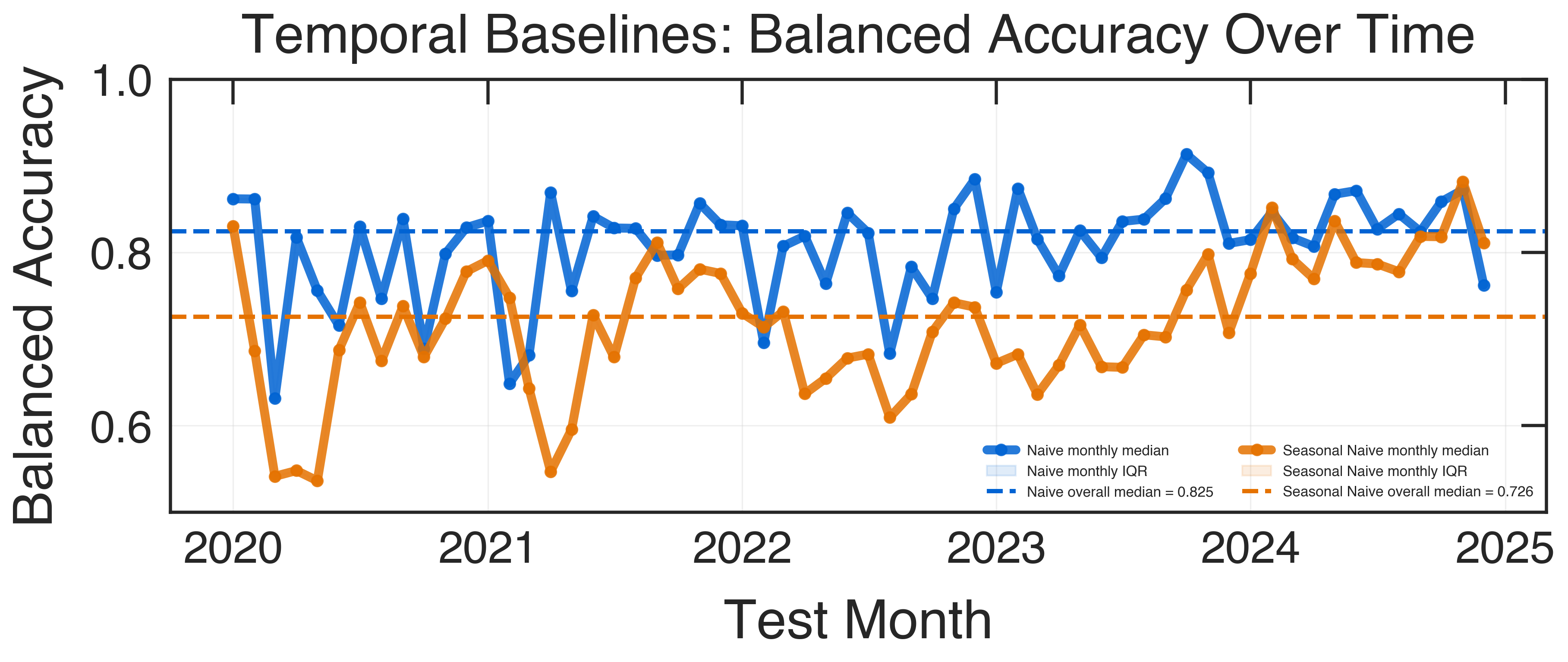}
    \caption{Temporal Baselines: Balanced Accuracy Over Time. Displays the chronological predictive stability of the persistence baselines over the 60-month evaluation period.}
    \label{fig:baseline_acc_time}
\end{figure}

\begin{figure}[htbp]
    \centering
    \includegraphics[width=\textwidth]{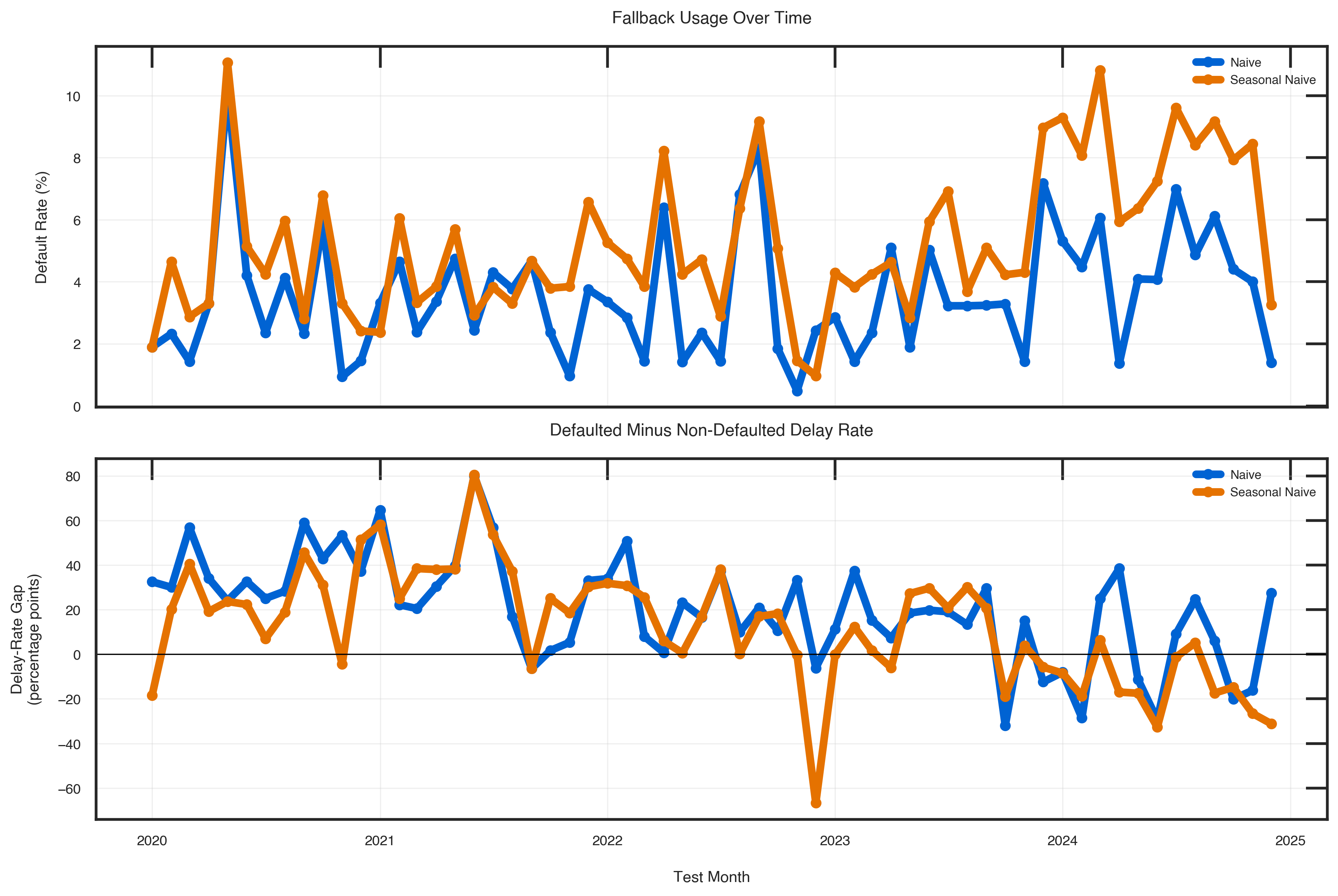}
    \caption{Survivorship-Bias Assessment. The top panel shows the Fallback Usage (default rate) over time due to missing historical trajectories. The bottom panel displays the Delay-Rate Gap between defaulted and non-defaulted predictions.}
    \label{fig:baseline_survivorship}
\end{figure}


\clearpage
\section{Kendall Rank Correlation}
\label{sec:apx:kendall}

\begin{table}[htbp]
    \centering
    \caption{Kendall rank correlation ($\tau$) results evaluating sequential SHAP feature importance stability for the XGBoost base model (Trajectory-based split, $\alpha=0.05$, $T=71$ sequential transitions).}
    \label{tab:kendall_shap_stability}
    \begin{tabular}{l c c c c c}
        \toprule
        \textbf{Feature Set} & \textbf{$N$ (Features)} & \textbf{Mean $\tau$} & \textbf{Min $\tau$} & \textbf{Max $\tau$} & \textbf{Significant Transitions ($p < 0.05$)} \\ 
        \midrule
        Topological Features & 13 & 0.540 & 0.128 & 0.769 & 59 / 71 \\
        Weight Features & 7 & 0.421 & -0.048 & 0.905 & 12 / 71 \\
        Weather Features & 14 & 0.449 & -0.209 & 0.846 & 48 / 71 \\
        Operational Features & 9 & 0.831 & 0.556 & 1.000 & 71 / 71 \\
        Topological + Weight & 20 & 0.479 & -0.158 & 0.705 & 63 / 71 \\
        Topological + Weight + Weather & 34 & 0.367 & -0.005 & 0.647 & 58 / 71 \\
        Topological + Weight + Operational & 29 & 0.594 & 0.281 & 0.749 & 71 / 71 \\
        All Features & 43 & 0.481 & 0.207 & 0.681 & 70 / 71 \\
        \bottomrule
    \end{tabular}
\end{table}

\begin{figure}[htbp]
    \centering
    
    \begin{subfigure}[b]{0.45\textwidth}
        \centering
        \includegraphics[width=\textwidth]{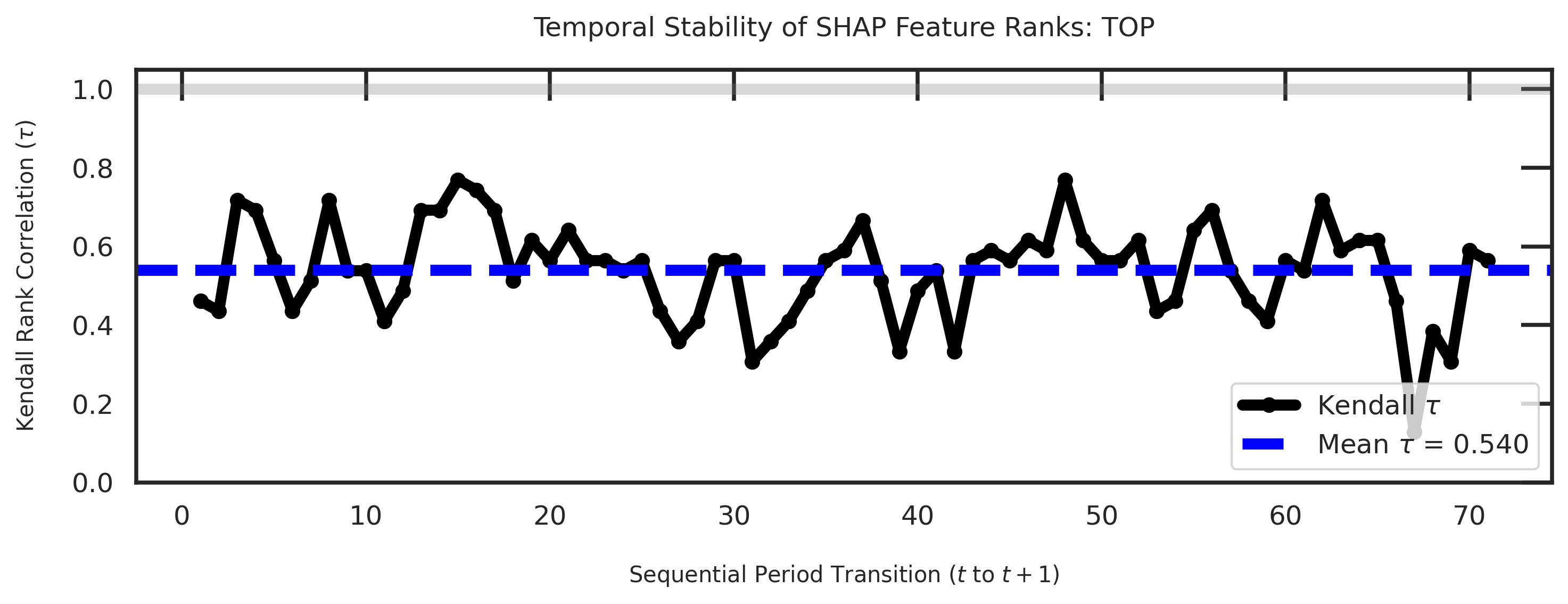}
        \caption{Topological Features}
        \label{fig:kendall_top}
    \end{subfigure}
    \hfill
    \begin{subfigure}[b]{0.45\textwidth}
        \centering
        \includegraphics[width=\textwidth]{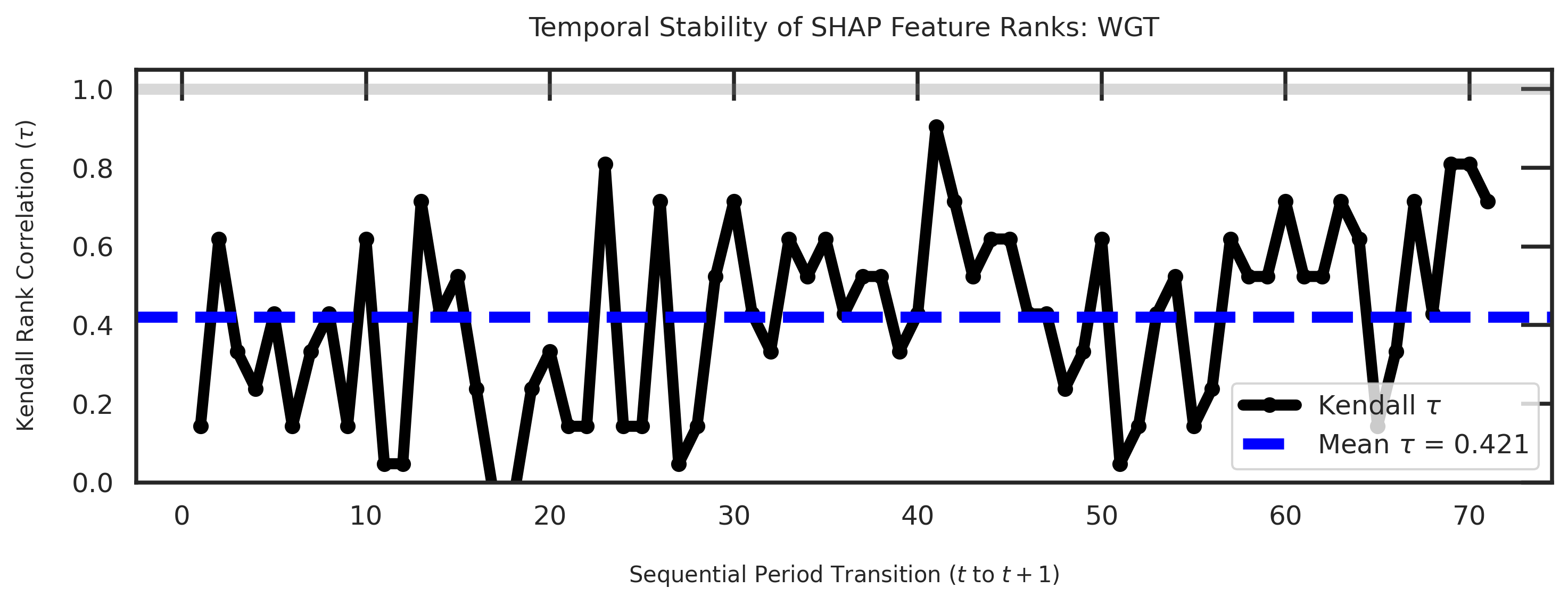}
        \caption{Weight Features}
        \label{fig:kendall_wgt}
    \end{subfigure}
    
    \vspace{0.15cm} 
    
    \begin{subfigure}[b]{0.45\textwidth}
        \centering
        \includegraphics[width=\textwidth]{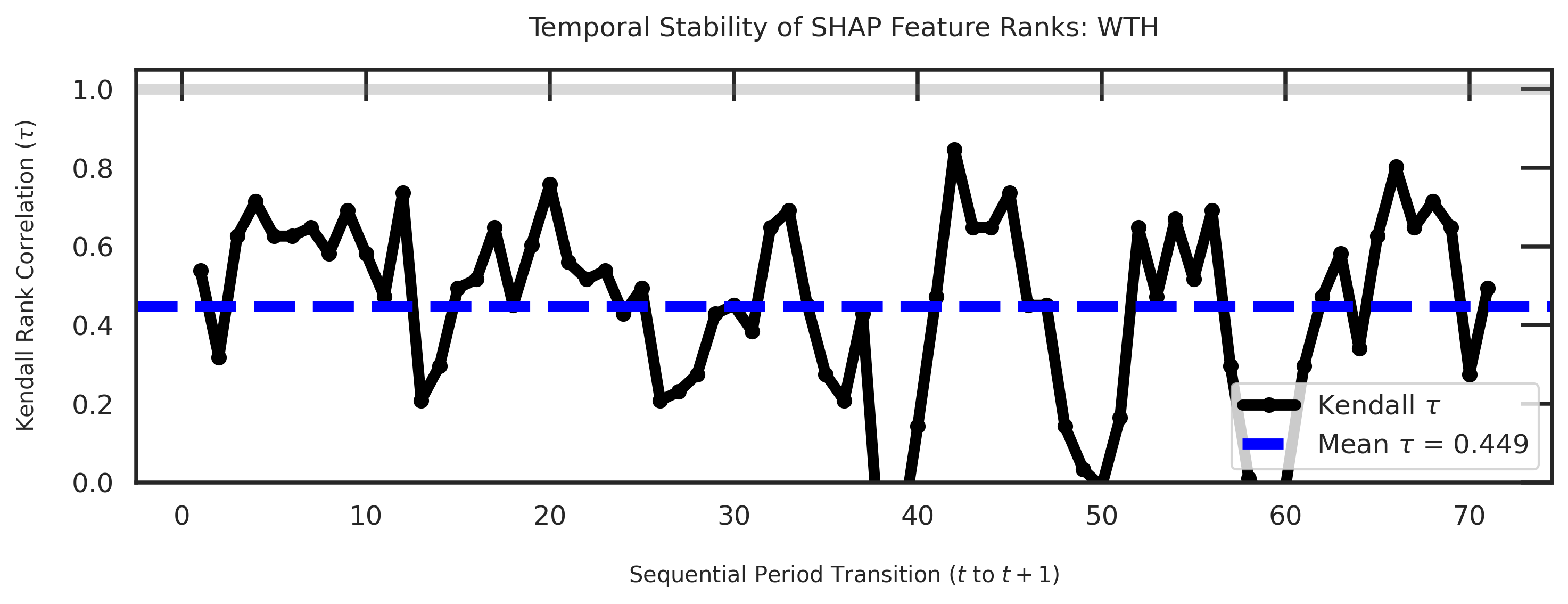}
        \caption{Weather Features}
        \label{fig:kendall_wth}
    \end{subfigure}
    \hfill
    \begin{subfigure}[b]{0.45\textwidth}
        \centering
        \includegraphics[width=\textwidth]{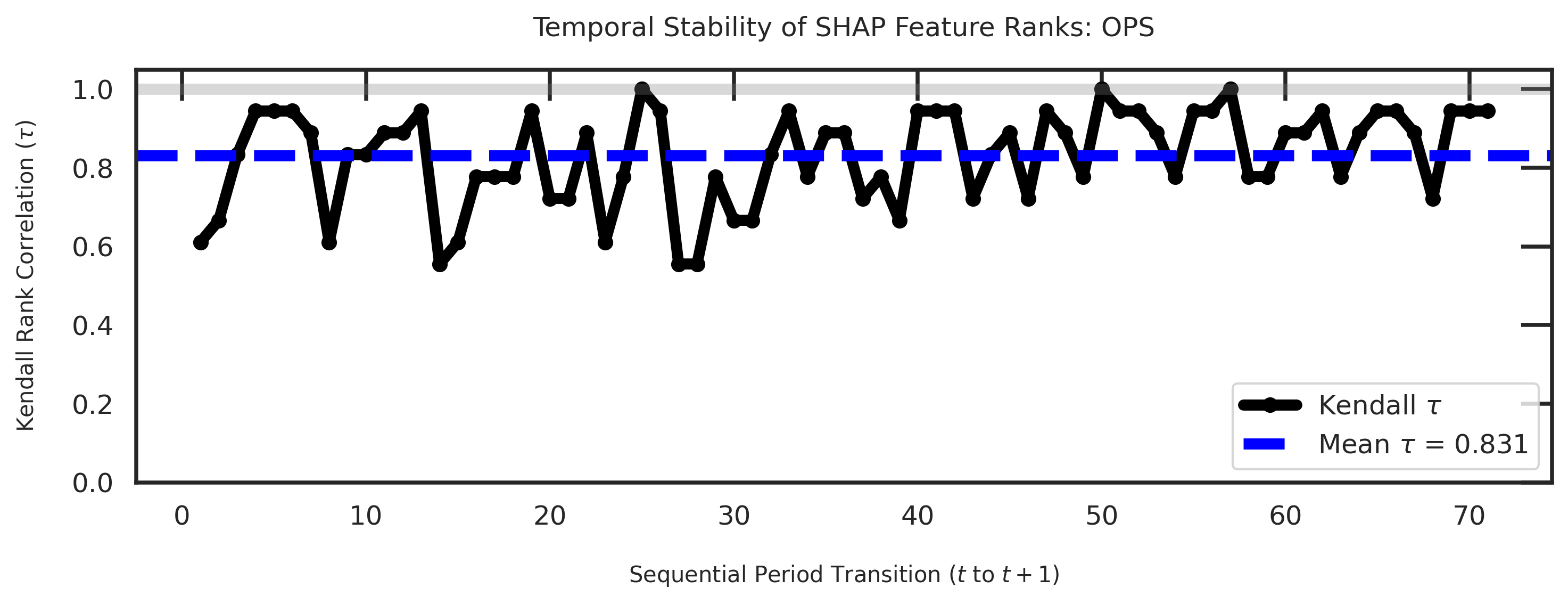}
        \caption{Operational Features}
        \label{fig:kendall_ops}
    \end{subfigure}
    
    \vspace{0.15cm}
    
    \begin{subfigure}[b]{0.45\textwidth}
        \centering
        \includegraphics[width=\textwidth]{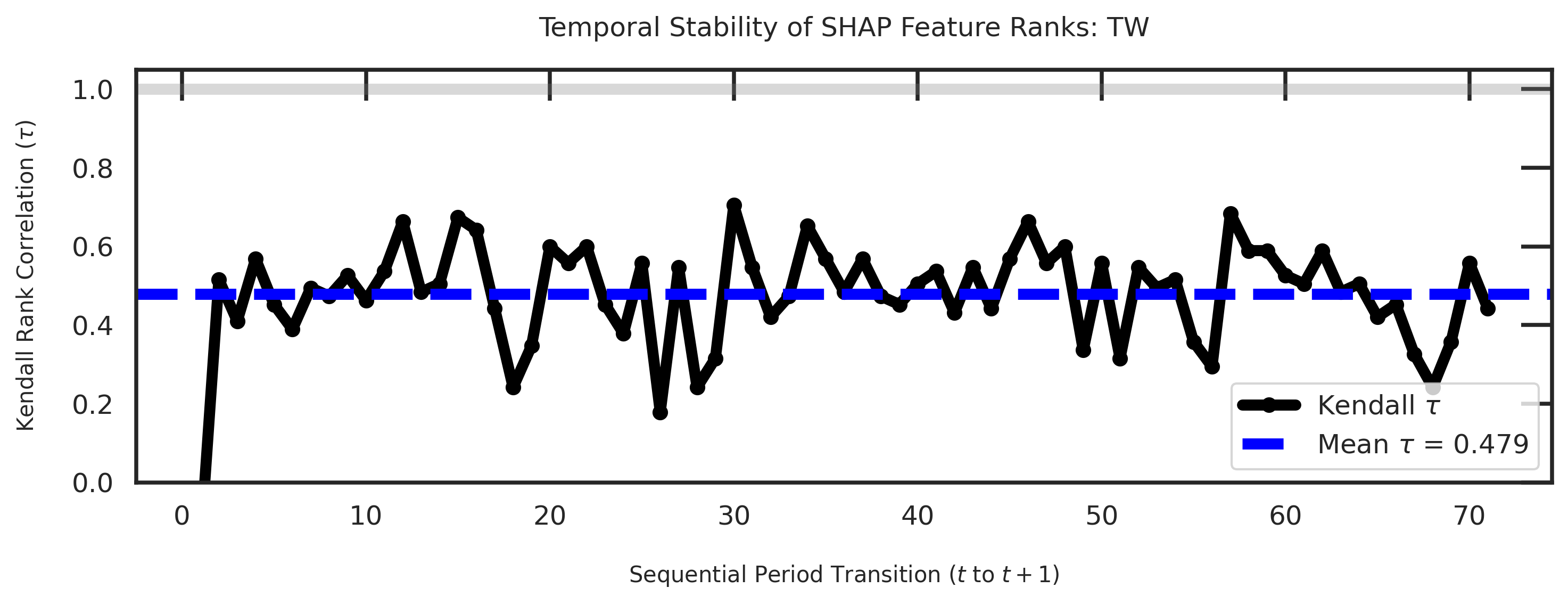}
        \caption{Topological + Weight}
        \label{fig:kendall_tw}
    \end{subfigure}
    \hfill
    \begin{subfigure}[b]{0.45\textwidth}
        \centering
        \includegraphics[width=\textwidth]{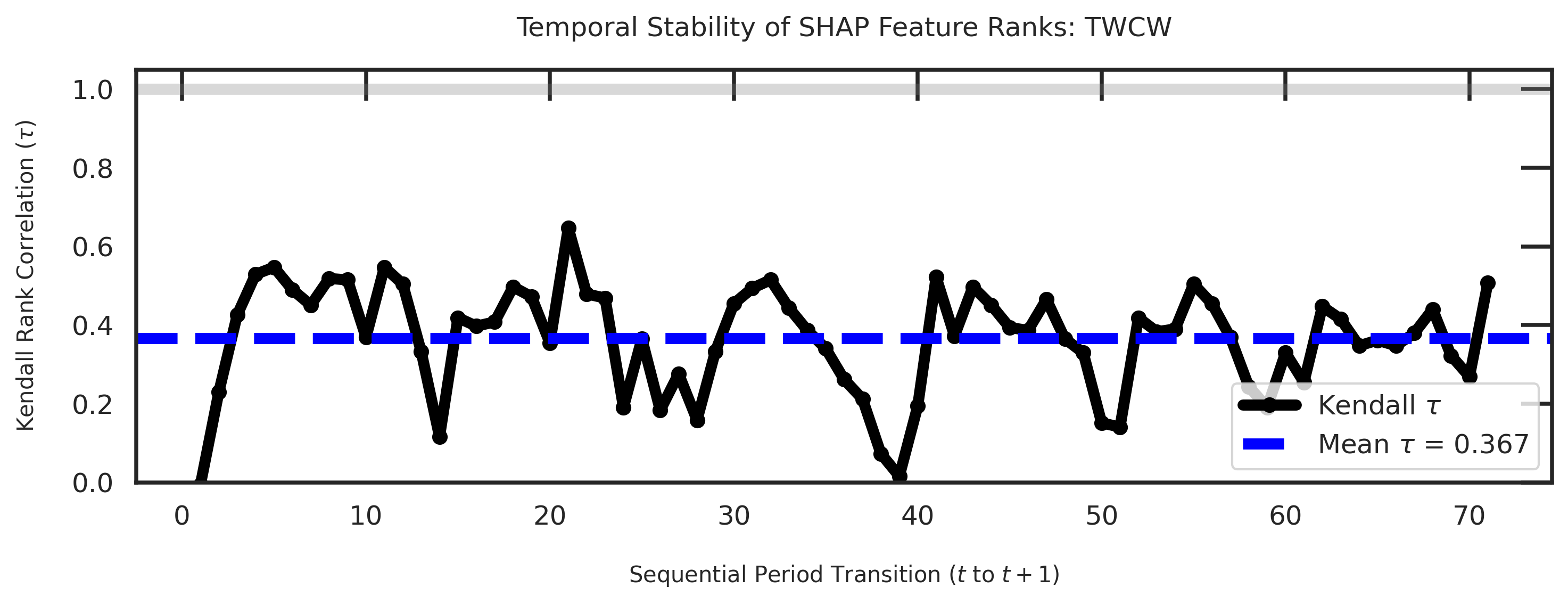}
        \caption{Topological + Weight + Weather}
        \label{fig:kendall_twcw}
    \end{subfigure}
    
    \vspace{0.15cm}
    
    \begin{subfigure}[b]{0.45\textwidth}
        \centering
        \includegraphics[width=\textwidth]{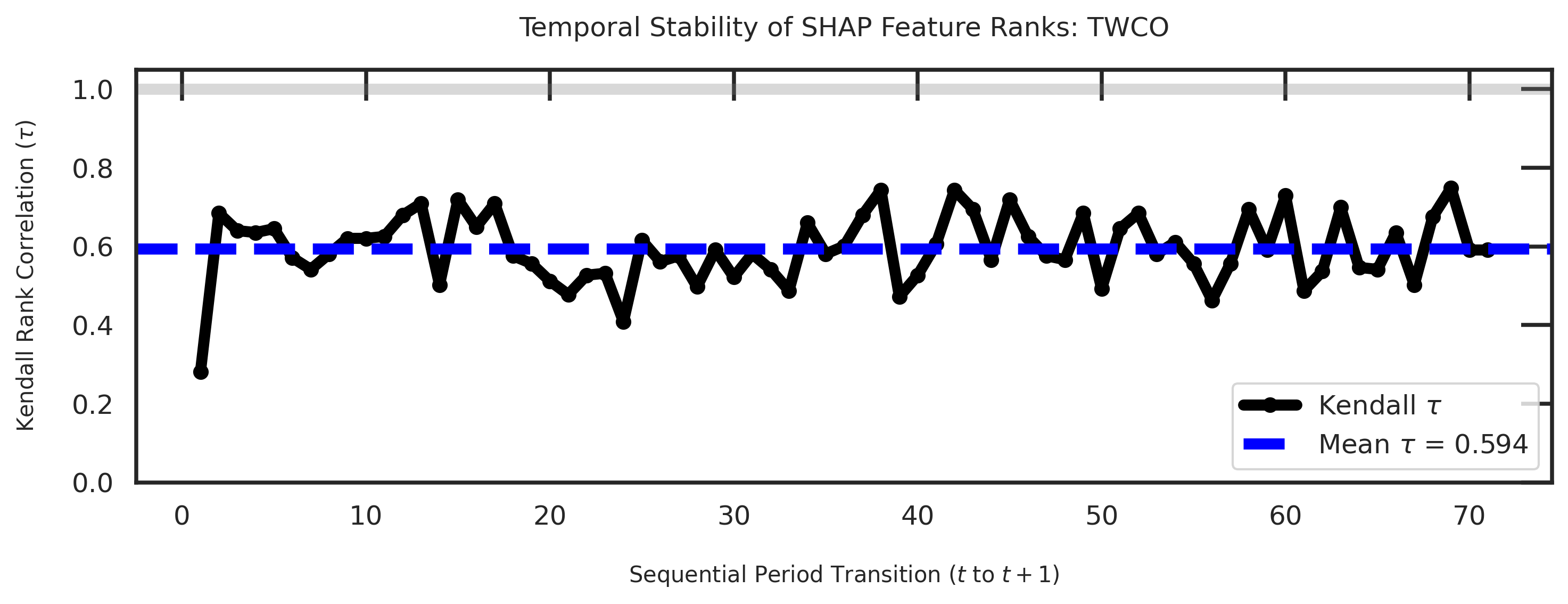}
        \caption{Topological + Weight + Operational}
        \label{fig:kendall_twco}
    \end{subfigure}
    \hfill
    \begin{subfigure}[b]{0.45\textwidth}
        \centering
        \includegraphics[width=\textwidth]{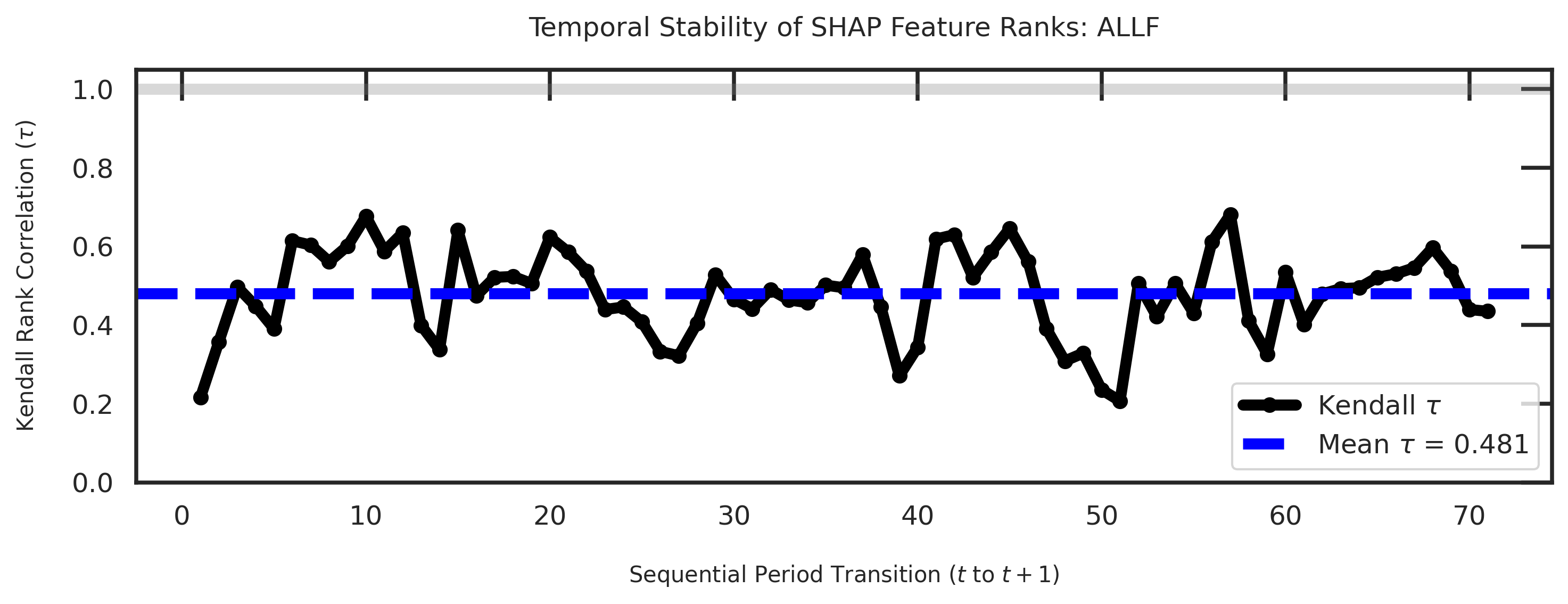}
        \caption{All Features}
        \label{fig:kendall_allf}
    \end{subfigure}
    
    \caption{Temporal trajectory of Kendall rank correlation coefficients across sequential transitions for all XGBoost feature configurations.}
    \label{fig:combined_kendall_stability}
\end{figure}


\clearpage
\section{Non Simultaneous Performance Across Months for the SHAP evaluated XGBoost model (trajectory based split)}
\label{sec:apx:performance_ot}
\begin{figure}[htbp]
    \centering
    
    \begin{subfigure}[b]{0.45\textwidth}
        \centering
        \includegraphics[width=\textwidth]{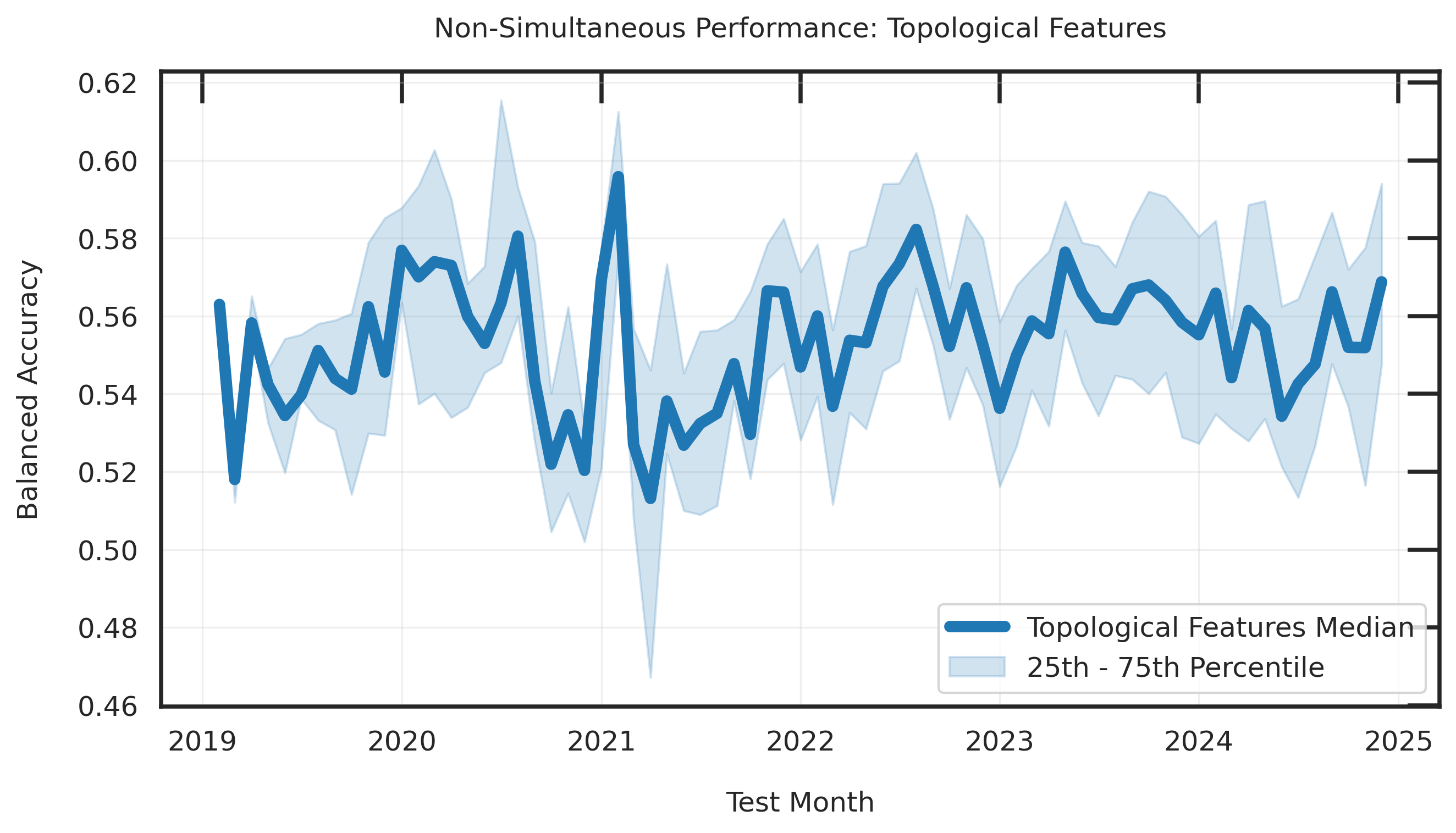}
        \caption{Topological Features}
        \label{fig:perf_ot_top}
    \end{subfigure}
    \hfill
    \begin{subfigure}[b]{0.45\textwidth}
        \centering
        \includegraphics[width=\textwidth]{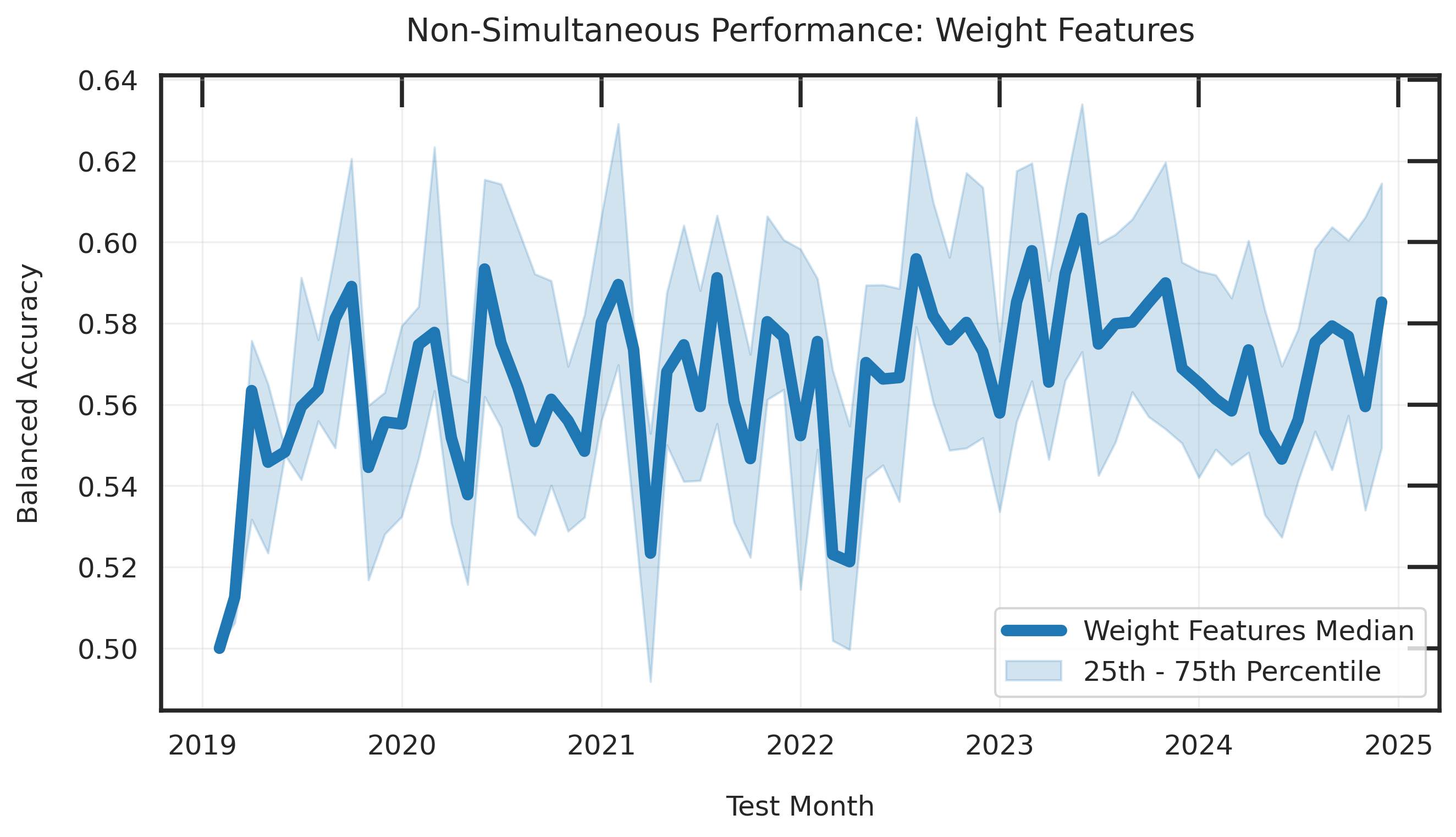}
        \caption{Weight Features}
        \label{fig:perf_ot_wgt}
    \end{subfigure}
    
    \vspace{0.15cm} 
    
    \begin{subfigure}[b]{0.45\textwidth}
        \centering
        \includegraphics[width=\textwidth]{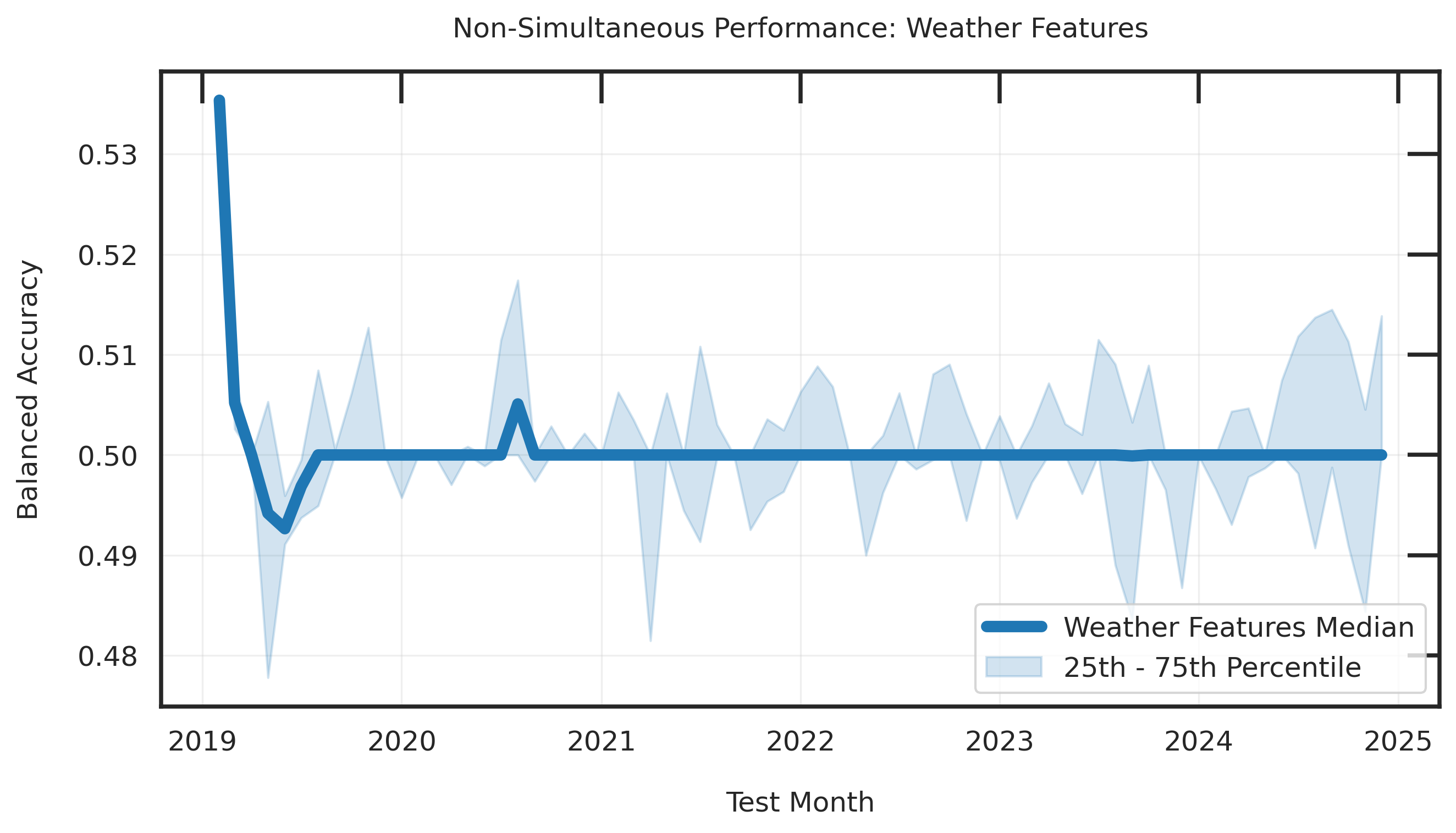}
        \caption{Weather Features}
        \label{fig:perf_ot_wth}
    \end{subfigure}
    \hfill
    \begin{subfigure}[b]{0.45\textwidth}
        \centering
        \includegraphics[width=\textwidth]{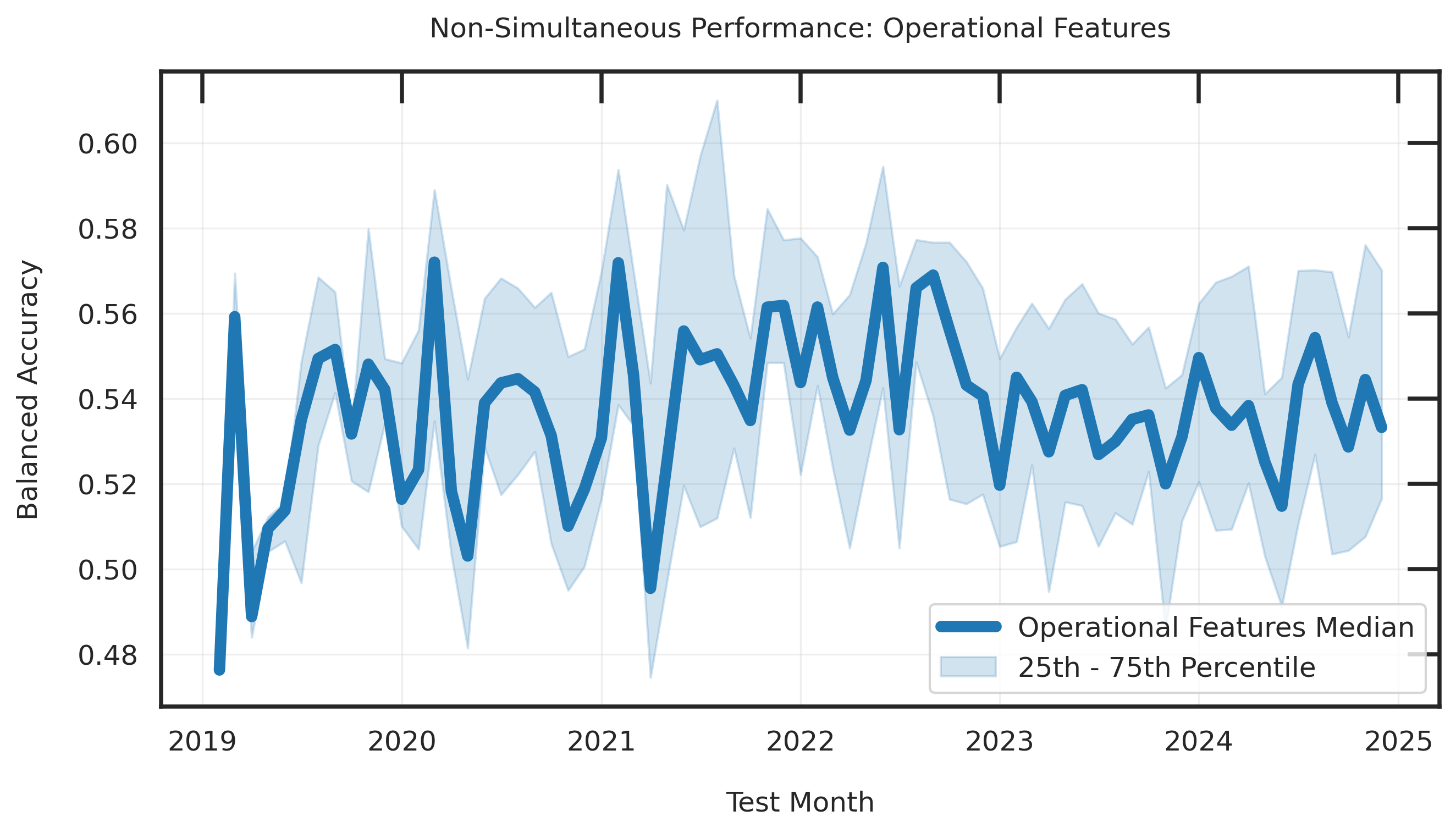}
        \caption{Operational Features}
        \label{fig:perf_ot_ops}
    \end{subfigure}
    
    \vspace{0.15cm}
    
    \begin{subfigure}[b]{0.45\textwidth}
        \centering
        \includegraphics[width=\textwidth]{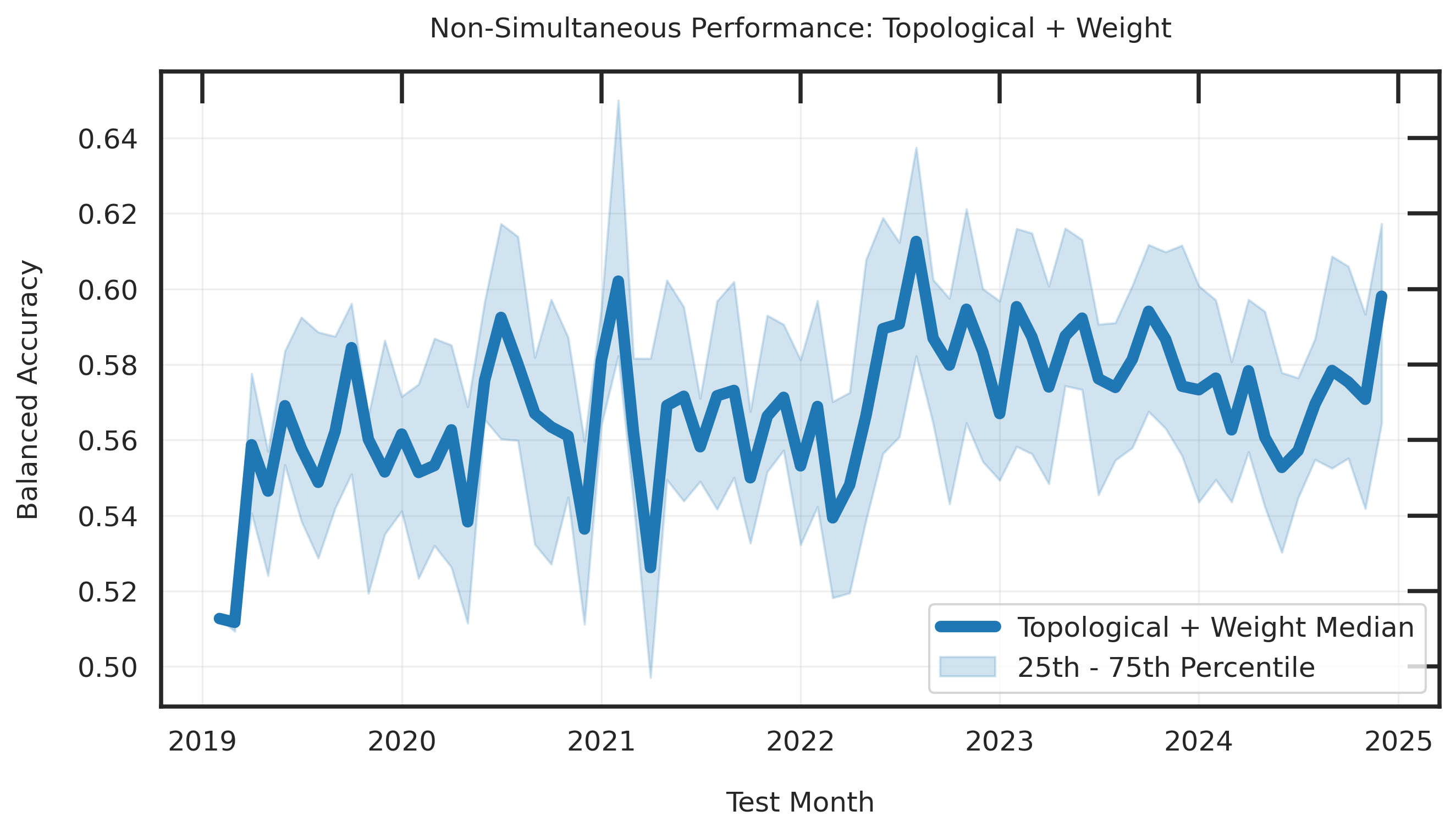}
        \caption{Topological + Weight}
        \label{fig:perf_ot_tw}
    \end{subfigure}
    \hfill
    \begin{subfigure}[b]{0.45\textwidth}
        \centering
        \includegraphics[width=\textwidth]{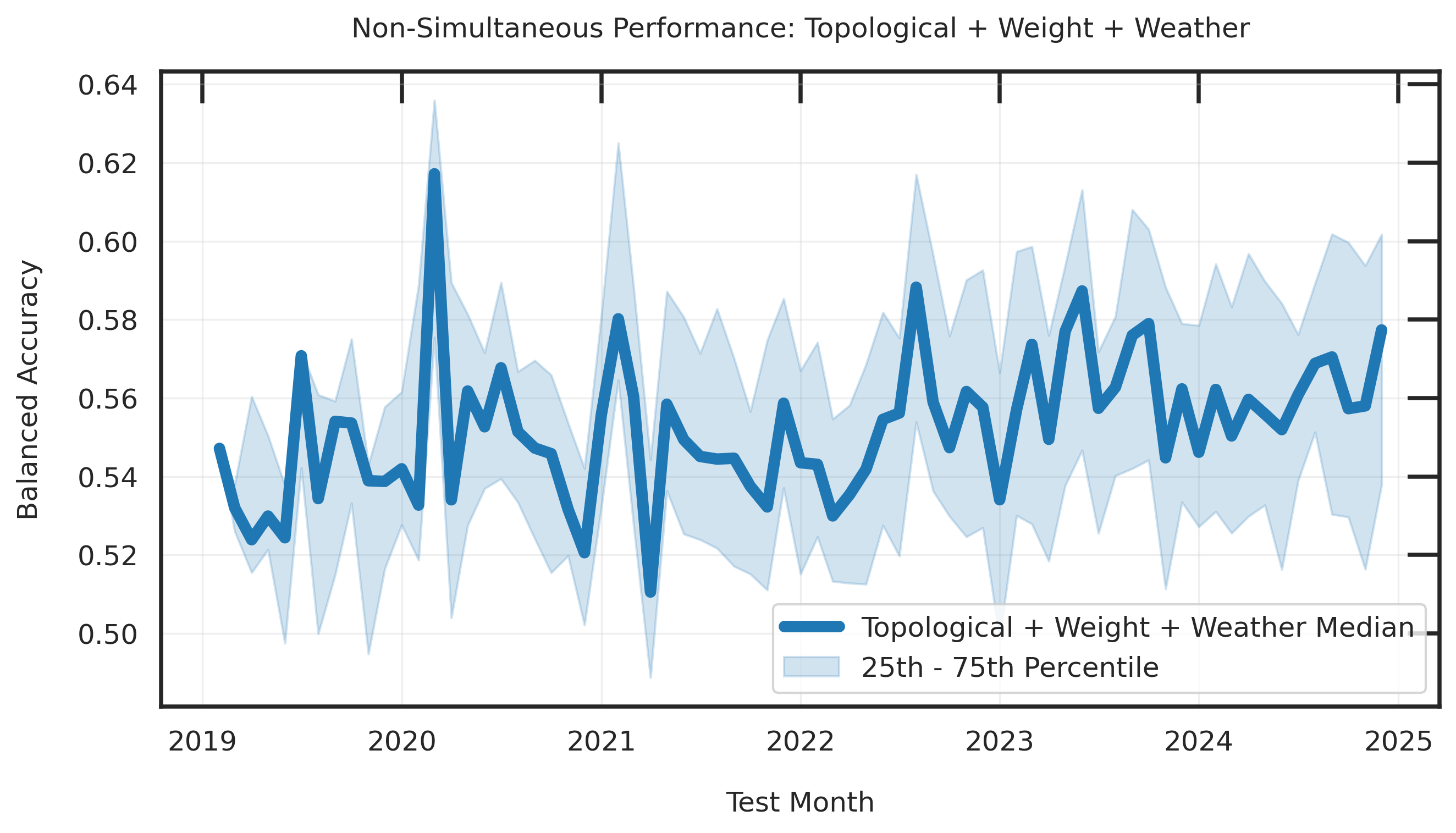}
        \caption{Topological + Weight + Weather}
        \label{fig:perf_ot_twcw}
    \end{subfigure}
    
    \vspace{0.15cm}
    
    \begin{subfigure}[b]{0.45\textwidth}
        \centering
        \includegraphics[width=\textwidth]{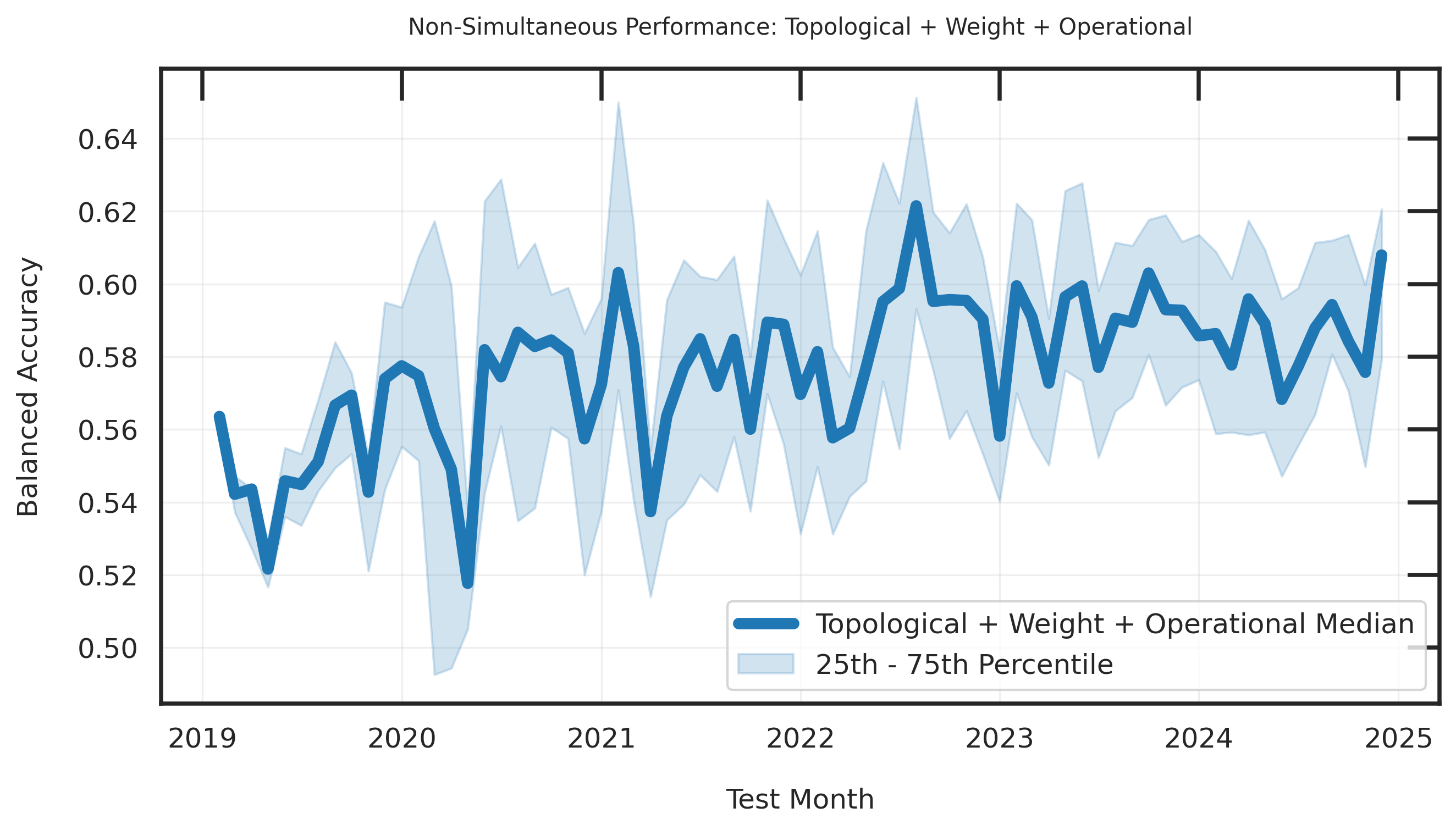}
        \caption{Topological + Weight + Operational}
        \label{fig:perf_ot_twco}
    \end{subfigure}
    \hfill
    \begin{subfigure}[b]{0.45\textwidth}
        \centering
        \includegraphics[width=\textwidth]{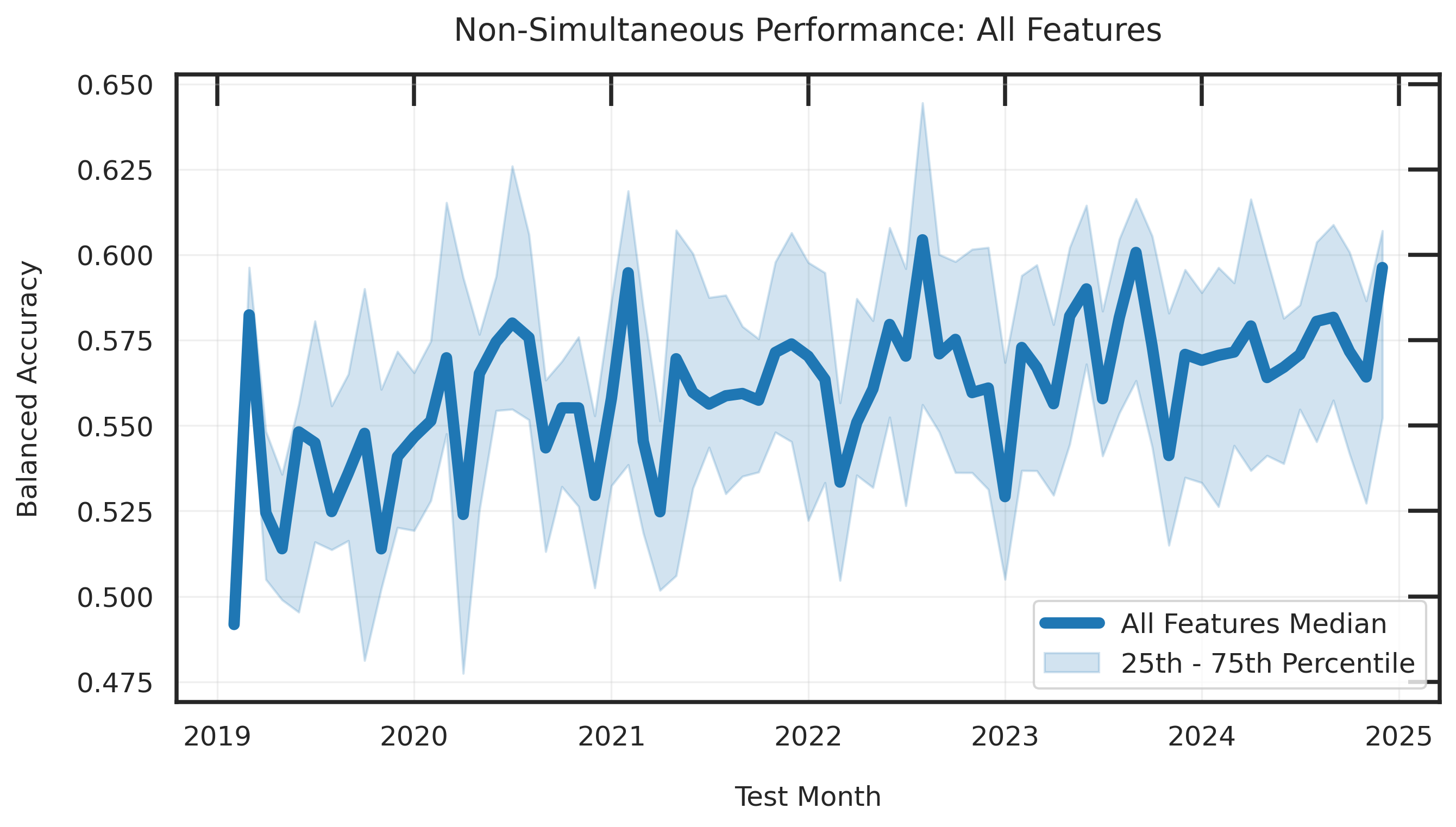}
        \caption{All Features}
        \label{fig:perf_ot_allf}
    \end{subfigure}
    
    \caption{Non-simultaneous balanced accuracy performance across months for all feature sets (XGBoost, trajectory0based split.}
    \label{fig:combined_performance_ot}
\end{figure}


\clearpage
\section{K-S Test Summary Tables}
\label{sec:apx:k-s}

\begin{table}[htbp]
\centering
\caption{Macro-Categorical Kolmogorov-Smirnov (K-S) Drift Summary. This table quantifies the frequency and magnitude (D-Statistic) of statistically significant distributional shifts across the primary feature categories.}
\label{tab:ks_macro_summary}
\begin{tabular}{lccc}
\toprule
\textbf{Category} & \textbf{Drift Frequency (\%)} & \textbf{Mean D-Stat} & \textbf{Max D-Stat} \\
\midrule
Operational (OPS) & 3.29 & 0.028 & 0.994 \\
Topological (TOP) & 6.94 & 0.045 & 0.190 \\
Weight (WGT)      & 1.41 & 0.043 & 1.000 \\
Weather (WTH)     & 84.81 & 0.645 & 1.000 \\
\bottomrule
\end{tabular}
\end{table}

\begin{table}[htbp]
\centering
\caption{Feature-Level Kolmogorov-Smirnov (K-S) Diagnostic Summary (Part 1: Topological and Weight Features). Details the significance rate of temporal drift across all 71 sequential transitions.}
\label{tab:ks_diagnostic_part1}
\resizebox{\textwidth}{!}{%
\begin{tabular}{llccc}
\toprule
\textbf{Feature} & \textbf{Category} & \textbf{Significance Rate (\%)} & \textbf{Mean D-Stat} & \textbf{Max D-Stat} \\
\midrule
\multicolumn{5}{c}{\textit{Topological Features (TOP)}} \\
\midrule
src\_degree & TOP & 4.23 & 0.045 & 0.079 \\
tgt\_degree & TOP & 1.41 & 0.044 & 0.073 \\
src\_avg\_distance\_in & TOP & 4.23 & 0.047 & 0.162 \\
common\_neighbors & TOP & 12.68 & 0.040 & 0.095 \\
jaccard\_coefficient & TOP & 12.68 & 0.045 & 0.100 \\
preferential\_attachment & TOP & 4.23 & 0.042 & 0.080 \\
adamic\_adar\_index & TOP & 14.08 & 0.046 & 0.095 \\
resource\_allocation\_index & TOP & 11.27 & 0.045 & 0.095 \\
src\_avg\_distance\_out & TOP & 4.23 & 0.044 & 0.168 \\
src\_avg\_distance\_total & TOP & 7.04 & 0.049 & 0.190 \\
tgt\_avg\_distance\_in & TOP & 4.23 & 0.045 & 0.154 \\
tgt\_avg\_distance\_out & TOP & 4.23 & 0.041 & 0.157 \\
tgt\_avg\_distance\_total & TOP & 5.63 & 0.046 & 0.182 \\
\midrule
\multicolumn{5}{c}{\textit{Weight Features (WGT)}} \\
\midrule
decayed\_edge\_weight & WGT & 1.41 & 0.044 & 0.994 \\
src\_w\_deg\_inbound & WGT & 1.41 & 0.045 & 0.968 \\
src\_w\_deg\_outbound & WGT & 1.41 & 0.042 & 1.000 \\
src\_w\_deg\_total & WGT & 1.41 & 0.044 & 1.000 \\
tgt\_w\_deg\_inbound & WGT & 1.41 & 0.044 & 1.000 \\
tgt\_w\_deg\_outbound & WGT & 1.41 & 0.041 & 1.000 \\
tgt\_w\_deg\_total & WGT & 1.41 & 0.041 & 1.000 \\
\bottomrule
\end{tabular}%
}
\end{table}

\begin{table}[htbp]
\centering
\caption{Feature-Level Kolmogorov-Smirnov (K-S) Diagnostic Summary (Part 2: Weather and Operational Features). Details the significance rate of temporal drift across all 71 sequential transitions.}
\label{tab:ks_diagnostic_part2}
\resizebox{\textwidth}{!}{%
\begin{tabular}{llccc}
\toprule
\textbf{Feature} & \textbf{Category} & \textbf{Significance Rate (\%)} & \textbf{Mean D-Stat} & \textbf{Max D-Stat} \\
\midrule
\multicolumn{5}{c}{\textit{Weather Features (WTH)}} \\
\midrule
src\_avg\_temperature\_2m & WTH & 100.00 & 0.878 & 1.000 \\
tgt\_avg\_temperature\_2m & WTH & 100.00 & 0.878 & 1.000 \\
src\_avg\_soil\_temperature & WTH & 100.00 & 0.913 & 1.000 \\
tgt\_avg\_soil\_temperature & WTH & 100.00 & 0.914 & 1.000 \\
src\_avg\_rain & WTH & 100.00 & 0.784 & 1.000 \\
tgt\_avg\_rain & WTH & 100.00 & 0.785 & 1.000 \\
src\_avg\_snowfall & WTH & 56.34 & 0.364 & 1.000 \\
tgt\_avg\_snowfall & WTH & 56.34 & 0.363 & 1.000 \\
src\_avg\_snow\_depth & WTH & 38.03 & 0.270 & 1.000 \\
tgt\_avg\_snow\_depth & WTH & 38.03 & 0.270 & 1.000 \\
src\_avg\_wind\_speed\_10m & WTH & 100.00 & 0.587 & 1.000 \\
tgt\_avg\_wind\_speed\_10m & WTH & 98.59 & 0.584 & 1.000 \\
src\_avg\_wind\_gusts\_10m & WTH & 100.00 & 0.721 & 1.000 \\
tgt\_avg\_wind\_gusts\_10m & WTH & 100.00 & 0.720 & 1.000 \\
\midrule
\multicolumn{5}{c}{\textit{Operational Features (OPS)}} \\
\midrule
distance & OPS & 0.00 & 0.013 & 0.042 \\
last\_active\_volume & OPS & 1.41 & 0.044 & 0.994 \\
months\_since\_last\_service & OPS & 1.41 & 0.033 & 0.994 \\
is\_new\_or\_stale\_route & OPS & 1.41 & 0.022 & 0.994 \\
ratio\_intercity & OPS & 2.82 & 0.038 & 0.090 \\
ratio\_intercity\_direct & OPS & 0.00 & 0.006 & 0.027 \\
ratio\_international & OPS & 0.00 & 0.003 & 0.021 \\
ratio\_operational\_exceptions & OPS & 11.27 & 0.043 & 0.124 \\
ratio\_sprinter & OPS & 11.27 & 0.046 & 0.106 \\
\bottomrule
\end{tabular}%
}
\end{table}


\clearpage
\section{Exploratory Data Analysis (EDA) Supplementary Figures}
\label{sec:apx:eda_supplementary}


\begin{figure}[H]
    \centering
    \begin{subfigure}[b]{\textwidth}
        \centering
        \includegraphics[width=0.40\linewidth]{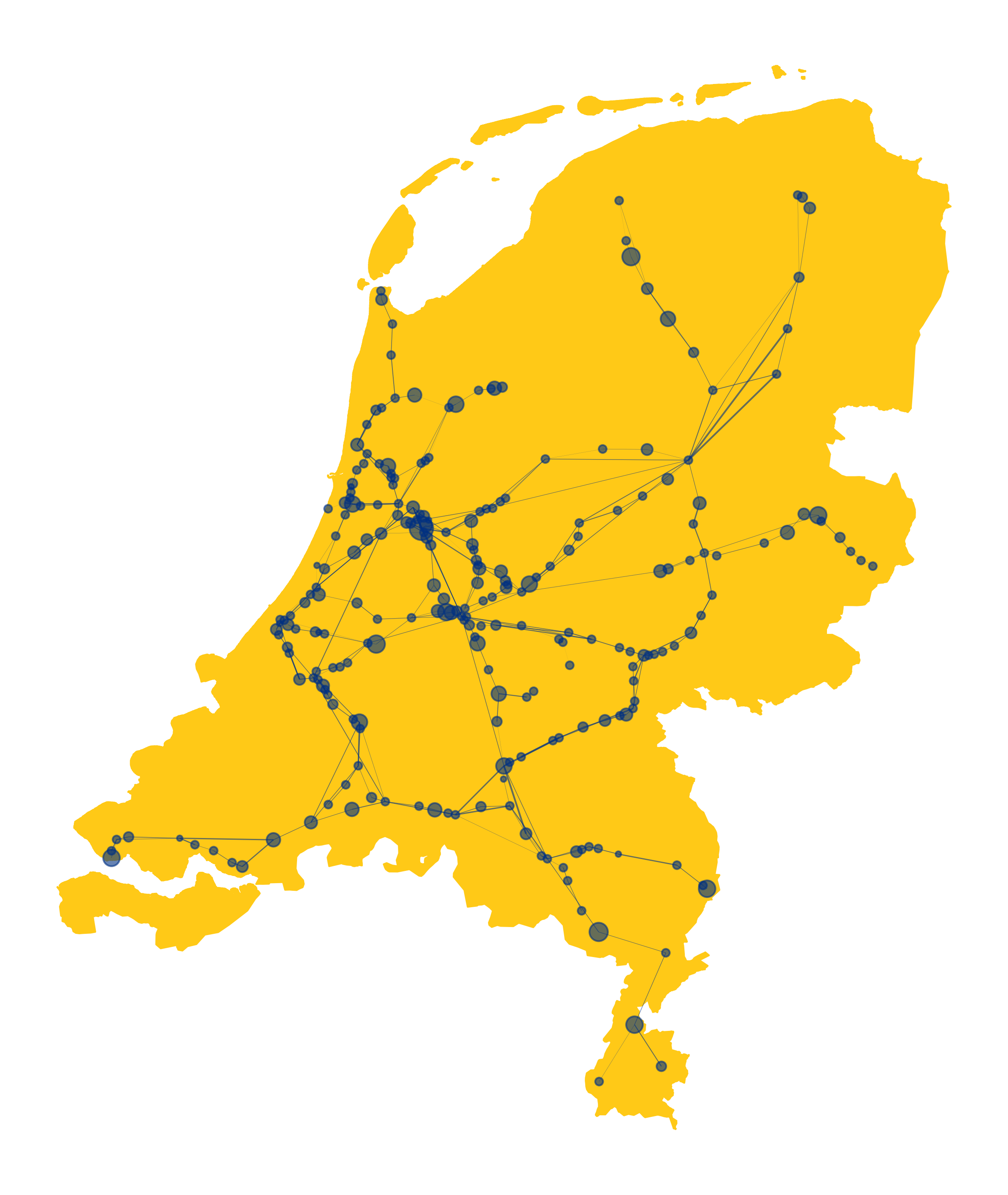}
        \caption{Geographical representation of the Dutch railway network at 2024-04, highlighting active stations and scheduled trajectory connections stops stations.}
        \label{fig:eda_map}
    \end{subfigure}


    \begin{subfigure}[b]{\textwidth}
        \centering
        \includegraphics[width=0.40\linewidth]{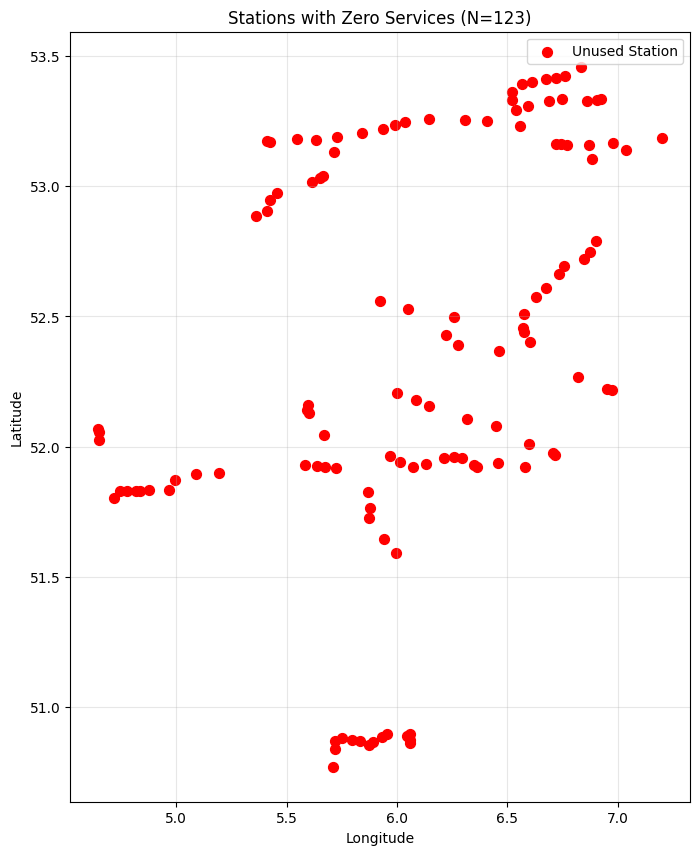}
        \caption{Stations from the reference stations dataset with no historical train railway passenger services operated by NS (i.e. present in stations dataset but not in historical services trajectories dataset).}
        \label{fig:eda_stations_noservices}
    \end{subfigure}
    \caption{Spatial overview of the Dutch railway network (April 2024) infrastructure and historical passenger service coverage.}
    \label{fig:eda_map_combined_grid}
\end{figure}

\begin{figure}[htbp]
    \centering
    \includegraphics[width=0.5\linewidth]{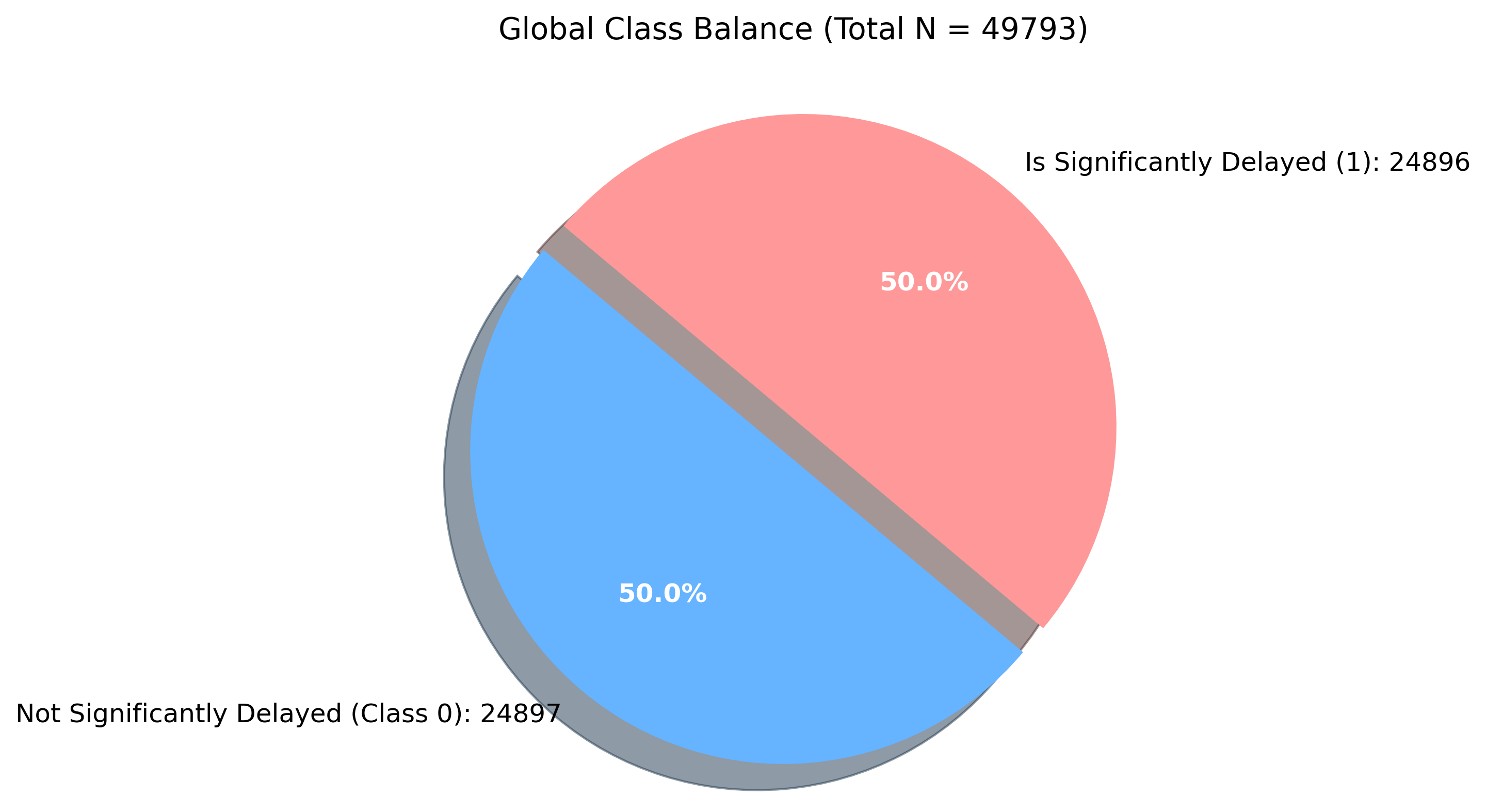}
    \caption{Binary Class Distribution with Target Label 'Is Significantly Delayed'}
    \label{fig:class_balance_pie}
\end{figure}

\begin{figure}[htbp]
    \centering
\includegraphics[width=0.5\linewidth]{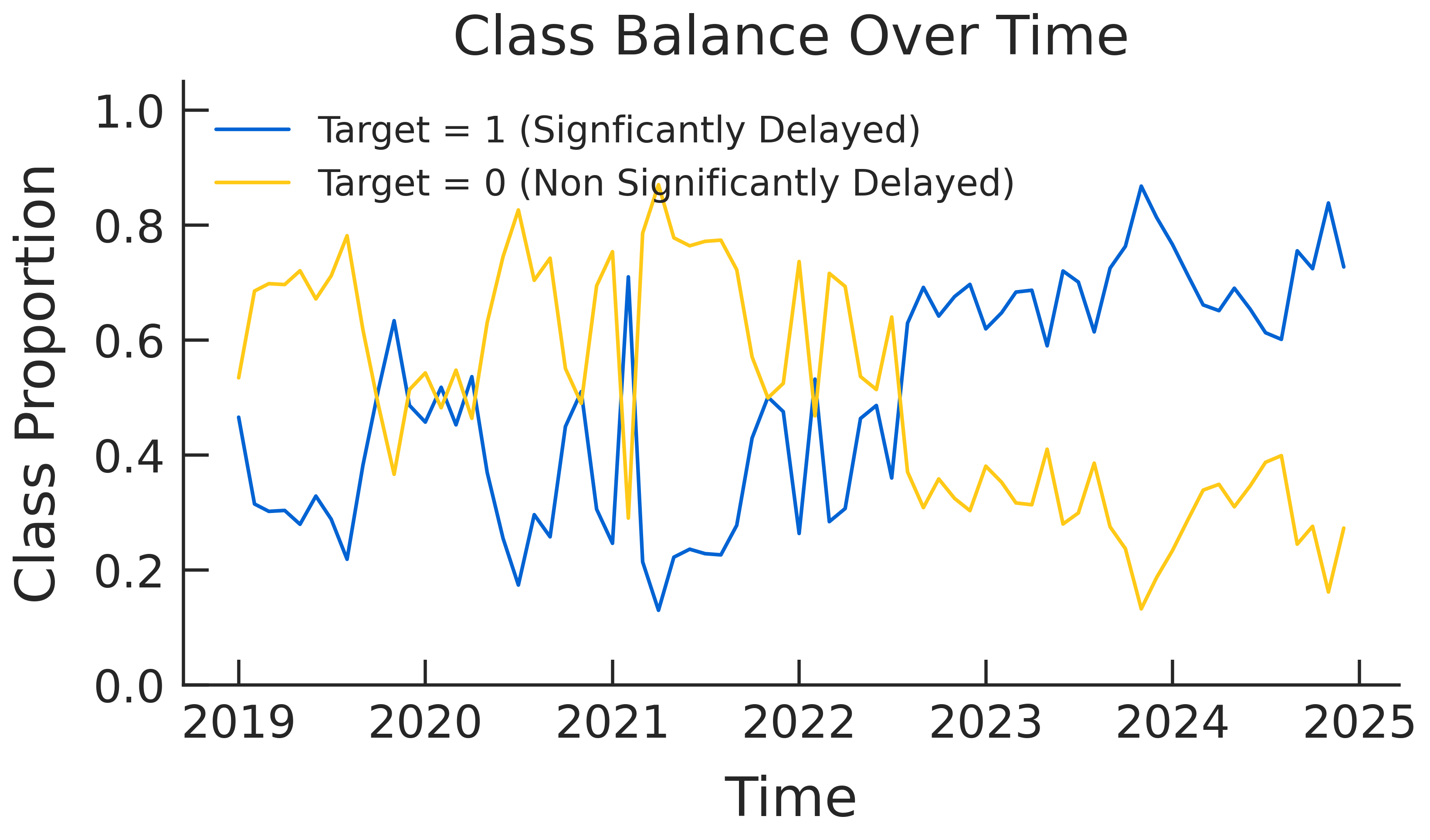}
    \caption{Monthly class balance illustrating the proportion of on-time versus significantly delayed edges over the observation period.}
    \label{fig:eda_class_balance_OT}
\end{figure}

\begin{figure}[htbp]
    \centering
    \includegraphics[width=0.5\linewidth]{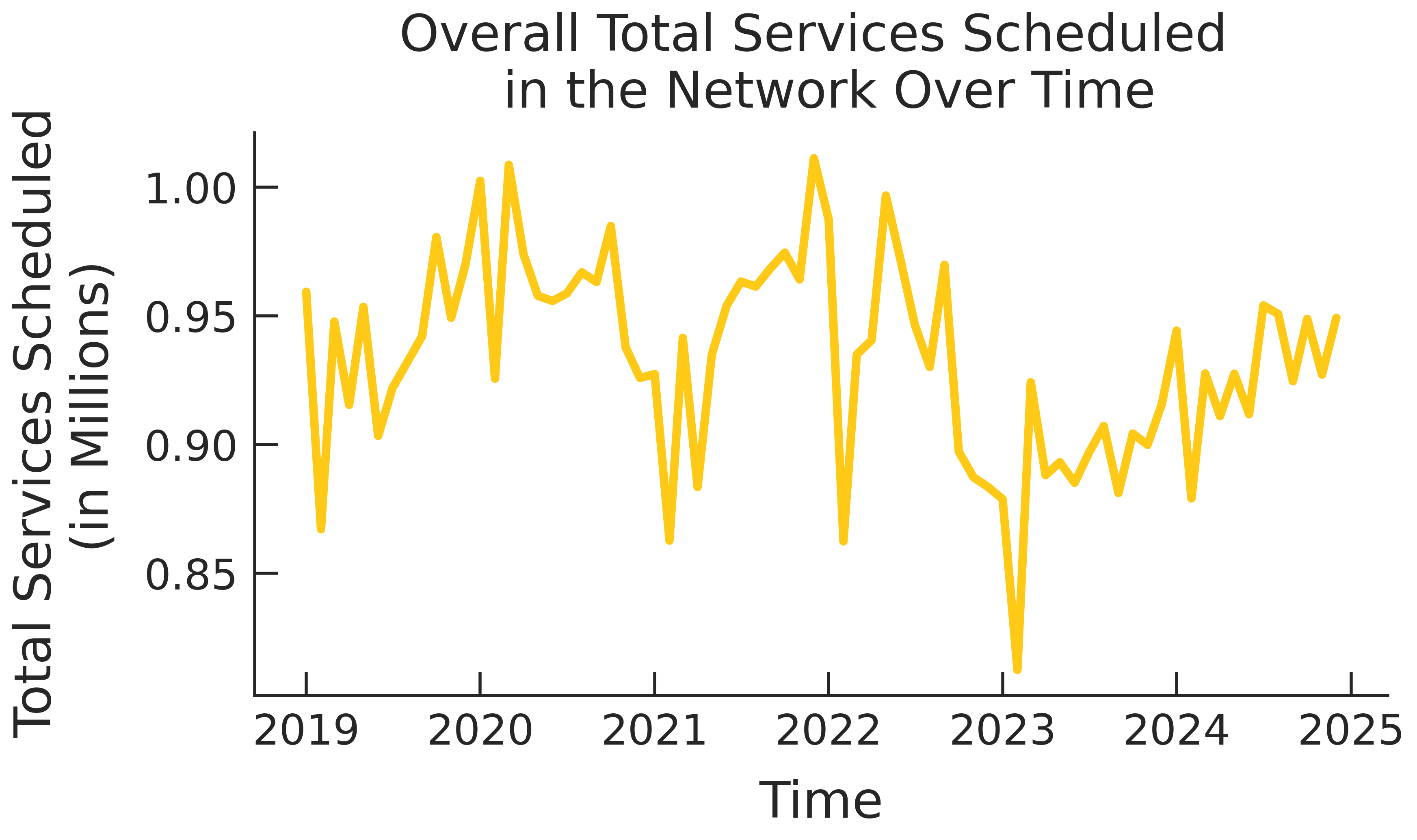}
    \caption{Total volume of scheduled train services aggregated per month across the entire railway network.}
    \label{fig:eda_total_services}
\end{figure}

\begin{figure}[htbp]
    \centering
    \includegraphics[width=0.5\linewidth]{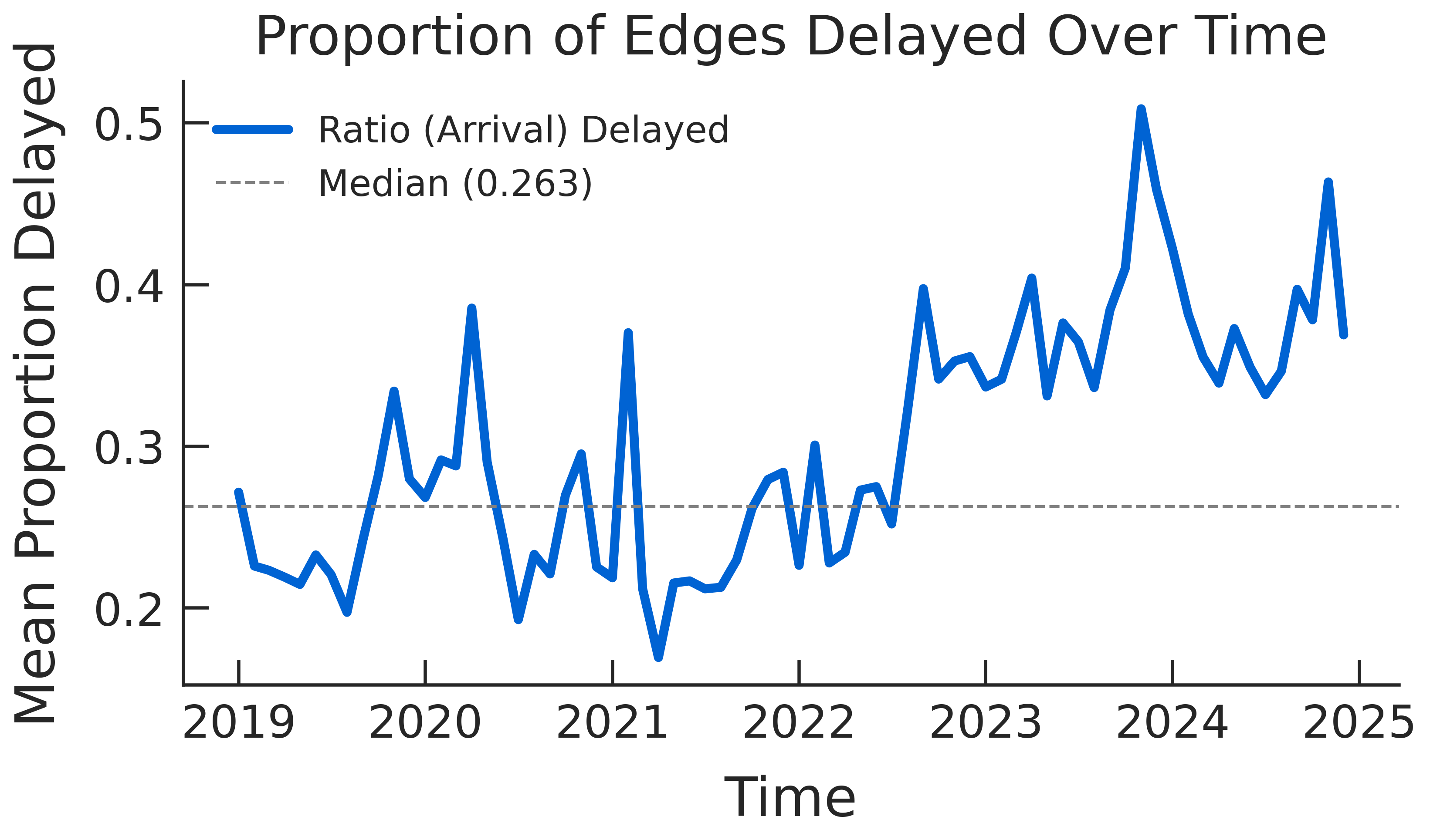}
    \caption{Mean proportion of delayed arrivals per edge over time, measured against the historical global median.}
    \label{fig:eda_ratio_delayed}
\end{figure}

\begin{figure}[htbp]
    \centering
    \includegraphics[width=0.5\linewidth]{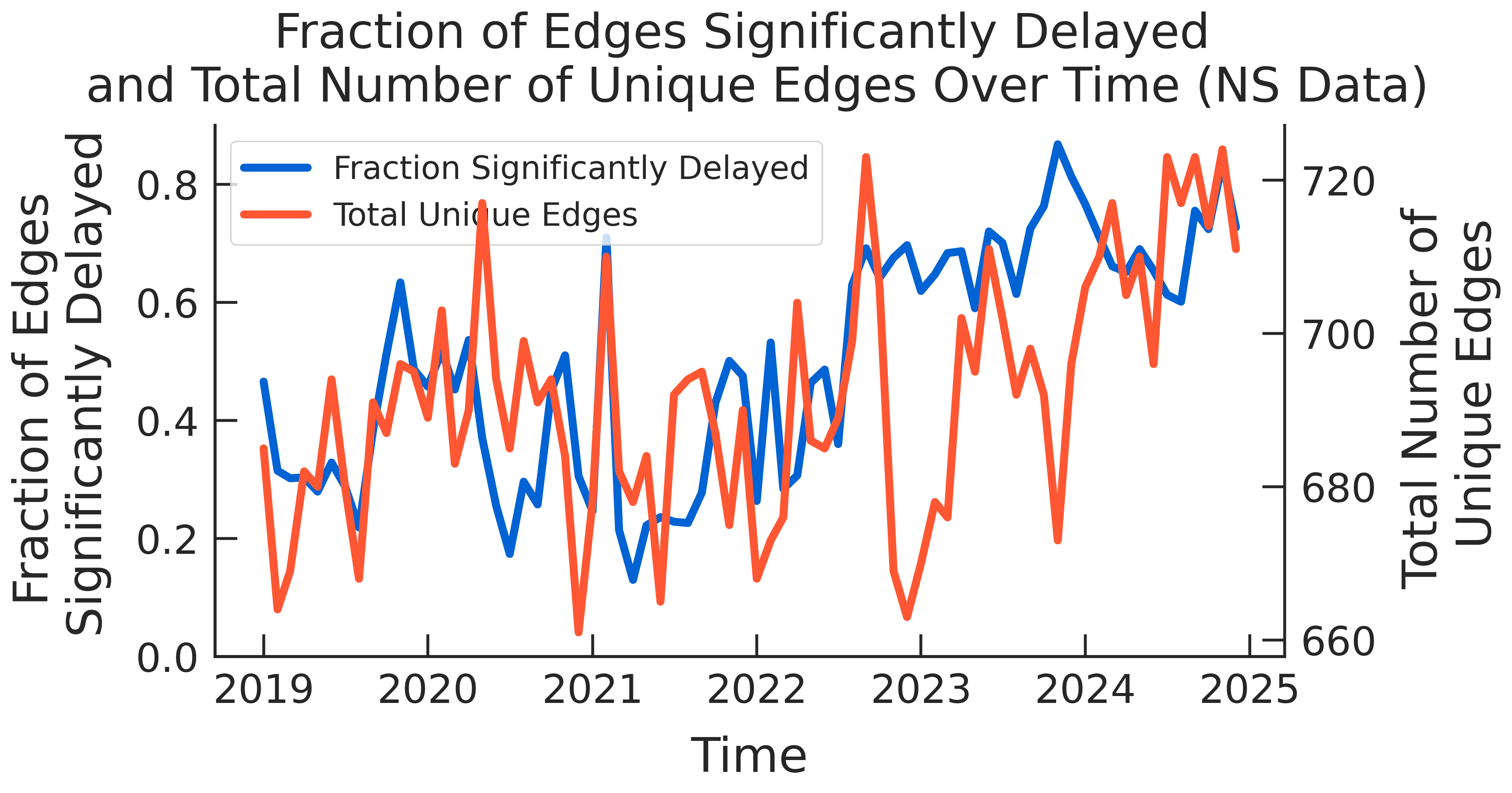}
    \caption{Distribution of the binary target label across unique edges within the network infrastructure of NS data.}
    \label{fig:eda_target_unique_edges}
\end{figure}

\begin{figure}[htbp]
    \centering
    \includegraphics[width=0.5\linewidth]{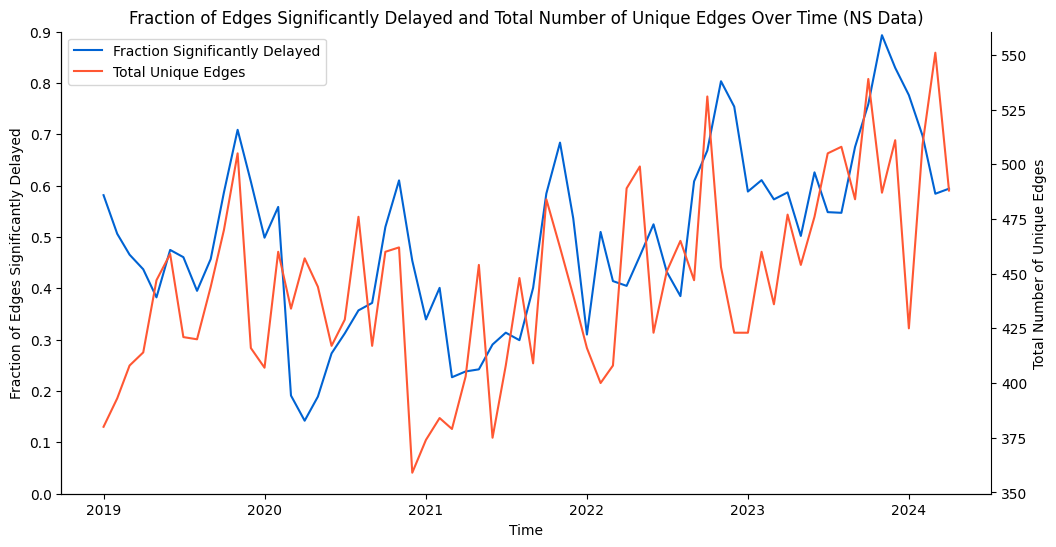}
    \caption{Distribution of the binary target label across unique edges within the network infrastructure utilizing only origin-destination as trajectories of NS data (Kämper's).}
    \label{fig:eda_kamper_unique_edges}
\end{figure}

\begin{figure}[htbp]
    \centering
    
    \begin{subfigure}[b]{\textwidth}
        \centering
        \includegraphics[width=0.5\linewidth]{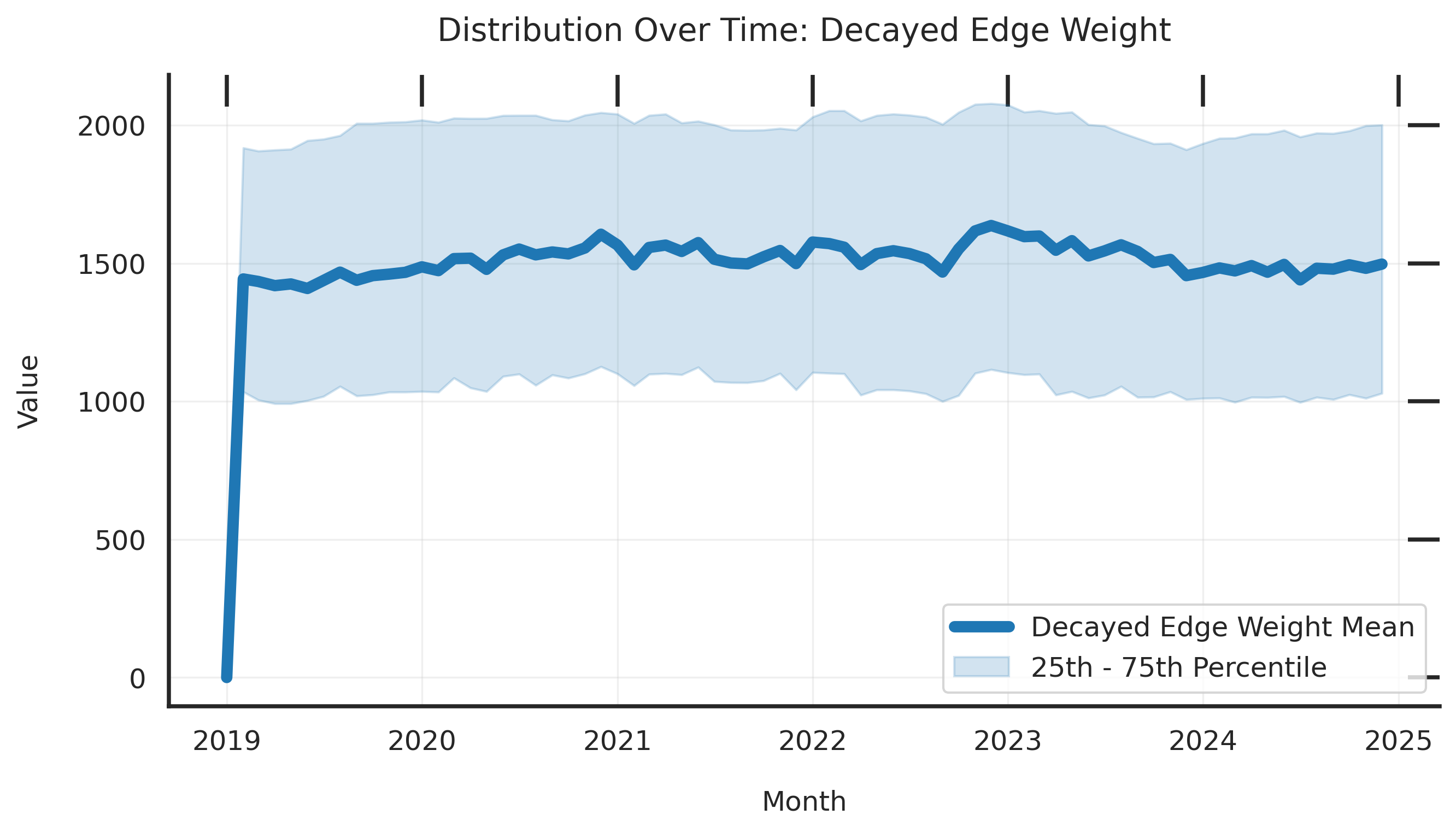}
        \caption{Decayed Edge Weight}
        \label{fig:eda_decayed_weight}
    \end{subfigure}
    
    \vspace{0.4cm} 
    
    \begin{subfigure}[b]{\textwidth}
        \centering
        \includegraphics[width=0.5\linewidth]{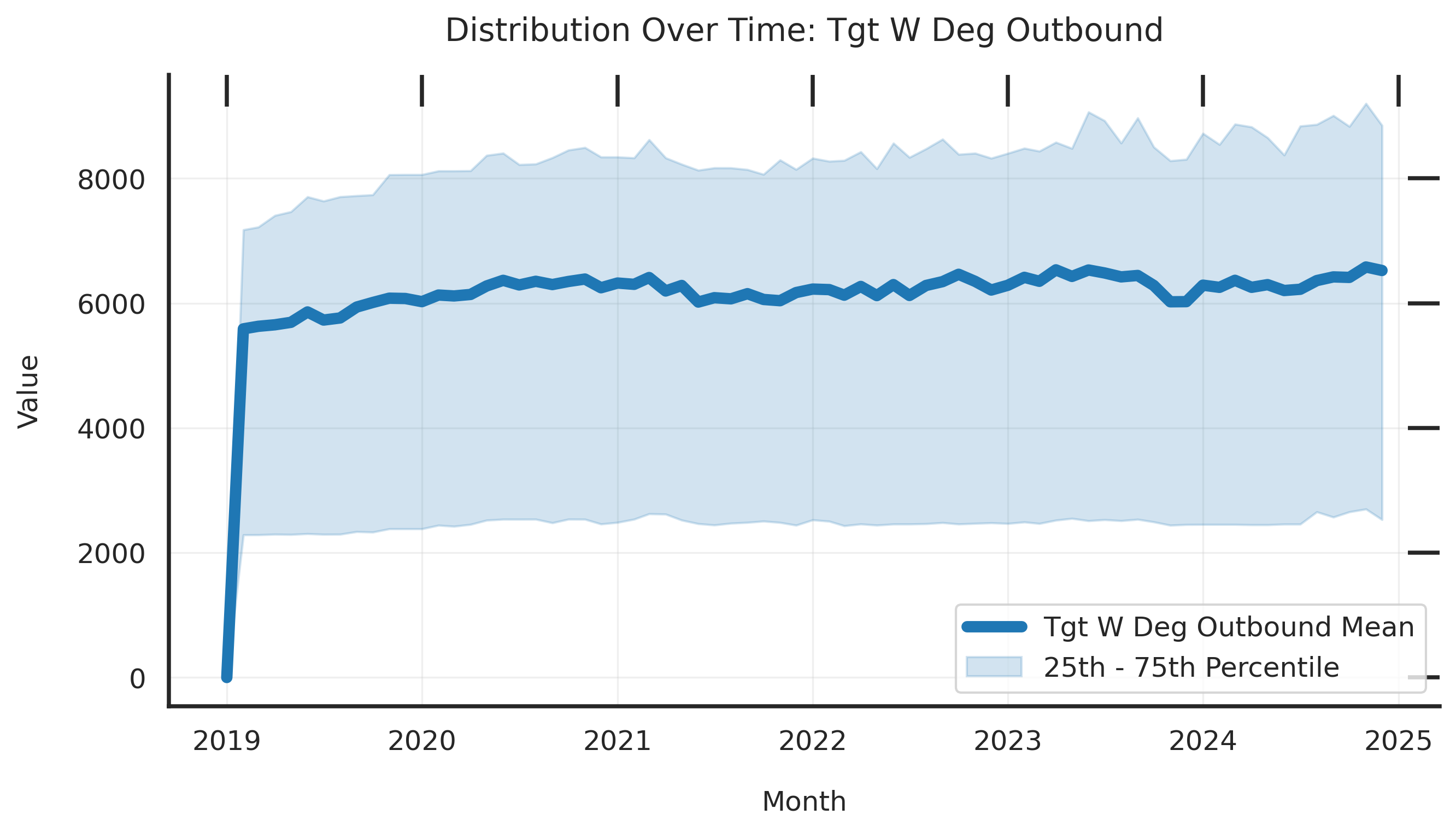}
        \caption{Target Weighted Outbound Degree}
        \label{fig:eda_tgt_outbound}
    \end{subfigure}
    
    \vspace{0.4cm} 
    
    \begin{subfigure}[b]{\textwidth}
        \centering
        \includegraphics[width=0.5\linewidth]{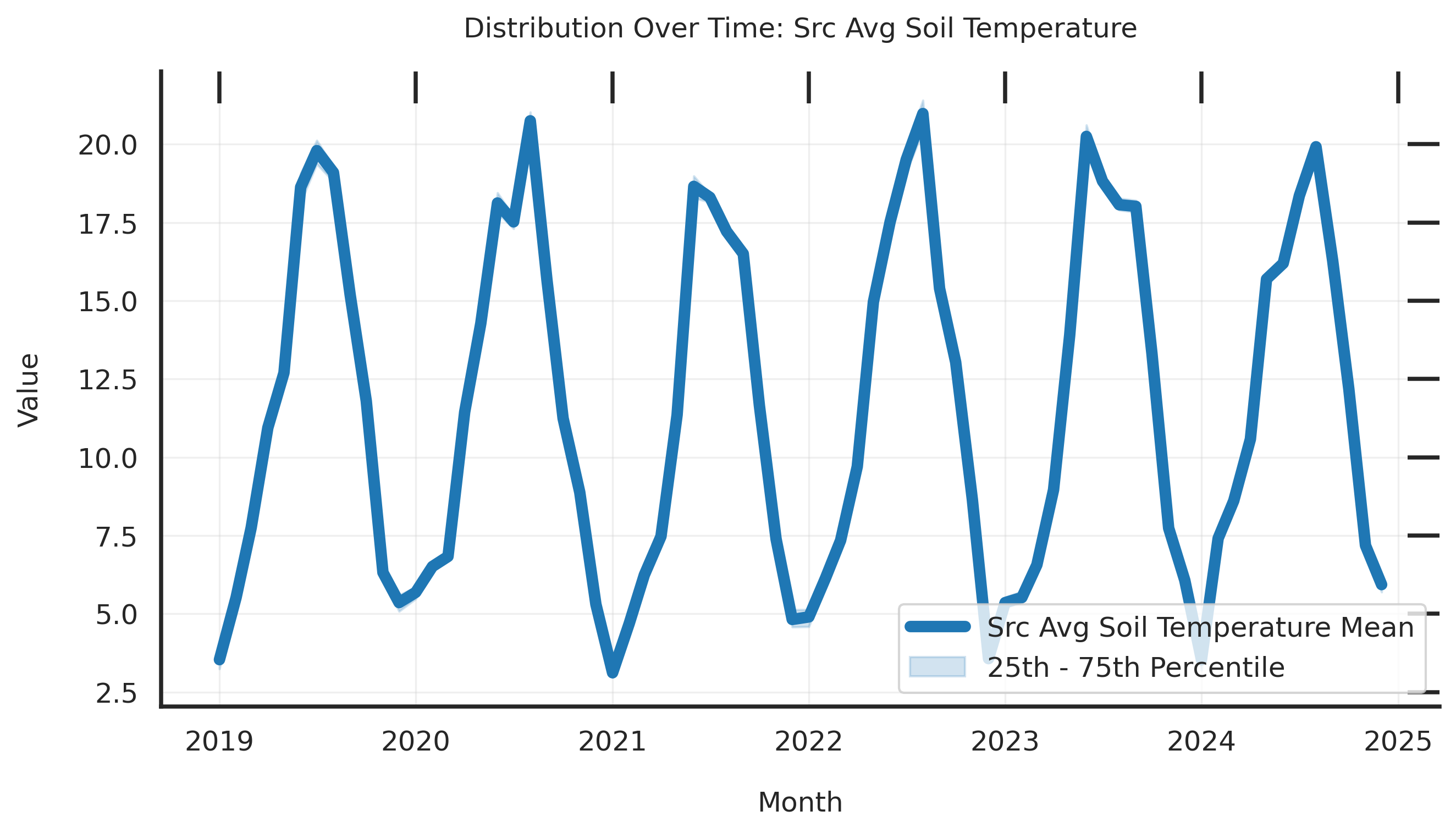}
        \caption{Source Average Soil Temperature}
        \label{fig:eda_src_soil_temp}
    \end{subfigure}
    
    \caption{Distribution over time for selected network and environmental features. Shaded regions denote the interquartile range (25th--75th percentiles). Note that variations in \texttt{src\_avg\_soil\_temperature} are present but extremely minor, causing the interquartile range to collapse and appear as a single line.}
    \label{fig:eda_features_ot}
\end{figure}


\clearpage
\section{Model Performance Distributions (XGBoost Trajectory-based split)}
\label{sec:apx:performance_distributions}

\begin{figure}[htbp]
    \centering

    \begin{subfigure}[b]{\textwidth}
        \centering
        \includegraphics[width=0.6\linewidth]{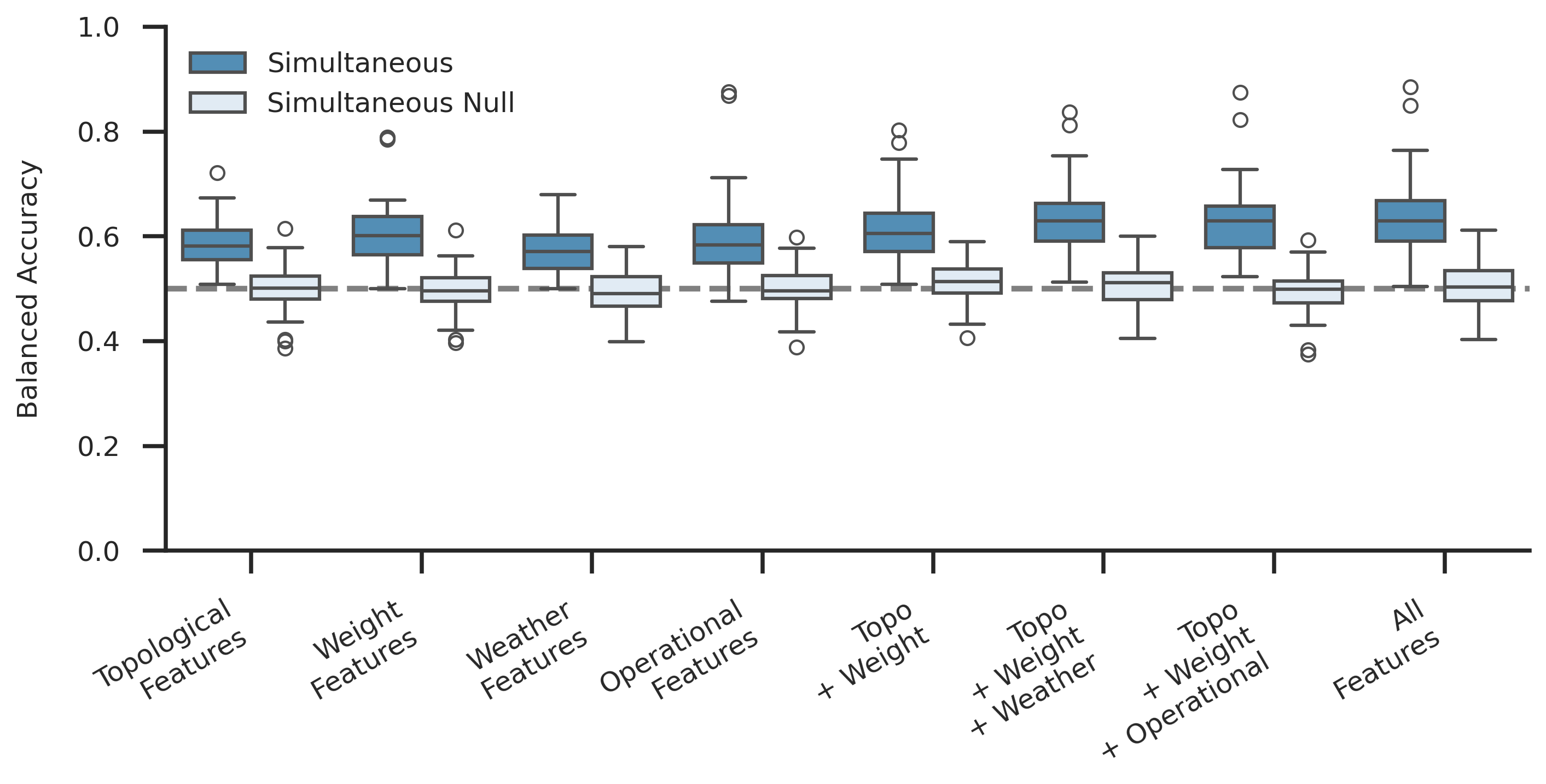}
        \caption{Box plot of balanced accuracy across all feature sets for Simultaneous testing (XGBoost, Trajectory-based split, Global Aggregation).}
        \label{fig:xgb_bp_sim}
    \end{subfigure}

    \vspace{0.75cm}
    \begin{subfigure}[b]{\textwidth}
        \centering
        \includegraphics[width=0.6\linewidth]{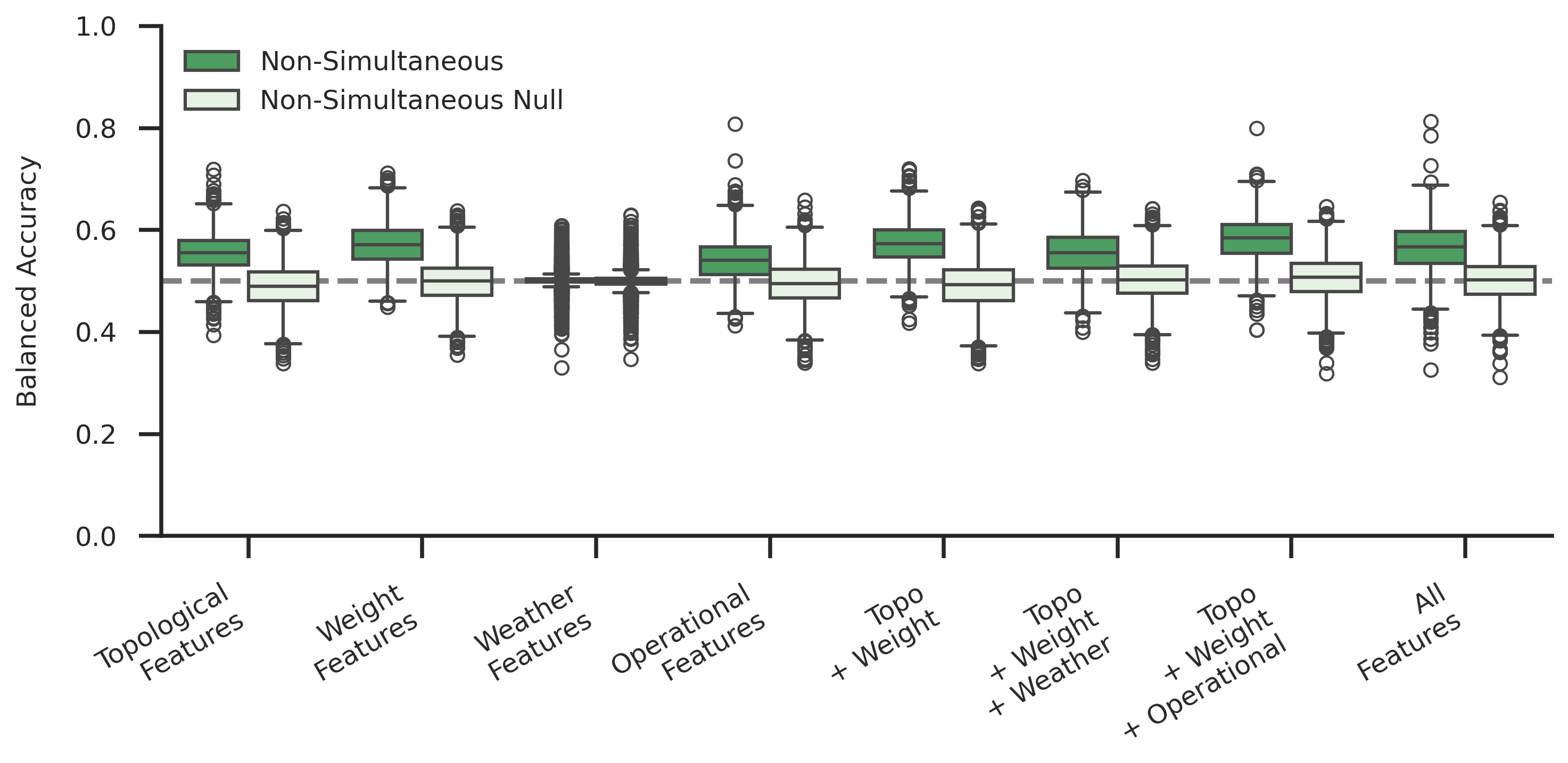}
        \caption{Figure 3: Box plot of balanced accuracy across all feature sets for Non-Simultaneous testing (XGBoost, Trajectory-based split, Global Aggregation).}
        \label{fig:xgb_bp_nonsim}
    \end{subfigure}
    \caption{Comparative analysis of balanced accuracy distributions across all evaluated feature sets under Non-Simultaneous and Simultaneous testing conditions.}
    \label{fig:xgb_bp_combined_grid}
\end{figure}

\clearpage
\section{Alternative Delay Thresholds: Absolute and Relative}
\label{sec:apx:alternative_thresholds}

\begin{figure}[htbp]
    \centering
    
    \begin{subfigure}[b]{0.48\textwidth}
        \centering
        \includegraphics[width=\textwidth]{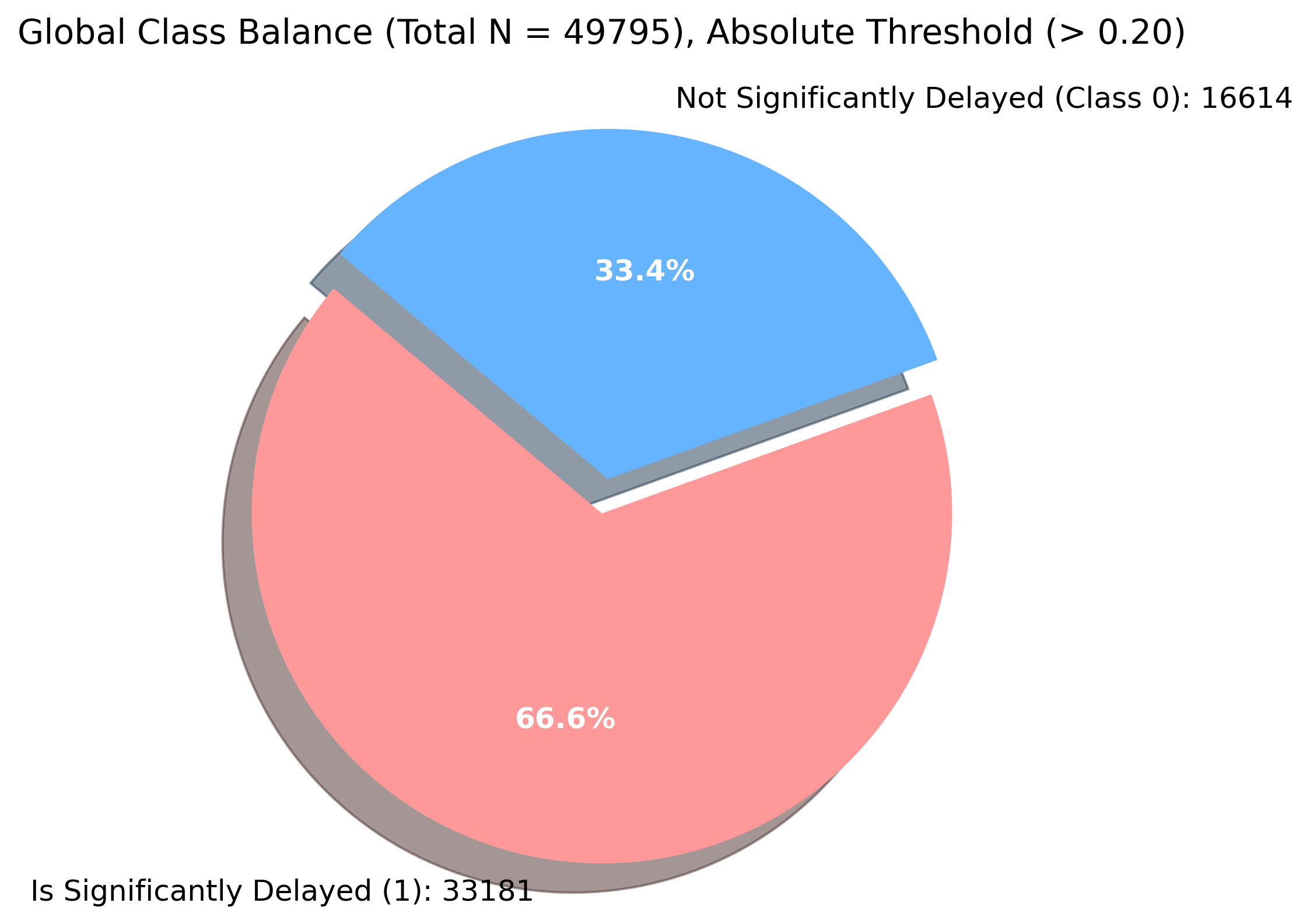}
        \caption{Global Class Balance Absolute Threshold}
        \label{fig:cb_pie_abs}
    \end{subfigure}
    \hfill
    \begin{subfigure}[b]{0.48\textwidth}
        \centering
        \includegraphics[width=\textwidth]{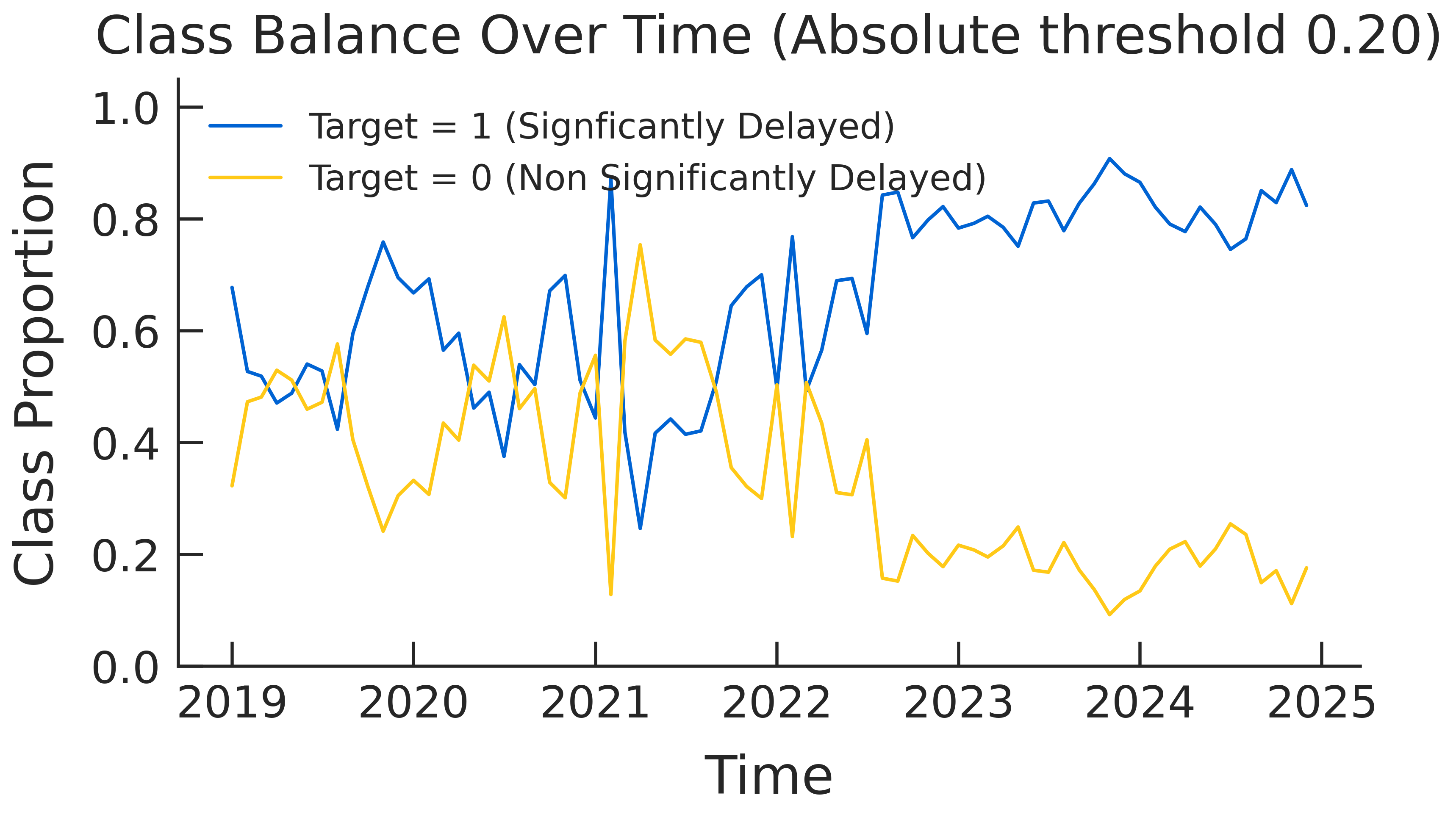}
        \caption{Class Balance Over Time Absolute Threshold}
        \label{fig:cb_ot_abs}
    \end{subfigure}
    
    \vspace{0.5cm} 
    
    \begin{subfigure}[b]{0.48\textwidth}
        \centering
        \includegraphics[width=\textwidth]{media/results/results_XGB/Boxplots_balanced_acc/Base_XGB_BP_Sim_ABS.png}
        \caption{XGBoost Balanced Accuracy Boxplots Simultaneous Testing Absolute Threshold}
        \label{fig:bp_sim_abs}
    \end{subfigure}
    \hfill
    \begin{subfigure}[b]{0.48\textwidth}
        \centering
        \includegraphics[width=\textwidth]{media/results/results_XGB/Boxplots_balanced_acc/Base_XGB_BP_NonSim_ABS.png}
        \caption{XGBoost Balanced Accuracy Boxplots Non-Simultaneous Testing Absolute Threshold}
        \label{fig:bp_nonsim_abs}
    \end{subfigure}
    
    \caption{Overview of class balance and XGBoost balanced accuracy performance utilizing the absolute delay threshold.}
    \label{fig:combined_abs_thresholds}
\end{figure}

\begin{figure}[htbp]
    \centering
    
    \begin{subfigure}[b]{0.48\textwidth}
        \centering
        \includegraphics[width=\textwidth]{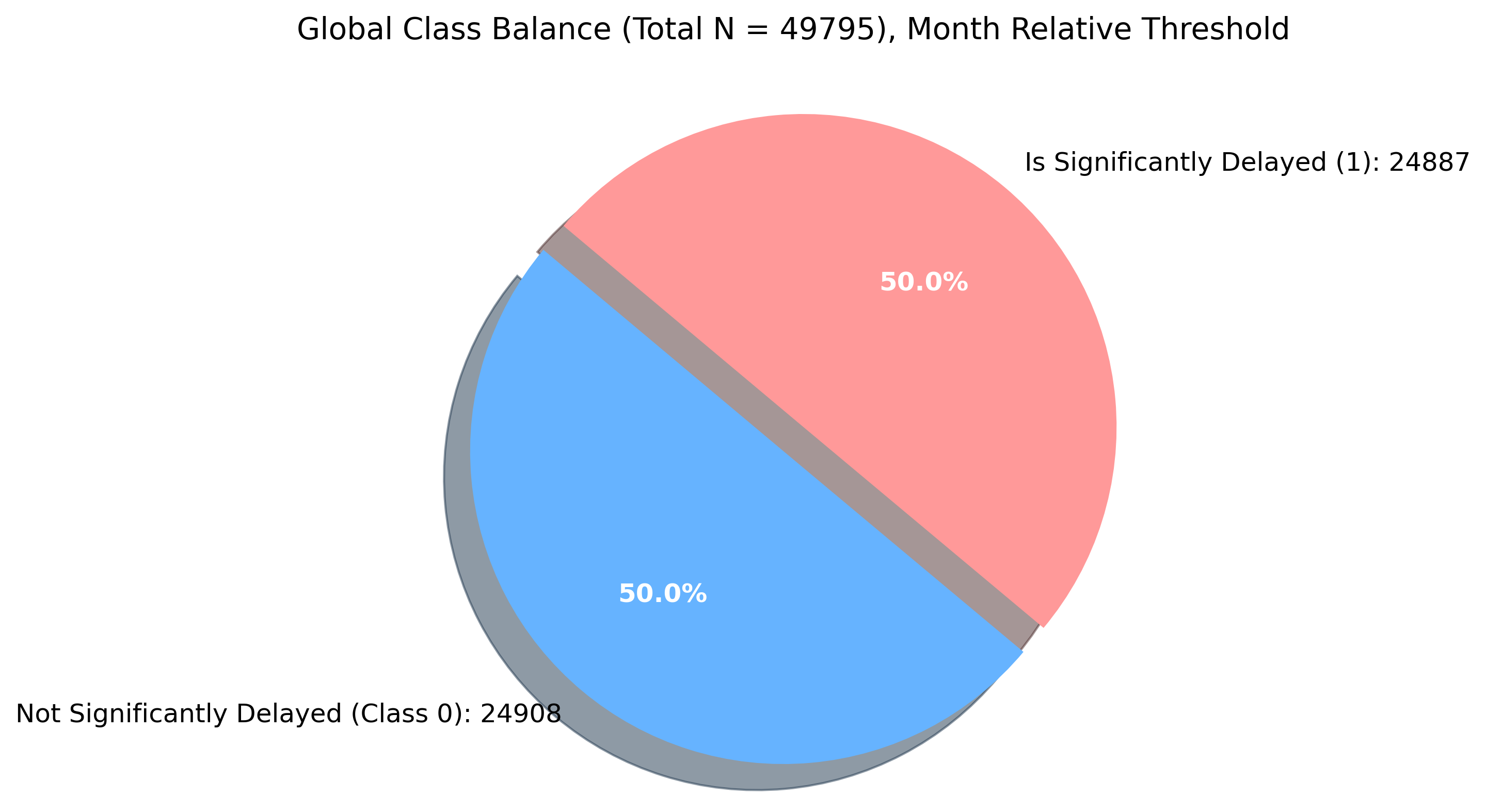}
        \caption{Global Class Balance Month Relative Threshold}
        \label{fig:cb_pie_rel}
    \end{subfigure}
    \hfill
    \begin{subfigure}[b]{0.48\textwidth}
        \centering
        \includegraphics[width=\textwidth]{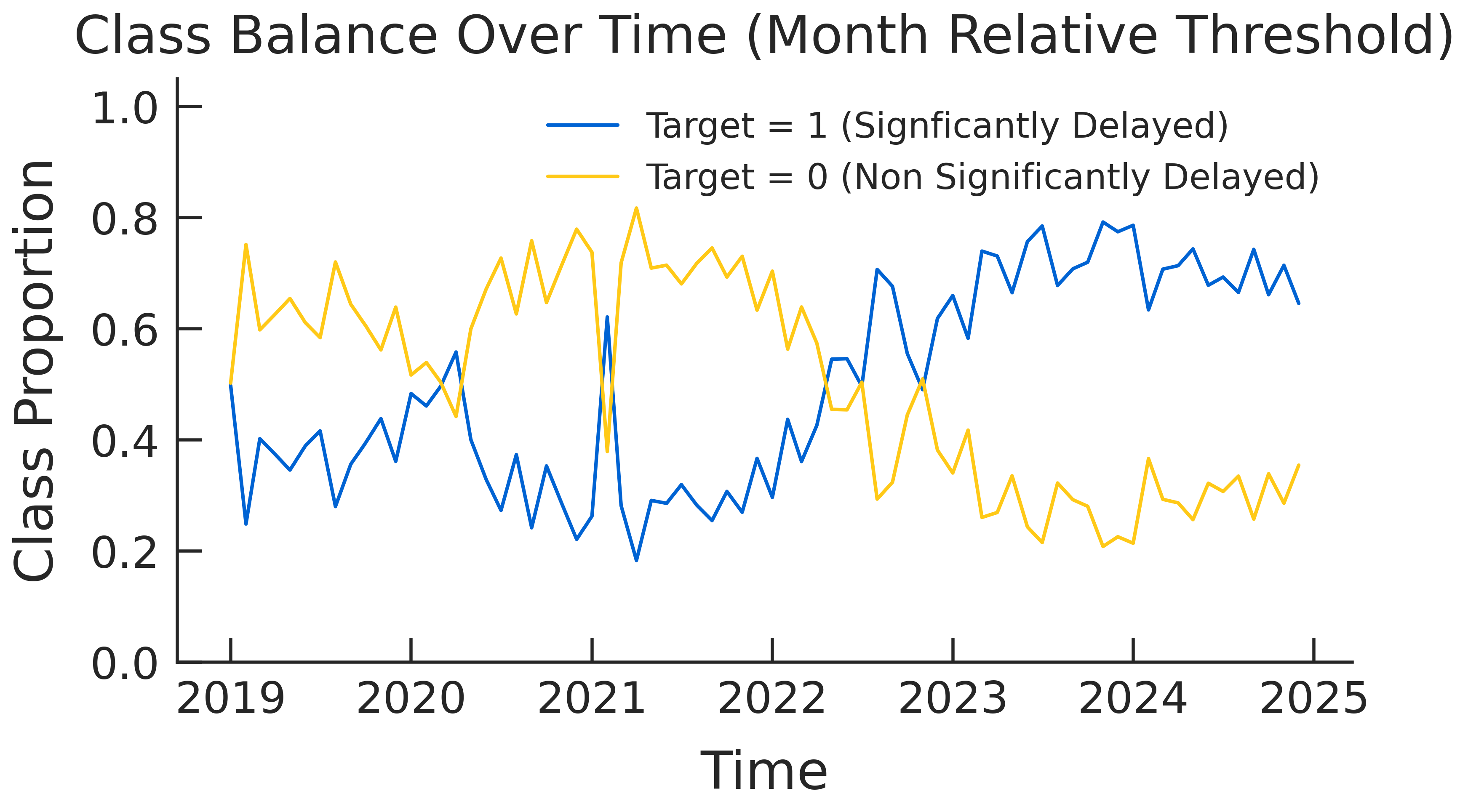}
        \caption{Class Balance Over Time Month Relative Threshold}
        \label{fig:cb_ot_rel}
    \end{subfigure}
    
    \vspace{0.5cm} 
    
    \begin{subfigure}[b]{0.48\textwidth}
        \centering
        \includegraphics[width=\textwidth]{media/results/results_XGB/Boxplots_balanced_acc/Base_XGB_BP_Sim_REL.png}
        \caption{XGBoost Balanced Accuracy Boxplots Simultaneous Testing Month Relative Threshold}
        \label{fig:bp_sim_rel}
    \end{subfigure}
    \hfill
    \begin{subfigure}[b]{0.48\textwidth}
        \centering
        \includegraphics[width=\textwidth]{media/results/results_XGB/Boxplots_balanced_acc/Base_XGB_BP_NonSim_REL.png}
        \caption{XGBoost Balanced Accuracy Boxplots Non-Simultaneous Testing Month Relative Threshold}
        \label{fig:bp_nonsim_rel}
    \end{subfigure}
    
    \caption{Overview of class balance and XGBoost balanced accuracy performance utilizing the month relative delay threshold.}
    \label{fig:combined_rel_thresholds}
\end{figure}

\clearpage
\section{Supplementary Operational Metrics}
\label{sec:apx:operational_metrics}

\begin{figure}[htbp]
    \centering
    \includegraphics[width=0.5\linewidth]{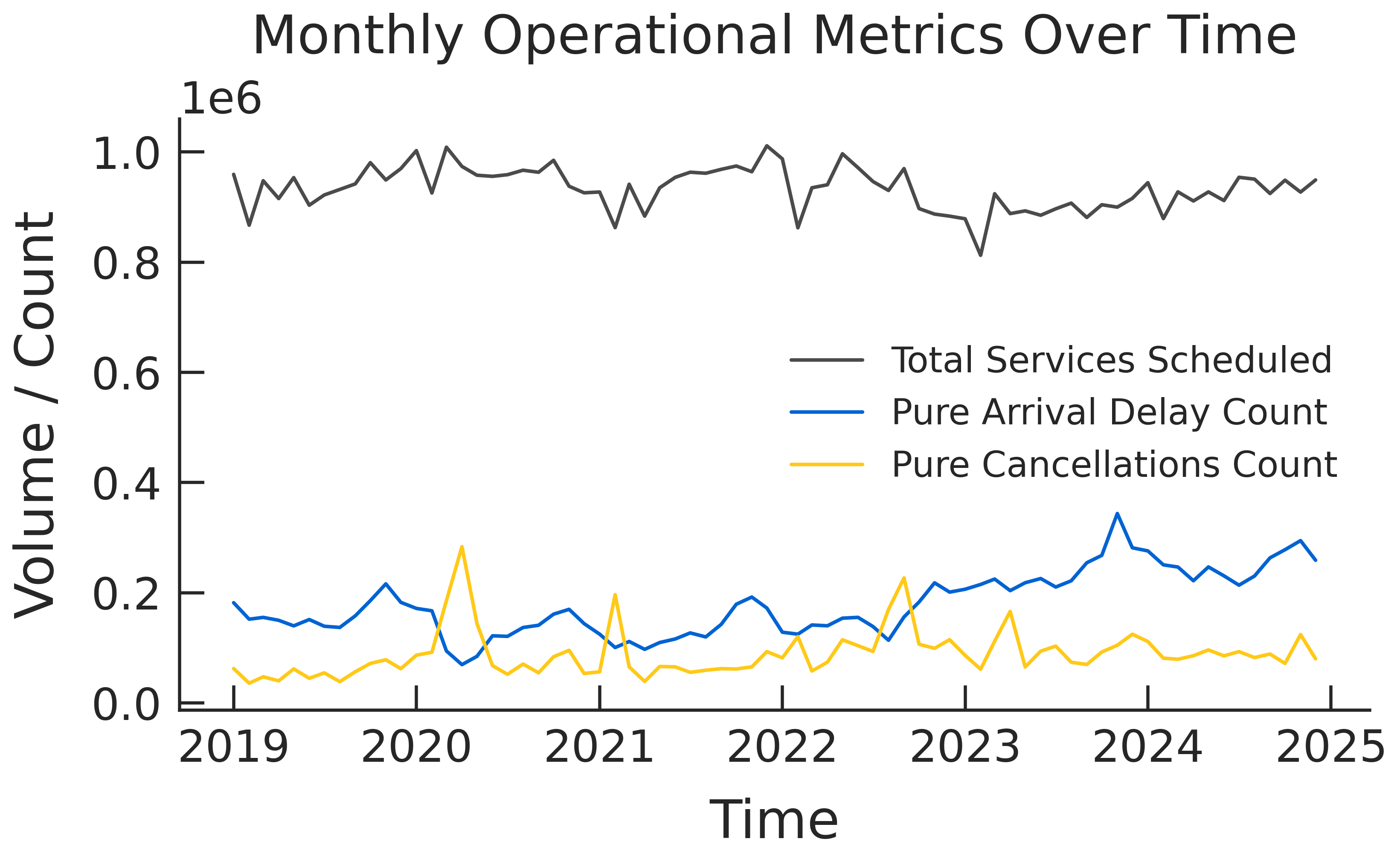}
    \caption{Time-Series of Monthly Operational Metrics: Pure Delays and Cancellations Compared Against Total Scheduled Service Volume.}
    \label{fig:pure_eda}
\end{figure}

\clearpage
\section{Feature Engineering and Hyperparameter Configurations}
\label{sec:apx:feature_hyperparams}

\begin{table}[htbp]
    \centering
    \caption{Overview of engineered feature sets utilized for trajectory delay prediction and their corresponding measurement units. Prefixes \textit{[src/tgt]} denote source or target station calculations.}
    \label{tab:feature_sets}
    \begin{tabular}{lc}
        \toprule
        \textbf{Feature Name} & \textbf{Measured In} \\
        \midrule
        
        \multicolumn{2}{l}{\textit{Topological Features (Unweighted)}} \\
        \texttt{[src/tgt] degree} & Count \\
        \texttt{[src/tgt] avg distance in} & Kilometers \\
        \texttt{[src/tgt] avg distance out} & Kilometers \\
        \texttt{[src/tgt] avg distance total} & Kilometers \\
        \texttt{common neighbors} & Count \\
        \texttt{jaccard coefficient} & Ratio (0--1) \\
        \texttt{preferential attachment} & Score \\
        \texttt{adamic adar index} & Score \\
        \texttt{resource allocation index} & Score \\
        
        \midrule
        \multicolumn{2}{l}{\textit{Weighted Topology Features}} \\
        \texttt{decayed edge weight} & Weight \\
        \texttt{[src/tgt] weighted degree inbound} & Weight \\
        \texttt{[src/tgt] weighted degree outbound} & Weight \\
        \texttt{[src/tgt] weighted degree total} & Weight \\
        
        \midrule
        \multicolumn{2}{l}{\textit{Environmental (Weather) Features}} \\
        \texttt{[src/tgt] mean temperature 2m} & $^\circ$C \\
        \texttt{[src/tgt] sum rain} & Millimeters \\
        \texttt{[src/tgt] sum snowfall} & Centimeters \\
        \texttt{[src/tgt] mean snow depth} & Meters \\
        \texttt{[src/tgt] mean wind speed 10m} & km/h \\
        \texttt{[src/tgt] max wind gusts 10m} & km/h \\
        \texttt{[src/tgt] avg soil temperature} & $^\circ$C \\

        \midrule
        \multicolumn{2}{l}{\textit{Operational Features}} \\
        \texttt{distance} & Kilometers \\
        \texttt{last active volume} & Count \\
        \texttt{months since last service} & Month Counts \\
        \texttt{is new or stale route} & Boolean (0/1) \\
        \texttt{ratio intercity} & Ratio (0--1) \\
        \texttt{ratio intercity\_direct} & Ratio (0--1) \\
        \texttt{ratio international} & Ratio (0--1) \\
        \texttt{ratio sprinter} & Ratio (0--1) \\
        \texttt{ratio operational exceptions} & Ratio (0--1) \\
        \bottomrule
    \end{tabular}
\end{table}


\begin{table}[htbp]
    \centering
    \caption{Topological Feature Definitions and Mathematical Formulations. The monthly railway network is formally defined as a directed graph $G_t = (V_t, E_t)$, where the nodes $V_t$ represent individual train stations and the directed edges $E_t$ denote active train trajectories. For any given node $v \in V_t$, $deg(v)$ and $wdeg(v)$ denote the standard degree and weighted degree, respectively. The neighborhood set is defined as $N(v)$, and the physical spatial distance between stations is denoted by $d(u, v)$. Within this framework, the physical distance between stations was utilized as the edge distance, while the decayed operational edge weight ($W_{\text{decayed}}$) previously defined in Section \ref{subsubsec:operational_features} was formally assigned as the structural edge weight $w(u,v)$ to map historical service density directly onto the network topology.}
    \label{tab:topological_features}
    \begin{tabular}{llp{9cm}}
        \toprule
        \textbf{Feature} & \textbf{Type} & \textbf{Formula / Description} \\
        \midrule
        Node Degree & Node-level & $deg(v) = |\{u \in V : (v, u) \in E\}|$ \\
        Out Weighted Degree & Node-level & $wdeg_{out}(v) = \sum_{u \in N(v)} w(v, u)$, where $w(v, u)$ is the edge weight. \\
        In Weighted Degree & Node-level & $wdeg_{in}(v) = \sum_{u \in N(v)} w(u, v)$, where $w(u, v)$ is the edge weight. \\
        Total Weighted Degree & Node-level & $wdeg_{total}(v) = wdeg_{out}(v) + wdeg_{in}(v)$ \\
        Out Average Distance & Node-level & $\frac{1}{|N(v)|} \sum_{u \in N(v)} d(v, u)$ \\
        In Average Distance & Node-level & $\frac{1}{|N(v)|} \sum_{u \in N(v)} d(u, v)$ \\
        Total Average Distance & Node-level & Sum of in and out average distances. \\
        Common Neighbours & Trajectory-level & $|\Gamma(u) \cap \Gamma(v)|$ \\
        Jaccard Coefficient & Trajectory-level & $\frac{|\Gamma(u) \cap \Gamma(v)|}{|\Gamma(u) \cup \Gamma(v)|}$ \\
        Preferential Attachment & Trajectory-level & $|\Gamma(u)| \cdot |\Gamma(v)|$ \\
        Adamic-Adar Index & Trajectory-level & $\sum_{w \in \Gamma(u) \cap \Gamma(v)} \frac{1}{\log |\Gamma(w)|}$ \\
        Resource Allocation Index & Trajectory-level & $\sum_{w \in \Gamma(u) \cap \Gamma(v)} \frac{1}{|\Gamma(w)|}$ \\
        \bottomrule
    \end{tabular}
\end{table}

\begin{table}[htbp]
\centering
\caption{Utilized Default Hyperparameter Configurations by Classifier}
\label{tab:classifier_hyperparameters}
\begin{tabular}{l p{11.5cm}}
\toprule
\textbf{Classifier} & \textbf{Hyperparameters} \\
\midrule
LogisticRegression (scikit-learn 1.8.0) & penalty: \{'l2'\}, C: \{1.0\}, l1\_ratio: \{0.0\}, fit\_intercept: \{True\}, intercept\_scaling: \{1\}, solver: \{'lbfgs'\}, max\_iter: \{1000\}, tol: \{0.0001\}, dual: \{False\}, class\_weight: \{None\}, random\_state: \{88\},  warm\_start: \{False\}, n\_jobs: \{None\} \\
RandomForestClassifier (scikit-learn 1.8.0) & n\_estimators: \{100\}, criterion: \{'gini'\}, max\_depth: \{None\}, min\_samples\_split: \{2\}, min\_samples\_leaf: \{1\}, min\_weight\_fraction\_leaf: \{0.0\}, max\_features: \{'sqrt'\}, max\_leaf\_nodes: \{None\}, min\_impurity\_decrease: \{0.0\}, bootstrap: \{True\}, oob\_score: \{False\}, n\_jobs: \{None\}, random\_state: \{88\}, class\_weight: \{None\}, ccp\_alpha: \{0.0\}, max\_samples: \{None\}, monotonic\_cst: \{None\} \\
XGBClassifier (xgboost 3.2.0) & objective: \{'binary:logistic'\}, n\_estimators: \{100\}, learning\_rate: \{0.3\}, max\_depth: \{6\}, booster: \{'gbtree'\}, tree\_method: \{'auto'\}, gamma: \{0\}, min\_child\_weight: \{1\}, max\_delta\_step: \{0\}, subsample: \{1\}, colsample\_bytree: \{1\}, colsample\_bylevel: \{1\}, colsample\_bynode: \{1\}, reg\_alpha: \{0\}, reg\_lambda: \{1\}, scale\_pos\_weight: \{1\}, base\_score: \{auto\}, random\_state: \{88\} \\
LGBMClassifier (lightgbm 4.6.0) & objective: \{'binary'\}, boosting\_type: \{'gbdt'\}, num\_leaves: \{31\}, max\_depth: \{-1\}, learning\_rate: \{0.1\}, n\_estimators: \{100\}, subsample\_for\_bin: \{200000\}, scale\_pos\_weight: \{1\}, min\_split\_gain: \{0.0\}, min\_child\_weight: \{0.001\}, min\_child\_samples: \{20\}, subsample: \{1.0\}, subsample\_freq: \{0\}, colsample\_bytree: \{1.0\}, reg\_alpha: \{0.0\}, reg\_lambda: \{0.0\}, random\_state: \{88\} \\
\bottomrule
\end{tabular}
\end{table}

\clearpage
\section{Year-over-Year (YoY) Temporal Feature Importance}
\label{sec:apx:yoy_feature_importance}


\begin{figure}[htbp]
    \centering
    \begin{subfigure}[b]{0.45\textwidth}
        \centering
        \includegraphics[width=\textwidth]{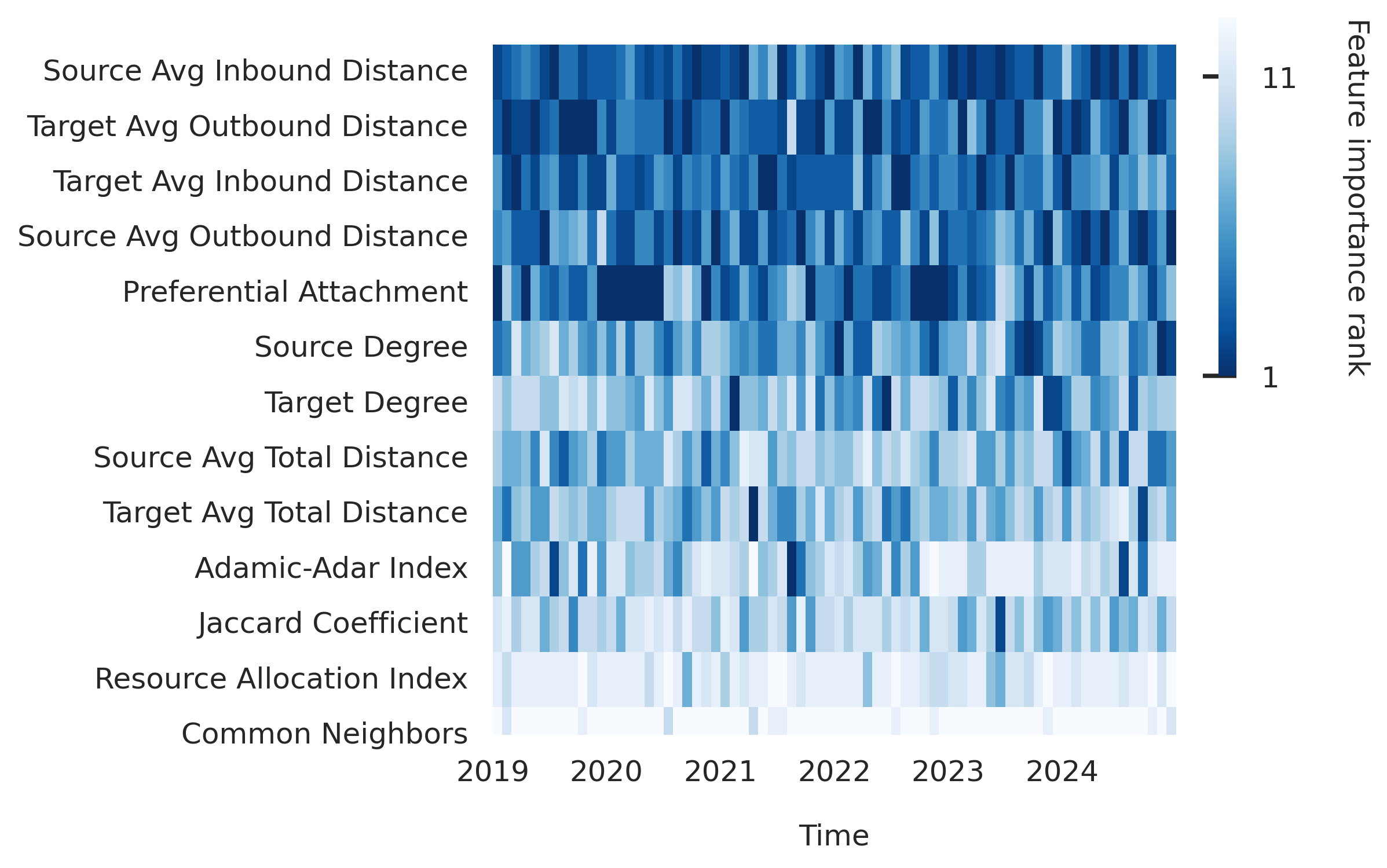}
        \caption{Topology}
        \label{fig:xgb_yoy_top}
    \end{subfigure}
    \hfill
    \begin{subfigure}[b]{0.45\textwidth}
        \centering
        \includegraphics[width=\textwidth]{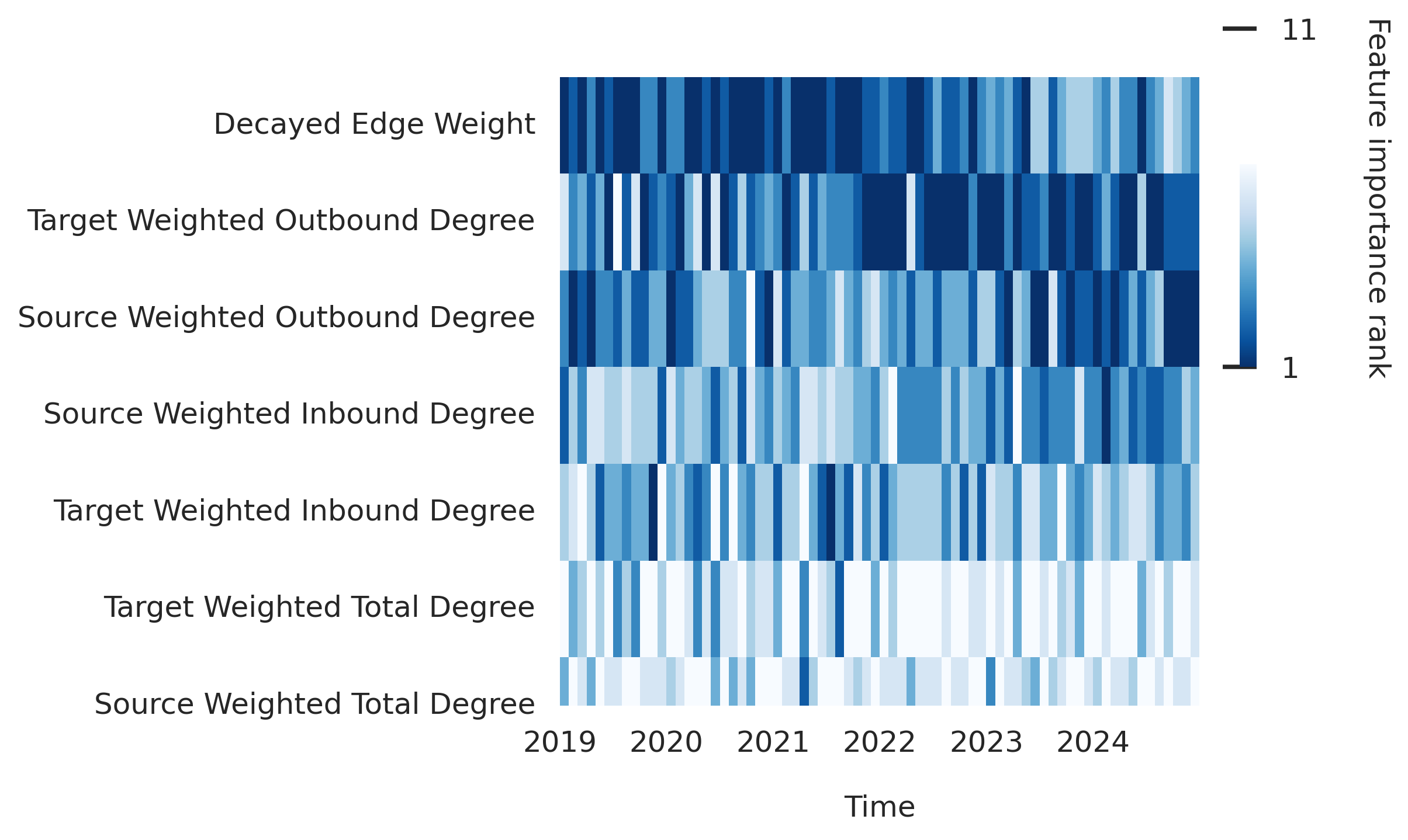}
        \caption{Weight}
        \label{fig:xgb_yoy_wgt}
    \end{subfigure}
    
    \vspace{0.5cm}
    
    \begin{subfigure}[b]{0.45\textwidth}
        \centering
        \includegraphics[width=\textwidth]{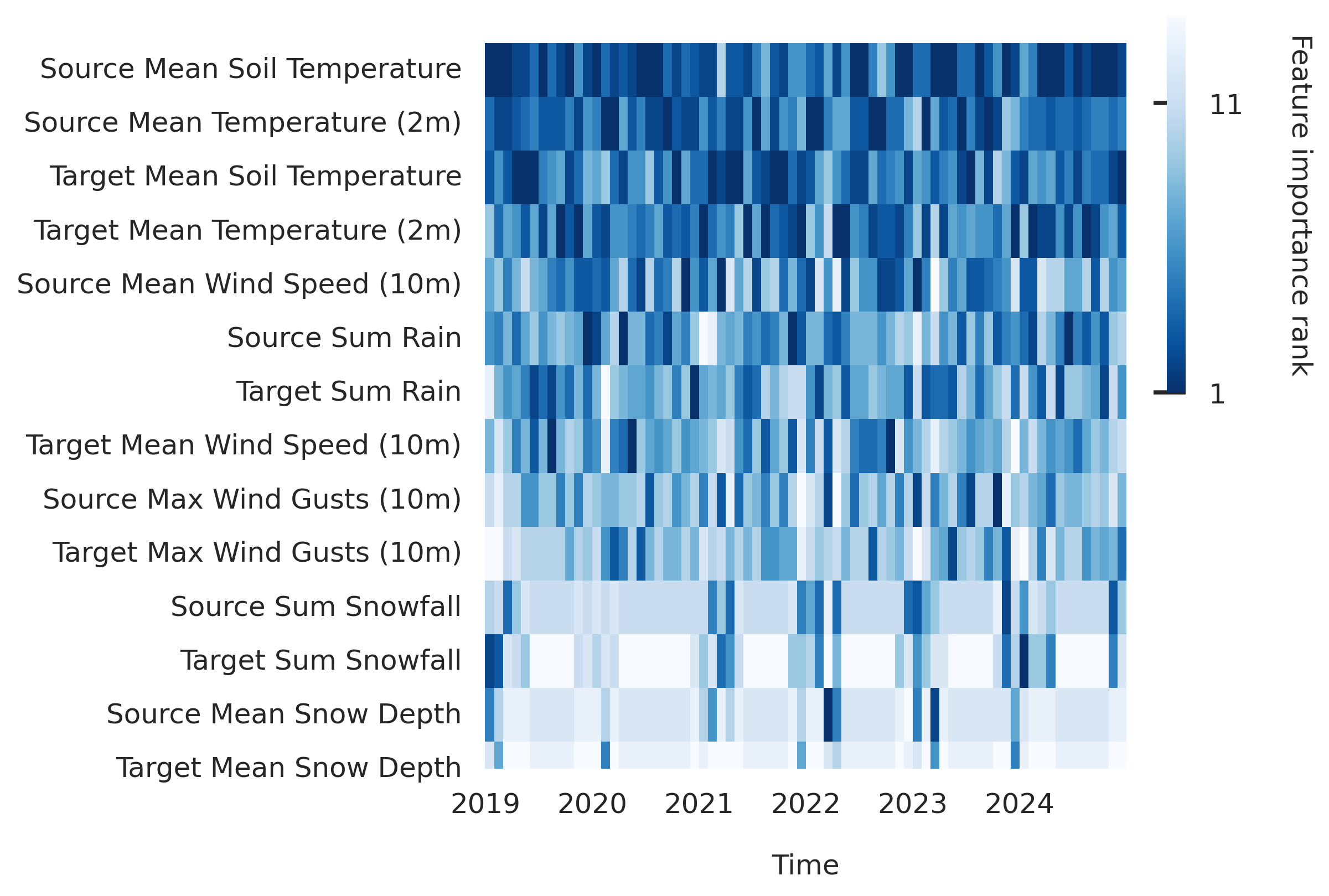}
        \caption{Weather}
        \label{fig:xgb_yoy_wth}
    \end{subfigure}
    \hfill
    \begin{subfigure}[b]{0.45\textwidth}
        \centering
        \includegraphics[width=\textwidth]{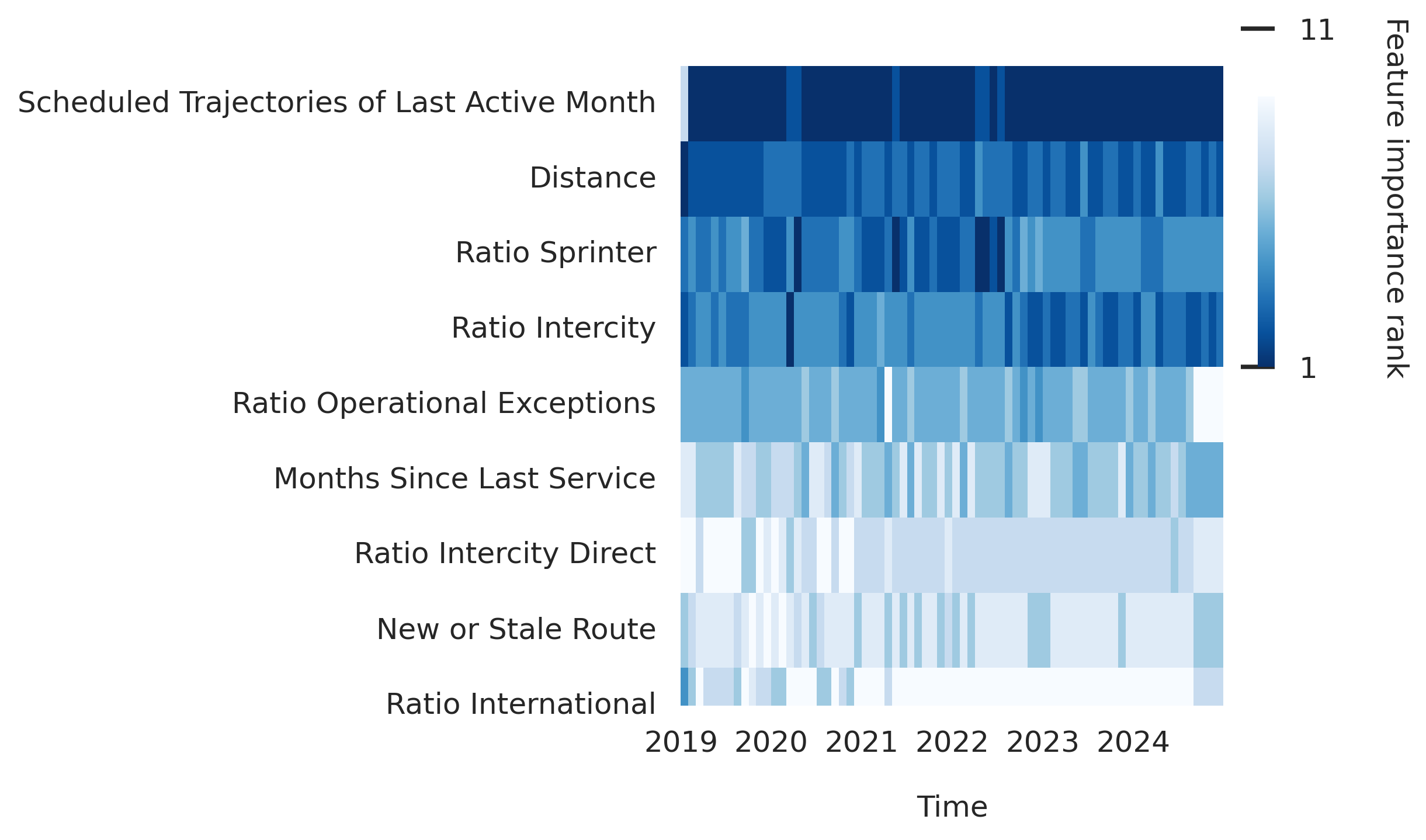}
        \caption{Operational}
        \label{fig:xgb_yoy_ops}
    \end{subfigure}
    \caption{Monthly Year-over-Year temporal feature importance for the XGBoost model evaluating isolated baseline feature sets.}
    \label{fig:xgb_yoy_core_grid}
\end{figure}

\begin{figure}[htbp]
    \centering
    \begin{subfigure}[b]{0.45\textwidth}
        \centering
        \includegraphics[width=\textwidth]{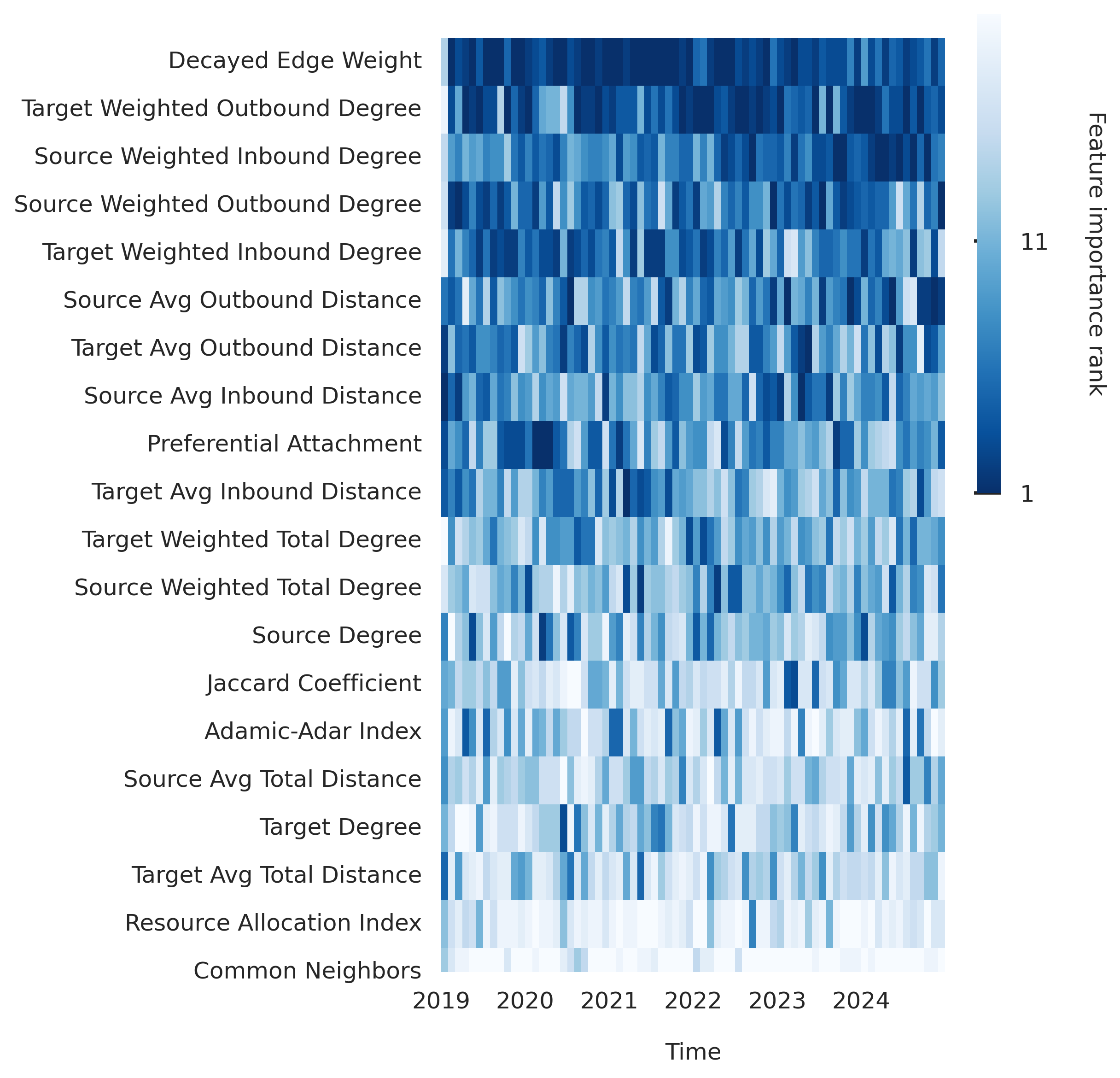}
        \caption{Topology + Weight}
        \label{fig:xgb_yoy_tw}
    \end{subfigure}
    \hfill
    \begin{subfigure}[b]{0.45\textwidth}
        \centering
        \includegraphics[width=\textwidth]{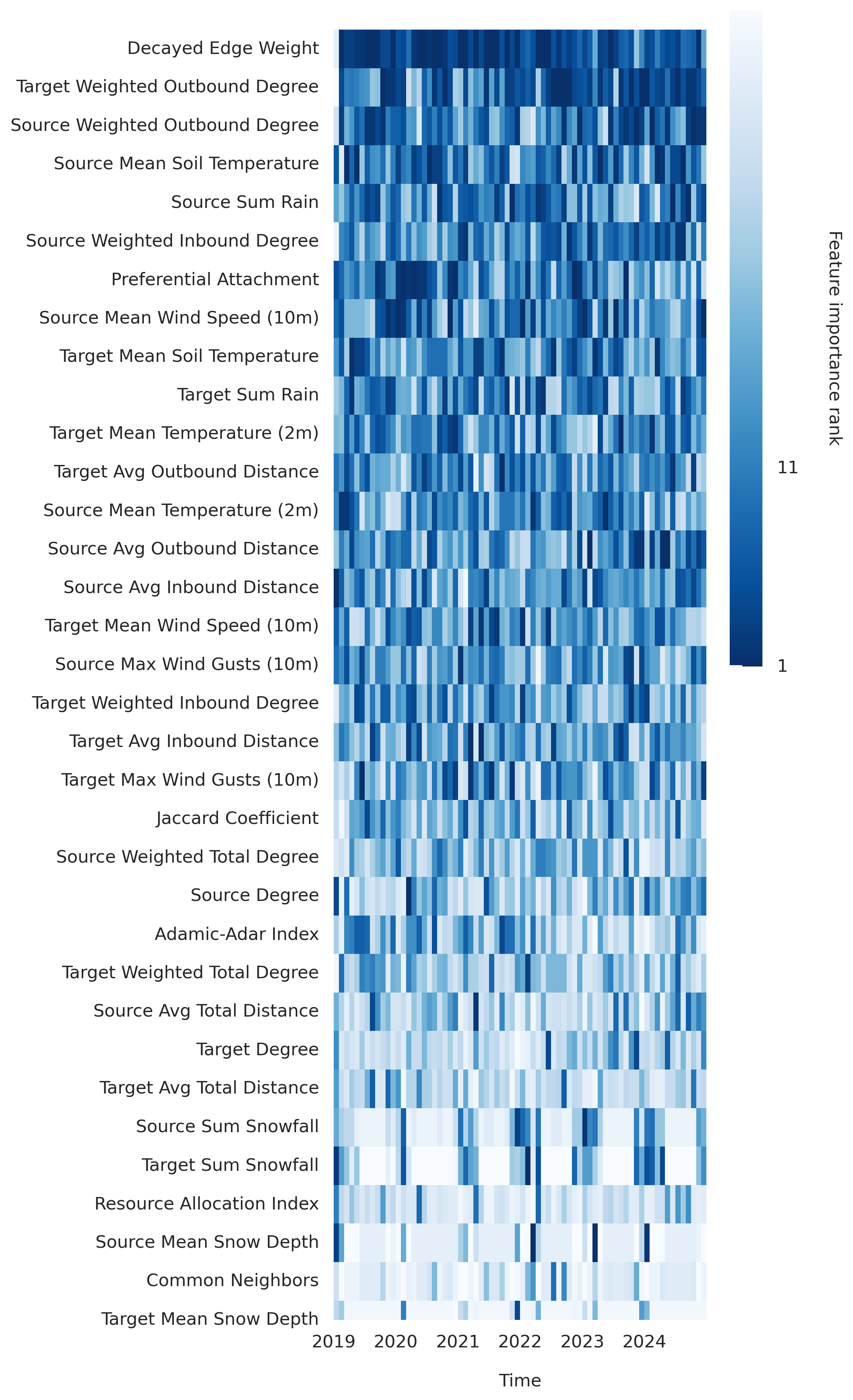}
        \caption{Topology + Weight + Weather}
        \label{fig:xgb_yoy_twcw}
    \end{subfigure}
    \caption{Monthly Year-over-Year temporal feature importance for the XGBoost model evaluating initial combined topological + weight and weather feature sets.}
    \label{fig:xgb_yoy_combined_grid_1}
\end{figure}

\begin{figure}[htbp]
    \centering
    \begin{subfigure}[b]{0.45\textwidth}
        \centering
        \includegraphics[width=\textwidth]{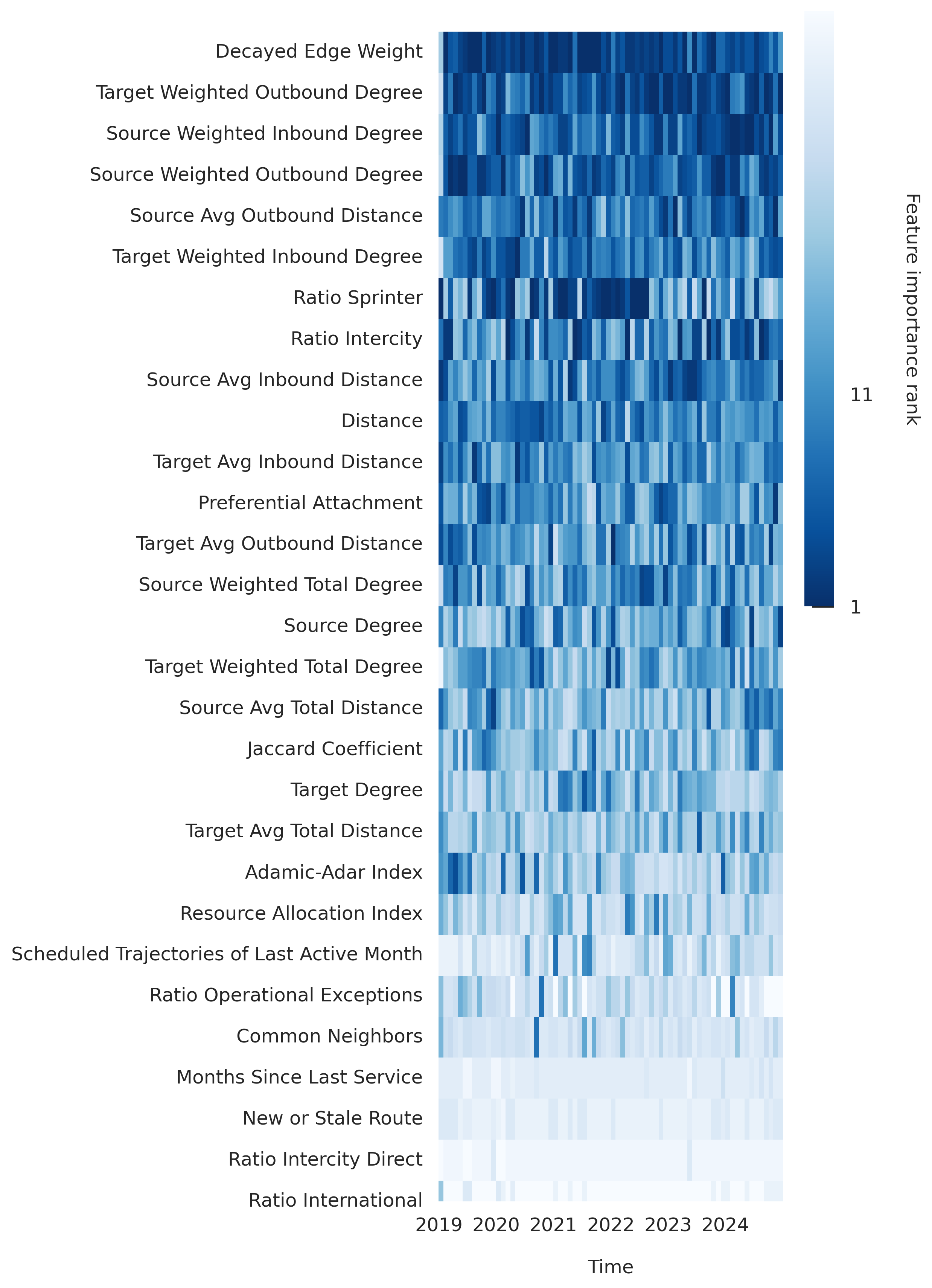}
        \caption{Topology + Weight + Operational}
        \label{fig:xgb_yoy_twco}
    \end{subfigure}
    \hfill
    \begin{subfigure}[b]{0.45\textwidth}
        \centering
        \includegraphics[width=\textwidth]{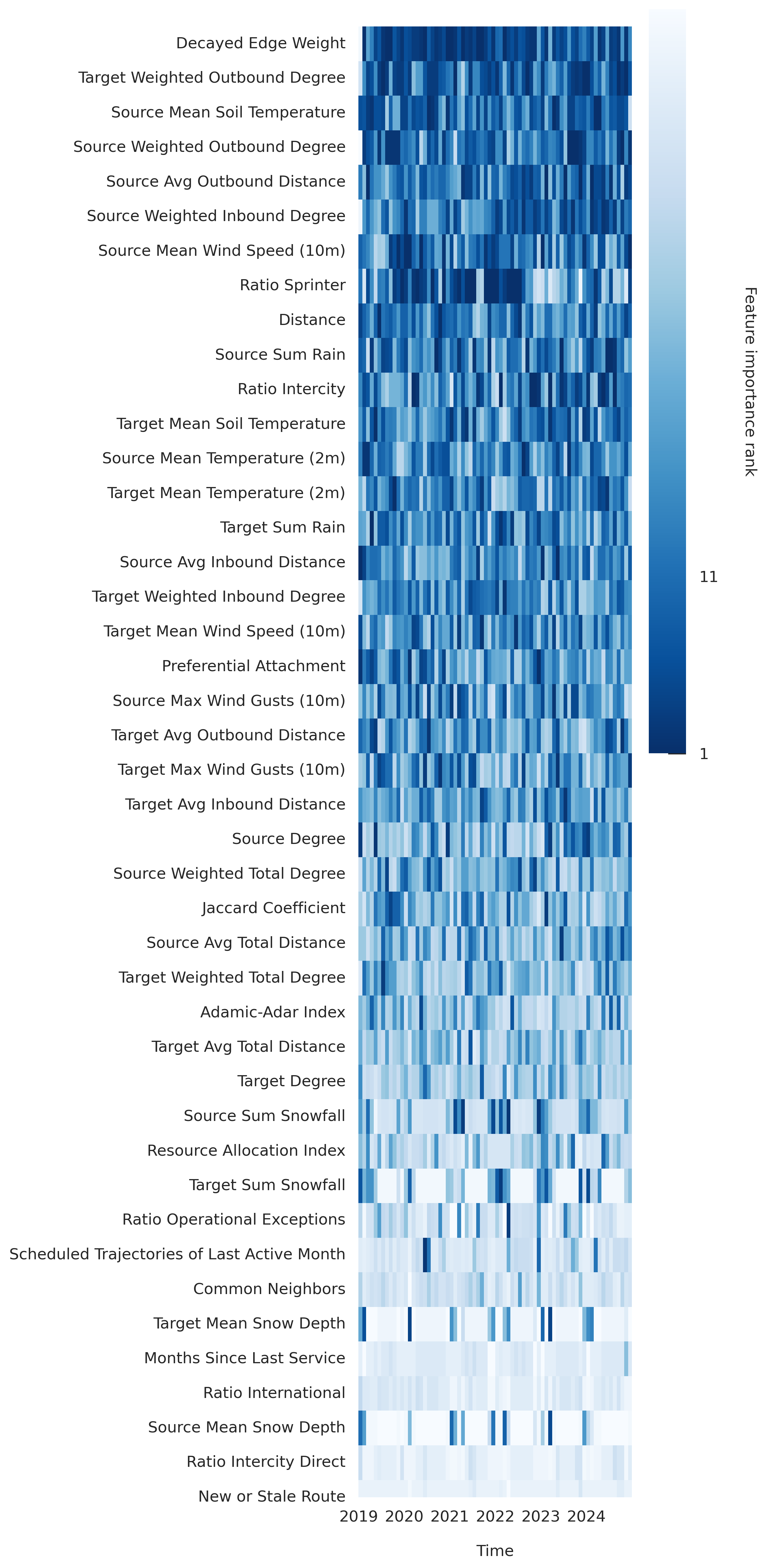}
        \caption{All Features}
        \label{fig:xgb_yoy_allf}
    \end{subfigure}
    \caption{Monthly Year-over-Year temporal feature importance for the XGBoost model evaluating combined topological + weight + operational feature and all feature set.}
    \label{fig:xgb_yoy_combined_grid_2}
\end{figure}
\clearpage
\section{Month-over-Month (MoM) Temporal Feature Importance}
\label{sec:apx:mom_feature_importance}

\begin{figure}[htbp]
    \centering
    \begin{subfigure}[b]{0.45\textwidth}
        \centering
        \includegraphics[width=\textwidth]{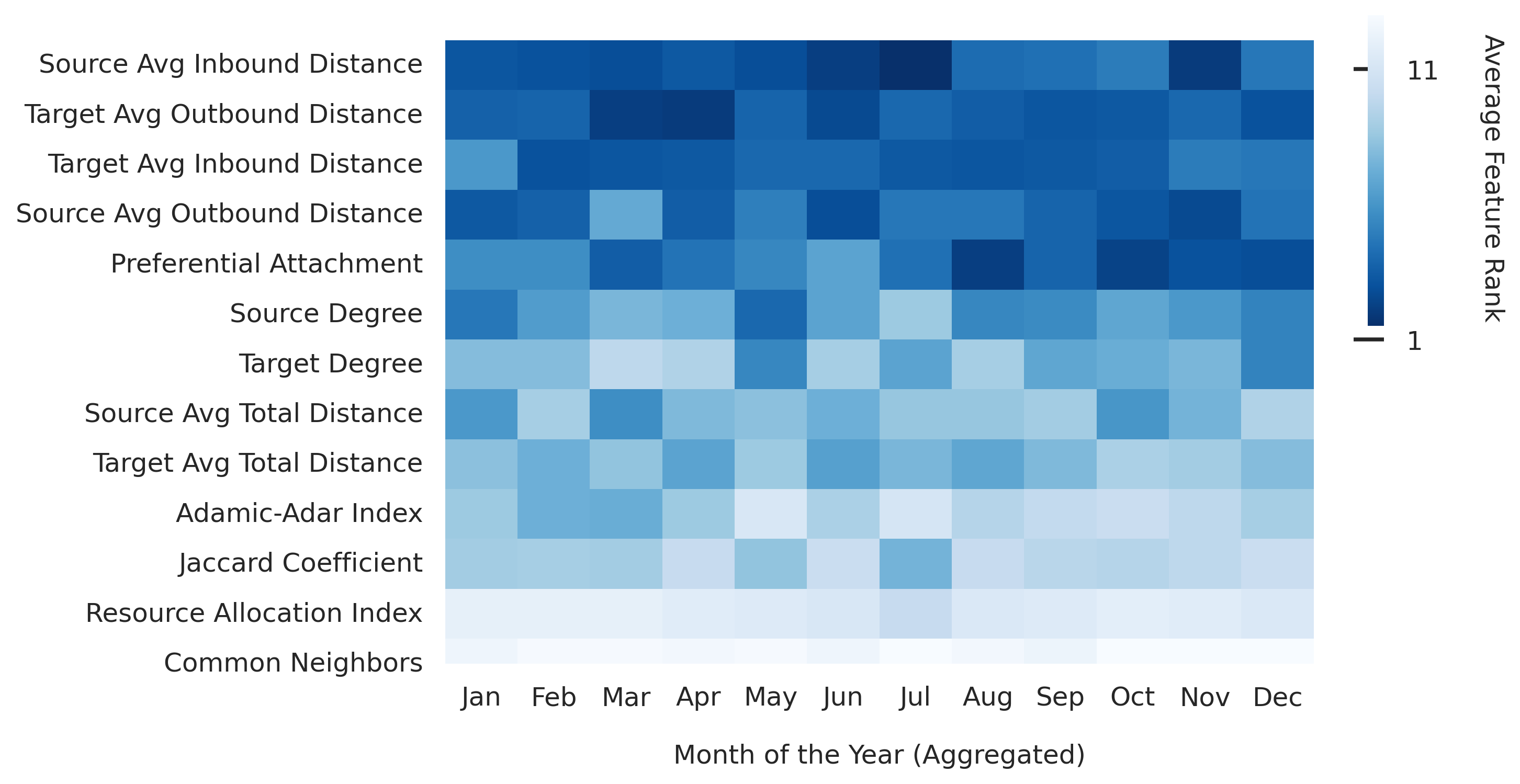}
        \caption{Topology}
        \label{fig:xgb_mom_top}
    \end{subfigure}
    \hfill
    \begin{subfigure}[b]{0.45\textwidth}
        \centering
        \includegraphics[width=\textwidth]{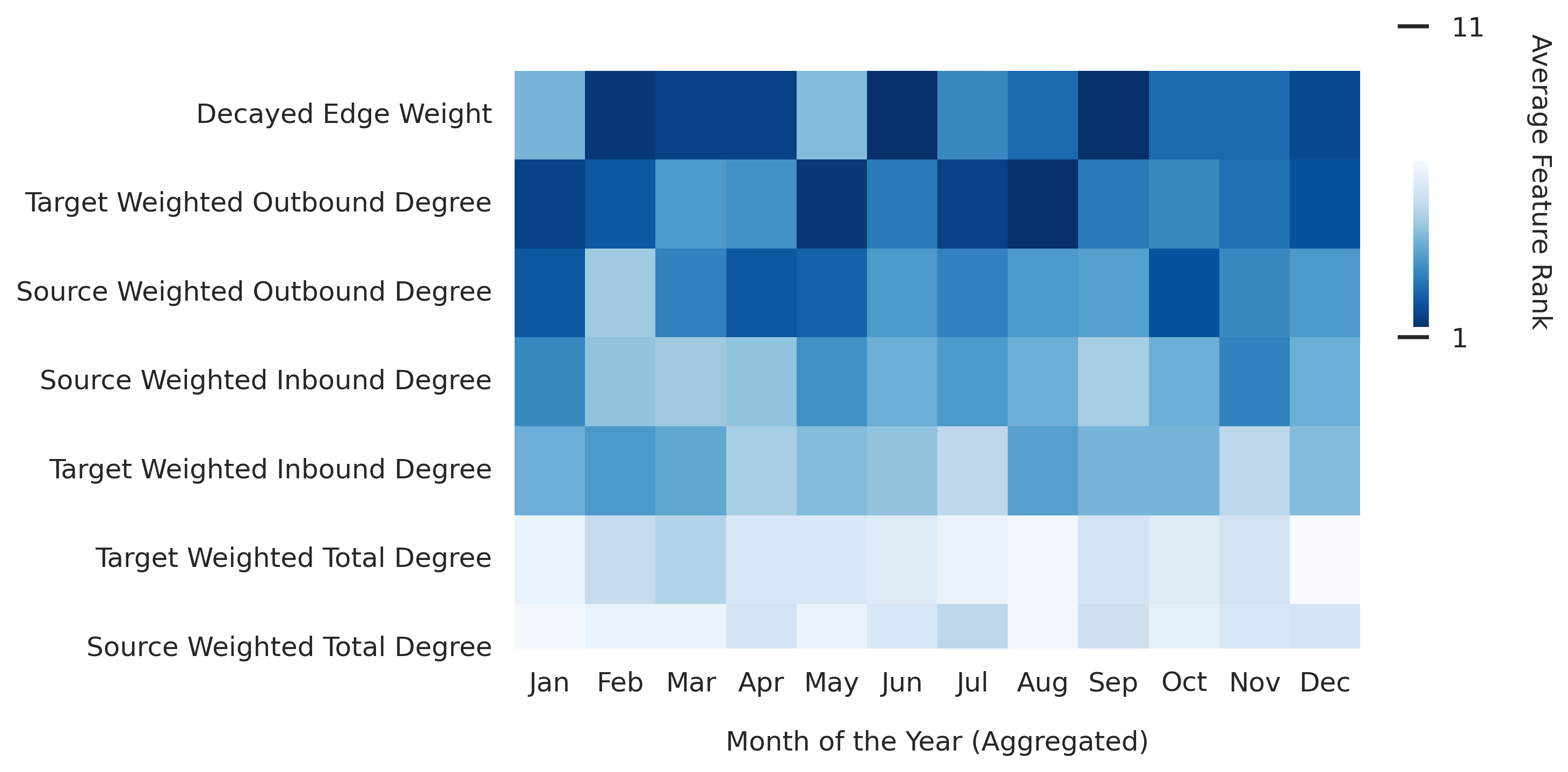}
        \caption{Weight}
        \label{fig:xgb_mom_wgt}
    \end{subfigure}
    
    \vspace{0.5cm}
    
    \begin{subfigure}[b]{0.45\textwidth}
        \centering
        \includegraphics[width=\textwidth]{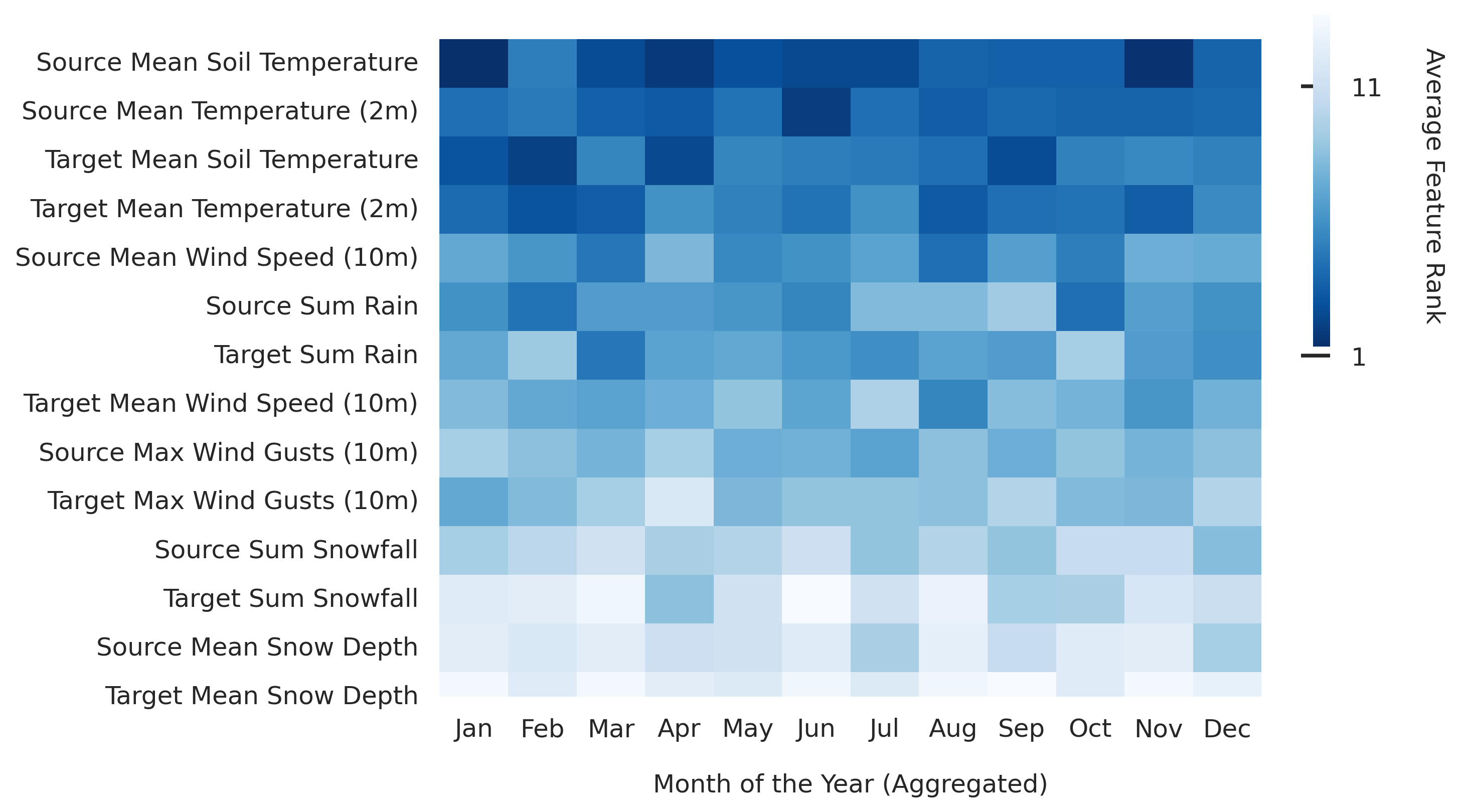}
        \caption{Weather}
        \label{fig:xgb_mom_wth}
    \end{subfigure}
    \hfill
    \begin{subfigure}[b]{0.45\textwidth}
        \centering
        \includegraphics[width=\textwidth]{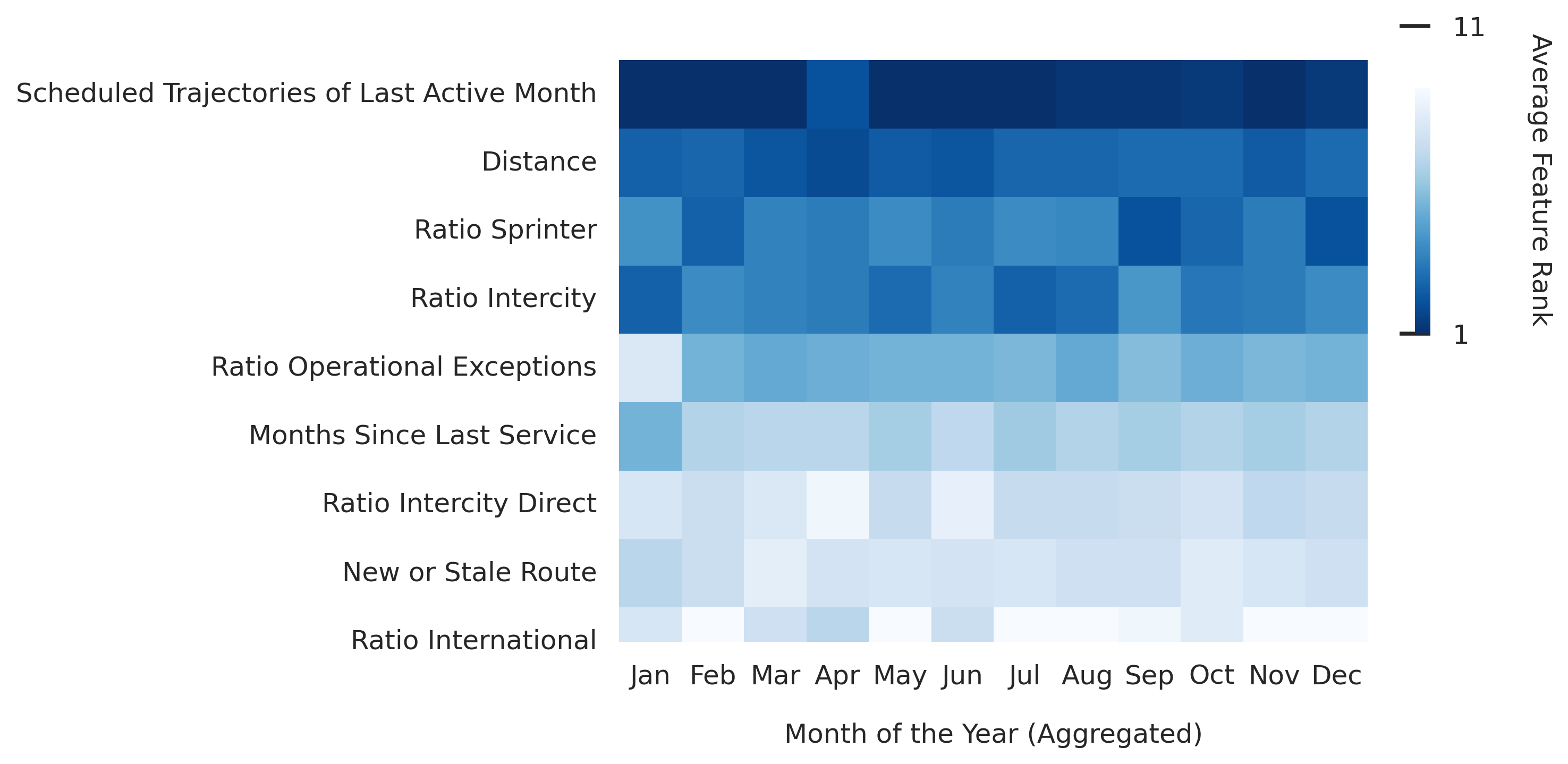}
        \caption{Operational}
        \label{fig:xgb_mom_ops}
    \end{subfigure}
    \caption{Aggregated (2019-2024) Month-over-Month SHAP feature importance for the XGBoost model evaluating isolated baseline feature sets.}
    \label{fig:xgb_mom_core_grid}
\end{figure}

\begin{figure}[htbp]
    \centering
    \begin{subfigure}[b]{0.45\textwidth}
        \centering
        \includegraphics[width=\textwidth]{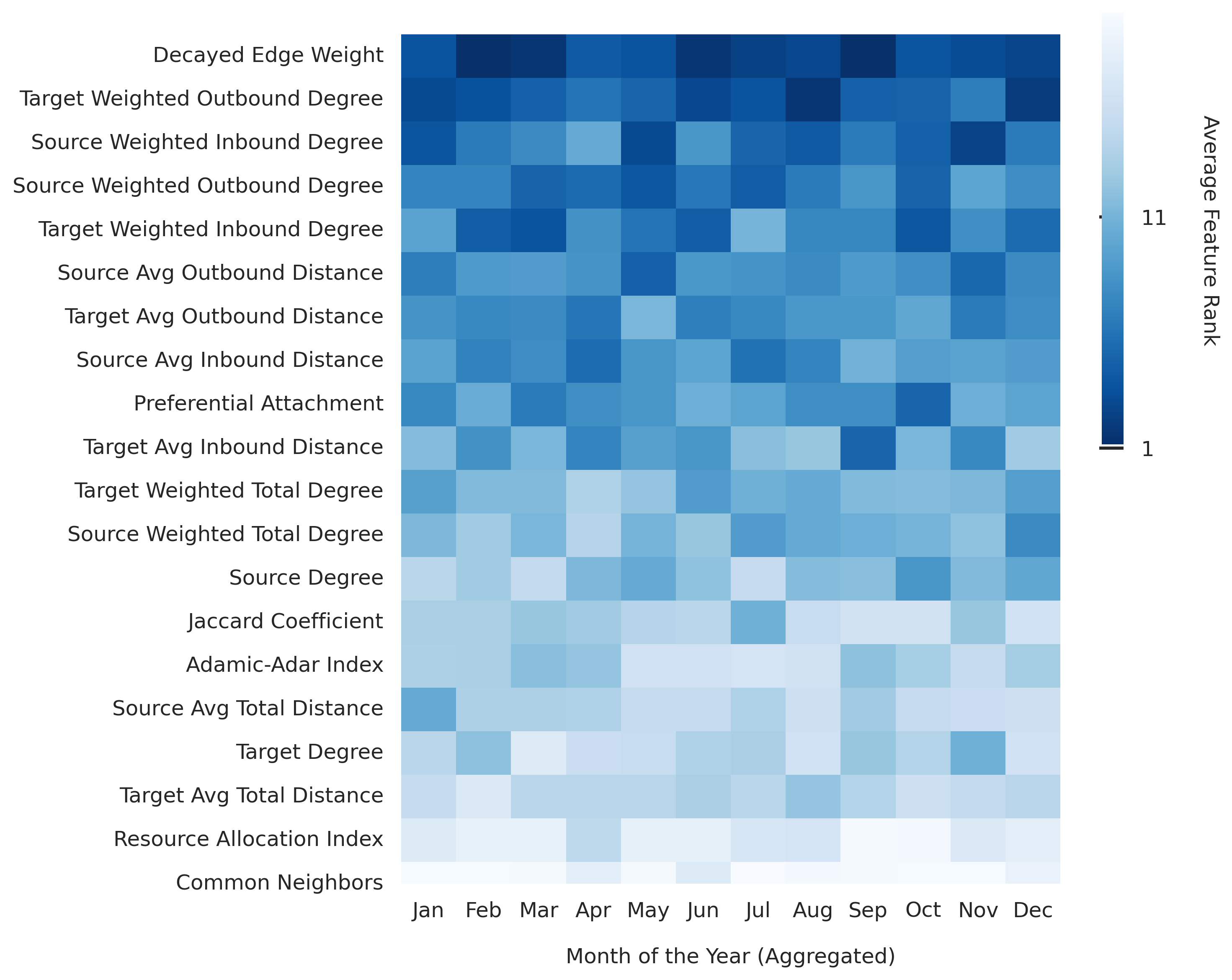}
        \caption{Topology + Weight}
        \label{fig:xgb_mom_tw}
    \end{subfigure}
    \hfill
    \begin{subfigure}[b]{0.45\textwidth}
        \centering
        \includegraphics[width=\textwidth]{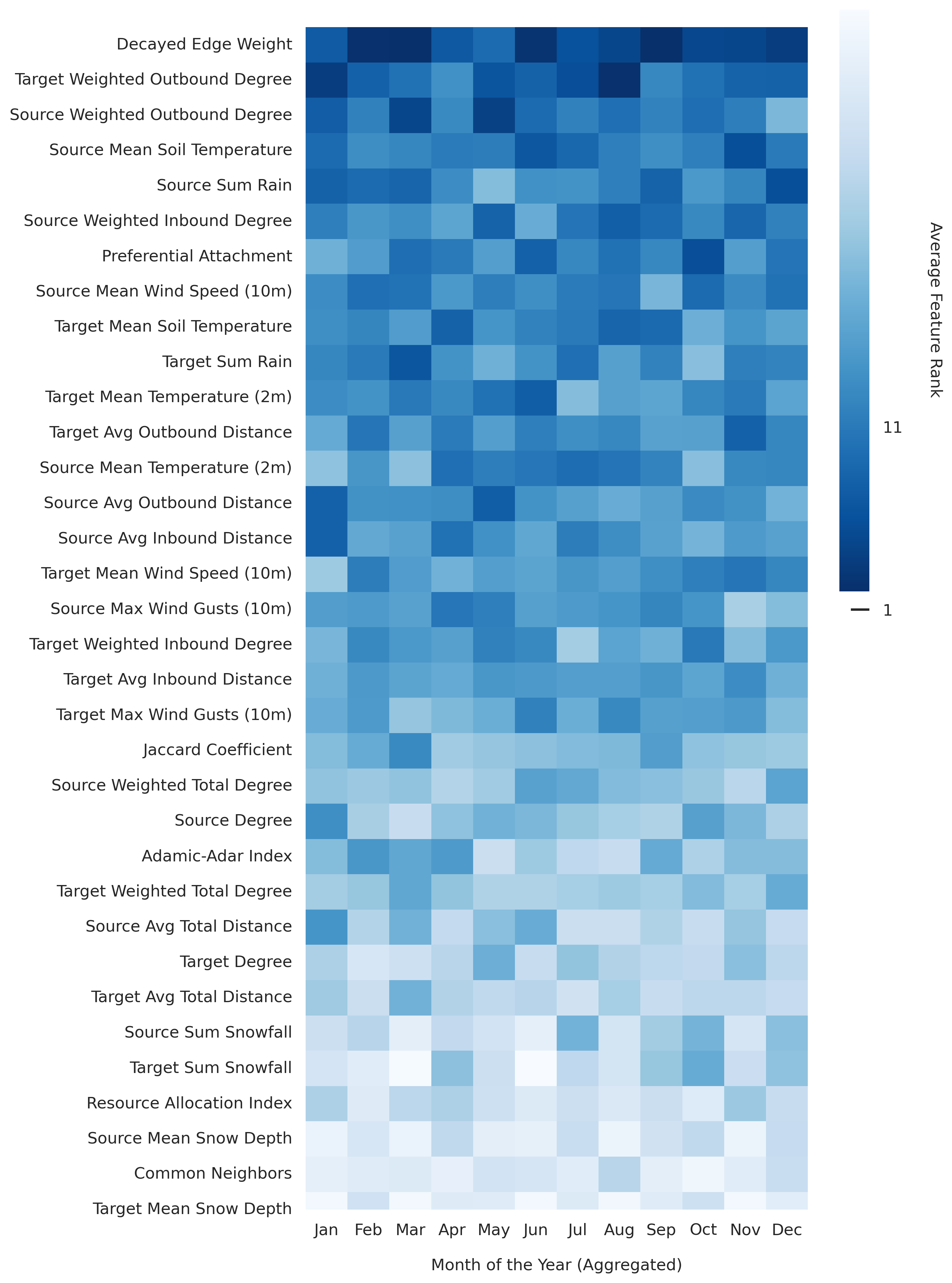}
        \caption{Topology + Weight + Weather}
        \label{fig:xgb_mom_twcw}
    \end{subfigure}
    \caption{Aggregated (2019-2024) Month-over-Month SHAP feature importance for the XGBoost model evaluating initial combined topological + weight and weather feature sets.}
    \label{fig:xgb_mom_combined_grid_1}
\end{figure}

\begin{figure}[htbp]
    \centering
    \begin{subfigure}[b]{0.45\textwidth}
        \centering
        \includegraphics[width=\textwidth]{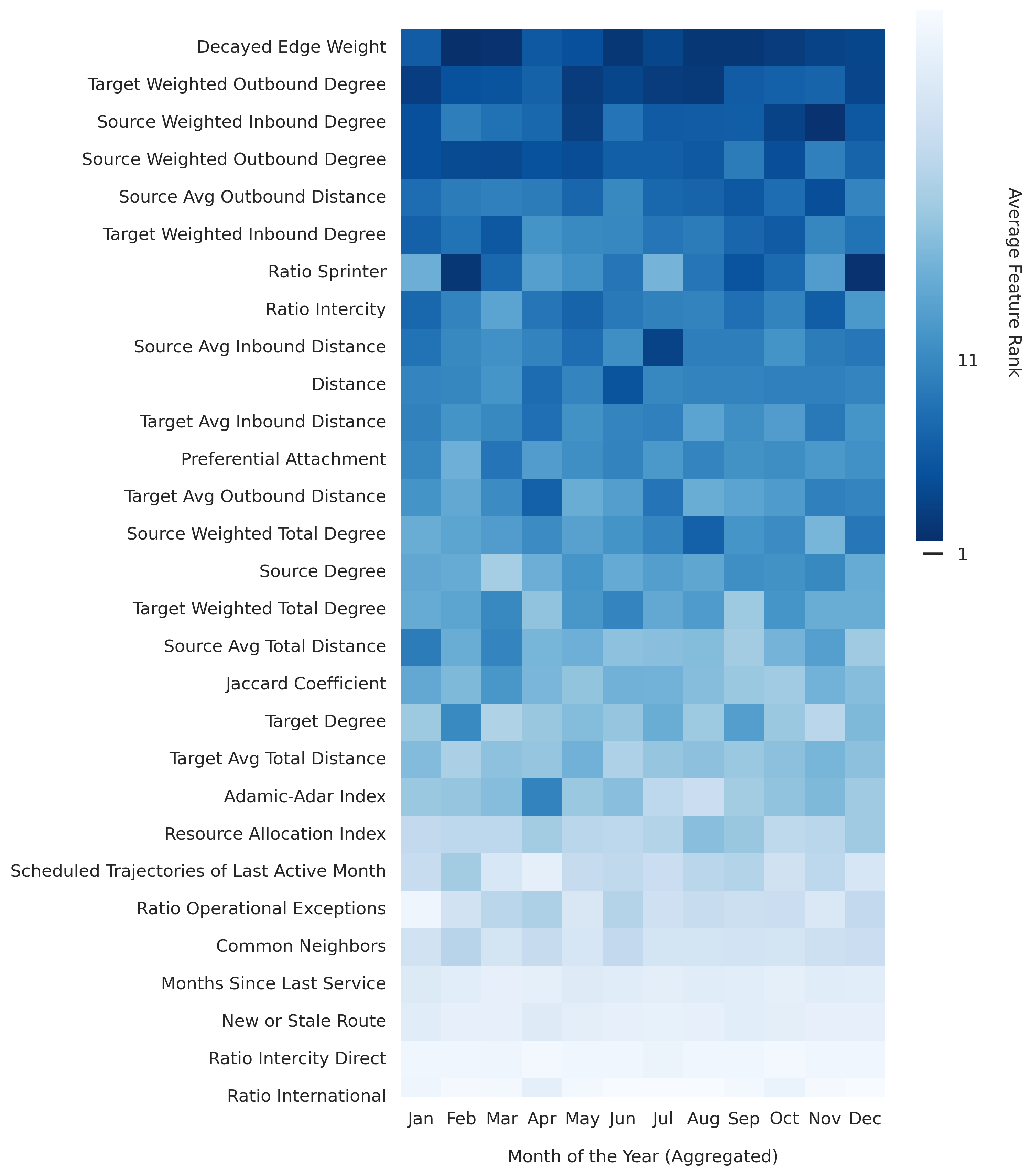}
        \caption{Topology + Weight + Operational}
        \label{fig:xgb_mom_twco}
    \end{subfigure}
    \hfill
    \begin{subfigure}[b]{0.45\textwidth}
        \centering
        \includegraphics[width=\textwidth]{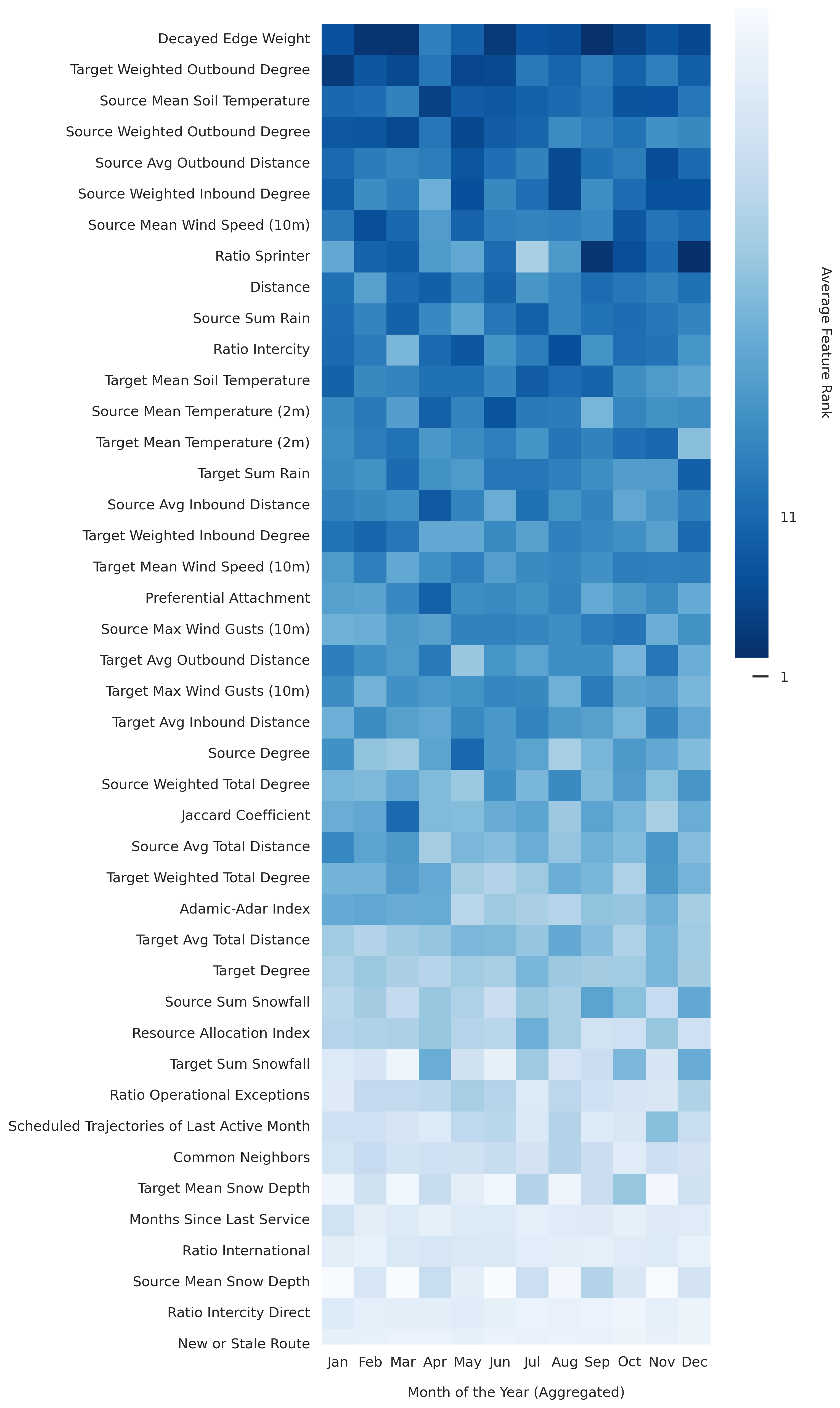}
        \caption{All Features}
        \label{fig:xgb_mom_allf}
    \end{subfigure}
    \caption{Aggregated (2019-2024) Month-over-Month SHAP feature importance for the XGBoost model evaluating combined operational features and all features set.}
    \label{fig:xgb_mom_combined_grid_2}
\end{figure}

\clearpage
\section{SHAP Value Feature Impact Distributions}
\label{sec:apx:shap_analysis}

\begin{figure}[htbp]
\centering
\begin{subfigure}[b]{0.45\textwidth}
\centering
\includegraphics[width=\textwidth]{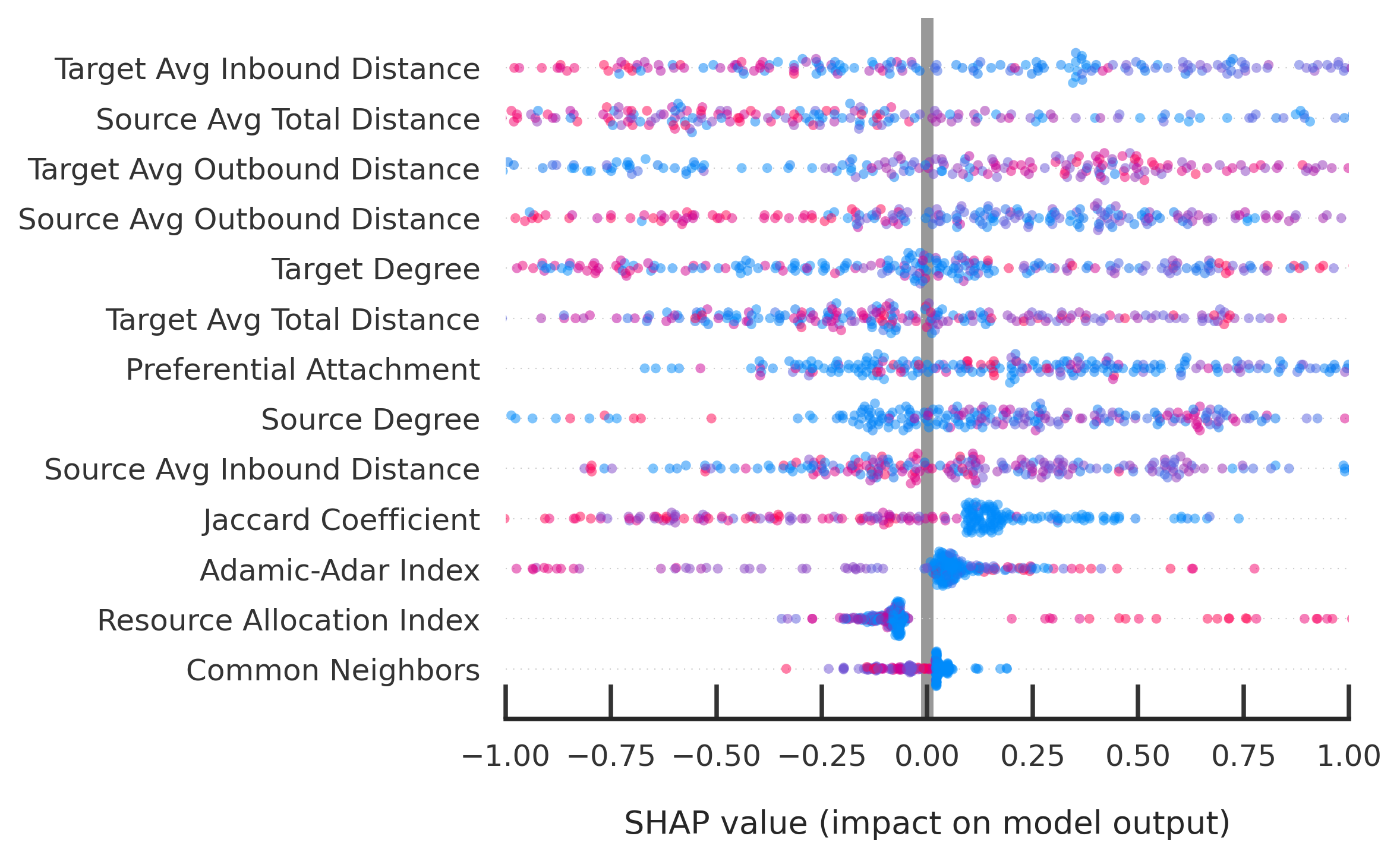}
\caption{Topology}
\label{fig:xgb_shap_top}
\end{subfigure}
\hfill
\begin{subfigure}[b]{0.45\textwidth}
\centering
\includegraphics[width=\textwidth]{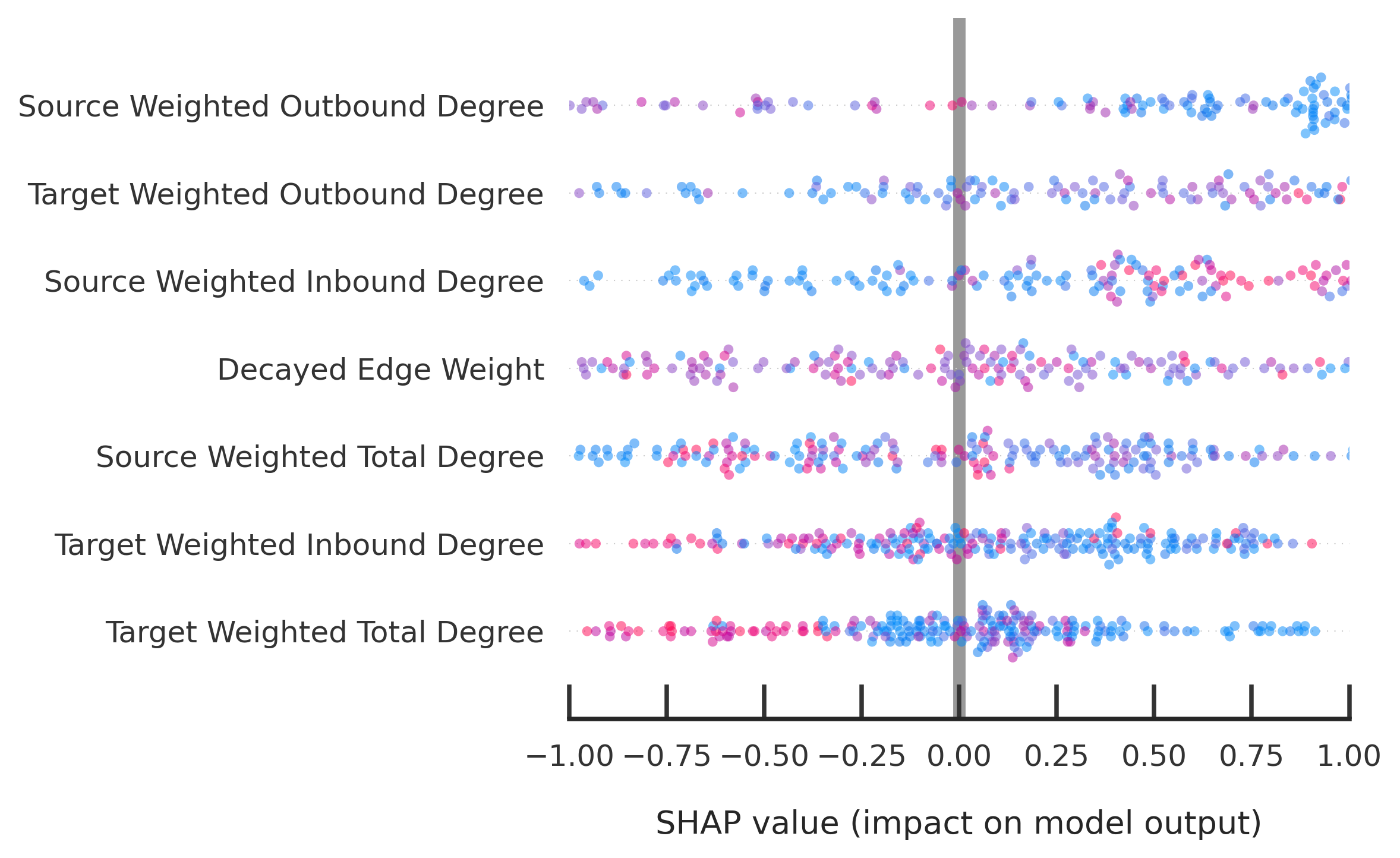}
\caption{Weight}
\label{fig:xgb_shap_wgt}
\end{subfigure}

\vspace{0.5cm}

\begin{subfigure}[b]{0.45\textwidth}
    \centering
    \includegraphics[width=\textwidth]{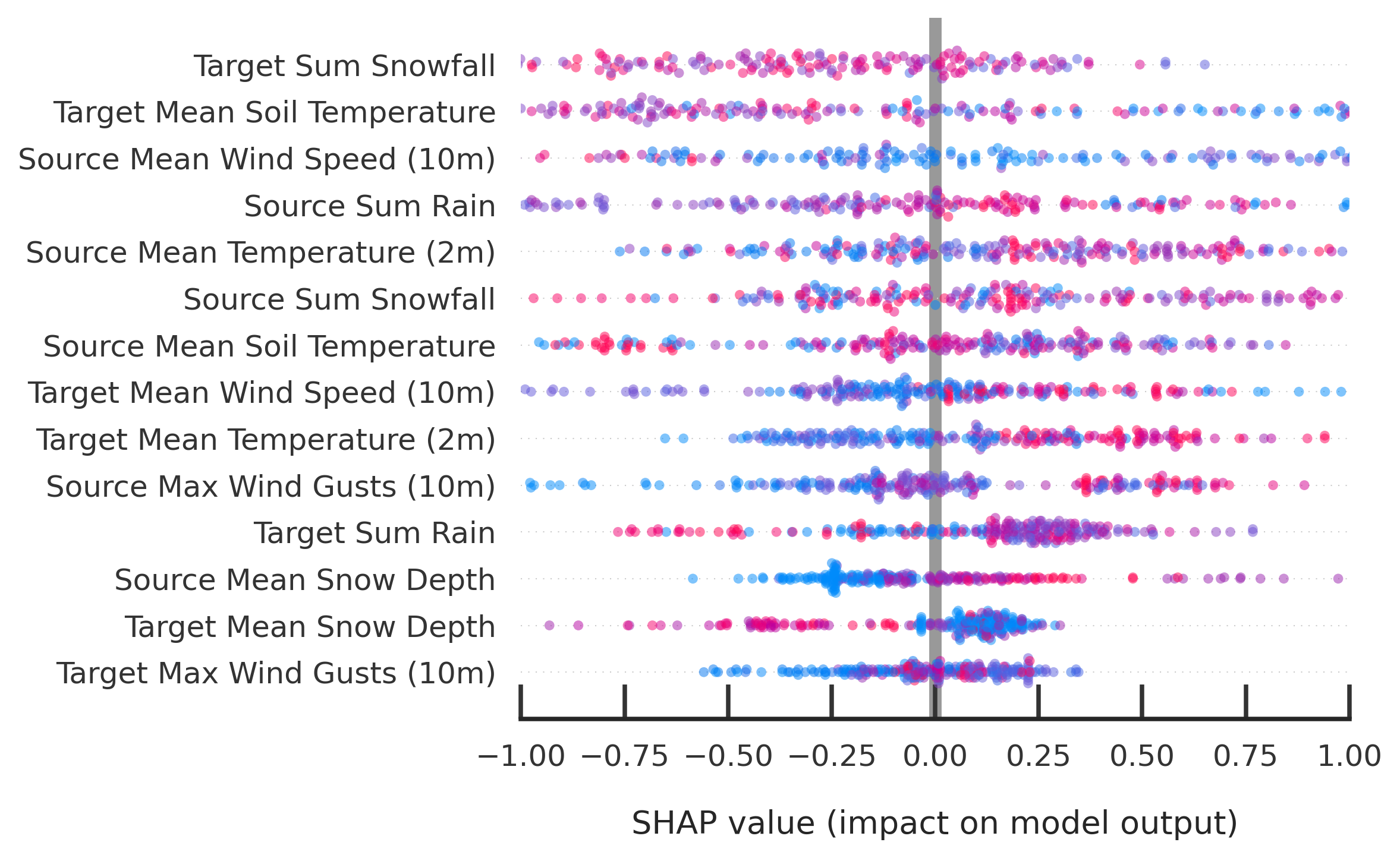}
    \caption{Weather}
    \label{fig:xgb_shap_wth}
\end{subfigure}
\hfill
\begin{subfigure}[b]{0.45\textwidth}
    \centering
    \includegraphics[width=\textwidth]{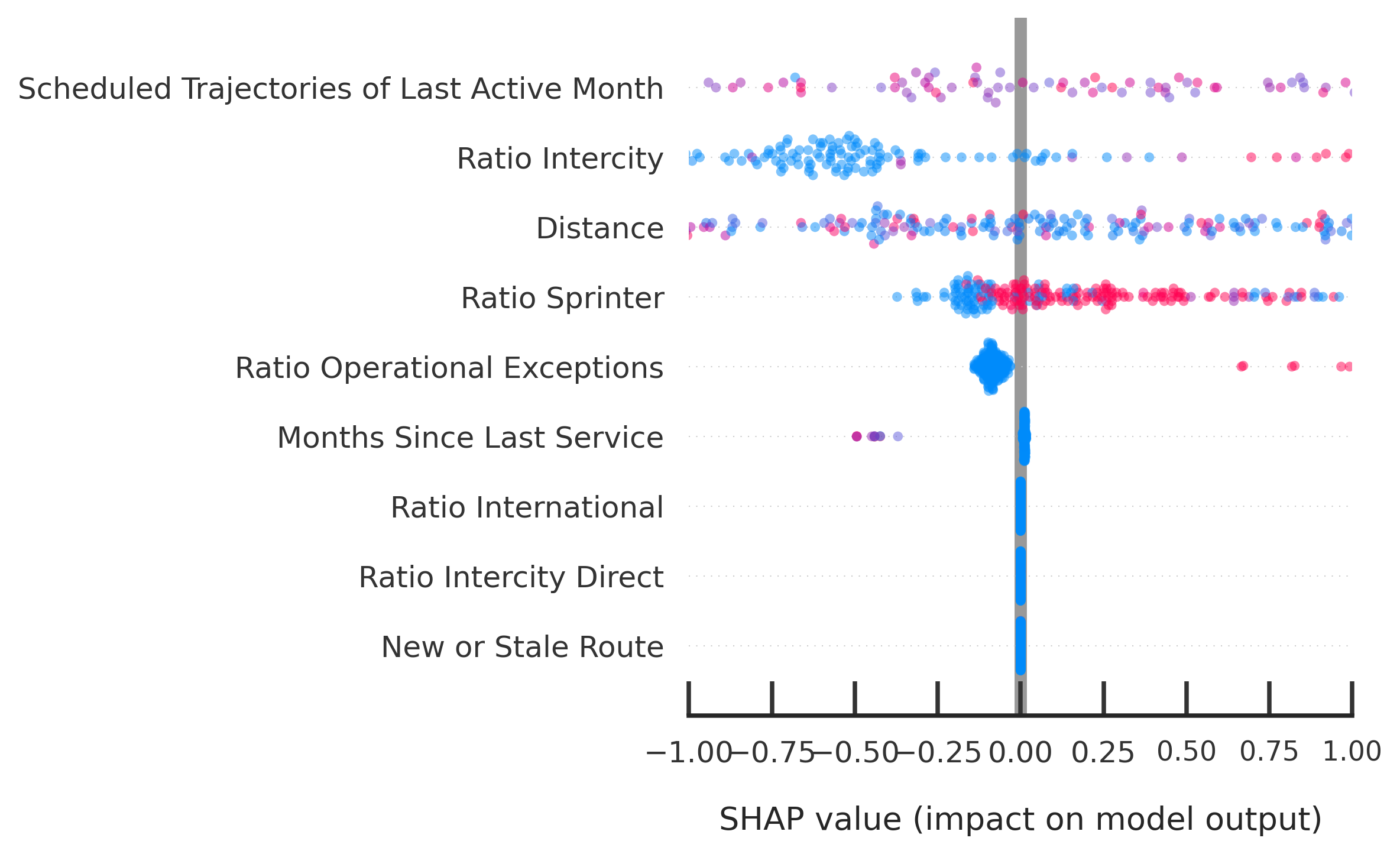}
    \caption{Operational}
    \label{fig:xgb_shap_ops}
\end{subfigure}
\caption{SHAP summary plots illustrating feature impact magnitudes for the XGBoost model utilizing individual feature sets of period 2024-01.}
\label{fig:xgb_shap_core_grid}
\end{figure}

\begin{figure}[htbp]
\centering
\begin{subfigure}[b]{0.45\textwidth}
\centering
\includegraphics[width=\textwidth]{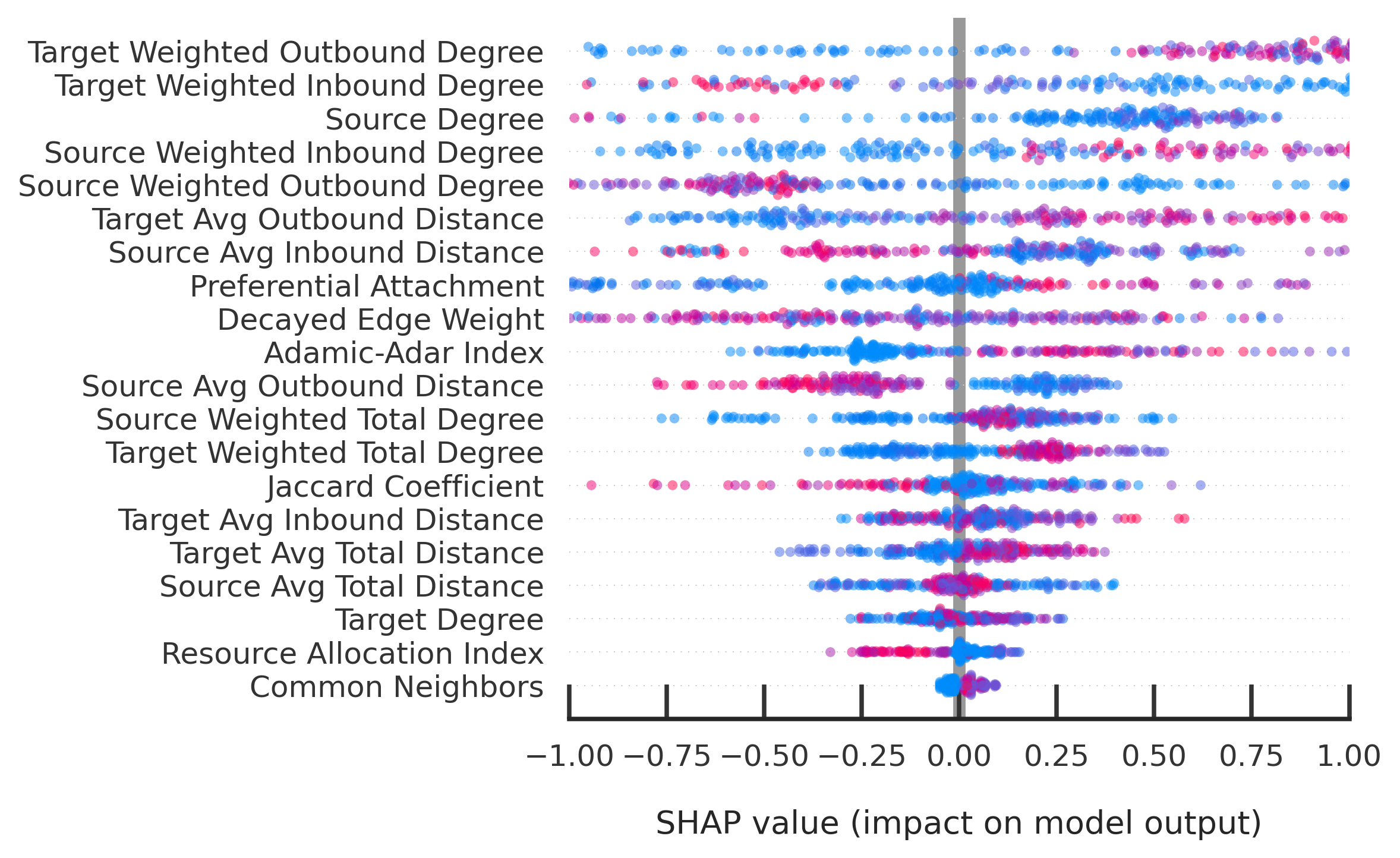}
\caption{Topology + Weight}
\label{fig:xgb_shap_tw}
\end{subfigure}
\hfill
\begin{subfigure}[b]{0.45\textwidth}
\centering
\includegraphics[width=\textwidth]{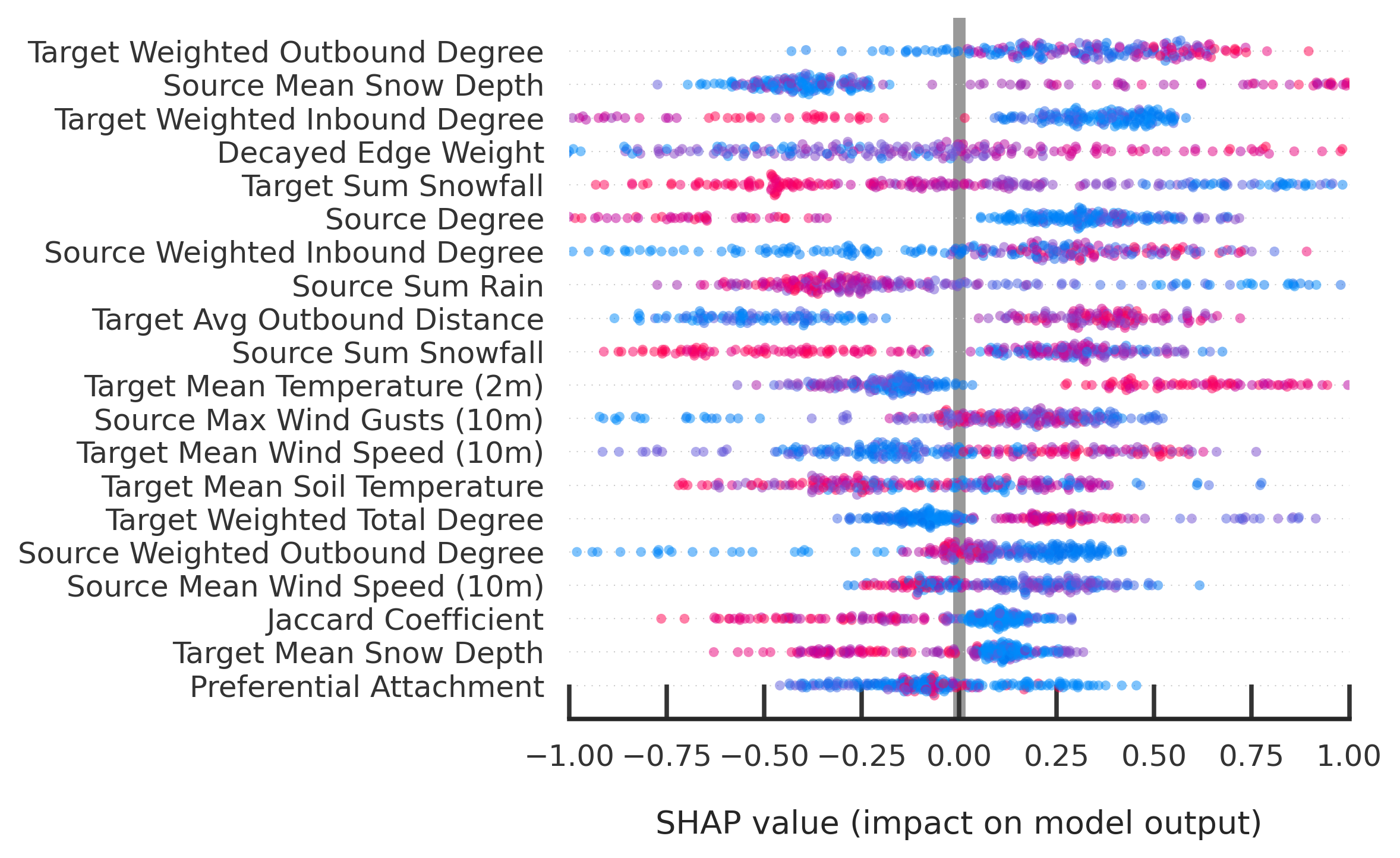}
\caption{Topological + Weight + Weather}
\label{fig:xgb_shap_twcw}
\end{subfigure}

\vspace{0.5cm}

\begin{subfigure}[b]{0.45\textwidth}
    \centering
    \includegraphics[width=\textwidth]{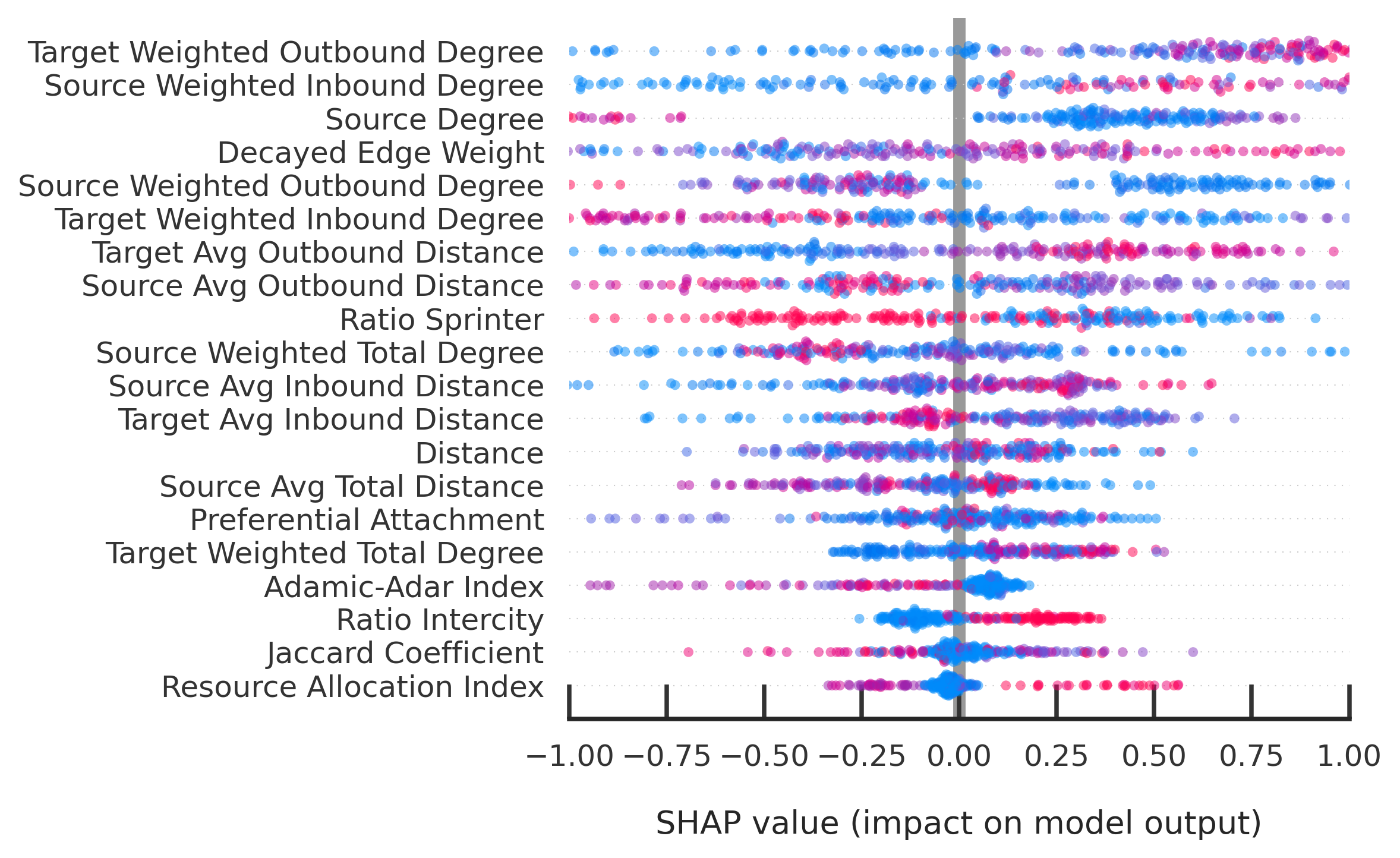}
    \caption{Topological + Weight + Operational}
    \label{fig:xgb_shap_twco}
\end{subfigure}
\hfill
\begin{subfigure}[b]{0.45\textwidth}
    \centering
    \includegraphics[width=\textwidth]{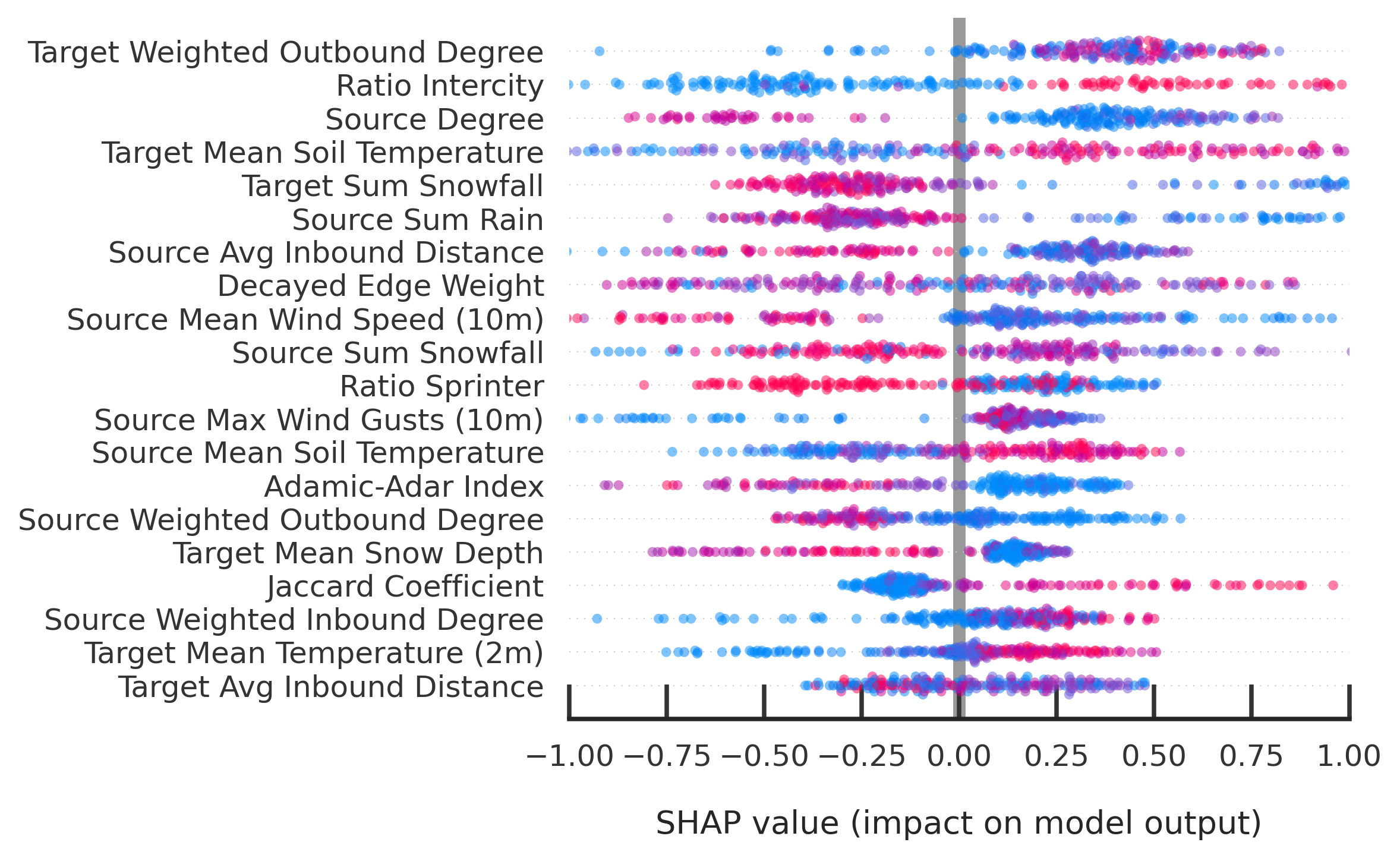}
    \caption{All Features}
    \label{fig:xgb_shap_allf}
\end{subfigure}
\caption{SHAP summary plots illustrating feature impact magnitudes for the XGBoost model utilizing combined feature sets of period 2024-01. Color indicates feature value (red = high, blue = low).}
\label{fig:xgb_shap_combined_grid}
\end{figure}

\clearpage
\section{Confusion Matrices}
\label{sec:apx:confusion_matrices}

\begin{figure}[H]
    \centering
    \begin{subfigure}[b]{0.48\textwidth}
        \centering
        \includegraphics[width=\textwidth, height=0.22\textheight, keepaspectratio]{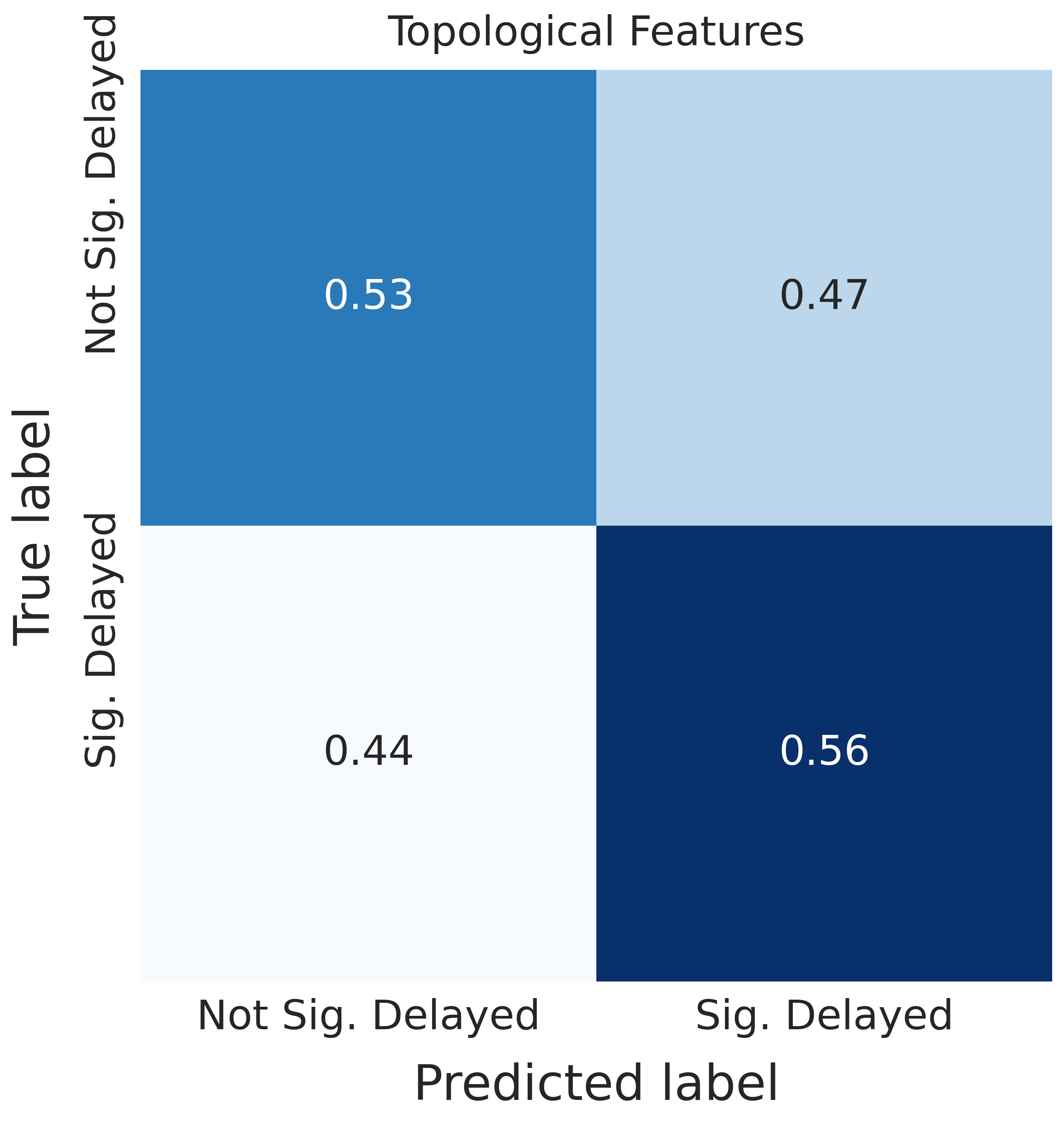}
        \caption{Simultaneous: Topological}
        \label{fig:cm_sim_top}
    \end{subfigure}
    \hfill
    \begin{subfigure}[b]{0.48\textwidth}
        \centering
        \includegraphics[width=\textwidth, height=0.22\textheight, keepaspectratio]{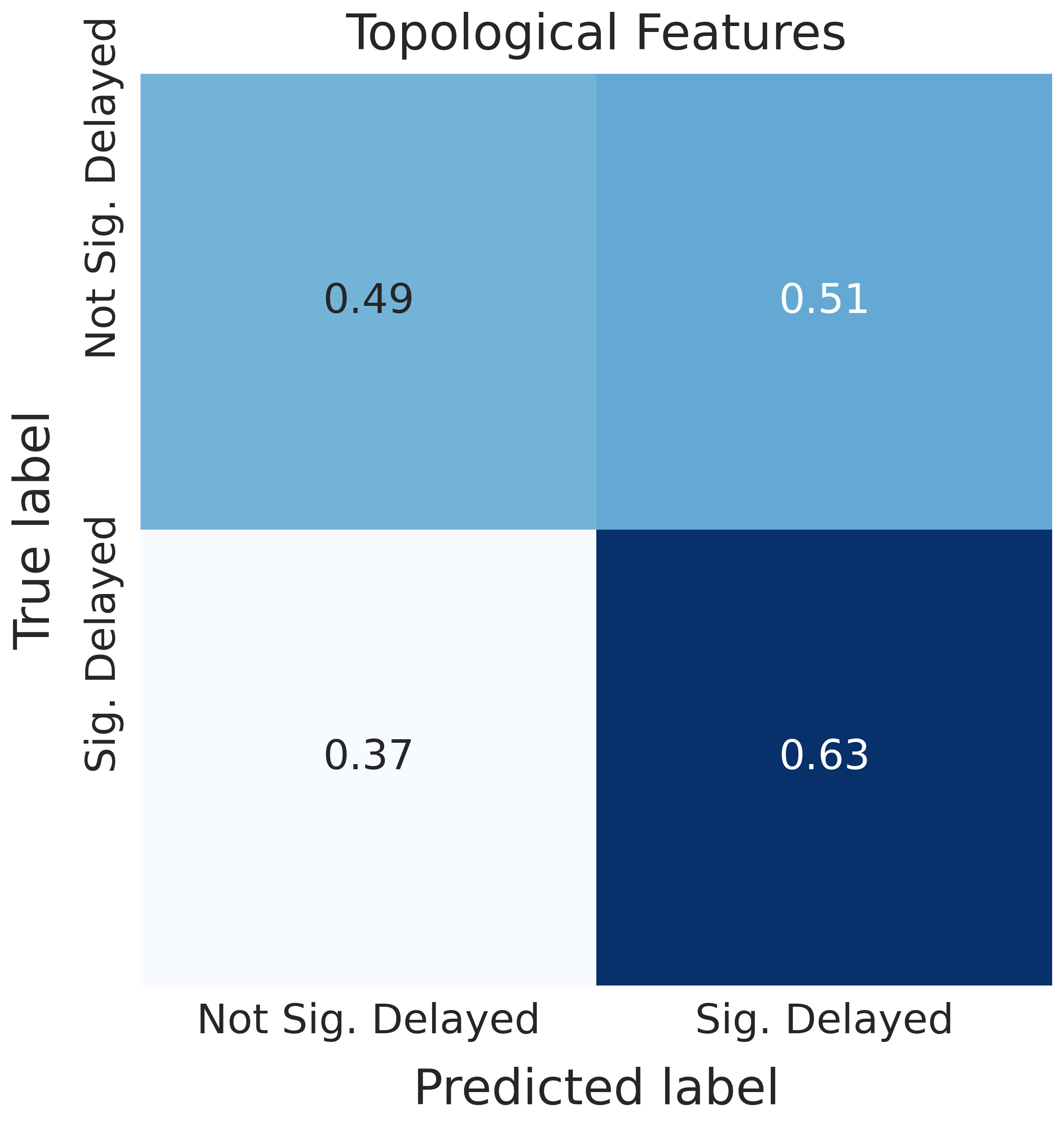}
        \caption{Non-Simultaneous: Topological}
        \label{fig:cm_nonsim_top}
    \end{subfigure}
    \vspace{0.2cm}

    \begin{subfigure}[b]{0.48\textwidth}
        \centering
        \includegraphics[width=\textwidth, height=0.22\textheight, keepaspectratio]{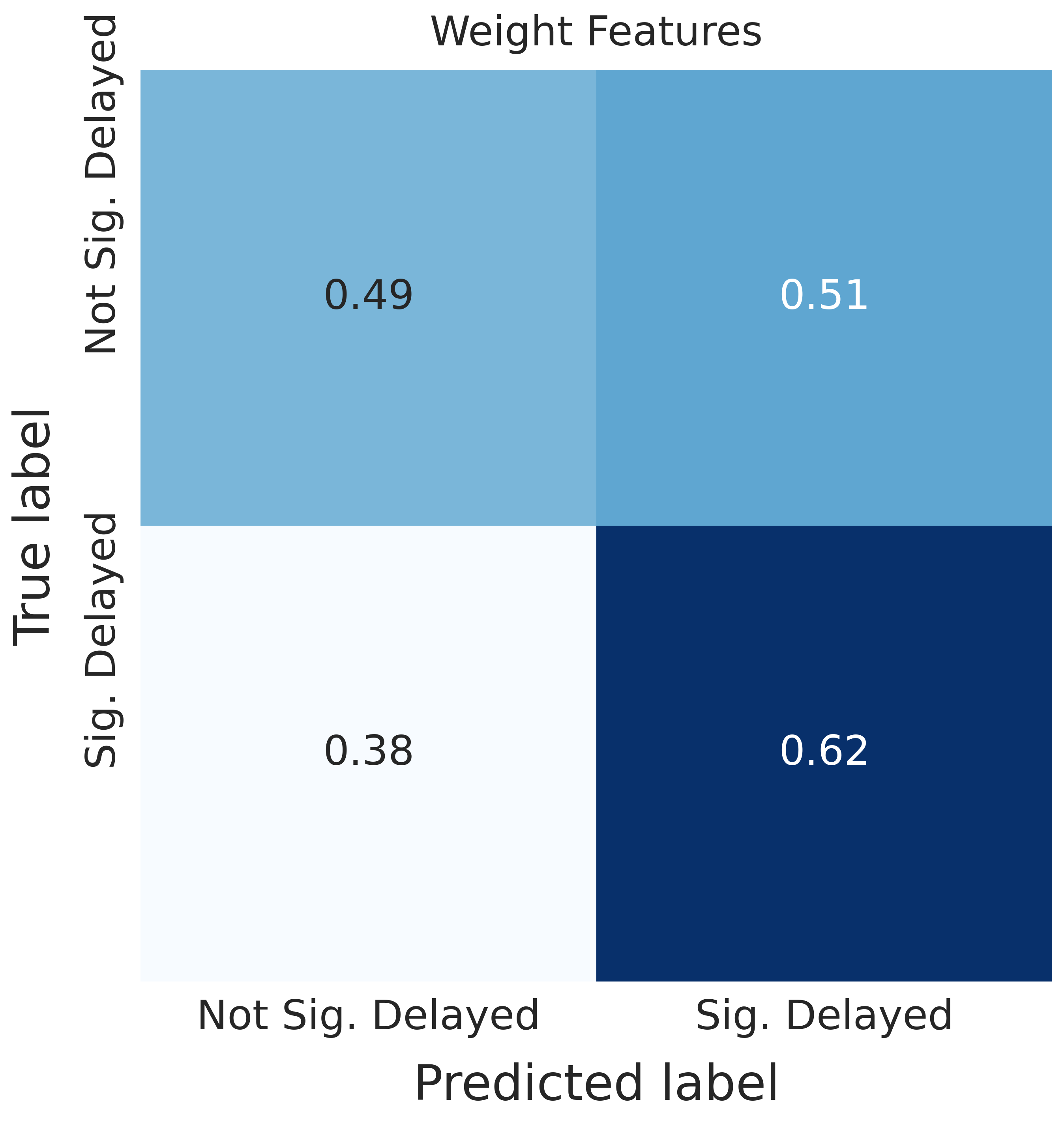}
        \caption{Simultaneous: Weight}
        \label{fig:cm_sim_wgt}
    \end{subfigure}
    \hfill
    \begin{subfigure}[b]{0.48\textwidth}
        \centering
        \includegraphics[width=\textwidth, height=0.22\textheight, keepaspectratio]{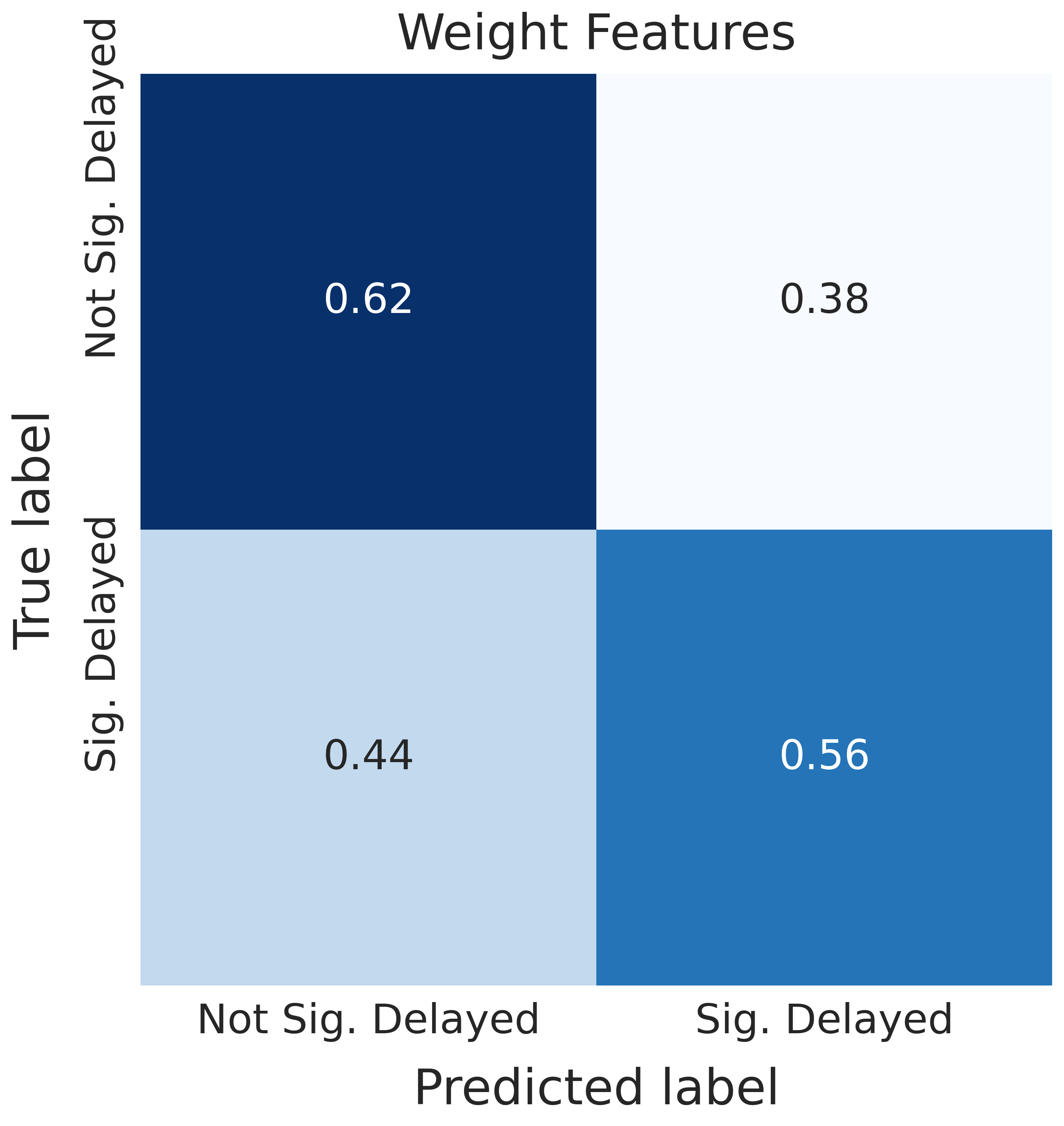}
        \caption{Non-Simultaneous: Weight}
        \label{fig:cm_nonsim_wgt}
    \end{subfigure}
    \vspace{0.2cm}

    \begin{subfigure}[b]{0.48\textwidth}
        \centering
        \includegraphics[width=\textwidth, height=0.22\textheight, keepaspectratio]{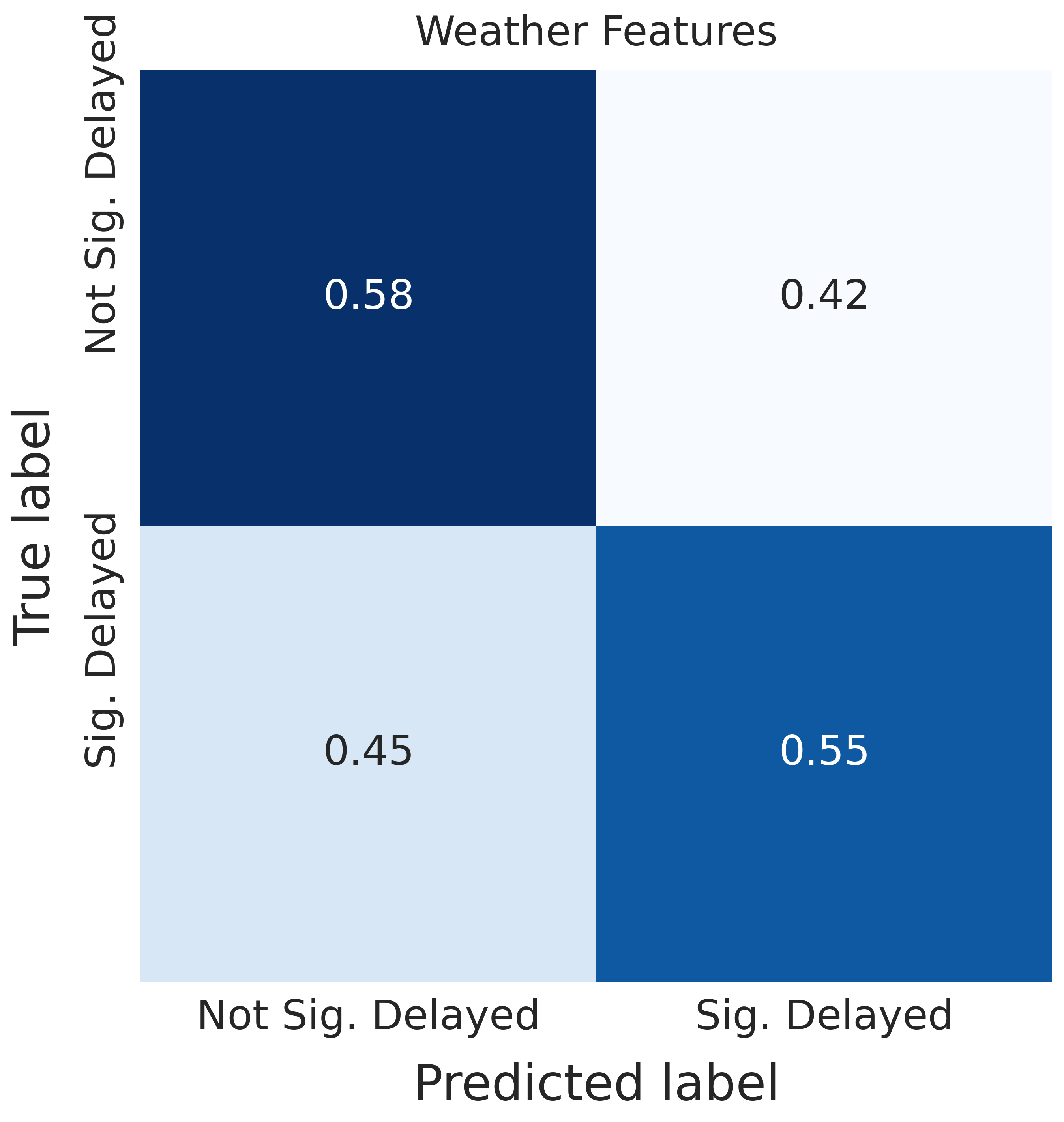}
        \caption{Simultaneous: Weather}
        \label{fig:cm_sim_wth}
    \end{subfigure}
    \hfill
    \begin{subfigure}[b]{0.48\textwidth}
        \centering
        \includegraphics[width=\textwidth, height=0.22\textheight, keepaspectratio]{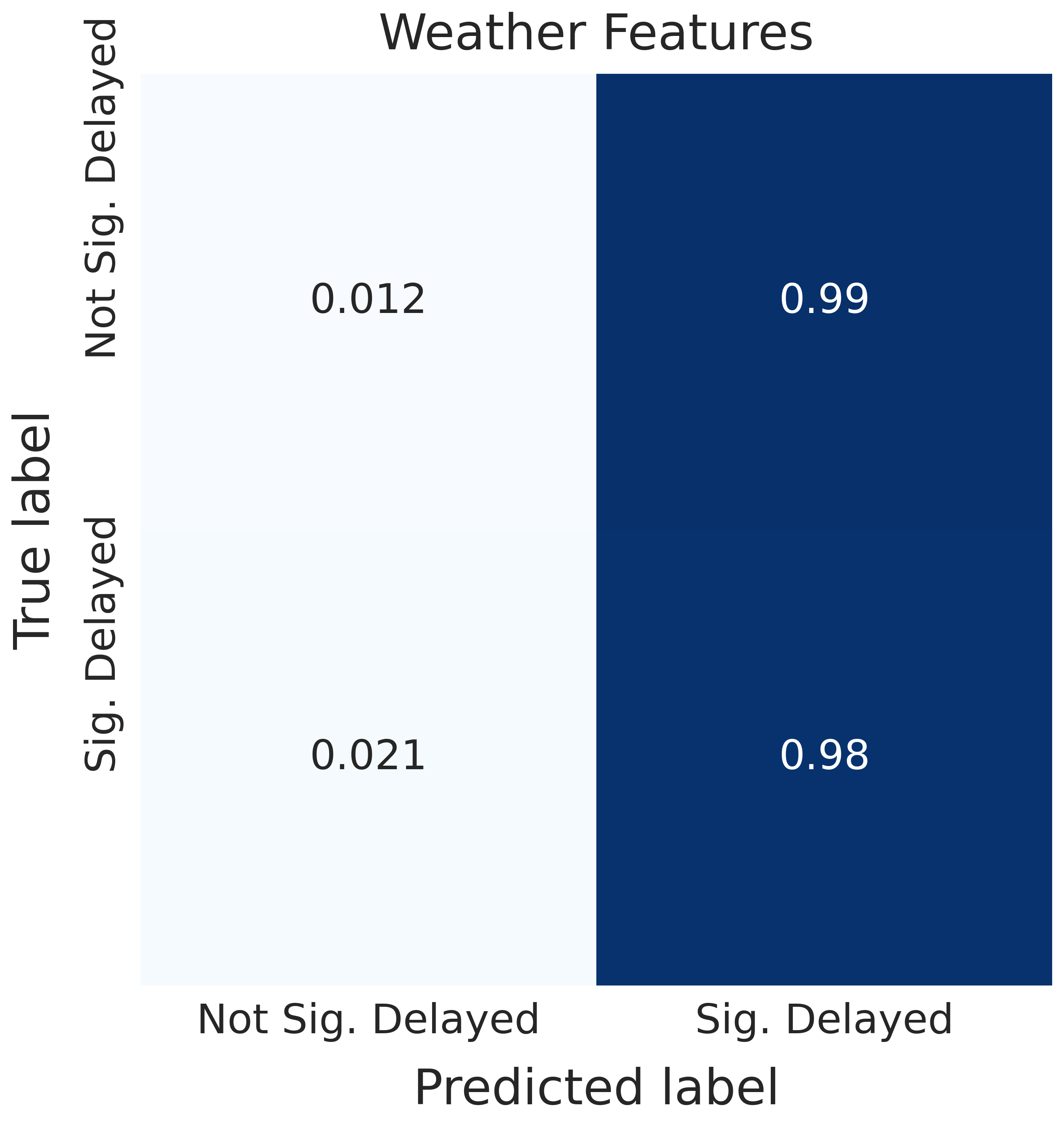}
        \caption{Non-Simultaneous: Weather}
        \label{fig:cm_nonsim_wth}
    \end{subfigure}
    \vspace{0.2cm}

    \begin{subfigure}[b]{0.48\textwidth}
        \centering
        \includegraphics[width=\textwidth, height=0.22\textheight, keepaspectratio]{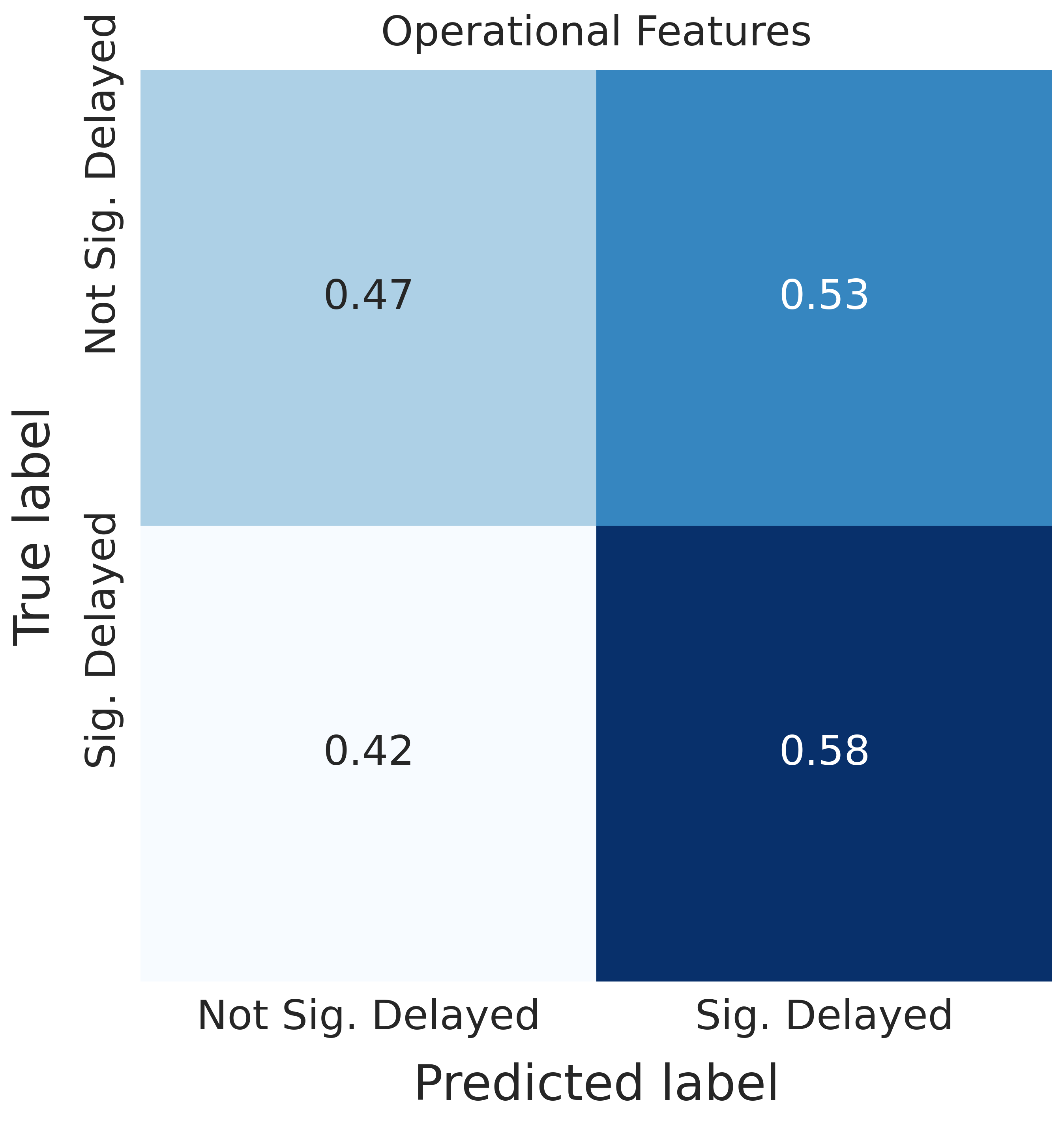}
        \caption{Simultaneous: Operational}
        \label{fig:cm_sim_ops}
    \end{subfigure}
    \hfill
    \begin{subfigure}[b]{0.48\textwidth}
        \centering
        \includegraphics[width=\textwidth, height=0.22\textheight, keepaspectratio]{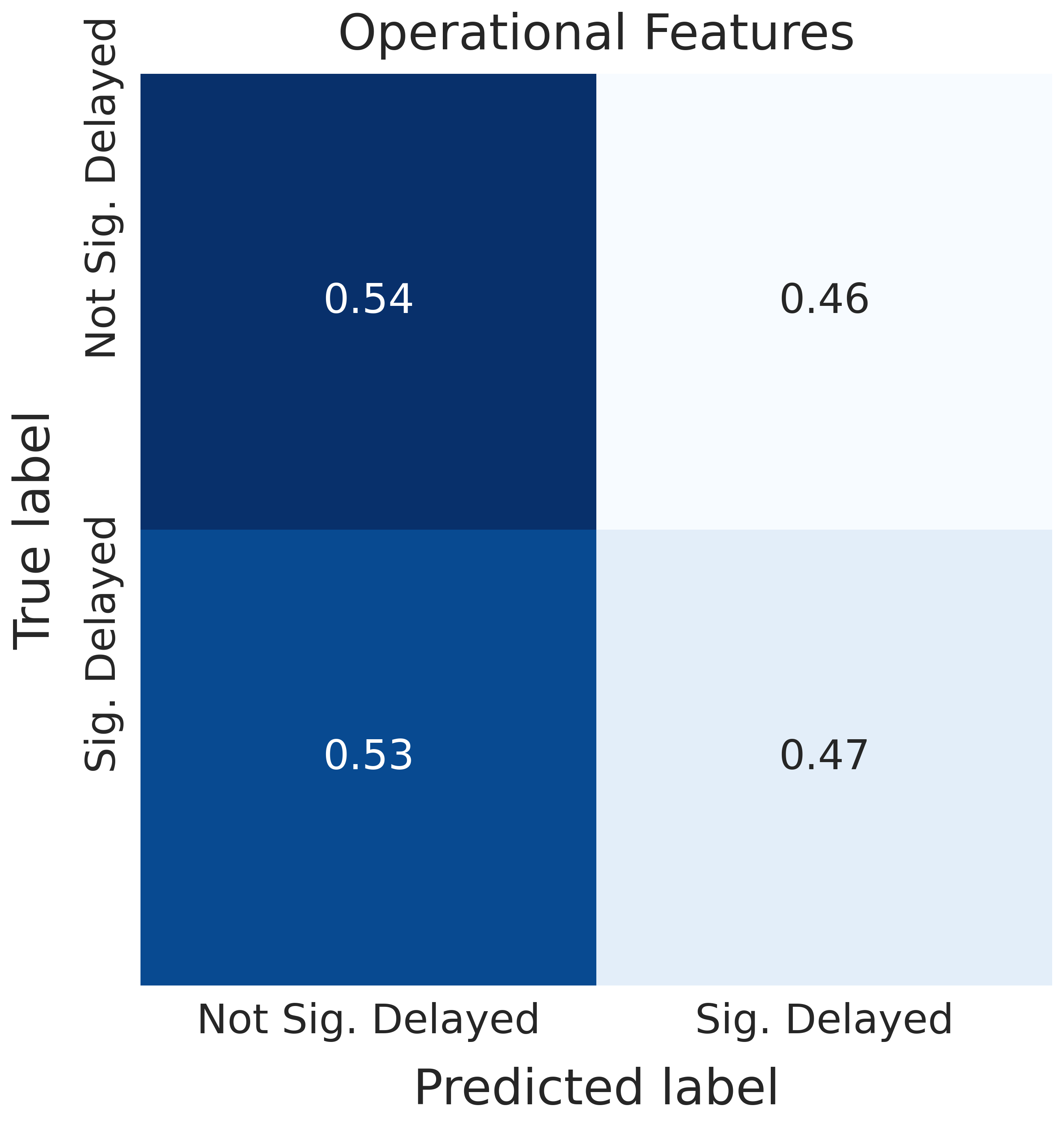}
        \caption{Non-Simultaneous: Operational}
        \label{fig:cm_nonsim_ops}
    \end{subfigure}
    \caption{Side-by-side confusion matrix comparisons for XGBoost models utilizing individual Topological, Weight, Weather, and Operational feature sets of period 2024-01.}
    \label{fig:cm_grid_individual_sets}
\end{figure}
\begin{figure}[htbp]
    \centering
    \begin{subfigure}[b]{0.48\textwidth}
        \centering
        \includegraphics[width=\textwidth, height=0.22\textheight, keepaspectratio]{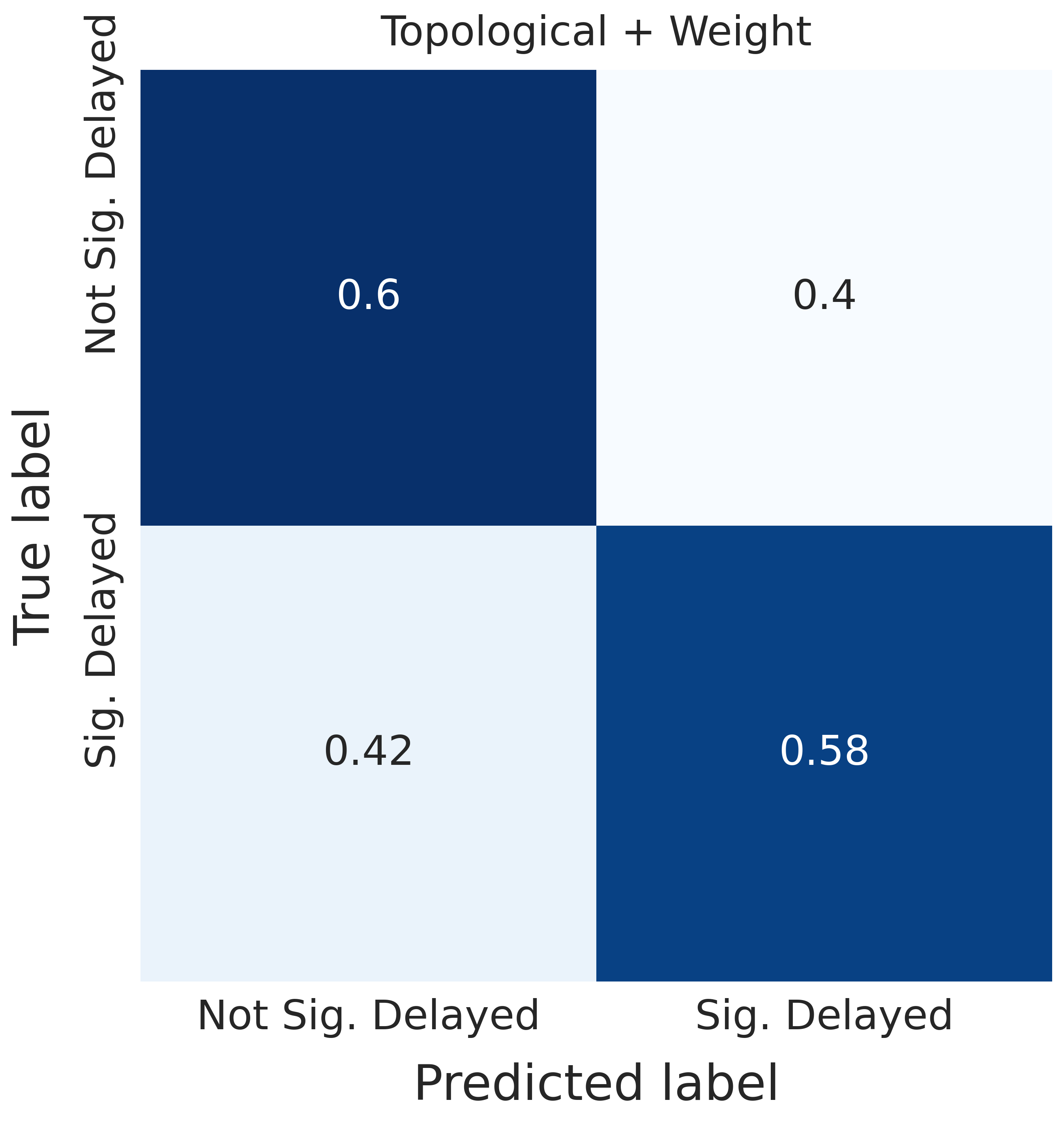}
        \caption{Simultaneous: Topology + Weight}
        \label{fig:cm_sim_tw}
    \end{subfigure}
    \hfill
    \begin{subfigure}[b]{0.48\textwidth}
        \centering
        \includegraphics[width=\textwidth, height=0.22\textheight, keepaspectratio]{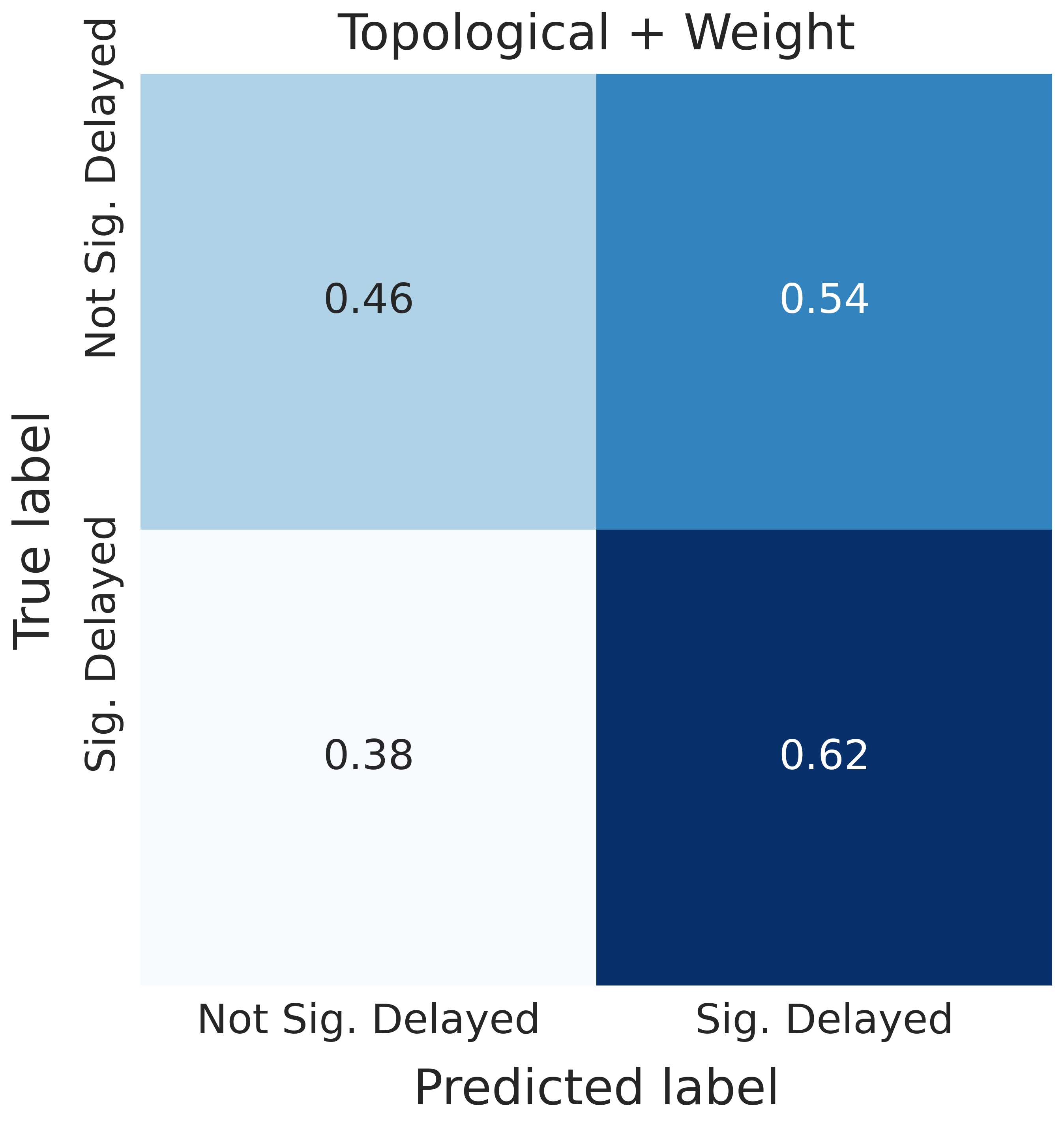}
        \caption{Non-Simultaneous: Topology + Weight}
        \label{fig:cm_nonsim_tw}
    \end{subfigure}
    \vspace{0.2cm}

    \begin{subfigure}[b]{0.48\textwidth}
        \centering
        \includegraphics[width=\textwidth, height=0.22\textheight, keepaspectratio]{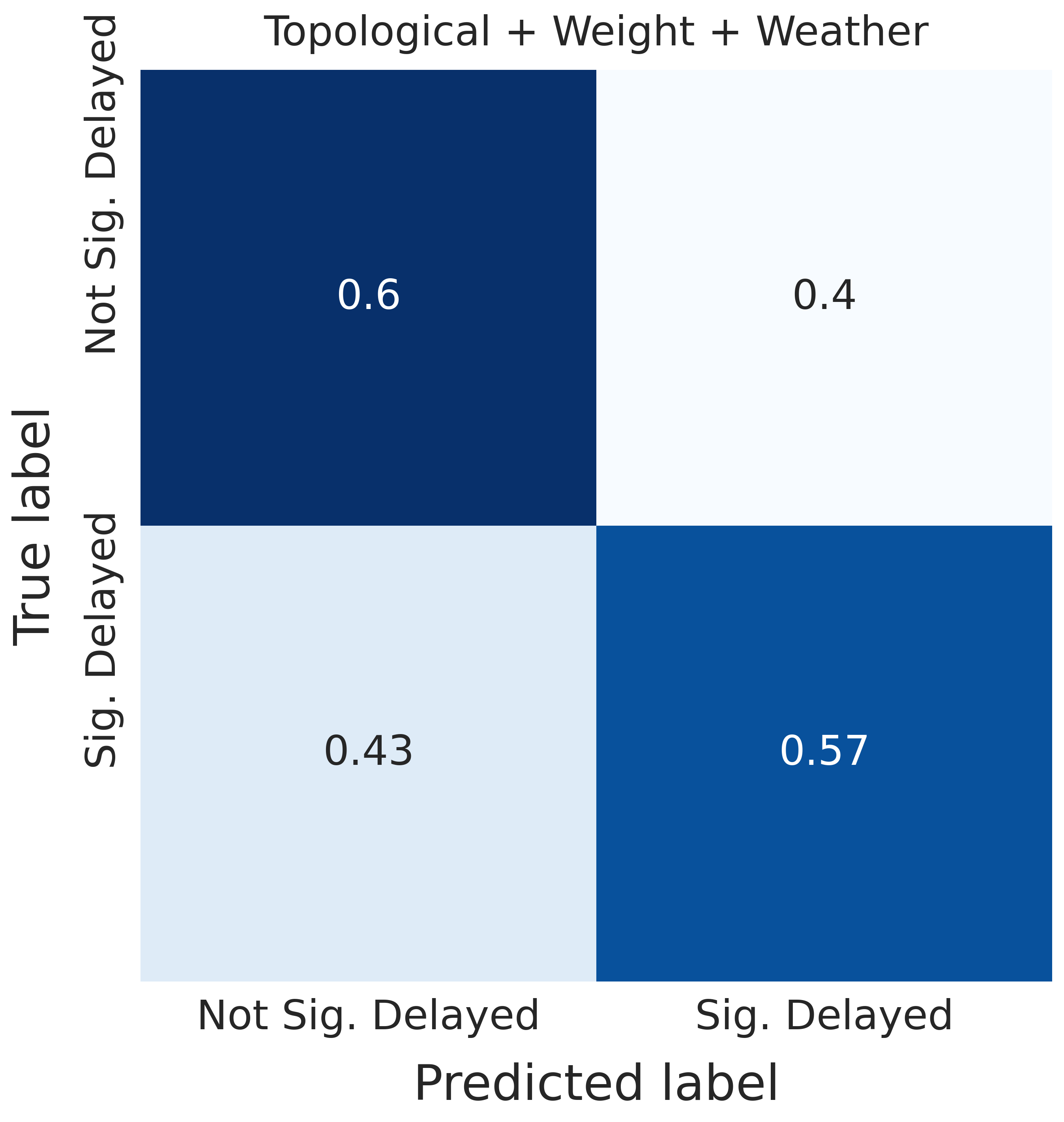}
        \caption{Simultaneous: Topology + Weight + Weather}
        \label{fig:cm_sim_twcw}
    \end{subfigure}
    \hfill
    \begin{subfigure}[b]{0.48\textwidth}
        \centering
        \includegraphics[width=\textwidth, height=0.22\textheight, keepaspectratio]{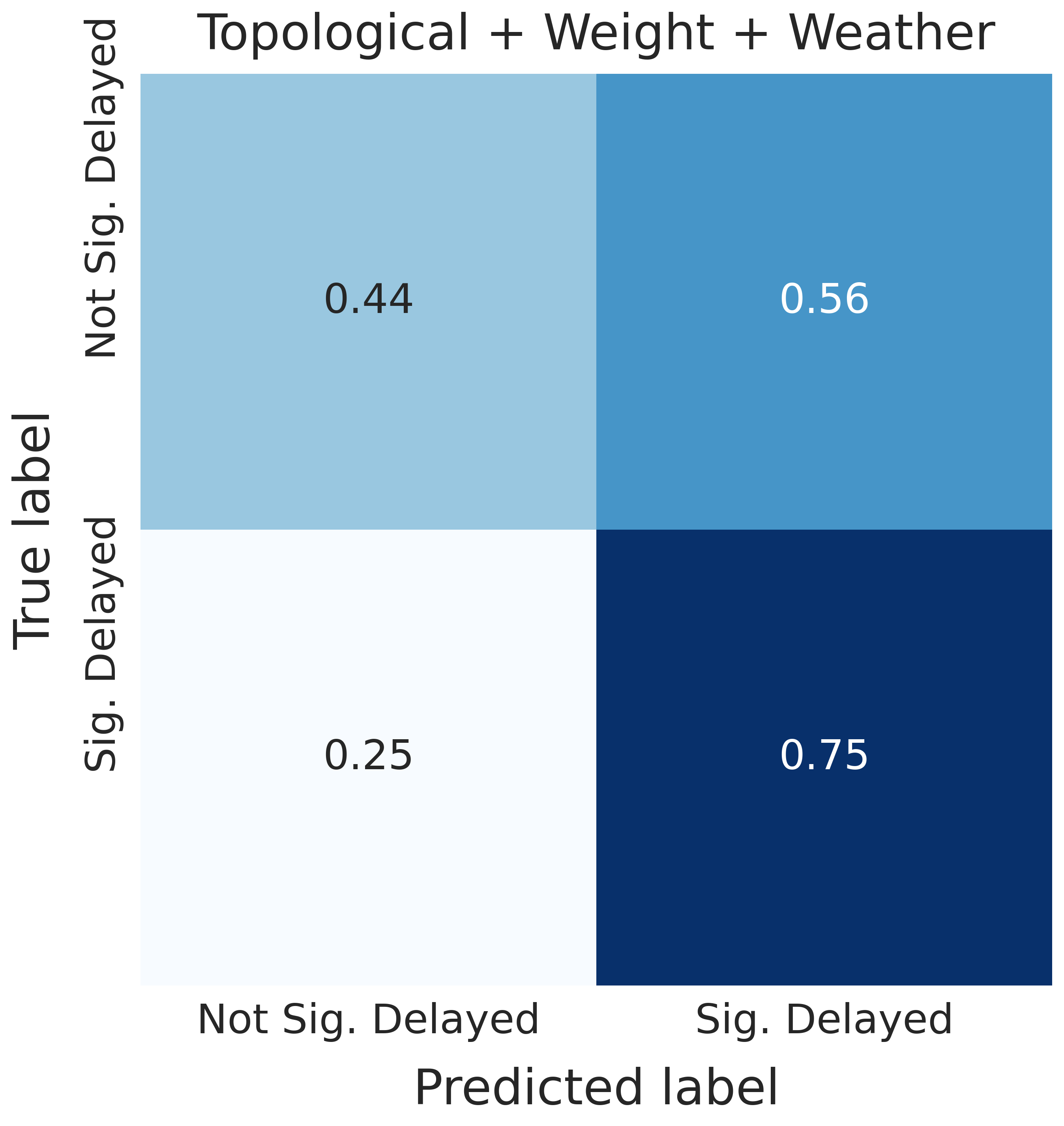}
        \caption{Non-Simultaneous: Topology + Weight + Weather}
        \label{fig:cm_nonsim_twcw}
    \end{subfigure}
    \vspace{0.2cm}

    \begin{subfigure}[b]{0.48\textwidth}
        \centering
        \includegraphics[width=\textwidth, height=0.22\textheight, keepaspectratio]{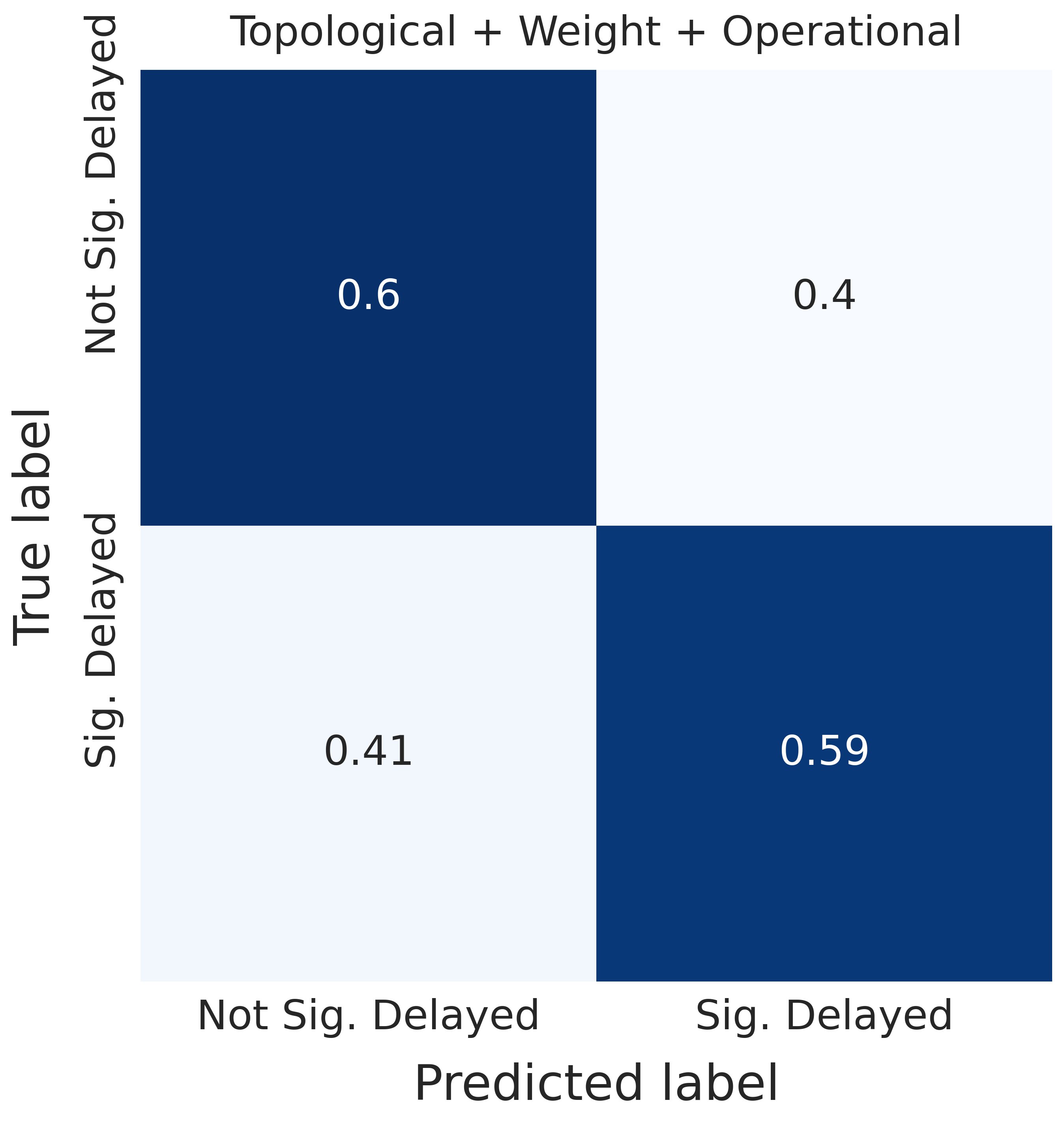}
        \caption{Simultaneous: Topology + Weight + Operational}
        \label{fig:cm_sim_twco}
    \end{subfigure}
    \hfill
    \begin{subfigure}[b]{0.48\textwidth}
        \centering
        \includegraphics[width=\textwidth, height=0.22\textheight, keepaspectratio]{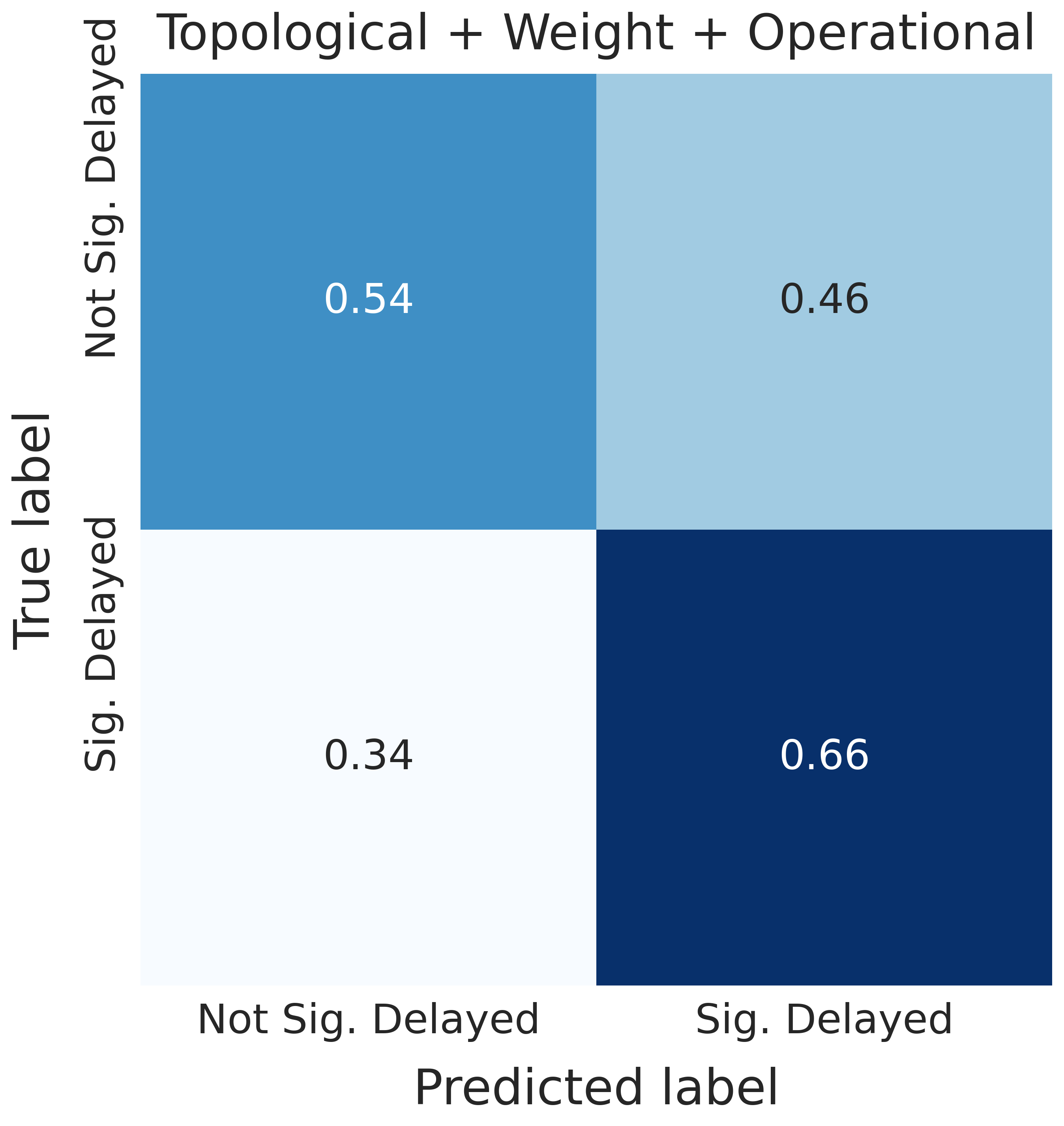}
        \caption{Non-Simultaneous: Topology + Weight + Operational}
        \label{fig:cm_nonsim_twco}
    \end{subfigure}
    \vspace{0.2cm}

    \begin{subfigure}[b]{0.48\textwidth}
        \centering
        \includegraphics[width=\textwidth, height=0.22\textheight, keepaspectratio]{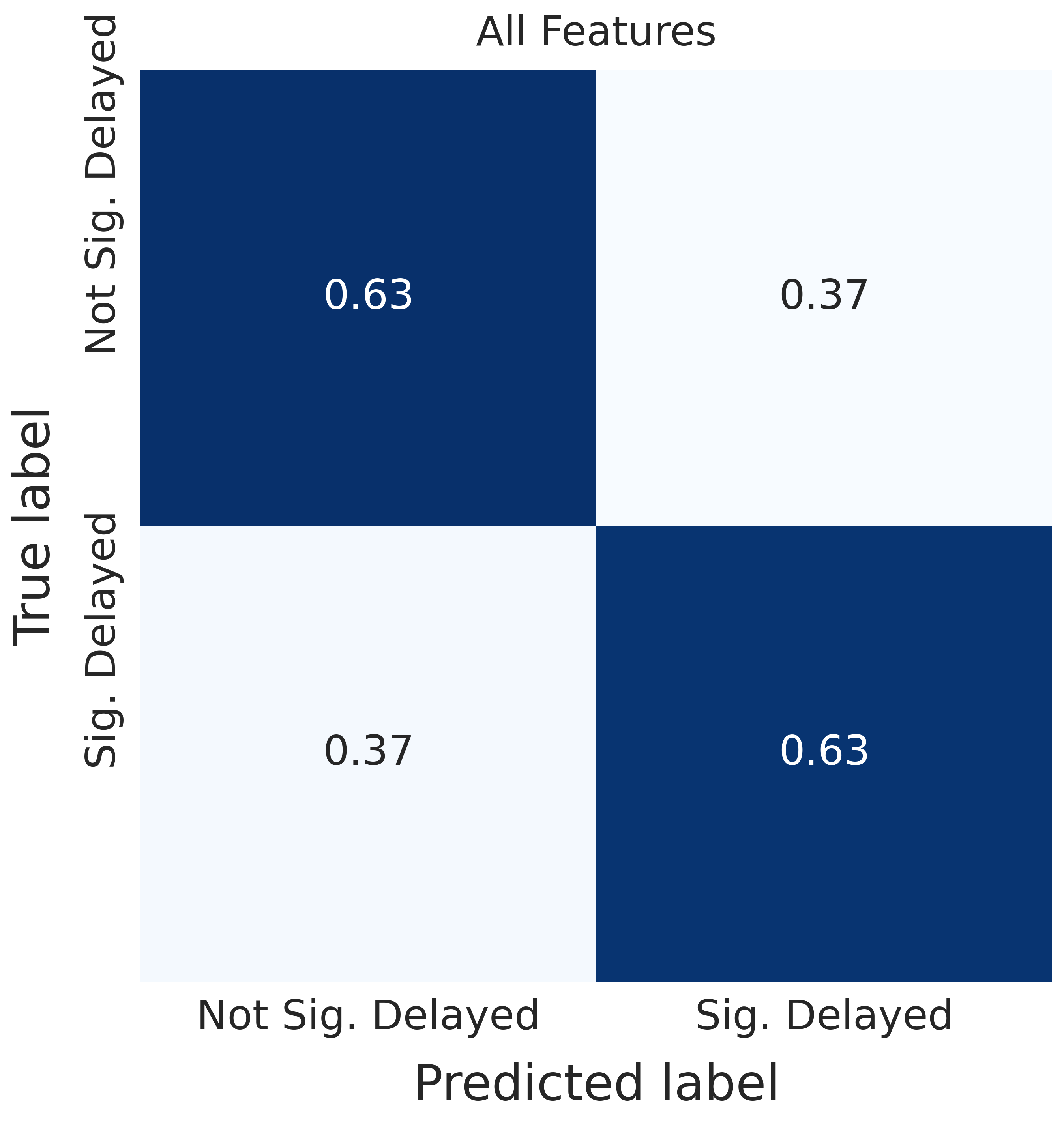}
        \caption{Simultaneous: All Features}
        \label{fig:cm_sim_allf}
    \end{subfigure}
    \hfill
    \begin{subfigure}[b]{0.48\textwidth}
        \centering
        \includegraphics[width=\textwidth, height=0.22\textheight, keepaspectratio]{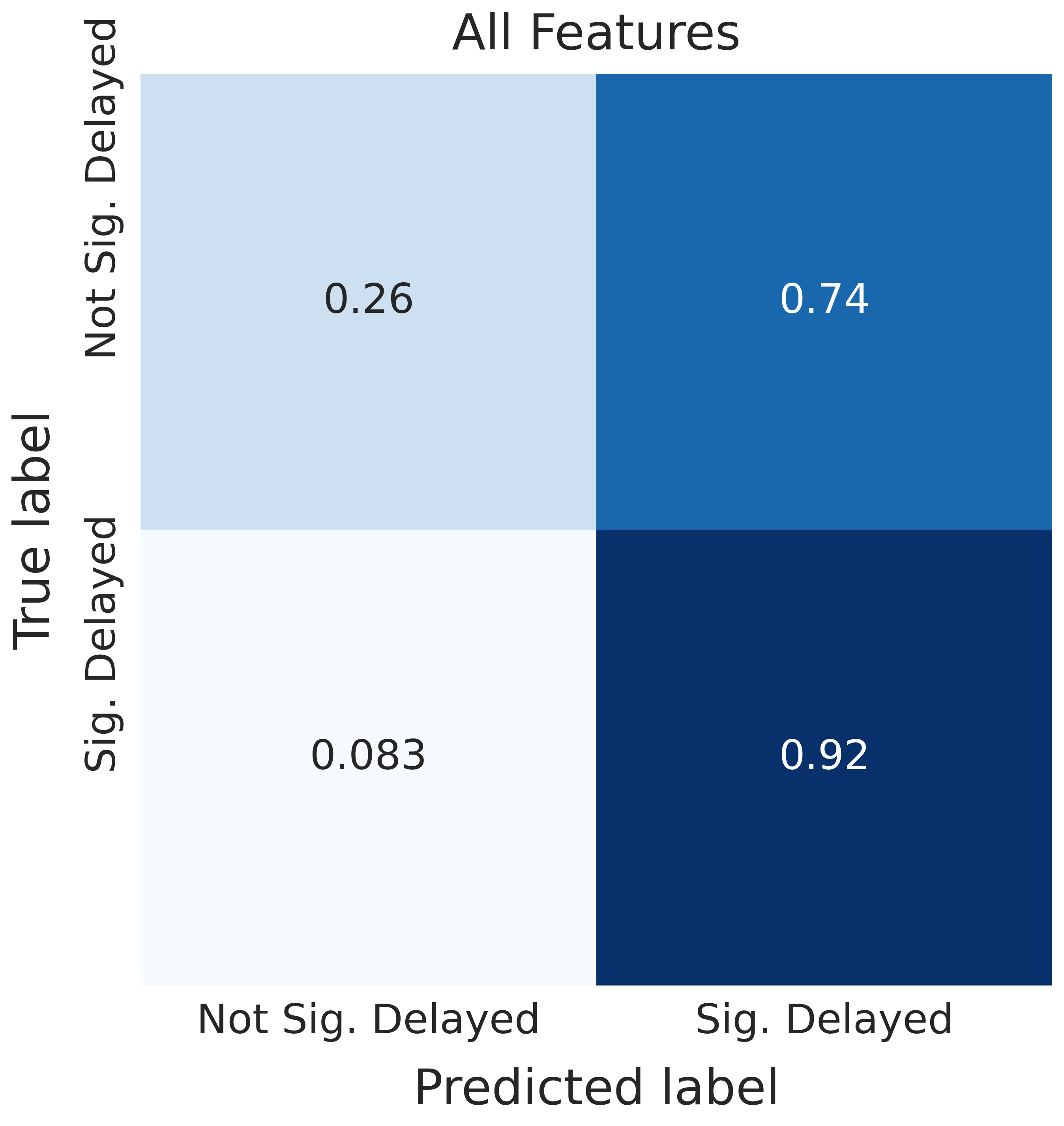}
        \caption{Non-Simultaneous: All Features}
        \label{fig:cm_nonsim_allf}
    \end{subfigure}
    \caption{Side-by-side confusion matrix comparisons for XGBoost models utilizing progressively combined feature sets and the complete feature space of period 2024-01.}
    \label{fig:cm_grid_combined_sets}
\end{figure}

\begin{figure}[htbp]
    \centering
    \begin{subfigure}[b]{0.48\textwidth}
        \centering
        \includegraphics[width=\textwidth]{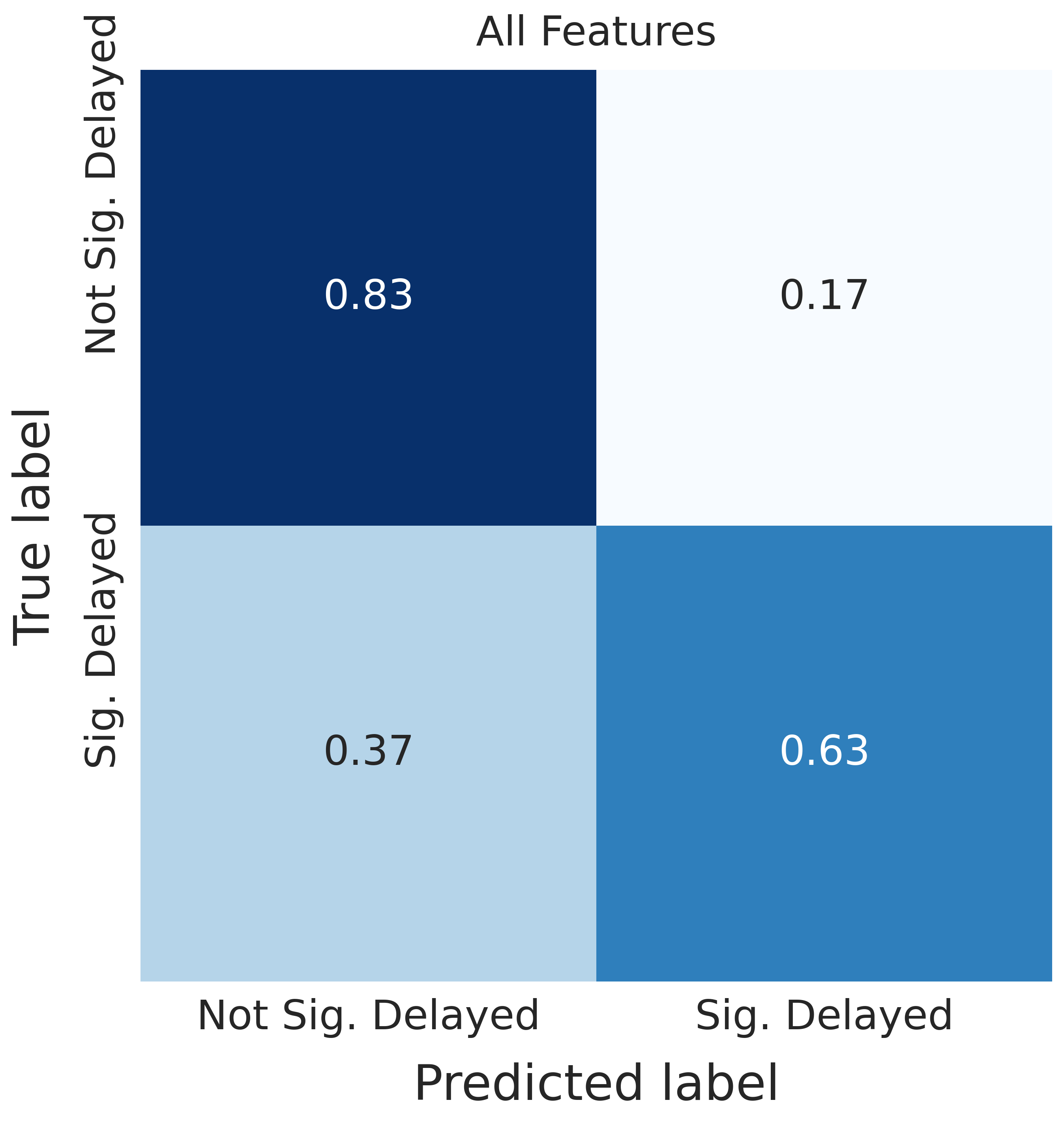}
        \caption{Simultaneous (Tuned Hyperparameters): All Features}
        \label{fig:cm_sim_tuned_allf}
    \end{subfigure}
    \hfill
    \begin{subfigure}[b]{0.48\textwidth}
        \centering
        \includegraphics[width=\textwidth]{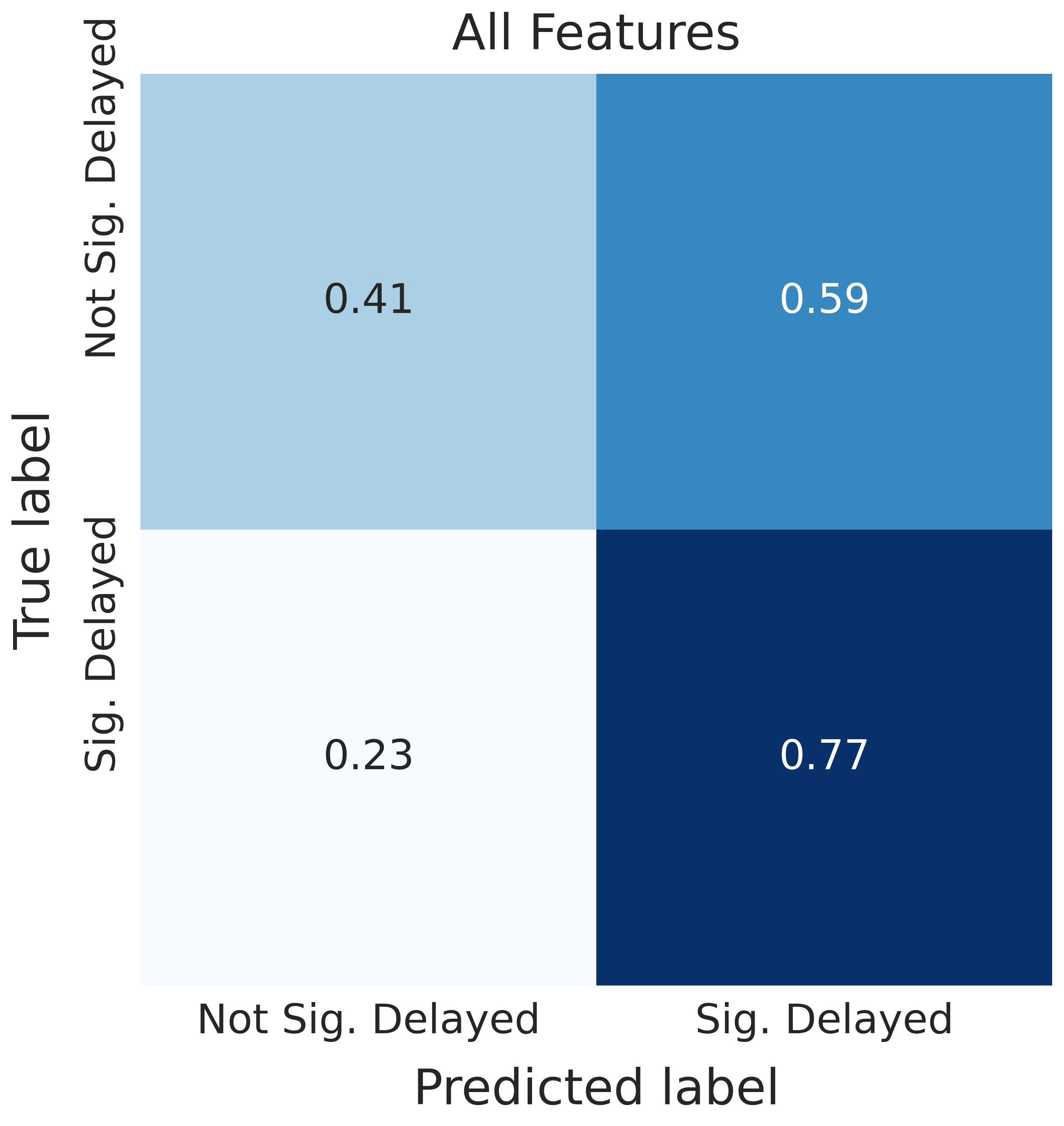}
        \caption{Non-Simultaneous (Tuned Hyperparameters): All Features}
        \label{fig:cm_nonsim_tuned_allf}
    \end{subfigure}
    \caption{Side-by-side confusion matrix comparisons for Tuned XGBoost models utilizing the complete feature space (Topological + Weight + Weather + Operational) of period 2024-01.}
    \label{fig:cm_grid_tuned_allf}
\end{figure}

\end{appendices}

\end{document}